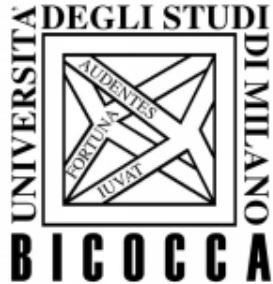

# A Model of Selective Advantage for the Efficient Inference of Cancer Clonal Evolution

Doctoral Thesis of:
Daniele Ramazzotti

Advisor: Professor Marco Antoniotti
Tutor: Professor Fabio Stella
Supervisor: Professor Giancarlo Mauri
Supervisor: Professor Bud Mishra
Reader: Ph.D Giulio Caravagna
Reader: Ph.D Alex Graudenzi

January 2013 - December 2015

*To my beloved Grandpas,*
*whose memory and affection*
*constantly dwell with me*

# ACKNOWLEDGMENTS

The work presented in my PhD thesis has been possible because of so many great people I worked with in the last three years to which I owe a lot and I want to thank here.

The prospect of working on understanding cancer evolution was proposed to me for the first time by my PhD advisor Professor Marco Antoniotti and Professor Giancarlo Mauri at the very start of my PhD in November 2012. At that time, Marco's group at the University of Milan Bicocca was starting a collaboration with Professor Bud Mishra at the New York University. This project turned out later on to be a great fit for me, changed my understanding of cancer and bioinformatics and made me a researcher. I want to thank for this and for the subsequent constant source of inspiration and guidance Marco, Giancarlo and Bud.

The "junior" components of the group arose from this joint collaboration deserve a special thank. In particular I thank Giulio Caravagna and Alex Graudenzi for their continuous daily presence, Loes Olde Loohuis for her priceless "remote" interactions especially in the first half of my PhD and, more recently, Luca De Sano whose technical skills and hard work have been crucial in the development of the algorithms. Giulio, Alex, Loes and Luca working with you paved the way for the achievements of this thesis and it has been for me a great possibility of growth.

Moreover, I need to thank my PhD tutor, Professor Fabio Stella for his always ready guidance when needed.

I also thank all my friends that always made me feel their presence even when I was on the other side of the world.

Lastly and foremost, I want to thank my family for the continuous support of all my choices and for their help through the last 3 years and especially my grandpas that were always in the front line in supporting me and whose memory will always be source of great energy.

It has been a great journey, thank you all for making this possible!

Recently, there has been a resurgence of interest in rigorous and scalable algorithms for efficient inference of cancer progression using genomic patient data. The motivations are manifold: (*i*) rapidly growing *NGS* and *single cell* data from cancer patients, (*ii*) long-felt need for novel Data Science and Machine Learning algorithms well-suited for inferring models of cancer progression, and finally, (*iii*) a desire to understand the temporal and heterogeneous structure of tumor so as to tame its natural progression through most efficacious therapeutic intervention. This thesis presents a multi-disciplinary effort to algorithmically and efficiently model tumor progression involving successive accumulation of genetic alterations, each resulting populations manifesting themselves with a novel *cancer phenotype.*

The framework presented in this work along with efficient algorithms derived from it, represents a novel and versatile approach for inferring cancer progression, whose accuracy and convergence rates surpass other existing techniques. The approach derives its power from many insights from, and contributes to, several fields including algorithms in machine learning, theory of causality, and cancer biology. Furthermore, a versatile and modular pipeline to extract ensemble-level progression models from cross-sectional sequenced cancer genomes is also proposed. The pipeline combines state-of-the-art techniques for sample stratification, driver selection, identification of fitness-equivalent exclusive alterations and progression model inference.

Furthermore, the results are rigorously validated using synthetic data created with realistic generative models, and empirically interpreted in the context of real cancer datasets; in the later case, biologically significant conclusions revealed by the reconstructed progressions are also highlighted. Specifically, it demonstrates also the pipeline's ability to reproduce much of the current knowledge on colorectal cancer progression, as well as to suggest novel experimentally verifiable hypotheses. Lastly, it also proves that the proposed framework can be applied, *mutatis mutandis*, in reconstructing the evolutionary history of cancer clones in single patients, as illustrated by an example with multiple biopsy data from clear cell renal carcinomas.

I

# CHAPTER 1

## INTRODUCTION

In the near future, cancer research is likely to become much more data-centric, primarily because of the rapid growth and ready availability of vast amount of cancer patient data, as well as because of advances in single-molecule single-cell technologies. Nonetheless, it remains impractical to track the tumor progression in any single patient over time, thus limiting the methods to work with data collected from biopsies of untreated tumors, although emerging technology for noninvasive analysis of circulating tumor cells and cell free DNA (in blood and urine) is beginning to paint an incomplete, but useful, picture[1]. Armed with the insights derived from such analysis, it would be possible to optimize therapy design (see CHA [117]) based on techniques of supervisory control theory, as well as to contribute insights for prevention, prognosis, treatment and drug design in new and possibly, unforeseen manners.

Motivated by the increased availability of genetic patient data, in this thesis we therefore focus on the problem of *reconstructing progression models* of cancer. In particular, we aim at inferring the plausible sequences of *genomic alterations* that, by a process of *accumulation*, selectively make a tumor fitter to survive, expand and diffuse (i.e., metastasize). Along the trajectories of progression, a tumor (monotonically) acquires or "activates" mutations in the genome, which, in turn, produce progressively more "viable" clonal subpopulations over the so-called *cancer evolutionary landscape* (see [128, 86, 186]). Knowledge of such progression models is very important for drug development and in therapeutic decisions. For example, it is known that for the same cancer type, patients in different stages of different progressions respond differently to different treatments.

Before moving on in the description of the framework proposed in this thesis, we now describe the adopted model of cancer evolution.

---

[1]For the sake of simplicity, we will limit our algorithmic studies to the ones dealing with data coming from untreated patients, derived from their initial biopsies and/or during "watchful waiting". For other datasets, the algorithm will require electronic medical records of treatments and tests. For the sake of a clear exposition we ignore these issues here.

## 1.1   A model of cancer evolution

Since the late seventies, evolutionary dynamics, with its interplay between variation and selection, has progressively provided the widely-accepted paradigm for the interpretation of cancer emergence and development [143, 52, 40]. Random alterations of an organism's (epi)genome can sometimes confer a functional *selective advantage* to certain cells, in terms of adaptability and ability to survive and proliferate. Since the consequent *clonal expansions* are naturally constrained by the availability of resources (metabolites, oxygen, etc.), further mutations in the emerging heterogeneous tumor populations are necessary to provide additional *fitness* of different kinds that allow survival and proliferation in the unstable micro environment. Such further advantageous mutations will eventually allow some of their sub-clones to outgrow the competing cells, thus enhancing tumor's heterogeneity as well as its ability to overcome future limitations imposed by the rapidly exhausting resources. Competition, predation, parasitism and cooperation are indeed often observed in co-existing cancer clones [128].

In the well-known vision of Hananah and Weinberg [77, 78], the phenotypic stages that characterize this multistep evolutionary process are called *hallmarks*. These can be acquired by cancer cells in many possible alternative ways, as result of a complex biological interplay at several spatio-temporal scales that is still only partially deciphered [86]. In this framework, we distinguish alterations driving the hallmark acquisition process (i.e., *drivers*) by activating *oncogenes* or inactivating *tumor suppressor genes*, from those that are transferred to sub-clones without increasing their fitness (i.e., *passengers*) [58]. Driver identification is a modern challenge of cancer biology, as distinct cancer types exhibit very different combinations of drivers, some cancers display mutations in hundreds of genes [186], and the majority of drivers is mutated at low frequencies ("long tail" distribution), not allowing their detection by examining the recurrence at the population-level [60]. One can also use the evolutionary models to characterize, what may be called, *anti-hallmarks* – the phenotypes that are possible by the variational processes, but rarely found to be selected. For instance, certain collections of driver mutations, whose individual members are often present in the patient genomes, are never seen jointly. These anti-hallmarks point to tumors' vulnerabilities, and thus, novel targets for therapeutic interventions.

Cancer clones harbour distinct types of "alterations". The *somatic* ones involve either few nucleotides or larger chromosomal regions, and are usually catalogued as mutations - i.e., *Single Nucleotide Variants* (SNVs) and *Structural Variants* (SVs) at multiple scales (insertions, deletions, inversions, translocations) – of which only some are detectable as *Copy Number Alterations* (CNAs), which appear to be most prevalent in many tumor types [195]. Also *epigenetic alterations*, such as DNA methylation and chromatin reorganization, play a key role in the process [9]. The overall picture is confounded by factors such as *genetic instability* [190], *aneuploidy, tumor heterogeneity* and *tumor-microenvironment* interplay [3], the latter involving stromal and immune-system cells with strong influence on the final effect of mutations [68]. Furthermore, spatial organization and tissue specificity play an essential role on tumor progression as

well [142][2].

In this scenario, genomic alterations are related to the phenotypic properties of tumor cells via the structure and dynamics of *functional pathways*, in a process which has been only partially characterized [185, 191, 92, 147]. In general, in fact, as there exist many equivalent ways to disrupt signaling and regulatory pathways, many mutations can provide equivalent fitness to cancer cells, leading to *alternative routes* to selective advantage across a population of tumors [141]. Practically, if multiple genes are equally functional for the same biological process, when any of those is altered the selection pressure on the others is diminished or even nullified [8]. Such genes, e.g., APC/CTNNB1 in colorectal cancer [64], therefore show a trend of *exclusivity* across a cohort – with few cases of co-occurrent alterations. The same applies when disruptive alterations hit on the same gene, e.g., PTEN's mutations and deletions in prostate cancer [67].

An immediate consequence of this state of affairs is the dramatic *heterogeneity* and *temporality* of cancer, both at the *inter-tumor* and at the *intra-tumor* levels [55]. The former manifests as different patients with the same cancer type can display few common alterations. This led to the development of techniques to stratify tumors into *subtypes* with different genomic signatures, prognoses and response to therapy [30]. The latter refers to the noteworthy genotypic and phenotypic variability among the cancer cells within a single neoplastic lesion, characterized by the coexistence of more than one cancer clones with distinct evolutionary histories [63].

Cancer heterogeneity poses a serious problem from the diagnostic and therapeutical perspective as, for instance, it is now acknowledged that a single biopsy might not be representative of other parts of the tumor, hindering the problem of devising effective treatment strategies [128]. Therefore, the quest for an extensive etiology of cancer heterogeneity and for the identification of cancer evolutionary trajectories is nowadays central to cancer research, and attempt to exploit the huge amount of sequencing data available through public projects such as *The Cancer Genome Atlas* (TCGA) [136].

Such projects involve an increasing number of *cross-sectional* (epi)genomic profiles collected via single biopsies of patients affected by various cancer types, which might be used to extract trends of cancer evolution across a population of samples. Higher resolution data such as *multiple samples* collected from the same tumor [63], as well as *single-cell* sequencing data [134], might be complementarily used to face the same problem within a specific patient. However, either the lack of public data or problems of accuracy and reliability, due to technical and technological issues, currently prevent a straightforward application [42].

Nonetheless, there is a serious conceptual gap in the understanding of how such temporal heterogeneous cancer data could be analyzed, since available Machine Learning algorithms are not well-suited for the purposes, primarily because of their stationarity assumptions regarding the underlying statistical distributions. To solve this problem rigorously, we build our foundations on the sound theory of probabilistic causality, originally proposed by Suppes [172] (see Chapter §2), and devise a framework, which for the first time algorithmizes Suppes' formulation, while taming it's efficiency satisfactorily, even for many complex situations that are specifically important in cancer studies (e.g., synthetic lethality or oncogene addiction). While the contributions of this thesis are primarily methodological – strongly supported by empirical studies using synthetic as well as some experimental genomic data – it is hoped to attract other algorithmicists to this problem and catalyze new directions of explorations.

---

[2]We mention that much attention has been recently casted on newly discovered cancer genes affecting global processes that are apparently not directly related to cancer development, such as cell signaling, chromatin and epigenomic regulation, RNA splicing, protein homeostasis, metabolism and lineage maturation [60].

## 1.2   Two facets of inferring cancer progression models

The aforementioned different perspectives regarding cancer progression lead to the different mathematical formulations of the problem of *inferring a cancer progression model* from genomic data, which we examine at length in this thesis [13]. Indeed, such models can either be focused to describe trends characteristics of a population, i.e. *ensemble-level*, or clonal progression in a *single-patient*. In general, both problems deal with understanding the *temporal ordering of somatic alterations* accumulating during cancer evolution, but use orthogonal perspectives and different input data – see Figure §1.1.

**Ensemble-level cancer evolution.**   It may seem desirable to extract a *probabilistic graphical model* (PGM) explaining the statistical trend of accumulation of somatic alterations in a population of $n$ cross-sectional samples collected from patients affected by a specific cancer. To make this problem independent of the experimental conditions in which tumors are gathered, we only consider the *list of alterations detected per sample* – thus, as 0/1 Bernoulli variables.

Much of the difficulty lies in estimating the true and unknown trends of *selective advantage* among genomic alterations in the data, from such observations. This hurdle is not unsurmountable, if we constrain the scope to only those alterations that are *persistent across tumor evolution in all sub-clonal populations*, since it yields a consistent model of a temporal ordering of mutations. Therefore, epigenetic and trascriptomic states, such as hyper and hypo-methylations or over and under expression, could only be used, provided that they are persistent thorough tumor development [159].

Historically, the linear colorectal progression by Vogelstein is an instance of a solution to the cancer progression modeling problem [184]. That approach was later generalized to accommodate *tree-models of branched evolution* [38, 39, 173, 12] and, later, generalized to the inference of directed acyclic graph (DAG) models by Beerenwinkel and others [10, 65, 131]. We contributed to this research program with two related algorithms: *CAncer PRogression Extraction with Single Edges* (CAPRESE, see [117] and §3) and *CAncer PRogression Inference* (CAPRI, see [158] and §4). Both techniques rely on Suppes' theory of probabilistic causation to define estimators of selective advantage [172], are robust to the presence of noise in the data and perform well even with limited sample sizes. The former algorithm exploits shrinkage-like statistics to extract a tree model of progression, the latter combines bootstrap and maximum likelihood estimation with regularization to extract general directed acyclic graphs (DAGs) that capture branched, independent and confluent evolution. Both algorithms represent the current state-of-the-art to approach this problem, as they outperform others in speed, scale and predictive accuracy.

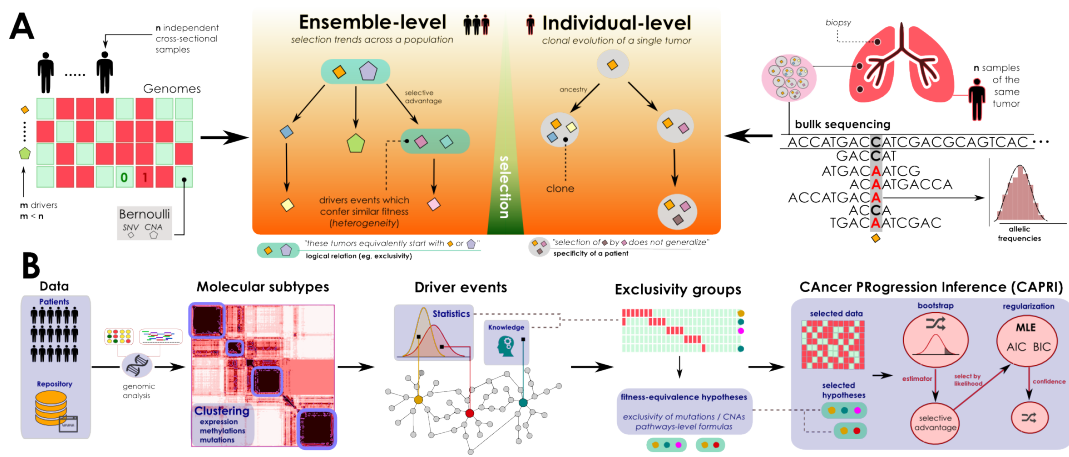

Figure 1.1: **(A) Problem statement.** (left) Inference of ensemble-level cancer progression models from a cohort of $n$ independent patients (cross-sectional data). Considering a list of somatic mutations or CNAs per patient (0/1 variables), a probabilistic graphical model for the temporal ordering of fixation and accumulation of such alterations is inferred in the input cohort. Sample size and tumor heterogeneity hardens the problem of extracting population-level trends, as this requires to account for patients' specificities such as multiple starting events. (right) For an individual tumor, its clonal phylogeny and prevalence is usually inferred from multiple biopsies or single-cell sequencing data. Phylogeny-tree reconstruction from an underlying statistical model of reads coverage or depths estimates alterations' prevalence in each clone, as well as ancestry relations. This problem is mostly hardened by the high intra-tumor heterogeneity and sequencing issues. **(B) A pipeline for ensemble-level inference.** The optimal pipeline includes several sequential steps to reduce tumor heterogeneity, before applying the CAPRI [158] algorithm. Available mutation, expression or methylation data are first used to stratify patients into distinct tumor molecular subtypes, usually by exploiting clustering tools. Then, subtype-specific alterations driving cancer initiation and progression should be identified with statistical tools and on the basis of prior knowledge. Next is the identification of the fitness-equivalent groups of mutually exclusive alterations across the input population, again done with computational tools or biological priors. Finally, CAPRI processes a set of relevant alterations and such groups. Via bootstrap and hypothesis-testing, CAPRI extracts a set of "selective advantage relations" among them, which is eventually narrowed down via maximizing likelihood fit with regularization (via BIC and AIC scores). The ensemble-level progression model is obtained by combining such relations in a graph, and its confidence is assessed via bootstrap.

**Clonal architecture in individual patients.**    At the time of this writing, technical and economical limitations of single-cell sequencing prevent a straightforward application of phylogeny inference algorithms to the reconstruction of the clonal evolutionary history of genomic alterations within a single tumor [135, 188].  Conversely, samples of cells collected from a single bulk tumor do not define an isogenic lineage [20] and most likely contain a large number of cells belonging to a collection of sub-clones resulting from the complex evolutionary history of the tumor, where the prevalence of a particular clone in time and its spatial distribution reflect its growth and proliferative fitness. To overcome hurdles such as this, many recent efforts have aimed at inferring the clonal signatures and prevalence in individual patients from sequencing data [63, 62].

The majority of attempts employ different strategies, usually based on Bayesian inference, to relate *allelic imbalance to cellular prevalence*, and benefits from multiple sample per patient, taken across time or space.  In particular, most tools usually process a set of read counts from a high-coverage sequencing experiment to estimate *Variant Allele Frequency* (VAF). Some of them are based on the VAF analysis of specific SNVs [130, 164].  Recent algorithms attempt to minimize the error between the observed and inferred mutation frequencies with distinct optimization procedures [91, 122, 48]. Other approaches support explicitly short-read data and different types of data, such as CNAs, SNVs and B-allele fractions (BAFs) [54].  Distinct techniques, instead, use genome-wide segmented read-depth information to determine mixtures of subclonal CNA profiles [144, 145], while others use a generative approach to deconvolve sequencing data to clonal architectures [196].  Clearly, any of these approaches gains precision from high-coverage sequencing data, since high read counts yield high confidence estimate of allele frequency.

As already discussed, several datasets are currently available that aggregate diverse cancer-patient data and report in-depth mutational profiles, including e.g., structural changes (e.g., inversions, translocations, copy-number variations) or somatic mutations (e.g., point mutations, insertions, deletions, etc.), see [136]. These data, by their very nature, only give a snapshot of a given tumor sample, mostly from biopsies of untreated tumor samples at the time of diagnoses.  As it still remains impractical to track the tumor progression in any single patient over time, thus limiting most analysis methods to work with *cross-sectional* data[3].

For this reason, in this thesis we focus on the problem of *reconstructing cancer progression models from cross-sectional data*.  As already stated, this problem is not new and, to the best of our knowledge, two threads of research starting in the late 90's have addressed it.  The first category of works examined mostly gene-expression data to reconstruct the temporal ordering of samples (see [121, 71]).  The second category of works aimed at inferring cancer progression models of increasing model-complexity, starting from the simplest tree models (see [38]) to more complex graph models (see [65]); see the next section for an overview of the state-of-the-art. Building on the works described in [117, 158], we present a novel and comprehensive framework of the second category that addresses this problem.

---

[3]Unlike longitudinal studies, these cross-sectional data are derived from samples that are collected at unknown time points, and can be considered as "static".

Moreover, the framework proposed in this thesis along with the described algorithms is part of the *TRanslational ONCOlogy* (TRONCO) package (see [4, 33, 5]). In summary (also see Chapter §2), starting from cross-sectional genomic data, such algorithms aim at reconstructing a probabilistic progression model by inferring "selectivity relations", where a mutation in a gene *A* "selects" for a later mutation in a gene *B*. These relations are depicted in a combinatorial graph and resemble the way a mutation exploits its "*selective advantage*" to allow its host cells to expand clonally. Among other things, a selectivity relation implies a putatively invariant temporal structure among the genomic alterations (i.e., *events*) in a specific cancer type[4]. In addition, these relations are expected to also imply "probability raising" for a pair of events in the following sense: namely, a selectivity relation between a pair of events here signifies that the presence of the earlier genomic alteration (i.e., the *upstream event*) that is advantageous in a Darwinian competition scenario increases the probability with which a subsequent advantageous genomic alteration (i.e., the *downstream event*) appears in the clonal evolution of the tumor. Thus the selectivity relation captures the effects of the evolutionary processes, and not just correlations among the events and imputed clocks associated with them. As an example, we show in Figure §1.2 the selectivity relation connecting a mutation of EGFR to the mutation of CDK, see §2.

Consequently, an inferred selectivity relation suggests mutational profiles in which certain samples (early-stage patients) display specific alterations only (e.g., the alteration characterizing the beginning of the progression), while certain other samples (e.g., late-stage patients) display a superset subsuming the early mutations (as well as alterations that occur subsequently in the progression).

Various kinds of genomic aberrations are suitable as input data, and include somatic point/indel mutations, copy-number alterations, etc., provided that they are *persistent*, i.e., once an alteration is acquired no other genomic event can restore the cell to the non-mutated (i.e., *wild type*) condition[5].

In what follows and in the rest of the thesis, we use the notations described below. *Atomic events*, in general, are denoted by small Roman letters, such as $a$, $b$, $c$, . . .; when it is clear from the context that the event in the model is, in fact, a genomic mutational event, we may refer to it directly using the standard biological nomenclature, e.g., BRCA1, BRCA2, etc. – it would be especially true, in the sections describing applications to real data. Patterns over events are mostly denoted by Greek letters, and their logical connectives with the usual "and" ($\wedge$), "or" ($\vee$) and "negation" ($\dot{\neg}$) symbols. Standard operations on sets are used as well.

We are not employing distinct notations to denote *observed probabilities* and prob-abilities in the model which we aim at inferring (i.e., the *"theoretical probabilities"*). Which quantity is being referred to, is made clear from the context. In the following, $\mathcal{P}(x)$ denotes the *probability* of $x$; $\mathcal{P}(x \wedge y)$, the *joint probability* of $x$ *and* $y$, which is naturally extended to the notation $\mathcal{P}(x \wedge y_1 \wedge \ldots \wedge y_n)$ for an arbitrary arity; and $\mathcal{P}(x \mid y)$, the *conditional probability* of $x$ *given* $y$. Here $x$ and $y$ are patterns over events.

---

[4]It has already been mentioned the known existence of various molecular subtypes within the same cancer.

[5]For instance, epigenetic alterations such as methylation and alterations in gene expression are not directly usable as input data for the algorithm. Notice that the selection of the relevant events is beyond the scope of this work and requires a further upstream pipeline, such as that provided, for instance, in [177, 186].

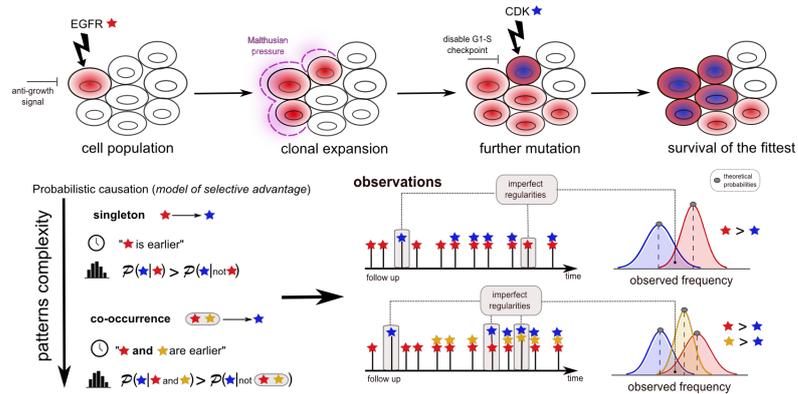

Figure 1.2: **Selectivity relation in tumor evolution.** In our framework, cancer patients' genomic cross-sectional data are examined to determine relationships among genomic alterations (e.g., somatic mutations, copy-number variations, etc.) that modulate the somatic evolution of a tumor. When it is concluded that aberration $a$ (say, an EGFR mutation) "selects for" aberration $b$ (say, a CDK mutation), such relations can be rigorously expressed using Suppes' conditions, which postulates that if $a$ selects $b$, then $a$ occurs before $b$ (*temporal priority*) and occurrences of $a$ raises the probability of emergence of $b$ (*probability raising*).

As with the discussion of selective advantage structures, we write $c \rhd e$, where $c$ and $e$ are events being modeled, in order to denote the selective advantage relation "$c$ has a selective influence on $e$". As we extend our presentation to general *patterns*, we generalize the notation to $\varphi \rhd e$ with the meaning generalized *mutatis mutandis*[6].

Before moving on, we now provide in the next Section an overview of the state-of-the-art on the problem of cancer progression models as for what pictured in this Chapter.

## 1.3   State of the art

For an extensive review on *cancer progression model reconstruction* we refer to the recent survey by [13]. In brief, progression models for cancer have been studied starting with the seminal work of [184] where, for the first time, cancer progression was described in terms of what could be interpreted as a *directed path* a directed path. [184] manually created a (colorectal) cancer progression from a genetic and clinical point of view. More rigorous and complex algorithmic and statistical automated approaches have appeared subsequently. As stated already, the earliest thread of research simply sought more generic progression models that could assume tree-like structures. The *oncogenetic tree model* captured evolutionary branches of mutations (see [38, 173]) by optimizing a *correlation*-based score. Another popular approach to reconstruct tree structures appears in [39]. Other general Markov chain models such as, e.g., [81] reconstruct more flexible probabilistic networks, despite a computationally expensive parameter estimation. Other results that extend tree representations of cancer evolution exploit mixture tree models, i.e., multiple oncogenetic trees, each of which can independently result in cancer development (see [12]). In general, all these methods are capable of modeling diverging temporal orderings of events in terms of branches, although the possibility of converging evolutionary paths is precluded.

---

[6]Note that the scope of this thesis is intentionally kept limited from further generalizing the "selective advantage patterns"; for instance, we are not dealing with any example of the form $\varphi_i \rhd \varphi_j$, where $\varphi$ could be any general pattern (including a complex causal pattern or a temporal pattern). This choice is justified in view of complexity, practicality, applicability and expressiveness in the context of cancer progression driven by somatic evolution.

To overcome this limitation, the most recent approaches tends to adopt Bayesian graphical models, i.e., Bayesian Networks (BN). In the literature, there have been two initial families of methods aimed at inferring the structure of a BN from data (see [101]). The first class of models seeks to explicitly capture all the conditional independence relations encoded in the edges and will be referred to as *structural approaches*; the methods in this family are inspired by the work on causal theories by Judea Pearl (see [150, 151, 171, 179]). The second class – *likelihood approaches* – seeks a model that maximizes the likelihood of the data (see [167, 79, 23]).

A more recent *hybrid approach* to learn a BN which combines the two families above by (*i*) constraining the search space of the valid solutions and, then, (*ii*) fitting the model with likelihood maximization (see [10, 65, 131]). A further technique to reconstruct progression models from cross-sectional data was introduced in [6], in which the transition probabilities between genotypes are inferred by defining a Moran process that describes the evolutionary dynamics of mutation accumulation. In [25] this methodology was extended to account for pathway-based phenotypic alterations.

This thesis is structured as follow. Chapter §2 together with Appendices §A and §B describe the theoretical foundations on which the presented framework is based. In Chapters §3 and §4 two efficient algorithms to respectively reconstruct tree-alike and directed acyclic graph models of cancer progression are described. Chapter §5 presents an R package which implements the described algorithms together with a series of functionalities to support the researcher through all the steps of the analysis of cancer progression. This package is then adopted in Chapter §6 where a detailed analysis based on a structured pipeline is performed for colorectal cancer. All the supplementary materials concerning to the previous Chapters are reported in the remaining Appendices. Finally, Chapter §7 concludes the thesis.

CHAPTER 2 ______________________________

|
|______________ MODELING CANCER CLONAL EVOLUTION

Based on the discussion of §1, in this Chapter we will formalize the adopted framework used to model cancer clonal evolution. In particular, we will define the proposed model of selective advantage through different levels of complexity.

## 2.1   A probabilistic model of selective advantage

Central to the model proposed in this thesis is Suppes' notion of *probabilistic causation* [172], which can be stated in the following terms: a selectivity relation[1] among two observables $i$ and $j$ is verified if (1) $i$ occurs earlier than $j$ – *temporal priority* (TP) – and (2) if the probability of observing $i$ raises the probability of observing $j$, i.e., $\mathcal{P}(j \mid i) > \mathcal{P}(j \mid \bar{i})$[2] – *probability raising* (PR), see §A.2 for a deeper discussion of the philosophical aspects of this theory.

Note that the definition of probability raising subsumes positive statistical dependency and mutuality (see, e.g., [117, 158]). But, it should be emphasized that the resulting relation (also termed prima facie causality) is purely observational and remains agnostic of any possible mechanistic cause-effect relation involving $i$ and $j$. When through this thesis we term any relation to be causal, we will use this word with such an interpretation, i.e., ignoring any mechanistic behaviour.

While Suppes' definition of probabilistic causation has known limitations in the context of general causality theory (see discussions in §A.2 and, e.g., [80, 98]), in the context of cancer evolution, this relation appropriately describes various features of *selective advantage* in somatic alterations that accumulate as tumor progresses.

Therefore, in this framework, we implement the temporal priority among events – condition (1) – as $\mathcal{P}(i) > \mathcal{P}(j)$, because it is intuitively sound to assume that the (cumulative) genomic events occurring earlier are the ones present in higher frequency in a dataset. In addition, condition (2) is implemented as is, that is by requiring that for each pair of observables $i$ and $j$ it holds that $\mathcal{P}(j \mid i) > \mathcal{P}(j \mid \bar{i})$. Taken together, these conditions give rise to a natural ordering relation among events, written "$i \rhd j$" and read as "$i$ has a selective influence on $j$". This relation is a *necessary* but *not sufficient* condition to capture the notion of selective advantage, hence additional constraints need to be imposed to filter any spurious correlation (see the discussions in §A.2). Spurious correlations are both intrinsic to the definition (e.g., if $i \rhd j \rhd w$ then also $i \rhd w$, which could be spurious) and to the model we aim at inferring, because of finite data as well as presence of observational noise.

---

[1]Suppes describes such a relation in terms of causality; however, here we avoid this terminology as we build on just two of his many axioms, which give rise to the notion of *prima-facie* causality.

[2]Please remind that we are considering cross-sectional data, hence without any explicit measurement of time which needs to be imputed.

Building on this framework, we aim at devising inference algorithms that capture the essential aspects of heterogeneous cancer progressions: *branching*, *independence* and *convergence* – all combining in a progression model.

Furthermore, the complexity of cancer requires modeling multiple non-trivial *patterns* of its progression: for a specific event, a pattern is defined as a specific combination of the closest upstream events that confers a selective advantage.

As an example, imagine a clonal subpopulation becoming fit – thus enjoying expansion and selection – once it acquires a further mutation of gene $c$, provided that it also has previously acquired a mutation in a gene in the upstream $a/b$ pathway. In terms of progression, we would like to capture these evolutionary trajectories: either $\{a, \neg b\}, \{\neg a, b\}$ or $\{a, b\}$ precedes $c$ (where $\neg$ denotes the absence of an event in the gene).

To formally take this into account, we augment our model of selection in a tumor with a language built from simple propositional logic formulas using the usual Boolean connectives: namely, "and" ($\wedge$), "or" ($\vee$) and "xor" ($\oplus$). These patterns can be described by formulæ in a propositional logical language, which can be rendered in *Conjunctive Normal Form* (CNF). A CNF formula $\varphi$ has the following syntax: $\varphi = \boldsymbol{c}_1 \wedge \ldots \wedge \boldsymbol{c}_n$, where each $\boldsymbol{c}_i$ is a *disjunctive clause* $\boldsymbol{c}_i = c_{i,1} \vee \ldots \vee c_{i,k}$ over a set of literals, each literal representing an event or its negation.

In this framework, we aim at reconstructing probabilistic graphical models of cancer progression. Given the above premises, this problem reduces to the following: for each input event $e$, assess a *set of selectivity patterns* $\{\varphi_1 \rhd e, \ldots, \varphi_k \rhd e\}$, filter the spurious ones, and combine the rest in a *direct acyclic graph* (DAG)[3], $G = (V, E)$, where the nodes are the atomic events (augmented, eventually, with logical symbols) and the edges represent selective advantage relations. Notice that, while we broke down the progression extraction into a series of sub-tasks, the problem still remains complex: patterns are unknown, potentially spurious, and exponential in formula size; moreover, data are noisy and patterns must allow for "imperfect regularities", rather than being strict[4].

To summarize, in this setting we can model complex progression trajectories with branches (i.e., events involved in various patterns), independent progressions (i.e., events without common ancestors) and convergence (via CNF formulas).

---

[3]A DAG is formed by a set of nodes and oriented edges connecting one node to another, such that there are no directed loops among them.

[4]This statement implies that there could be samples – i.e., patients or tumor cells – contradicting a pattern which still remains valid at a population level. For this reason a pattern $x \wedge y \rhd z$ is sometimes called a "noisy and".

For the sake of clarity, in the next sections we develop the presentation in steps of successively increasing complexity of the selective advantage patterns: e.g., going from singleton (i.e., "atomic") patterns, to co-occurrance patterns consisting of atomic events, up to patterns in *Conjunctive Normal Forms* (CNF) (e.g., [('burning cigarette' ∧ 'dried wood' ) ∨ ('lightning' ∧ 'no rain') ▷ 'forest fire'])[5].

## 2.2   Singleton prima facie topologies

When at most a single incoming edge is assigned to each event (i.e., an event has at most one *unique parent*: $\forall_{e \in V} \exists!_{c \in V} c \triangleright e$), we term this causal structure *singleton prima facie topology*, a special and important case of the most general prima facie topology structures. Note once again that the general model can be represented as a direct acyclic graph (DAG) where each edge represents a prima facie relation between a parent and its child. In the special case of the singleton prima facie topology, such a graph is a *tree* or, more generally, a *forest* when there are disconnected components. Thus, each progression tree induces a distribution of observing a subset of the mutations in a cancer sample (see [117] for a detailed discussion).

In [117] the following propositions (summarized in Figure §2.1 and discussed in details in §3) were shown to hold for singleton prima facie topologies (see §3 and §C for a detailed description and the related proofs), and used to derive an algorithm to infer tree (forest) models of cancer progression.

**Statistical dependence.**   Whenever probability raising holds between two events $c$ and $e$, then the events are *statistically dependent* in a positive sense, i.e.,

$$\mathcal{P}(e \mid c) > \mathcal{P}(e \mid \bar{c}) \iff \mathcal{P}(e \land c) > \mathcal{P}(e)\mathcal{P}(c) \,. \tag{2.1}$$

**Mutuality.**   If $c$ is a probability raiser for $e$, then so is the converse, i.e.,

$$\mathcal{P}(e \mid c) > \mathcal{P}(e \mid \bar{c}) \iff \mathcal{P}(c \mid e) > \mathcal{P}(c \mid \bar{e}) \,. \tag{2.2}$$

**Natural ordering.**   For any two events $c$ and $e$ such that $c$ is a probability raiser for $e$, a "natural" ordering arises to disentangle a causality relation, i.e.,

$$\mathcal{P}(c) > \mathcal{P}(e) \iff \frac{\mathcal{P}(e \mid c)}{\mathcal{P}(e \mid \bar{c})} > \frac{\mathcal{P}(c \mid e)}{\mathcal{P}(c \mid \bar{e})} \,. \tag{2.3}$$

---

[5]The statement above, which is expressed for conveniency in Disjunctive Normal Form and could be automatically translated in CNF, may also be shortened as 'burning cigarette' ▷ 'forest fire.' The intended interpretation is that, 'burning cigarette' is an *insufficient but non-redundant part of an unnecessary but sufficient causal condition* (INUS) for 'forest fire,' as originally suggested by the philosopher J. Mackie.

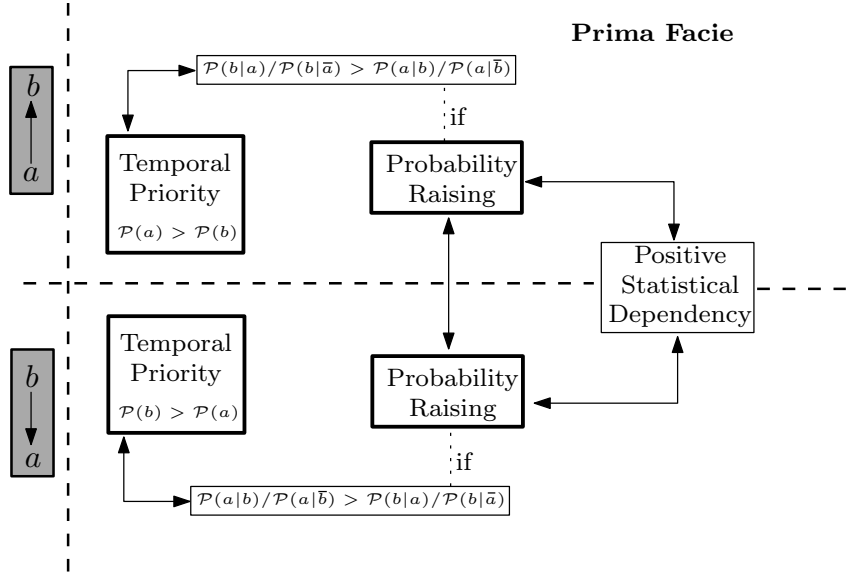

Figure 2.1: **Prima facie properties.** Properties of Suppes' definition of probabilistic causation: $c$ is a *prima facie cause* of $e$ *if it is a probability raiser of $e$, and it occurs more frequently.*

Putting together all these properties, it is straightforward to derive the following characterization of singleton prima facie relations: $c$ is said to be a *singleton prima facie cause* of $e$ *if $c$ is a probability raiser of $e$, and it occurs more frequently,* i.e.,

$$c \triangleright e \iff \mathcal{P}(e \mid c) > \mathcal{P}(e \mid \bar{c}) \quad \wedge \quad \mathcal{P}(c) > \mathcal{P}(e). \tag{2.4}$$

Consequently to this definition, we observe that (see also the discussions in §A.2) it is *necessary but not sufficient* to identify the accumulative processes (path or branch) and, thus, to solve the considered problem. In fact, as it can be easily observed in the Figure §2.2, black arrows make this definition necessary, while red arrows (*spurious*, resulting, e.g., from *transitivities*, because of the singleton hypothesis) make the condition insufficient. We remark that red arrows will *always* be present to indicate potential *genuine* causes corresponding to actual selective advantage relations. Thus, a correct inferential algorithm will have to select the real relations among the potential genuine ones, within a subset of the prima facie causes.

A further discussion about spurious connections is now warranted. As deeply described in §A.2, spurious causes may manifest through *spurious correlation* or *chance*. In the infinite sample size limit the "law of large numbers" eliminates the effect of chance; in other words, with large enough sample, chance by itself will not suffice to satisfy Suppes' conditions. Instead, the former situation for spuriousness depends on the topology to be reconstructed, and might appear under observation like a prima-facie/genuine selective advantage relation in disguise, even with an infinite sample size (purple edges, for which the "temporal direction" has no causal interpretation, as it depends on the data and topology). For these reasons, a singleton prima facie topology asymptotically will not contain *false negatives* (i.e., all the actual selective advantage relations are modeled in the topology) but it might contain *false positives* (red or purple edges, as Suppes' prima facie conditions are not sufficient).

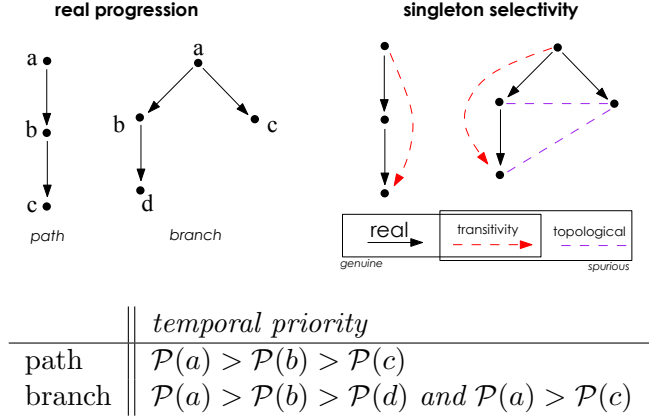

| | temporal priority |
|---|---|
| path | $\mathcal{P}(a) > \mathcal{P}(b) > \mathcal{P}(c)$ |
| branch | $\mathcal{P}(a) > \mathcal{P}(b) > \mathcal{P}(d)$ and $\mathcal{P}(a) > \mathcal{P}(c)$ |

Figure 2.2: **Singleton prima facie topology.** Example of a linear path and branching model (left) and corresponding singleton selectivity patterns with infinite sample size (right). All the genuine connections are shown (red and black, directed by the temporal priority), as well as edges (purple, undirected) which might be suggested by the topology (or observations, if data were finite).

## 2.3   Co-occurence prima facie patterns

We now denote by a Boolean conjunctive clause, a propositional formula composed of conjunctions of a set of literals: $\boldsymbol{c} = c_1 \wedge \cdots \wedge c_n$, which implies that $n$ events $c_1, \ldots, c_n$ have occurred (in some unspecified order) so as to collectively lead to effect $e$ (graphically pictured as in Figure §2.3), and we assume that each $c_i$ $(1 \leq i \leq n)$ is an atomic event.

Suppes' notion of probabilistic causation can be naturally extended to such co-occurence patterns as in the following definition:

**Definition 1** (Co-occurence probabilistic causation). *For any conjunctive event* $\boldsymbol{c} = c_1 \wedge \ldots \wedge c_n$ *and* $e$, *occurring respectively at times* $\{t_{c_i} \mid i = 1, \ldots, n\}$ *and* $t_e$, *under the mild assumptions that* $0 < \mathcal{P}(c_i), \mathcal{P}(e) < 1$, *for any* $i$, *the conjunctive event* $\mathbf{c}$ *is a* prima facie *conjunctive cause of* $e$ $(\mathbf{c} \triangleright e)$ *if all of its components* $c_i$ *occur* before *the effect and their occurrences collectively* raises the probability *of the effect, i.e.,*

$$\max\{t_{c_1}, \ldots, t_{c_n}\} < t_e \quad \text{and} \quad \mathcal{P}(e \mid \boldsymbol{c}) > \mathcal{P}(e \mid \overline{\boldsymbol{c}}) \,. \tag{2.5}$$

$$\text{where} \quad \begin{cases} \mathcal{P}(e \mid \boldsymbol{c}) = \mathcal{P}(e \mid c_1 \wedge \cdots \wedge c_n) \\ \mathcal{P}(e \mid \overline{\boldsymbol{c}}) = \mathcal{P}(e \mid \overline{c_1 \wedge \cdots \wedge c_n}) = \mathcal{P}(e \mid \overline{c}_1 \vee \cdots \vee \overline{c}_n) \,. \end{cases}$$

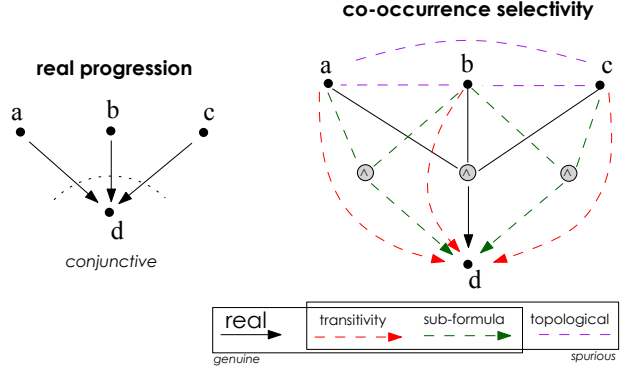

| | temporal priority |
|---|---|
| co-occurrence | $\mathcal{P}(a) > \mathcal{P}(d)$ and $\mathcal{P}(b) > \mathcal{P}(d)$ and $\mathcal{P}(c) > \mathcal{P}(d)$ |

Figure 2.3: **Co-occurrence Selectivity Patterns.** Example of conjunctive model ($a$ *and* $b$ *and* $c$). The co-occurrence selectivity pattern is shown, with all true patterns and infinite sample size. The topology is augmented by logical connectives; green arrows are spurious patterns emerging from the structure of the true pattern $a \wedge b \wedge c \triangleright d$.

This extension simply follows the semantics of co-occurence patterns, which states that *all causes* must occur *before* their effect, thus justifying the choice of picking the latest event, in time, prior to $e$ as a generalization: namely, the $\max\{\cdot\}$ operation applied to the preceeding events. Clearly, this definition retains the semantics of singleton prima facie relations unchanged, as it is a special case with $\boldsymbol{c} = c$ and $\max\{t_{c_i}\} = t_c$. Unfortunately, as before, it still has the same weakness and it is *necessary but not sufficient* to identify co-occurrence patterns.

The properties of singleton prima facie topologies also extend appropriately to co-occurrence topologies – see §C and §D for the related proofs, along with all the other properties and theorems that appear in this thesis.

**Proposition 1.** *The properties of* statistical dependence, mutuality *and* natural ordering *for singleton patterns are still valid for co-occurence patterns.*

In this case some caution must be exercised in distinguishing between prima facie singleton or co-occurance patterns. As shown in Figure §2.3, in fact, for any co-occurance pattern in the real world ($a$ *and* $b$ *and* $c$) the following conjunctive clauses

$$a \wedge b \triangleright d \qquad\qquad a \wedge c \triangleright d \qquad\qquad b \wedge c \triangleright d$$

as well as the singleton patterns $a \triangleright d$, $b \triangleright d$ and $c \triangleright d$, are prima facie. The singleton patterns can be *spurious* or *transitive*, as in Figure §2.2. But this time, we will also have *spurious sub-formulas*, i.e., the conjunctive clauses that are *syntactically strictly sub-formulas* of $a \wedge b \wedge c \triangleright d$, that is the only genuine formula we would like to infer. Notice that as in branch processes, topology-dependent spurious causes might also appear because of spurious correlations; in the Figure §2.2, we have not shown other potential spurious relations, as what we depict to make it visualizable is just a one-level conjunctive network. These selective advantage relations could include general *spurious formulas* constituting of a sub-formula and any of its parents. Similarly, spurious relations due to chance will vanish asymptotically as sample size grows to infinity. Summarizing, we note that a co-occurence topology, just as in the singleton patterns framework, will not contain false negatives (i.e., all real world selective advantage relations will be modeled in the topology) but it might also contain, depending on the real world topology, false positives (red, green or purple edges).

Before concluding, we note that the total number of potential formulas and transitivities is *exponential* in the size of $|G| = n$, that is

$$\sum_{i=1}^{n-1} \binom{n-1}{i} = 2^{n-1} - 1 \, .$$

Notice that this is a lower bound accounting only for the level of the connective, and is expected to grow further when more complex real world accumulative processes are considered. Finally, as shown in Figure §2.2, the number of spurious relations due to the topology (purple edges), are quadratic in the formula size, being

$$2\binom{n-1}{2} = (n-1)(n-2) \, .$$

This complexity hints at the fact that an exhaustive search of all the possible conjunctive formula is not feasible, in general.

## 2.4  Generalization to conjunctive normal form

We consider next a formula in *conjunctive normal form* (CNF)

$$\varphi = \boldsymbol{c}_1 \wedge \ldots \wedge \boldsymbol{c}_n,$$

where each $\boldsymbol{c}_i$ is a *disjunctive clause* $\boldsymbol{c}_i = c_{i,1} \vee \ldots \vee c_{i,k}$ over a set of literals and each literal represents an event (a Boolean variable) or its negation. By following analogous arguments as the ones used before, we can define $\varphi \rhd e$ as follows.

**Definition 2** (CNF probabilistic causation). *For any CNF formula $\varphi$ and $e$, occurring respectively at times $t_\varphi$ and $t_e$, under the mild assumptions that $0 < \mathcal{P}(\varphi), \mathcal{P}(e) < 1$, $\varphi$ is a* prima facie *cause of $e$ if*

$$t_\varphi < t_e \quad and \quad \mathcal{P}(e \mid \varphi) > \mathcal{P}(e \mid \overline{\varphi}) \, . \tag{2.6}$$

As stated before, also this definition is *necessary but not sufficient* to identify selective advantage relations, hence lacking the power to solve the considered problem.

Clearly, in this case, the number of prima facie (including both genuine and spurious) relations grows combinatorially much more rapidly than the simplest case of a unique conjunctive clause (see §2.3); this situation is rather alarming, since even the simplest case already produces an exponentially large set of prima facies causes in terms of the number of events. In this case, in fact, further causal relations emerge as a result of mixing events from all the clauses of $\varphi$. CNF formulas follow analogous properties as singleton and co-occurrence topologies, as shown below.

**Proposition 2.** *The properties of* statistical dependence*, mutuality and* natural ordering *for singleton and co-occurence prima facie topologies also extend to CNF formulas* mutatis mutandis.

We conclude this Section with two final comments about CNF formulas: their relation with background contexts (see §A.2), and the notion of timing in Definition §2.

The first comment concerns Cartwright's idea of background contexts as a conjunction of independent factors §A.2. For illustrative purposes, consider the formula $(a \wedge b) \vee c \rhd d$, which is in *disjunctive normal form* (DNF). If, for example, we were to evaluate the claim $a \rhd d$, the (unique) background context would be the atomic event $c$, being relevant to evaluate the claim $a$ causes $d$. A symmetric situation holds, when we were to evaluate $b \rhd d$. In light of this discussion note that, if we convert the formula to its CNF analogue $(a \vee c) \wedge (b \vee c) \rhd d$, we need to correctly interpret the roles of sub-formulas $a \vee c$ and $b \vee c$ in identifying a background context, $c$. It follows immediately that, for any CNF formula, the atomic events of all the disjunctive clauses in the equivalent DNF formula provide all the possible background contexts à-la-Cartwright.

The second comment concerns timing in the real world. Consider the CNF formula above, denote it as $\varphi$ and recall that Definition §2 requires $t_\varphi < t_d$. One might wonder whether a trivial time-ordering relation exists, whose complexity is linear with respect to all the operators in $\varphi$. Were it so, we would be able to parse $\varphi$ into its constituents, and recursively express the temporal relations as a direct function of those relations that hold for its sub-formulas. Unfortunately, this appears not to be the case, except when the underlying syntax is restricted to certain specific operators (e.g., conjunctions). Thus appropriate care must be taken in implementing a model of time. Thus, an algorithm, working on the illustrative example of the previous paragraph, cannot conclude any ordering about $t_{a \vee c}$, $t_{b \vee c}$ and $t_d$, solely by looking at the marginal probabilities of their atomic events – instead it must gather the correct information for certain sub-formulas at the level of their connective (the $\vee$ in this case). A general rule that avoids these difficulties and devises a correct and efficient timing-inference algorithms, may be stated as follows: it is *always safe* to model probabilistic causation in terms of whole formulas, while permitting *compositional* reasoning over sub-formulas is operable only when the syntax is restricted to certain Boolean connectives. Further related comments appear in §3 and §4, where we describe the algorithmic implementations of the described framework.

Next we will describe efficient algorithmic implementations of the framework presented in this Chapter.

SINGLETON MODELS OF CANCER PROGRESSION

In this Chapter we will present an algorithm for the efficient inference of singleton models of cancer progression. As a reference, see [117].

## 3.1 Problem setting

Assuming that we have a set $G$ of $n$ mutations (*events*, in probabilistic terminology) and $s$ samples, we represent a cross-sectional dataset as an $s \times n$ binary matrix in which an entry $(k,l) = 1$ if the mutation $l$ was observed in sample $k$, and 0 otherwise. The problem we solve here is the one of extracting a set of edges $E$ yielding a progression tree $\mathcal{T} = (G \cup \{\diamond\}, E, \diamond)$ from this matrix which, we remark, only implicitly provides information of progression timing. The root of $\mathcal{T}$ is modeled using a (special) event $\diamond \notin G$ such that *heterogenous progression paths* or *forests* can be reconstructed. More precisely, we aim at reconstructing a *rooted tree* that satisfies: (*i*) each node has at most one incoming edge, (*ii*) the root has no incoming edges (*iii*) there are no *cycles*.

Each progression tree induces a distribution of observing a subset of the mutations in a cancer sample that can be formalized as follows:

**Definition 3** (Tree-induced distribution)**.** *Let $\mathcal{T}$ be a tree and $\alpha : E \to [0,1]$ a labeling function denoting the independent probability of each edge, $\mathcal{T}$ generates a distribution where the probability of observing a sample with the set of alterations $G^* \subseteq G$ is*

$$\mathcal{P}(G^*) = \prod_{e \in E'} \alpha(e) \cdot \prod_{\substack{(u,v) \in E \\ u \in G^*, v \notin G}} \Big[ 1 - \alpha(u,v) \Big] \tag{3.1}$$

*where all events in $G^*$ are assumed to be reachable from the root $\diamond$, and $E' \subseteq E$ is the set of edges connecting the root to the events in $G^*$.*

We would like to emphasize two properties related to the tree-induced distribution. First, the distribution subsumes that, given any oriented edge $(a \to b)$, an observed sample contains alteration $b$ with probability $\mathcal{P}(a)\mathcal{P}(b)$, that is the probability of observing $b$ after $a$. For this reason, if $a$ causes $b$, the probability of observing $a$ will be greater than the probability of observing $b$ accordingly to the temporal priority principle which states that all causes must precede, in time, their effects [160].

Second, the input dataset is a set of samples generated, ideally, from an unknown distribution induced by an unknown tree or forest that we aim at reconstructing. However, in some cases, it could be that no tree exists whose induced distribution generates *exactly* those input data. When this happens, the set of observed samples slightly diverges from any tree-induced distribution. To model these situations a notion of *noise* can be introduced, which depends on the context in which data are gathered. Adding noise to the model complicates the reconstruction problem.

**The *oncotree* approach.** In [38] Desper *et al.* developed a method to extract progression trees, named *"oncotrees"*, from static CNV data. In [173], Szabo *et al.* extended the setting of Desper's reconstruction problem to account for both *false positives* and *negatives* in the input data. In the oncotrees, nodes represent alterations and edges correspond to possible progressions from one event to the next.

The reconstruction problem is exactly as described above, and each tree is rooted in the special event $\diamond$. The choice of which edge to include in a tree is based on the estimator

$$w_{a \to b} = \log \left[ \frac{\mathcal{P}(a)}{\mathcal{P}(a) + \mathcal{P}(b)} \cdot \frac{\mathcal{P}(a \wedge b)}{\mathcal{P}(a)\mathcal{P}(b)} \right], \tag{3.2}$$

which assigns to each edge $a \to b$ a weight accounting for both the relative and joint frequencies of the events – thus measuring *correlation*. The estimator is evaluated after including $\diamond$ to each sample of the dataset. In this definition the rightmost term is the (symmetric) *likelihood ratio* for $a$ and $b$ occurring together, while the leftmost is the asymmetric *temporal priority* measured by rate of occurrence. This implicit form of timing assumes that, if $a$ occurs *more often* than $b$, then it likely occurs *earlier*, thus satisfying

$$\frac{\mathcal{P}(a)}{\mathcal{P}(a) + \mathcal{P}(b)} > \frac{\mathcal{P}(b)}{\mathcal{P}(a) + \mathcal{P}(b)}.$$

An oncotree is the rooted tree whose total weight (i.e., sum of all the weights of the edges) is maximized, and can be reconstructed in $O(|G|^2)$ steps using Edmond's algorithm [43]. By construction, the resulting graph is a proper tree rooted in $\diamond$: each event occurs only once, *confluences* are absent, i.e., any event is caused by at most one other event. This method has been used to derive progressions for various cancer datasets e.g., [93, 85, 157]), and even though several methods that extend this framework exists (e.g., [39, 12, 65]), to the best of our knowledge, it is currently the only method that aims to solve exactly the same problem as the one investigated here and thus provide a benchmark to compare against.

## 3.2    A probabilistic approach to selective advantage

We briefly recall the approach to probabilistic causation, on which this method is based. For an extensive discussion on this topic we refer to our previous discussion in §2.

In his seminal work [172], Suppes proposed a set of conditions that are necessary for any causal claim, that is, for any pair of two events $c$ and $e$, occurring respectively at times $t_c$ and $t_e$, under the mild assumptions that $0 < \mathcal{P}(c), \mathcal{P}(e) < 1$, the event $c$ is a *prima facie cause* of the event $e$ if it occurs *before* the effect and the cause *raises the probability* of the effect, i.e.,

$$t_c < t_e \quad \text{and} \quad \mathcal{P}(e \mid c) > \mathcal{P}(e \mid \bar{c}) \,.$$

As already discussed in §2, the above conditions are not, in general, sufficient to claim that event $c$ is a cause of event $e$. In fact a prima facie cause is either *genuine* or *spurious*. In the latter case, the fact that the conditions hold in the observations is due either to coincidence or to the presence of a certain third *confounding factor*, related both to $c$ and to $e$ [172]. Genuine causes, instead, satisfy Suppe's criteria and are not screened-off by any confounding factor (also see §C). However, they need not be direct causes. See Figure §3.1.

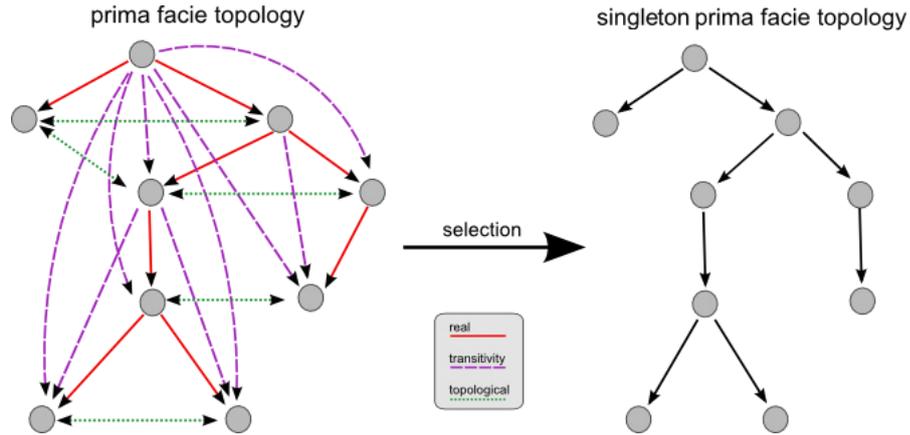

Figure 3.1: **Prima facie topology.** Example of prima facie topology where all edges $(a, b)$ represent prima facie causes, i.e., $a$ is a probability raiser of $b$ and it occurs more frequently. On the right, we filter out spurious selective advantage relations and select only the real ones among the genuine, yielding a singleton prima facie topology.

Note that we consider cross-sectional data where no information about $t_c$ and $t_e$ is available, hence in this reconstruction setting we are restricted to consider solely the *probability raising* (PR) property, i.e., $\mathcal{P}(e \mid c) > \mathcal{P}(e \mid \bar{c})$, which makes it harder to discriminate among genuine and spurious causes. Now we review some of its properties.

**Proposition 3** (Statistical dependence)**.** *Whenever the* PR *holds between two events $a$ and $b$, then the events are* statistically dependent *in a positive sense, i.e.,*

$$\mathcal{P}(b \mid a) > \mathcal{P}(b \mid \overline{a}) \iff \mathcal{P}(a \wedge b) > \mathcal{P}(a)\mathcal{P}(b) \,. \tag{3.3}$$

This and the next proposition are well-known facts of the PR; their derivation as well as the proofs of all the results we discuss here, are shown in §C.

Notice that the opposite implication holds as well: when the events $a$ and $b$ are still dependent but in a negative sense, i.e., $\mathcal{P}(a \wedge b) < \mathcal{P}(a)\mathcal{P}(b)$, the PR does not hold, i.e., $\mathcal{P}(b \mid a) < \mathcal{P}(b \mid \overline{a})$.

We would like to use the asymmetry of the PR to determine whether a pair of events $a$ and $b$ satisfy a selective advantage relation so to place $a$ before $b$ in the progression tree but, unfortunately, the PR satisfies the following property.

**Proposition 4** (Mutuality)**.** $\mathcal{P}(b \mid a) > \mathcal{P}(b \mid \overline{a}) \iff \mathcal{P}(a \mid b) > \mathcal{P}(a \mid \overline{b}) \,.$

That is, if $a$ raises the probability of observing $b$, then $b$ raises the probability of observing $a$ too.

Nevertheless, in order to determine the order among pair of genetic events, we can use the *degree of confidence* in the estimate of probability raising to decide the direction of their relationship. In other words, if $a$ raises the probability of $b$ *more* than the other way around, then $a$ is a more likely singleton prima facie cause of $b$ than $b$ of $a$. Notice that this is sound as long as each event has *at most* one cause; otherwise, *frequent late events* with more than one predecessor, which are rather common in biological progressive phenomena, should be treated differently. As mentioned, the PR is not symmetric, and the *direction* of probability raising depends on the relative frequencies of the events. We make this asymmetry precise in the following proposition.

**Proposition 5** (Natural ordering)**.** *For any two events $a$ and $b$ such that the probability raising $\mathcal{P}(a \mid b) > \mathcal{P}(a \mid \overline{b})$ holds, we have*

$$\mathcal{P}(a) > \mathcal{P}(b) \iff \frac{\mathcal{P}(b \mid a)}{\mathcal{P}(b \mid \overline{a})} > \frac{\mathcal{P}(a \mid b)}{\mathcal{P}(a \mid \overline{b})} \,. \tag{3.4}$$

That is, given that the PR holds between two events, $a$ raises the probability of $b$ *more* than $b$ raises the probability of $a$, if and only if $a$ is observed more frequently than $b$. Also notice that we use the ratio to assess the PR inequality. The proof of this proposition is technical and can be found in §C.

From this result it follows that if we measure the timing of an event by the rate of its occurrence (that is, $\mathcal{P}(a) > \mathcal{P}(b)$ implies that $a$ happens before $b$), this notion of PR subsumes the same notion of temporal priority induced by a tree. We also remark that this is also the temporal priority made explicit in the coefficients of Desper's method [38]. Given these results, we define the following notion of singleton prima facie causaluty.

**Definition 4.** *We state that $a$ is a prima facie cause of $b$ if $a$ is a probability raiser of $b$, and it occurs more frequently: $\mathcal{P}(b \mid a) > \mathcal{P}(b \mid \overline{a})$ and $\mathcal{P}(a) > \mathcal{P}(b)$.*

We term *prima facie topology* a directed acyclic graph (over some events) where each edge represents a prima facie cause. When at most a single incoming edge is assigned to each event (i.e., an event has at most a *unique cause*, in the real world), we term this structure *singleton prima facie topology*. Intuitively, this last class of topologies correspond to the trees or, more generally forests when they have disconnected components, that we aim at reconstructing.

Before moving on to introducing the algorithm proposed here, let us once again recall the definition of *selective advantage*, its role in the definition of the reconstruction problem and some of its limitations. As already mentioned, it may be that for some prima facie cause $c$ of an event $e$, there is a third event $a$ prior to both, such that $a$ causes $c$ and ultimately $c$ causes $e$. Alternatively, $a$ may cause both $c$ and $e$ independently, and the selective advantage relationship observed from $c$ to $e$ is merely *spurious*. In the context of singleton topologies, namely when it is assumed that each event has at most a unique predecessor, the aim is to filter out the spurious edges from a general prima facie topology, so to extract a singleton prima facie structure (see Figure §3.1).

Proposition §5 summarizes Suppes basic notion of prima facie causality, while it is ignoring deeper discussions of causationthat aim at distinguishing between actual genuine and spurious causes (see §A), e.g. screening-off, background context, d-separation [22, 150, 80]. For our purposes however, the above definition is sufficient when (*i*) all the significant events are considered, i.e., all the genuine causes are observed as in a closed-world assumption, and (*ii*) we aim at extracting the *order* of progression among them (or determine that there is no apparent relation), rather than extracting causalities *per se*.

Finally, we recall a few algebraic requirements necessary for our framework to be well-defined. First of all, the PR must be computable: every mutation $a$ should be observed with probability strictly $0 < \mathcal{P}(a) < 1$. Moreover, we need each pair of mutations $(a, b)$ to be *distinguishable* in terms of PR, that is, for each pair of mutations $a$ and $b$, $\mathcal{P}(a \mid b) < 1$ or $\mathcal{P}(b \mid a) < 1$ similarly to the above condition. Any non-distinguishable pair of events can be merged as a single composite event. From now on, we will assume these conditions to be verified.

## 3.3   Results and discussion

We will now present the method in details.

### 3.3.1   Extracting progression trees

The CAPRESE reconstruction method is described in Algorithm §1. The algorithm is similar to Desper's and Szabo's algorithms, the main difference being an alternative weight function based on a shrinkage-like estimator.

**Definition 5** (Shrinkage-like estimator)**.** *We define the* shrinkage-like estimator $m_{a \to b}$ *of the confidence in the causationrelationship from $a$ to $b$ as*

$$m_{a \to b} = (1 - \lambda)\alpha_{a \to b} + \lambda\beta_{a \to b}\,, \tag{3.5}$$

*where* $0 \leq \lambda \leq 1$ *and*

$$\alpha_{a \to b} = \frac{\mathcal{P}(b \mid a) - \mathcal{P}(b \mid \overline{a})}{\mathcal{P}(b \mid a) + \mathcal{P}(b \mid \overline{a})} \qquad \qquad \beta_{a \to b} = \frac{\mathcal{P}(a \wedge b) - \mathcal{P}(a)\mathcal{P}(b)}{\mathcal{P}(a \wedge b) + \mathcal{P}(a)\mathcal{P}(b)}\,. \qquad (3.6)$$

This estimator is similar in spirit to a shrinkage estimator (see [45]) and, in fact, it combines a normalized version of the PR, i.e., the *raw estimate* $\alpha$, with a *correction factor* $\beta$ (which in this case is a correlation-based measure of temporal distance among events), to estimate the confidence of each selective advantage relationship. The parameter $\lambda$ is the analogous of the *shrinkage coefficient* and can have a Bayesian interpretation in terms of a measure of the belief that $a$ and $b$ are causally relevant to one another and of the evidence that $a$ raises the probability of $b$. In the absence of a closed form solution for the optimal value of $\lambda$, one may rely on cross-validation of simulated data. The power of shrinkage (and the shrinkage-like estimator) lies in the possibility of determining an optimal value for $\lambda$ to balance the effect of the correction factor on the raw model estimate to ensure optimal performances on ill posed instances of the inference problem. A crucial difference, however, between our estimator and classical shrinkage, is that this estimator aims at improving the performance of the *overall* reconstruction process, not limited to the performance of the estimator itself as is the case in shrinkage. That is, the metric $m$ induces an ordering to the events reflecting the confidence for their causation. Furthermore, since we make no assumption about the underlying distribution, we learn it empirically by cross-validation. In the next sections we show that the shrinkage-like estimator is an effective way to get such an ordering especially when data are noisy. In CAPRESE we use a pairwise matrix version of the estimator.

**The raw estimator and the correction factor.** By considering only the raw estimator $\alpha$, we would include an edge ($a \to b$) in the tree consistently in terms of ($i$) Definition §4 and ($ii$) if $a$ is the best probability raiser for $b$. When $\mathcal{P}(a) = \mathcal{P}(b)$ the events $a$ and $b$ are indistinguishable in terms of temporal priority, thus $\alpha$ is not sufficient to decide their causal relation, if any. This intrinsic ambiguity is unlikely in practice even if, in principle, it is possible. Notice that this formulation of $\alpha$ is a monotonic normalized version of the PR ratio.

**Proposition 6** (Monotonic normalization)**.** *For any two events $a$ and $b$ we have*

$$\mathcal{P}(a) > \mathcal{P}(b) \iff \frac{\mathcal{P}(b \mid a)}{\mathcal{P}(b \mid \overline{a})} > \frac{\mathcal{P}(a \mid b)}{\mathcal{P}(a \mid \overline{b})} \iff \alpha_{a \to b} > \alpha_{b \to a}\,. \qquad (3.7)$$

This raw model estimator satisfies $-1 \leq \alpha_{a \to b}, \alpha_{b \to a} \leq 1$: when it tends to $-1$ the pair of events appear disjointly (i.e., they show an anti-causation pattern), when it tends to $0$ no causation or anti-causation can be inferred and the two events are statistically independent and, when it tends to $1$, the causation relationship between the two events is genuine. Therefore, $\alpha$ provides a quantification of the degree of confidence for a PR selective advantage relationship. In fact, for any given possible causation edge $(a, b)$, the term $\mathcal{P}(b \mid \overline{a})$ gives an estimate of the *error rate* of $b$, therefore the numerator of the raw

---

**Algorithm 1:** CAPRESE: a tree-alike reconstruction with a shrinkage-like estimator

---

1: consider a set of $n$ genetic events $G$ plus a special event $\diamond$, added to each sample of the dataset;

2: define a $m \times n$ matrix $M$ where each entry contains the shrinkage-like estimator

$$m_{i \rightarrow j} = (1 - \lambda) \cdot \frac{\mathcal{P}(j \mid i) - \mathcal{P}(j \mid \bar{i})}{\mathcal{P}(j \mid i) + \mathcal{P}(j \mid \bar{i})} + \lambda \cdot \frac{\mathcal{P}(i \wedge j) - \mathcal{P}(i)\mathcal{P}(j)}{\mathcal{P}(i \wedge j) + \mathcal{P}(i)\mathcal{P}(j)}$$

according to the observed probability of the events $i$ and $j$;

3: [PR causation] define a tree $\mathcal{T} = (G \cup \{\diamond\}, E, \diamond)$ where $(i \rightarrow j) \in E$ for $i, j \in G$ if and only if:

$$m_{i \rightarrow j} > 0 \quad \text{and} \quad m_{i \rightarrow j} > m_{j \rightarrow i} \quad \text{and} \quad \forall i' \in G, \, m_{i,j} > m_{i',j} \, .$$

4: [Independent progressions filter] define $G_j = \{x \in G \mid \mathcal{P}(x) > \mathcal{P}(j)\}$, replace edge $(i \rightarrow j) \in E$ with edge $(\diamond \rightarrow j)$ if, for all $x \in G_j$, it holds

$$\frac{1}{1 + \mathcal{P}(j)} > \frac{\mathcal{P}(x)}{\mathcal{P}(x) + \mathcal{P}(j)} \frac{\mathcal{P}(x \wedge j)}{\mathcal{P}(x)\mathcal{P}(j)} \, .$$

---

model $\alpha$ provides an estimate of how often $b$ is actually caused by $a$. The $\alpha$ estimator is then normalized to range between $-1$ and $+1$.

However, $\alpha$ does not provide a general criterion to disambiguate among genuine causes of a given event. We show a specific case in which $\alpha$ is not a sufficient estimator. Let us consider, for instance, a causal linear path: $a \rightarrow b \rightarrow c$. In this case, when evaluating the candidate parents $a$ and $b$ for $c$ we have: $\alpha_{a \rightarrow c} = \alpha_{b \rightarrow c} = 1$, so $a$ and $b$ are genuine causes of $c$, though we would like to select $b$, instead of $a$. Accordingly, we can only infer that $t_a < t_c$ and $t_b < t_c$, i.e., a partial ordering, which does not help to disentangle the relation among $a$ and $b$ with respect to $c$.

In this case, the $\beta$ coefficient can be used to determine which of the two genuine causes occurs closer, in time, to $c$ ($b$, in the example above). In general, such a correction factor provides information on the *temporal distance* between events, in terms of statistical dependency. In other words, the higher the $\beta$ coefficient, the closer two events are. Therefore, when dealing with noisy data and limited sample sizes, there are situations where, by using the $\alpha$ estimator alone, we could infer a wrong transitive edge to be the most likely cause even in the presence of the real cause. For this reason, we reduce the $\alpha$ estimator to the correction factor $\beta$, which, for each given edge $(a, b)$, is normalized within $-1$ and $(1 - \max[\mathcal{P}(a), \mathcal{P}(b)])/(1 + \max[\mathcal{P}(a), \mathcal{P}(b)]) < +1$.

The shrinkage-like estimator $m$ then results in the combination of the raw PR estimator $\alpha$ and of the $\beta$ correction factor, which respects the temporal priority induced by

$\alpha$.

**Proposition 7** (Coherence in dependency and temporal priority). *The $\beta$ correction factor is* symmetrical *and subsumes the same notion of dependency of the raw estimator $\alpha$, that is*

$$\mathcal{P}(a \wedge b) > \mathcal{P}(a)\mathcal{P}(b) \ \Leftrightarrow \alpha_{a \to b} > 0 \Leftrightarrow \beta_{a \to b} > 0 \quad and \quad \beta_{a \to b} = \beta_{b \to a} \,. \tag{3.8}$$

**The independent progressions filter.** As in Desper's approach, we also add a *root $\diamond$* with $\mathcal{P}(\diamond) = 1$ in order to separate different progression paths, i.e., the different sub-trees rooted in $\diamond$. CAPRESE initially builds a unique tree by using the estimator; typically, the most likely event will be at the top of the progression even if there may be rare cases where more than one event has no valid parent, in these cases we would already be reconstructing a forest. In the reconstructed tree, all the edges represent the most confident prima facie cause, although some of those could still be spurious causes. Then the correlation-like weight between any node $j$ and $\diamond$ is computed as

$$\frac{\mathcal{P}(\diamond)}{\mathcal{P}(\diamond) + \mathcal{P}(j)} \frac{\mathcal{P}(\diamond \wedge j)}{\mathcal{P}(\diamond)\mathcal{P}(j)} = \frac{1}{1 + \mathcal{P}(j)} \,.$$

If this quantity is greater than the weight of $j$ with each upstream connected element $i$, we consider the best prima facie cause of $j$ to be a spurious cause and we substitute the edge $(i \to j)$ with the edge $(\diamond \to j)$. Note that in this work we are ignoring deeper discussions of probabilistic causationthat aim at distinguishing between actual genuine causes and spurious causes. Instead, we remove spurious causes by using a filter based on correlation because the probability raising of the omnipresent event $\diamond$ is not well defined (see Methods). In addition, we remark that the evaluation for an edge to be a genuine or a spurious cause takes into account all the given events. Because of this, if events are added or removed from the dataset, the same edge can be defined to be genuine or spurious as the set of events included in the model is varied arbitrarily. However, since we do not consider the problem of selecting the set of progression events, we assume that all and only the relevant events for the problem at hand are already known a priori and included in the model.

Finally, note that this filter is indeed implying a non-negative threshold for the shrinkage-like estimator, when an edge is valid.

**Theorem 1** (Independent progressions). *Let $G^* = \{a_1, \ldots, a_k\} \subset G$ a set of $k$ prima facie causes for some $b \notin G^*$, and let $a^* = \max_{a_i \in G^*}\{m_{a_i \to b}\}$. The reconstructed tree by CAPRESE contains edge $\diamond \to b$ instead of $a^* \to b$ if, for all $a_i \in G^*$*

$$\mathcal{P}(a_i, b) < \mathcal{P}(a_i)\mathcal{P}(b)\frac{1}{1 + \mathcal{P}(b)} + \frac{\mathcal{P}(b)^2}{1 + \mathcal{P}(b)} \,. \tag{3.9}$$

The proof of this theorem can be found in §C. What this theorem suggests is that, in principle, by examining the level of statistical dependency of each pair of events, it would be possible to determine how many trees compose the reconstructed forest. Furthermore, the theorem suggests that CAPRESE could be defined by first processing the independent progressions filter, and then using $m$ to build the independent progression trees in the forest.

To conclude, the algorithm reconstructs a well defined tree (or, more in general, forest).

**Theorem 2** (Algorithm correctness). *CAPRESE reconstructs a well defined tree $\mathcal{T}$ without disconnected components, transitive connections and cycles.*

Additionally, asymptotically with the number of samples, the reconstructed tree is the correct one.

**Theorem 3** (Asymptotic convergence). *Let $T = (G \cup \{\diamond\}, E, \diamond)$ be the forest to reconstruct from a set of $s$ input samples, given as the input matrix $D$. If $D$ is strictly sampled from the distribution induced by $T$ and infinite samples are available, i.e., $s \to \infty$, CAPRESE with $\lambda \to 0$ correctly reconstructs $T$.*

The proof of these Theorems are also in §C. These theorems considered datasets where the observed and theoretical probabilities match, because of $s \to \infty$. However, data often contains false positives and negatives (i.e., data are noisy) and resistance to their effects is desirable in any inferential technique. With this in mind, we prove a corollary of the theorem analoguos to a result appearing in [173].

**Corollary 1** (Uniform noise). *Let the input matrix $D$ be strictly sampled from the distribution induced by $T$ with sample size $s \to \infty$, but let it be corrupted by noise levels of false positives $\epsilon_+$ and false negatives $\epsilon_-$. Let $p_{\min} = \min_{x \in G}\{\mathcal{P}(x)\}$, CAPRESE correctly reconstructs $T$ for $\lambda \to 0$ whenever*

$$\epsilon_+ < \sqrt{p_{\min}}(1 - \epsilon_+ - \epsilon_-)$$

*and $\epsilon_+ + \epsilon_- < 1$.*

Essentially, this corollary states that CAPRESE (and so the estimator $m$) is robust against a noise affecting all samples equally. Also, the fact that it holds for $\lambda \to 0$ is sound with the theory of shrinkage estimators for which, asymptotically, the corrector factor is not needed to regularize the ill posed problem.

### 3.3.2   Optimal shrinkage-like coefficient

Theorem §3 and Corollary §1 state that with infinite samples and mild constraints on the false positive/negative rates we get optimal results with $\lambda \to 0$. Precisely, for the uniform noise model that we applied to synthetic data, we have $\epsilon_+ = \epsilon_- = \nu/2$, thus the hypothesis required by Corollary §1 is

$$\nu < \frac{\sqrt{p_{\min}}}{1/2 + \sqrt{p_{\min}}}\,.$$

For $p_{min} = 0.05$, which we set in data generation (see §C), this inequality implies correct reconstruction for $\nu < 0.3$ (a 15% error rate), with infinite samples. However, we are interested in performance and the optimal value of $\lambda$ in situations in which we have finite sample sizes as well. Here, we empirically estimate the optimal $\lambda$ value, both in the case of trees and forests, as a function of noise and sample size. In the next section, we assess performance of our algorithm empirically.

In Figure §3.2, we show the variation of the performance of CAPRESE as a function of $\lambda$, for datasets with 150 samples generated from tree topologies. The optimal value, i.e., lowest Tree Edit Distance (TED), for noise-free datasets (i.e., $\nu = 0$) is obtained for $\lambda \to 0$, whereas for the noisy datasets a series of U-shaped curves suggests a unique optimum value for $\lambda \to 1/2$, immediately observable for $\nu < 0.15$. Identical results are obtained when dealing with forests (not shown here). In addition, further experiments with $n$ varying around the typical sample size ($n = 150$) show that the optimal $\lambda$ is largely insensitive to the dataset size (see Figure §3.3). Thus we have limited our analysis to datasets with the typical sample size that is characteristic of data currently available.

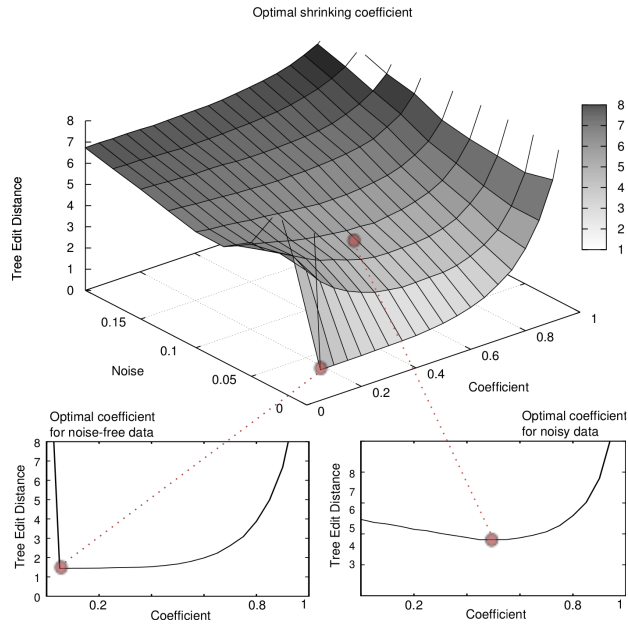

Figure 3.2: **Optimal shrinkage-like coefficient for reconstruction performance.** We show here the performance in the reconstruction of trees (TED surface) with $n = 150$ samples as a function of the shrinkage-like coefficient $\lambda$. Notice the global optimal performance for $\lambda \to 0$ when $\nu \to 0$ and for $\lambda \approx 1/2$ when $\nu > 0$.

Summarizing, Figures §3.2 and §3.3 suggest that for sample size below 250 without false positives and negatives the PR raw estimate $\alpha$ suffices to drive reconstruction to good results (TED is 0 with 250 samples), i.e.,

$$m_{a \to b} \overset{\lambda \to 0}{\approx} \alpha_{a \to b} \tag{3.10}$$

which is obtained by setting $\lambda$ to a very small value, e.g. $10^{-2}$, in order to consider at least a small contribution of the correction factor too. Conversely, when $\nu > 0$, the best performance is obtained by averaging the shrinkage-like effect, i.e.,

$$m_{a \to b} \overset{\lambda = 1/2}{=} \frac{\alpha_{a \to b}}{2} + \frac{\beta_{a \to b}}{2} \,. \tag{3.11}$$

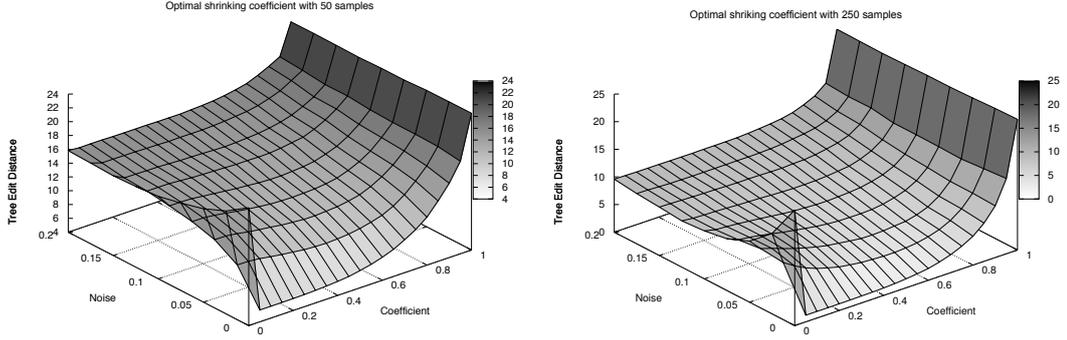

Figure 3.3: **Optimal $\lambda$ with datasets of different size.** We show the analogous of Figure §3.2 with 50 and 250 samples. The estimation of the optimal shrinkage-like coefficient $\lambda$ appears to be irrespective of the sample size.

These results suggest that, in general, a unique optimal value for the shrinkage-like coefficient can be determined, even in situations not captured by Theorem §3 and Corollary §1.

### 3.3.3   Performance measure and synthetic datasets

To evaluate the performance of the algorithm, named CAPRESE, i.e., CAncer PRogression Extraction with Single Edge, proposed here to infer singleton prima facie topologies, we made substantial use of *synthetic data* as a function of dataset size and the false positive and negative rates. Many distinct synthetic datasets were created for this purpose, as explained below. The algorithm's performance was measured in terms of *Tree Edit Distance* (TED, [197]), i.e., the minimum-cost sequence of node edit operations (relabeling, deletion and insertion) that transforms the reconstructed trees into the ones generating the data. The choice of this evaluation measure is motivated by the fact that we are interested in the *structure* behind the progressive phenomenon of cancer evolution and, in particular, we are interested in a measure of the genuine causes that we miss and of the spurious causes that we fail to recognize (and eliminate). Also, since topologies with similar distributions can be structurally different we choose to measure performance using structural distance rather than a distance in terms of distributions. Within the realm of 'structural metrics' however, we have also evaluated the performance with the *Hamming Distance* [76], another commonly used structural metric, and we obtained analogous results (see §C).

**Synthetic data generation and experimental setting.** We generated synthetic data by sampling from various random trees constrained to have depth $\log(|G|)$, since wide branches are harder to reconstruct than straight paths, and by sampling event probabilities in $[0.05, 0.95]$ (see §C).

Unless explicitly specified, in all the experiments we used 100 distinct random trees (or forests, accordingly to the test to perform) of 20 events each. This seems a fairly reasonable number of events and is in line with the usual size of reconstructed trees, e.g., [166, 69, 116, 148]. The *scalability* of the techniques was tested against the number of samples by ranging $|G|$ from 50 to 250, with a step of 50, and by replicating 10 independent datasets for each parameters setting (see the caption of the figures for details).

We included a form of *noise* in generating the datasets, in order to account for (*i*) the realistic presence of *biological noise* (such as the one provided by bystander mutations, genetic heterogeneity, etc.) and (*ii*) *experimental errors*. A noise parameter $0 \leq \nu < 1$ denotes the probability that any event assumes a random value (with uniform probability), after sampling from the tree-induced distribution. Algorithmically this process generates, on average, $|G|\nu/2$ random entries in each sample (e.g., with $\nu = 0.1$ we have, on average, one error per sample). We wish to assess whether these noisy samples can mislead the reconstruction process, even for low values of $\nu$. Notice that assuming a uniformly distributed noise may appear simplistic since some events may be more robust, or easy to measure, than others. However, introducing in the data both *false positives* (at rate $\epsilon_+ = \nu/2$) and *negatives* (at rate $\epsilon_- = \nu/2$) makes the inference problem substantially harder, and was first investigated in [173].

In the next section, we refer to datasets generated with rate $\nu > 0$ as noisy synthetic dataset. In the numerical experiments, $\nu$ is usually discretized by 0.025, (i.e., 2.5% noise).

## Performance of CAPRESE compared to *oncotrees*

An analogue of Theorem §3 holds for Despers's oncotrees (Theorem 3.3, [38]), and an analogue of Corollary §1 holds for Szabo's extension with uniform noise (Reconstruction Theorem 1, [173]). Thus, with infinite samples both approaches reconstruct the correct trees/forests. With finite samples and noise, however, their performance may show different patterns, as speed of convergence may vary. We investigate this issue in the current section.

In Figure §3.4 we compare the performance of CAPRESE with oncotrees, for the case of noise-free synthetic data with the optimal shrinkage-like coefficient: $\lambda \to 0$, equation (§3.10). Since Szabo's algorithm is equivalent to Desper's without false negatives and positives, we rely solely on Szabo's implementation [173]. In Figure §3.5 we show an example of reconstructed tree where CAPRESE infers the correct tree while oncotrees mislead an edge.

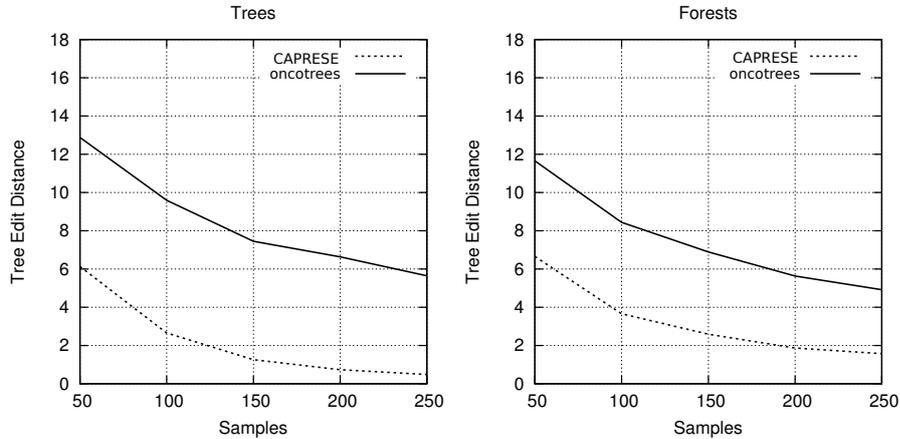

Figure 3.4: **Comparison on noise-free synthetic data.** Performance of CAPRESE (dashed line) and oncotrees (full line) in average TED when data are generated by random trees (left) and forests (right). In this case $\nu = 0$ (no false positives/negatives) and $\lambda \to 0$ in the estimator $m$.

In general, one can observe that the TED of CAPRESE is, on average, always bounded above by the TED of oncotrees, both in the case of trees and forests. For trees, with 50 samples the average TED of CAPRESE is around 6, whereas for Desper's technique it is around 13. The performance of both algorithms improves as long as the number of samples is increased: CAPRESE has the best performance (i.e., TED $\approx 0$) with 250 samples, while oncotrees have TED around 6. When forests are considered, the difference between the performance of the algorithms reduces slightly, but also in this case CAPRESE clearly outperforms oncotrees.

Notice that the improvement due to the increase in the sample size seems to reach a *plateau*, and the initial TED for our estimator seems rather close to the plateau value. This empirical analysis suggests that CAPRESE has already good performances with few samples, a favorable adjoint to Theorem §3. This result has some important practical implications, particularly considering the scarcity of available biological data.

In Figure §3.6 we extend the comparison to *noisy* datasets. In this case, we used the optimal shrinkage-like coefficient: $\lambda \to 1/2$, equation (§3.11). The results confirm what observed without false positives and negatives, as CAPRESE outperforms oncotrees up to $\nu = 0.15$, for all the sizes of the sample sets. In §C, similar plots for the noise-free case are shown.

We can thus draw the conclusion that our algorithm performs better with finite samples and noise, since less samples are required to get good performances and a higher resistance to false positives and negatives is shown.

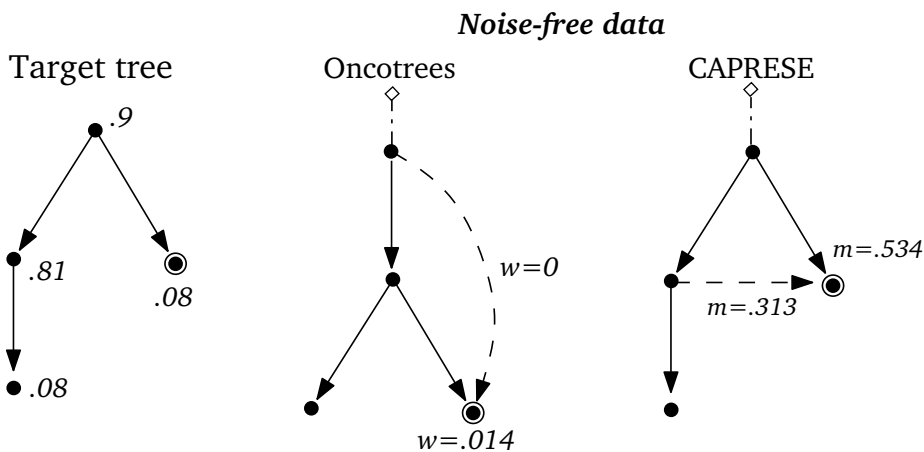

Figure 3.5: **Example of reconstructed trees.** Example of reconstruction from a dataset with 100 samples generated by the left tree (the theoretical probabilities are shown, i.e., the doubly-circled event appears in a sample with probability .08), with $\nu = 0$. In the sampled dataset oncotrees mislead the parent of the doubly-circled mutation ($w = 0$ for the true edge and $w = 0.014$ for the wrong one). CAPRESE infers the correct parent (the values of the estimator $m$ with $\lambda = 1/2$ are shown, similar results are obtained for $\lambda \to 1$).

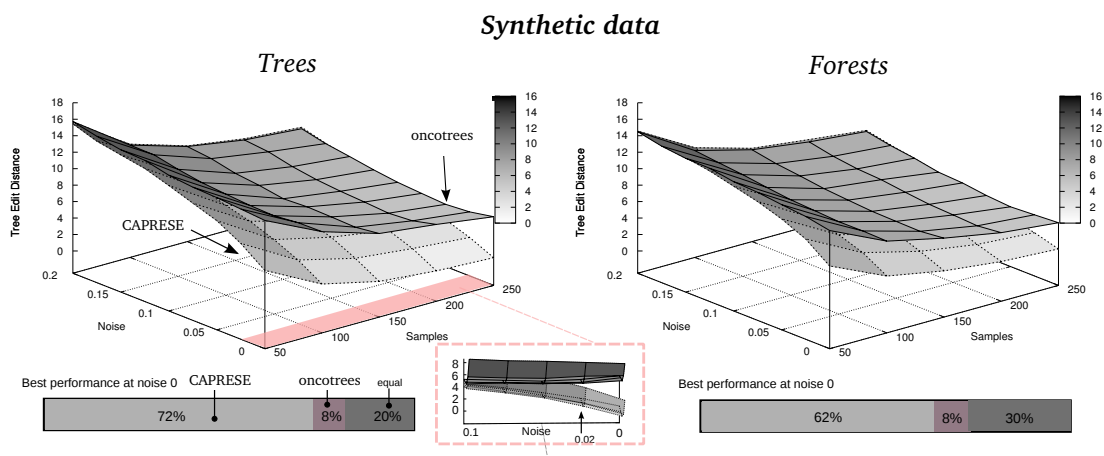

Figure 3.6: **Reconstruction with noisy synthetic data and $\lambda = 1/2$.** Performance of CAPRESE and oncotrees as a function of the number of samples and noise $\nu$. According to Figure §3.2 the shrinkage-like coefficient is set to $\lambda = 1/2$. The magnified image shows the convergence to Desper's performance for $\nu \approx 0.1$. The barplot represents the percentage of times the best performance is achieved at $\nu = 0$.

**Performance of CAPRESE compared to *Conjunctive Bayesian Networks***

Inspired by Desper's seminal work, Beerenwinkel and others developed methods to estimate the constraints on the order in which mutations accumulate during cancer progression, using a probabilistic graphical model called *Conjuntive Bayesian Networks* (CBN) [10, 65]. While the goal of this research was to reconstruct *direct acyclic graphs* and not trees per se, evidence presented in [72] suggests that, in the absence of noise, these models perform better than oncotrees even at reconstructing *trees*. For this reason, we performed experiments similar to the ones suggested above, comparing CAPRESE to the extension of CBN called *hidden-CBN* (h-CBN) that accounts for noisy genotype observations [65]. This method combines CBNs with a simulated annealing algorithm for structure search and a denoising of the genotypes via the maximum a posteriori estimates to compute the most likely progression. One aspect that complicates a comparison between CAPRESE and (h-)CBN is that the methods assume different models. For example, at the heart of CBN is a monotonicity assumption (i.e., an event can only occur if all its predecessors have occurred) not assumed by CAPRESE. Despite the differences between the model assumptions, we present a comparison between the methods in §C, indicating that CAPRESE not only outperform oncotrees, but h-CBNs as well. In particular, this suggests that CAPRESE converges much faster than h-CBNs with respect to the sample size, also in the presence of noise.

We also analyze the rate of *false positives/negatives* reconstructed by CAPRESE when (synthetic) data are sampled from DAGs ( §C). The rate of *false positives* goes to 0 as the sample size increases, implying that CAPRESE is capable of reconstructing a tree subsumed by the underlying causal DAG topology. In addition, the number of *false negatives* approaches a value proportional to the connectivity of the model from which the data was generated. This is expected, since CAPRESE will assign at most one cause to each considered event. However, it should be noted that further experiments with samples selected from a wider array of topologies should be performed to confirm these results and compare both methods in full. While not within the scope of this specific work.

### 3.3.4   Case studies

In the next subsections we apply CAPRESE to real cancer data obtained by *Comparative Genomic Hybridization* (CGH) and *Next Generation Sequencing* (NGS). This shows the potential application of reconstruction techniques to various types of mutational profiles and various cancers.

**Performance on cancer CGH datasets**

We test our reconstruction approach on a real *ovarian cancer* dataset made available within the oncotree package [38]. The data were collected through the public platform SKY/M-FISH [100], used to allow investigators to share molecular cytogenetic data. The data was obtained by using the CGH technique on samples from *papillary serous cystade-* *nocarcinoma* of the ovary. This technique uses fluorescent staining to detect CNV data at the resolution of chromosome arms. While the recent emergence of NGS approaches make the dataset itself rather outdated, the underlying principles remain the same and the dataset provides a valid test-case for our approach. The seven most commonly occurring events are selected from the 87 samples, and the set of events are the following gains and losses on chromosomes arms $G = \{8q+, 3q+, 1q+, 5q-, 4q-, 8p-, Xp-\}$ (e.g., $4q-$ denotes a deletion of the $q$ arm of the $4^{th}$ chromosome).

In Figure §3.7 we compare the trees reconstructed by the two approaches. Our technique differs from Desper's by predicting the causal sequence of alterations

$$8q+ \; \rightarrow \; 8p- \; \rightarrow \; Xp-,$$

when used either $\lambda \to 0$ or $\lambda = 1/2$. Notice that among the samples in the dataset some are not generated by the distribution induced by the recovered tree, thus comparing the reconstruction for both $\lambda$s is necessary.

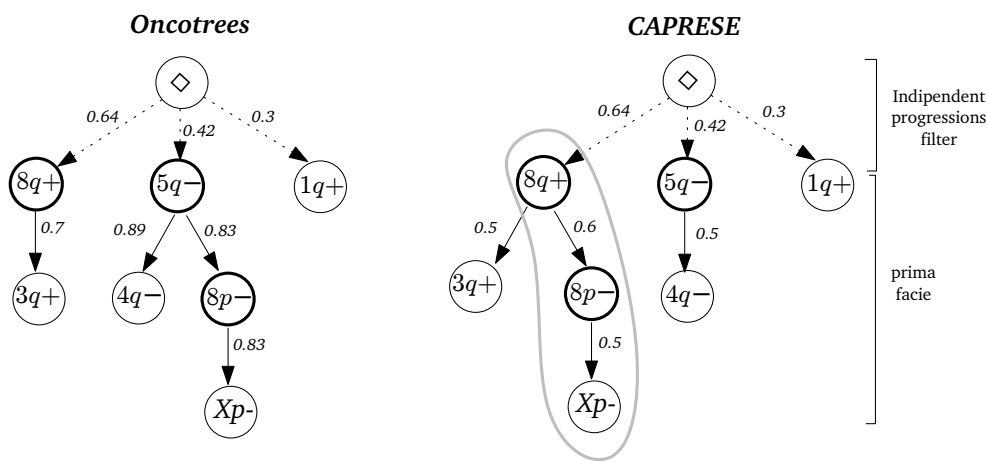

Figure 3.7: **Reconstruction of ovarian cancer progression.** Trees reconstructed by oncotrees and CAPRESE (with $\lambda \to 0$, with $\lambda = 1/2$ the same tree is reconstructed). The set of CGH events considered are gains on $8q$, $3q$ and $1q$ and losses on $5q$, $4q$, $8p$ and $Xp$. Events on chromosomes arms containing the key genes for ovarian cancer are in bolded circles. In the left tree all edge weights are the observed probabilities of events. In the right the full edges are the causation inferred with the PR and the weights represent the scores used by CAPRESE. Weights on dashed lines are as in the left tree.

At this point, we do not have a biological interpretation for this result. However, we do know that common cancer genes reside in these regions, e.g. the tumor suppressor gene PDGFR on $5q$ and the oncogene MYC on $8q$), and loss of heterozygosity on the short arm of chromosome 8 is quite common (see, e.g., http://www.genome.jp/kegg/). Recently, evidence has been reported that $8p$ contains many cooperating cancer genes [193].

In order to assign a confidence level to these inferences we applied both parametric and non-parametric *bootstrapping methods* to these results. Essentially, these tests consist of using the reconstructed trees (in the parametric case), or the probability observed in the dataset (in the non-parametric case) to generate new synthetic datasets, and then reconstructs again the progressions (see, e.g., [46] for an overview of these methods and [174] for the use of bootstrap for evalutating the confidence of oncogenetic trees.).  The confidence is given by the number of times the trees in Figure §3.7 are reconstructed from the generated data. A similar approach can be used to estimate the confidence of every edge separately. For oncotrees the *exact tree* is obtained 83 times out of 1000 non-parametric resamples, so its estimated confidence is 8.3%. For CAPRESE the confidence is 8.6%.  In the parametric case with false positive and false negative error rates of 0.21 and 0.027, following [173], the confidence of oncotrees is 17% while the confidence of our method is much higher: 32%. When error rates are forced to 0 the confidence of oncotrees raises to 86.6% and 90.9% respectively.

For the non-parametric case, edges confidence is shown in Table §3.1. Most notably, our algorithm reconstructs the inference $8q+ \rightarrow 8p-$ with high confidence (confidence 62%, and 26% for $5q- \rightarrow 8p-$), while the confidence of the edge $8q+ \rightarrow 5q-$ is only 39%, almost the same as $8p- \rightarrow 8q+$ (confidence 40%).  The confidences are similar with either $\lambda \rightarrow 0$ or $\lambda = 1/2$.

Oncotrees (overall confidence 8.3%)

| $\rightarrow$ | $8q+$ | $3q+$ | $5q-$ | $4q-$ | $8p-$ | $1q+$ | $Xp-$ |
|---|---|---|---|---|---|---|---|
| $\diamond$ | **.99** | .06 | **.51** | .22 | .004 | **.8** | .06 |
| $8q+$ | 0 | **.92** | .08 | 0.16 | 0.4 | .02 | .007 |
| $3q+$ | .002 | 0 | .04 | 0 | 0 | .09 | .04 |
| $5q-$ | .001 | .002 | 0 | **.52** | **.39** | .009 | .16 |
| $4q-$ | 0 | 0 | .27 | 0 | .14 | .05 | .11 |
| $8p-$ | 0 | 0 | .07 | .08 | 0 | .004 | **.59** |
| $1q+$ | 0 | 0 | 0 | .004 | 0 | 0 | 0 |
| $Xp-$ | 0 | 0 | .003 | .003 | .04 | .01 | 0 |

CAPRESE (overall confidence 8.6%)

| $\rightarrow$ | $8q+$ | $3q+$ | $5q-$ | $4q-$ | $8p-$ | $1q+$ | $Xp-$ |
|---|---|---|---|---|---|---|---|
| $\diamond$ | **.99** | .06 | **.51** | .22 | .004 | **.8** | .06 |
| $8q+$ | 0 | **.92** | .06 | .16 | **.62** | .01 | .008 |
| $3q+$ | .002 | 0 | .03 | .002 | 0 | .09 | .04 |
| $5q-$ | .001 | .002 | 0 | **.5** | .26 | .009 | .17 |
| $4q-$ | 0 | 0 | .29 | 0 | .09 | .05 | .12 |
| $8p-$ | 0 | 0 | .07 | .08 | 0 | .004 | **.59** |
| $1q+$ | 0 | 0 | 0 | .004 | 0 | 0 | 0 |
| $Xp-$ | 0 | .001 | .003 | .004 | .01 | .01 | 0 |

Table 3.1: **Estimated confidence for ovarian progression.** Frequency of edge occurrences in the non-parametric bootstrap test, for the trees shown in Figure §3.7. Colors represent confidence: light gray is $< 0.4$, mid gray is in the range $[0.4, 0.8]$ and dark gray is $> 0.8$. Bold entries are the edges recovered by the algorithms.

**Analysis of other CGH datasets.** We report the differences between the reconstructed trees also based on datasets of gastrointestinal and oral cancer ([69, 148] respectively). In the case of gastrointestinal stromal cancer, among the 13 CGH events considered in [69] (gains on $5p$, $5q$ and $8q$, losses on $14q$, $1p$, $15q$, $13q$, $21q$, $22q$, $9p$, $9q$, $10q$ and $6q$), oncotrees identify the path progression

$$1p- \to 15q- \to 13q- \to 21q-$$

while CAPRESE reconstructs the branch

$$1p- \to 15q- \qquad\qquad 1p- \to 13q- \to 21q - \ .$$

In the case of oral cancer, among the 12 CGH events considered in [148] (gains on $8q$, $9q$, $11q$, $20q$, $17p$, $7p$, $5p$, $20p$ and $18p$, losses on $3p$, $8p$ and $18q$), the reconstructed trees differ since oncotrees identifies the path

$$8q+ \to 20q+ \to 20p+$$

while our algorithm reconstructs the path

$$3p- \to 7p+ \to 20q+ \to 20p + \ .$$

These examples show that CAPRESE provides important differences in the reconstruction compared to oncotrees.

## Performance on cancer NGS datasets

In this section we show the application of reconstruction techniques to the validation of a specific relation among recurrent mutations involved in *atypical Chronic Myeloid Leukemia* (aCML).

In [154] Piazza *et al.* used high-throughput *exome sequencing technology* to identity somatically acquired mutations in 64 aCML patients, and found a previously unidentified recurring *missense point mutation* hitting SETBP1. By re-sequencing SETBP1 in samples with aCML and other common human cancers, they found that around 25% of the aCML patients tested positive for SETBP1, while most of the other types of tumors were negative. Assessing the possible relationship between SETBP1 variants and the mutations in many driver aCML oncogenes such as (e.g., ASXL1, TET2, KRAS, etc.) no significant association or mutual exclusion with SETBP1 was found but for ASXL1, which was frequently mutated together with SETBP1, hinting at a potential relation among the events. In particular, ASXL1 was presenting either a *non-sense point* or a *indel* type of somatic mutation.

Hence, we reconstructed aCML progression models from the datasets provided in [154], with the goal of assessing a *potential selective advantage relation* between mutated SETBP1 and ASXL1. A more extensive analysis is postponed, as we only seek to clearly illustrate the functionalities of the algorithm here.

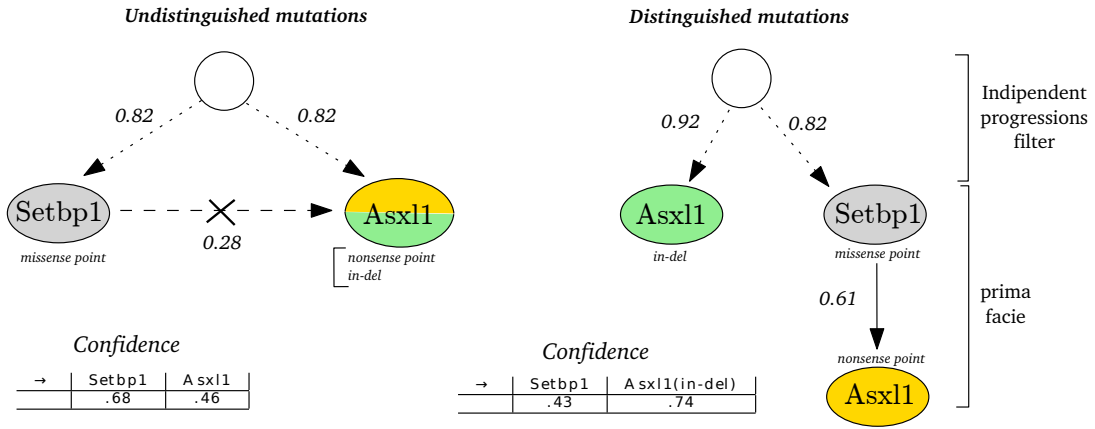

Figure 3.8: **The setbp1-asxl1 relation in atypical Chronic Myeloid Leukemia.** Progression models where ASXL1 indel and non-sense point are merged (left) and separate (right) suggest that a missense point mutation hitting SETBP1 can cause a non-sense point mutation in ASXL1, that the observed ASXL1 mutations might be independent and that indel ASXL1 is an early event with high confidence.

As a first case (Figure §3.8, left), we treated the ASXL1 missense point and indel mutations as indistinguishable, and we merged the two events in the dataset. Afterwards, we separated the two types of mutations for ASXL1 (Figure §3.8, right).

In particular, it is interesting to notice that, when the ASXL1 mutations are considered equivalent, the inference suggests that the mutations belong to two independent progression paths (i.e., the independent progression filters "breaks" every potential selective advantage relation among the events). Conversely, when the mutations are kept separate, the progression model suggests that: (i) the missense point mutation hitting SETBP1 can cause a non-sense point mutation in ASXL1 and (ii) the observed ASXL1 mutations seems to be independent. Concerning edges confidence, as before assessed via non-parametric bootstrap, it is worth noting that the confidence in the indel ASXL1 mutation being an early event raises consistently in the latter case.

All in all, it seems that a progression model allows to test the significance of the association firstly observed in [154] and also refines the knowledge by suggesting a specific causal and temporal relations among events. With this this in mind, ad-hoc sequencing experiments might be set up to assess these predictions, eventually providing a strong evidence that could be used to, e.g., synthesize a progression-specific ACML-effective drug.

# MORE COMPLEX MODELS OF CANCER PROGRESSION

In this Chapter we will present an algorithm that is capable of efficiently inferring complex models of cancer progression such as directed acyclic graphs. As a reference, see [158].

## 4.1 Problem setting

We define a *progression DAG* as a directed acyclic graph $\mathcal{D} = (N, \pi)$, where $N \subseteq \mathcal{U}$ is the set of nodes (e.g., selected from a universe $\mathcal{U}$ of *mutations* or propositional formulas) and $\pi : N \to \wp(N)$ is a function, which associates with each node $j$ its *parents* $\pi(j) \subseteq N$. We wish to study the cases where such a DAG can be seen as a model for the following classes of *selectivity patterns*, expressed in conjunctive normal form (CNF). The symbol $\triangleright$ stands for the *selectivity relation*.

**Definition 6** (DAG patterns). *A $\mathcal{D} = (N, \pi)$ is a model for models the* patterns

$$\bigcup_{j \in N} \left\{ (\boldsymbol{c}_1 \wedge \ldots \wedge \boldsymbol{c}_n) \triangleright j \mid \pi(j) = \{\boldsymbol{c}_1, \ldots, \boldsymbol{c}_n\} \right\},$$

*where $\boldsymbol{c}_1 \wedge \ldots \wedge \boldsymbol{c}_n$ is a CNF formula (each clause $\boldsymbol{c}_j$ is one of two kinds: either an atomic event or a disjunction of events).*

Each DAG represents an induced *distribution* of observing a subset of the considered events in a set of samples (i.e., the probability of observing certain genetic alterations in a group of patients or cells representing their *mutational profile*).

**Definition 7** (DAG-induced distribution). *Let $\mathcal{D} = (N, \pi)$ be a DAG and $\alpha : N \to [0, 1]$ a labeling function, $\mathcal{D}$ generates a distribution where the probability of observing $N^* \subseteq N$ events is*

$$\mathcal{P}(N^*) = \prod_{x \in N^*} \alpha(x) \cdot \prod_{y \in N \setminus N^*} \left[ 1 - \alpha(y) \right] \tag{4.1}$$

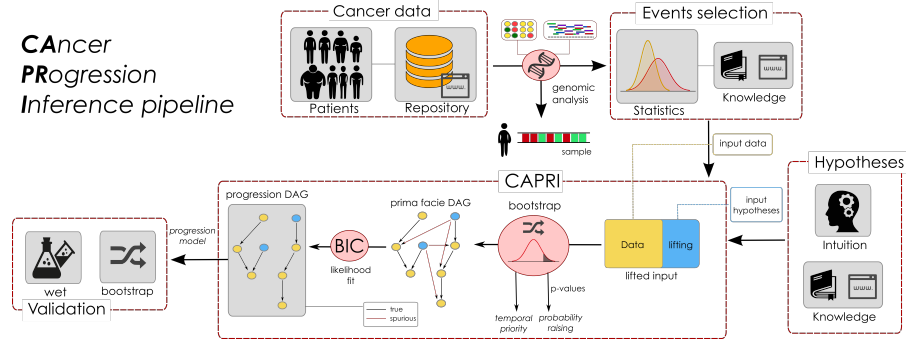

Figure 4.1: **Data processing pipeline for cancer progression inference.** We sketch a pipeline to best exploit CAPRI's ability to extract cancer progression models from cross-sectional data. Initially, one collects *experimental data* (which could be accessible through publicly available repositories such as TCGA) and performs *genomic analyses* to derive profiles of, e.g., somatic mutations or Copy-Number Variations for each patient. Then, statistical analysis and biological priors are used to select events relevant to the progression and imputable by CAPRI - e.g., *driver mutations*. To exploit CAPRI's supervised execution mode (see Methods) one can use further statistics and priors to generate *patterns of selective advantage* - , e.g, hypotheses of mutual exclusivity. CAPRI can extract a progression model from these data and assess various *confidence* measures on its constituting relations - e.g., (non-)parametric bootstrap and hypergeometric testing. *Experimental validation* concludes the pipeline.

*whenever* $x \in N^*$, $\pi(x) \subset N^*$, *and 0 otherwise.*

Notice that this definition, as expected, is equivalent to the one used in [10] and retains a tree-induced distribution such as those used in [117, 38, 173]. Further, notice that a sample which contains an event but not all of its parents has a zero probability, thus subsuming the conjunctive interpretation of DAGs, as the result of compositional reasoning to infer co-occurrence patterns. These kinds of samples, which represent "irregularities" with respect to $\mathcal{D}$, might be generated when adding false positives/negatives to the sampling strategy.

## 4.2 A novel efficient framework

Building on the framework described in the §2, we now describe the implementation of my framework for CAncer PRogression Inference (named CAPRI)'s building blocks. Notice that, in general, the inference of cancer progression models requires a complex *data processing pipeline*, as summarized in Figure §4.1; its architecture optimally exploits CAPRI's efficiency.

**Assumptions.** CAPRI relies on the following assumptions: *i)* Every pattern is expressible as a propositional CNF formula; *ii)* All events are persistent, i.e., an acquired mutation cannot disappear; *iii)* All relevant events in tumor progression are observable, with the observations describing the progressive phenomenon in an essential manner (i.e., *closed world* assumption, in which all events 'driving' the progression are detectable); *iv)* All the events have non-degenerate observed probability in $(0, 1)$; *v)* All events are distinguishable, in the following sense: input alterations produce different profiles across input samples.

Assumptions *i-ii)* relate to the framework derived in previous section, while *iii)* imposes an onerous burden on the experimentalists, who must select the *relevant* genomic events to model[1]. Assumption *iv)* relates instead to the statistical distinguishability of the input events (see the next section on CAPRI's Data Input).

**Trading Complexity for Expressivity.** To automatically extract the patterns that underly a progression model, one may try to adopt a brute-force method of enumerating and testing all possibilities. This strategy is computationally intractable, however, since the number of (distinct) (sub)formulæ grows exponentially with the number of events included in the model. Therefore, we need to exploit certain properties of the $\rhd$ relation whenever possible, and trade expressivity for complexity in other cases, as explained below.

Note that *singleton* and *co-occurrence* ($\wedge$) types of patterns are amenable to *compositional reasoning*: if $i_1 \wedge \ldots \wedge i_k \rhd j$ then, for any $p = 1, \ldots, k$, $i_p \rhd j$. This observation leads to the following straightforward strategy of evaluating every conjunctive (and henceforth singleton) relation using a pairwise-test for the selectivity relation (see Figure §2.2).

Unfortunately, it is easy to see that this reasoning fails to generalize for CNF patterns: e.g., when the pattern contains disjunctive operators ($\vee$). As an example, consider pattern $a \vee b \rhd c$, in a cancer where $\{a, \neg b\}$ progression to $c$ is more prevalent than $\{\neg a, b\}$ and $\{a, b\}$. In this case, considering sub-formulas only we might find $a \rhd c$ but miss $b \rhd c$ because the probability of mutated $b$ is smaller than that of $c$, thus invalidating condition (1) of relation $\rhd$. Notice that in extreme situations, when the data is very noisy, the algorithm may even "invert" the selectivity relation to $c \rhd b$.

This difficulty is not a peculiarity of my framework, but rather intrinsic to the problem of extracting complex "causal networks" (see, [151, 150, 98]). To handle this situation, CAPRI adapts a strategy that trades complexity for expressivity: the resulting inference procedure, Algorithm §2, can be executed in two modes: unsupervised and supervised. In the former, inferred patterns of confluent progressions are constrained to co-occurrence types of relations, in the latter CAPRI can test more complex patterns, i.e., disjunctive or "mutual exclusive" ones, provided they are given as prior hypotheses.

---

[1]Theoretically, this assumption - common to other Bayesian learning problems - is necessary to prove CAPRI's ability to extract the exact model in the optimal case of infinite samples. Practically, as all *relevant* events are hardly selectable a priori and sample size is finite, further statistics can be used to select the most relevant driver alterations – see also Section §4.3. Nonetheless, CAPRI can provide significant results even if this assumption is not or cannot be verified.

In both cases, CAPRI's complexity – studied in next sections – is quadratic both in the number of events and hypotheses.

**Data Input (Step 1).**  CAPRI (sett, Algorithm §2) requires an input set $G$ of $n$ events, i.e., genomic alterations, and $m$ cross-sectional samples, represented as a dataset in an $m \times n$ binary matrix $D$, in which an entry $D_{i,j} = 1$ if the event $j$ was observed in sample $i$, and 0 otherwise. Assumption $iv$) is satisfied when all columns in $D$ differ - i.e., the alteration profiles yield different observations.

Optionally, a set of $k$ input hypotheses $\Phi = \{\varphi_1 \rhd e_1, \dots, \varphi_k \rhd e_k\}$, where each $\varphi_i$ is a well-formed[2] CNF formula. Note that we advise that the algorithm be used in the following regime [3]: $k + n \ll m$.

**Data Preprocessing (Lifting, step 2).**  When input hypotheses are provided (e.g., by a domain expert), CAPRI first performs a *lifting operation* over $D$ to permit direct inference of complex selectivity relations over a joint representation, which involve input events as well as the hypotheses. Lifting operation evaluates each input CNF formula – for all input hypotheses in $\Phi$ – and outputs a augmented matrix $D(\Phi)$ to be processed further as in step 1. As an example, consider hypothesis $a \oplus b \rhd c$ augmented input matrix $D$ is:

$$D(\Phi) = \begin{bmatrix} a & b & c & a \oplus b \rhd c \\ \hline 1 & 1 & 1 & 1 \oplus 1 = \mathbf{0} \\ 1 & 0 & 1 & 1 \oplus 0 = \mathbf{1} \\ 0 & 1 & 0 & 0 \oplus 1 = \mathbf{1} \\ 1 & 0 & 1 & 1 \oplus 0 = \mathbf{1} \end{bmatrix}.$$

Note that the first row (profile $\{a, b, c\}$) contradicts the hypothesis, while all other rows support it.

**Selectivity Topology (steps 3, 4, 5).**  We exploit a compositional approach to test CNF hypotheses as follows: the disjunctive relations are grouped, and treated as if they were individual objects in $G$. For example, when a formula $\varphi \rhd d$ where $\varphi = (a \vee b) \wedge c$ is considered, we assess $\varphi \rhd d$ as whether $(a \vee b) \rhd d$ and $c \rhd d$ hold – with the proviso that we treat $(a \vee b)$ as an individual event. Formally, with clauses $(\varphi)$ we denote the disjunctive clauses in a CNF formula.

Nodes in the reconstruction are all input events together with all the disjunctive clauses of each input formula $\varphi$.

Edges in the reconstructed DAG are patterns that satisfy both conditions (1) and (2) of the selectivity relation $\rhd$. Formally, CAPRI includes an edge between two nodes $\varphi$ and $j$ only if both $\Gamma_{\varphi,j} = \mathcal{P}(\varphi) - \mathcal{P}(j)$ and $\Lambda_{\varphi,j} = \mathcal{P}(j \mid \varphi) - \mathcal{P}(j \mid \overline{\varphi})$ are strictly positive. Note that $\varphi$ can be both a disjunctive clause as well as a singleton event. A function $\pi(\cdot)$ assigns a parent to each node that is not an input formula. Note that this approach works efficiently by nature of the augmented representation of $D$. The reconstructed DAG contains all the true positive patterns, with respect to $\triangleright$, plus spurious instances of $\triangleright$ which CAPRI subsequently removes in step 6 (see §D for a proof of this statement).

---

[2]Formally, we require that $\varphi_i \not\sqsubseteq e_i$, where $\sqsubseteq$ represents the usual *syntactical* ordering relation among atomic events and formulas, and disallows for example $a \vee b \rhd a$.

[3]In the current biomedical setting, the number of samples ($m$) is usually in the hundreds, while number of possible mutations ($n$) and hypotheses ($k$), absent any pre-processing, could be large, thus violating the assumption; in these cases, we rely on various commonly used pre-preprocessing filters to limit $n$ to driver mutations, and $k$ to simple hypotheses involving the driver mutations. However, in the future as the number of samples increases, we envision a more agnostic application.

Note that $\mathcal{D}$ can be readily interpreted as a probabilistic graphical model, once it is augmented with a labeling function $\alpha : N \to [0,1]$, where $N$ is the set of nodes – i.e., the genetic alterations – such that $\alpha(i)$ is the *independent probability* of observing mutation $i$ in a sample, whenever *all of its parent* mutations (i.e., $\pi(i)$) are observed (if any). Thus $\mathcal{D}$ induces a *distribution* of observing a subset of events in a set of samples (i.e., a probability of observing a certain *mutational profile* in a patient).

**Maximum Likelihood Fit (step 6).** As the selectivity relation provides only a *necessary* condition, we must filter out all of its *spurious instances* that might have been included in $\mathcal{D}$ (i.e., the possible *false positives*).

For any selectivity relation, spurious claims contribute to a reduction in the *likelihood-fit*[4] relative to true patterns. Thus, a standard maximum-likelihood fit can be used to select and prune the selectivity DAG (including a *regularization term* to avoid overfitting[5]). Here, we adopt the *Bayesian Information Criterion* (BIC), which implements *Occam's razor* by combining log-likelihood fit with a *penalty criterion* proportional to the log of the DAG size via *Schwarz Information Criterion* (see [167]). The BIC score is defined as follows.

$$\text{BIC}\left(\mathcal{D},\, D(\Phi)\right) = \mathcal{LL}\left(\mathcal{D},\, D(\Phi)\right) - \frac{\log m}{2}\dim(\mathcal{D}). \tag{4.2}$$

Here, $D(\Phi)$ is the augmented input matrix, $m$ denotes the number of samples and $\dim(\mathcal{D})$ is the number of parameters in the model $\mathcal{D}$. Because, in general, $\dim(\cdot)$ depends on the number of parents each node has, it is a good metric for model complexity. Moreover, since each edge added to $\mathcal{D}$ increases model complexity, the regularization term based on $\dim(\cdot)$ favors graphs with fewer edges and, more specifically, fewer parents for each node.

At the end of this step, $\mathcal{D}$ and the labeling function are modified accordingly, based on the result of BIC regularization. By collecting all the incoming edges in a node it is possible to extract the patterns, which have been selected by CAPRI as the positive ones.

---

[4]The maximum-likelihood estimation (MLE) is a method for estimating the parameters of a statistical model given data. In general, given a dataset and its underlying statistical model, the maximum likelihood estimation aim at selecting the set of values of the model parameters (in the setting of this thesis, a set of arcs of a graphical model) that maximizes the likelihood function. Intuitively, this maximizes the *agreement* of the selected model given the observed data. See also §B.

[5]In principle other regularisation strategies common to Bayesian learning could be used, e.g., Akaike information criterion (see [23] and references therein). In this task, we prefer to work with BIC which, in general, trades model complexity to reduce false positives rate.

---

**Algorithm 2:** *CAncer PRogression wenference* (CAPRI)

1: **Input:** A set of events $G = \{g_1, \ldots, g_n\}$, a matrix $D \in \{0,1\}^{m \times n}$ and $k$ CNF causal claims
$\Phi = \{\varphi_1 \rhd e_1, \ldots, \varphi_k \rhd e_k\}$ where, for any $i$, $e_i \not\sqsubseteq \varphi_i$ and $e_i \in G$;

2: [*Lifting*] Define the *lifting of $D$ to $D(\Phi)$* as the augmented matrix

$$D(\Phi) = \left[ \begin{array}{ccc|ccc} D_{1,1} & \ldots & D_{1,n} & \varphi_1(D_{1,\cdot}) & \ldots & \varphi_k(D_{1,\cdot}) \\ \vdots & \ddots & \vdots & \vdots & \ddots & \vdots \\ D_{m,1} & \ldots & D_{m,n} & \varphi_1(D_{m,\cdot}) & \ldots & \varphi_k(D_{m,\cdot}) \end{array} \right].$$

by adding a column for each $\varphi_i \rhd c_i \in \Phi$, with $\varphi_i$ evaluated row-by-row. Define then the coefficients
$\Gamma_{i,j} = \mathcal{P}(i) - \mathcal{P}(j)$ and $\Lambda_{i,j} = \mathcal{P}(j \mid i) - \mathcal{P}(j \mid \hat{i})$ pairwise over $D(\Phi)$;

3: [*DAG nodes*] Define the set of nodes $N = G \cup \left( \bigcup_{\varphi_i} \text{clauses}\,(\varphi_i) \right)$ which contains both input events
and the disjunctive clauses in every input formula of $\Phi$.

4: [*DAG edges*] Define a parent function $\pi$ where $\pi(j \notin G) = \emptyset$ – avoid edges incoming in a formula [6]
and

$$\pi(j \in G) = \{i \in G \mid \Gamma_{i,j}, \Lambda_{i,j} > 0\}$$
$$\cup \{\text{clauses}\,(\varphi) \mid \Gamma_{\varphi,j}, \Lambda_{\varphi,j} > 0,\ \varphi \rhd j \in \Phi\}.$$

Set the DAG to $\mathcal{D} = (N, \pi)$.

5: [*DAG labeling*] Define the labeling $\alpha$ as follows

$$\alpha(j) = \begin{cases} \mathcal{P}(j), & \text{if } \pi(j) = \emptyset \text{ and } j \in G; \\ \mathcal{P}(j \mid i_1 \wedge \ldots \wedge i_n), & \text{if } \pi(j) = \{i_1, \ldots, i_n\}. \end{cases}$$

6: [*Likelihood fit*] Filter out all spurious causes from $\mathcal{D}$ by likelihood fit with the regularization
*BIC* score and set $\alpha(j) = 0$ for each removed edge.

7: **Output:** the DAG $\mathcal{D}$ and $\alpha$;

---

**Inference Confidence: Bootstrap and Statistical Testing.** To infer confidence
intervals of the selectivity relations $\rhd$, CAPRI employs *bootstrap with rejection resampling* by estimating a distribution of the marginal and joint probabilities as follows.
For each event, ($i$) CAPRI samples with repetitions rows from the input matrix $D$
(bootstrapped dataset), ($ii$) CAPRI next estimates the distributions from the observed
probabilities, and finally, ($iii$) CAPRI rejects values which do not satisfy $0 < \mathcal{P}(i) < 1$
and $\mathcal{P}(i \mid j) < 1 \vee \mathcal{P}(j \mid i) < 1$, and iterates restarting from ($i$). We stop when we
have, for each distribution, at least $K$ values (in this case $K = 100$). Any inequality
(i.e., checking temporal priority and probability raising) is estimated using the non-parametric Mann-Whitney U test[7] with $p$-values set to 0.05. We compute confidence $p$-values for both temporal priority and probability raising using this test, which need not assume Gaussian distributions for the populations.

---

[6]Although CAPRI is equipped with bootstrap testing it is still possible to encounter various degenerate situations. In particular, for some pair of events it could be that temporal priority cannot be satisfactorily resolved, i.e. there is no significant *p*-value for any edge orientation. Thus, loops might be present in the inferred prima facie topology. Nonetheless, some of these could be still disentangled by probability raising, while some might remain, albeit rarely. To remove such edges we suggest to proceed as follows: ($i$) sort these edges according to their *p*-value (considering both temporal priority and probability raising), ($ii$) scan the sorted list in decreasing order of confidence, ($iii$) remove an edge if it forms a loop.

Once a DAG $\mathcal{D}$ is inferred both *parametric and non-parametric bootstrapping methods* can be used to assign a confidence level to its respective pattern and to the overall model. Essentially, these tests consist of using the reconstructed model (in the parametric case), or the probabilities observed in the dataset (in the non-parametric case) to generate new synthetic datasets, which are then reused to reconstruct the progressions (see, e.g., [45] for an overview of these methods). The confidence is estimated by the number of times the DAG or any instance of $\triangleright$ is reconstructed from the generated data.

**Complexity, Correctness and Expressivity.** CAPRI has the following asymptotic complexity (see §D): (*i*) Without input hypotheses the execution is self-contained and polynomial in the size of $D$. (*ii*) In addition to the above cost, CAPRI tests input hypotheses of $\Phi$ at a polynomial cost in the size of $|\Phi|$. In this case, however, its complexity may range over many orders of magnitude depending on the structural complexity of the input set $\Phi$ consisting of hypotheses. An empirical analysis of the execution time of CAPRI and the competing techniques on synthetic datasets is provided in §D.

CAPRI is a *sound and complete* algorithm, and its expressivity in terms of the inferred patterns is proportional to the hypothesis set $\Phi$ which, in turn, determines the complexity of the algorithm. With a proper set of input hypothesis, CAPRI can infer all (and only) the true patterns from the data, filtering out all the spurious ones (Theorem 2, §D). Without hypotheses, besides singleton and co-occurrence, no other patterns can be inferred (see Figure §2.2). Also, some of these claims might be spurious in general for more complex (and unverified) CNF formula (see §D).

## 4.3 Results and discussion

To determine CAPRI's relative accuracy (true-positives and false-negatives) and performance compared to the state-of-the-art techniques for *network inference*, we performed extensive *simulation experiments*. From a list of potential competitors of CAPRI, we selected: *Incremental Association Markov Blanket* (IAMB, [179]), the *PC algorithm* (see [171]), *Bayesian Information Criterion* (BIC, [167]), *Bayesian Dirichlet with likelihood equivalence* (BDE, [79]) *Conjunctive Bayesian Networks* (CBN, [65]) and *Cancer Progression Inference with Single Edges* (CAPRESE, [117]). These algorithms constitute a rich landscape of structural methods (IAMB and PC), likelihood scores (BIC and BDE) and hybrid approaches (CBN and CAPRESE).

Also, we applied CAPRI to the analysis of an atypical Chronic Myeloid Leukemia dataset of somatic mutations  with data based onsee [154].

---

[7]The Mann-Whitney U test is a rank-based non-parametric statistical hypothesis test that can be used as an alternative to the Student's t-test and is particularly useful if data are not normally distributed.

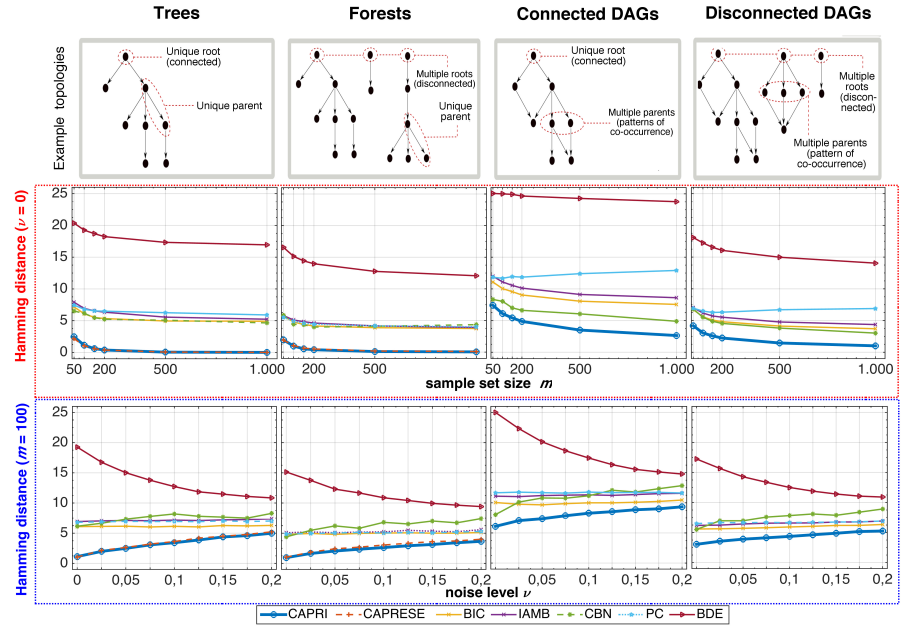

Figure 4.2: **Comparative Study**. Performance and accuracy of CAPRI (unsupervised execution) and other algorithms, IAMB, PC, BIC, BDE, CBN and CAPRESE, were compared using synthetic datasets sampled by a large number of randomly parametrized progression models – *trees*, *forests*, *connected* and *disconnected DAGs*, which capture different aspects of confluent, branched and heterogenous cancer progressions. For each of those, 100 models with $n = 10$ events were created and 10 distinct datasets were sampled by each model. Datasets vary by number of samples ($m$) and level of noise in the data ($\nu$) – see the Supplementary Information file for details. (*Red box*) Average *Hamming distance* (HD) – with 1000 runs – between the reconstructed and the generative model, as a function of dataset size ($m \in \{50, 100, 150, 200, 500, 1000\}$), when data contain no noise ($\nu = 0$). The lower the HD, the smaller is the total rate of mis-inferred selectivity relations among events. (*Blue box*) The same is shown for a fixed sample set size $m = 100$ as a function of noise level in the data ($\nu \in \{0, 0.025, 0.05, \cdots, 0.2\}$) so as to account for input *false positives* and *negatives*. Seesee §D for more extensive results on precision and recall scores and also including additional combinations of noise and samples as well as experimental settings.

### 4.3.1 Synthetic data

We performed extensive tests on a large number of *synthetic datasets* generated by randomly parametrized progression models with distinct key features, such as the presence/absence of: (1) *branches*, (2) *confluences with patterns of co-occurrence*, (3) *independent progressions* (i.e., composed of disjoint sub-models involving distinct sets of events). Accordingly, we distinguish four classes of generative models with increasing complexity and the following features:

|     | trees | forests | connected DAGs | disconnected DAGs |
|-----|-------|---------|----------------|-------------------|
| (1) | ✓     | ✓       | ✓              | ✓                 |
| (2) | ✗     | ✗       | ✓              | ✓                 |
| (3) | ✗     | ✓       | ✗              | ✓                 |

The choice of these different type of topologies is not a mere technical exercise, but rather it is motivated, in this application of primary interest, by *heterogeneity of cancer cell types* and *possibility of multiple cells of origin*.

To account for *biological noise* and *experimental errors* in the data we introduce a parameter $\nu \in (0, 1)$ which represents the probability of each entry to be random in $D$, thus representing a *false positive* ($\epsilon_+$) and a *false negative* rate ($\epsilon_-$): $\epsilon_+ = \epsilon_- = \nu/2$. The noise level complicates the inference problem, since samples generated from such topologies will likely contain sets of mutations that are correlated but causally irrelevant.

To have reliable statistics in all the tests, 100 distinct progression models per topology are generated and, for each model, for every chosen combination of sample set size $m$ and noise rate $\nu$, 10 different datasets are sampled (see §D for my synthetic data generation methods).

Algorithmic performance was evaluated using the metrics *Hamming distance* (HD), *precision* and *recall*, as a function of dataset size, $\epsilon_+$ and $\epsilon_-$. HD measures the *structural similarity* among the reconstructed progression and the generative model in terms of the minimum-cost sequence of node edit operations (inclusion and exclusion) that transforms the reconstructed topology into the generative one[8]. Precision and recall are defined as follows: *precision* = TP/(TP + FP) and *recall* = TP/(TP + FN), where TP are the *true positives* (number of correctly inferred true patterns), FP are the *false positives* (number of spurious patterns inferred) and FN are the *false negatives* (number of true patterns that are *not* inferred). The closer both precision and recall are to 1, the better.

In Figure §4.2 we show the performance of CAPRI and of the competing techniques, in terms of Hamming distance, on datasets generated from models with 10 events and all the four different topologies. In particular, we show the performance: (*i*) in the case of noise-free datasets, i.e., $\nu = 0$ and different values of the sample set size $m$ and (*ii*) in the case of a fixed sample set size, $m = 100$ (size that is likely to be found in currently available cancer databases, such as TCGA (see [136])) and different values of the noise rate $\nu$. As is evident from Figure §4.2 CAPRI outperforms all the competing techniques with respect to all the topologies and all the possible combinations of noise rate and sample set size, in terms of average Hamming distance (with the only exception of CAPRESE in the case of tree and forests, which displays a behavior closer to CAPRI's). The analyses on precision and recall display consistent results (see §D). In other words, we demonstrate on the basis of extensive synthetic tests that CAPRI requires a much lower number of samples than the other techniques in order to converge to the real generative model and also that it is much more robust even in the presence of significant amount of noise in the data, irrespective of the underlying topology.

---

[8]This measure corresponds to the sum of false positives and false negative and, for a set of $n$ events, is bounded above by $n(n-1)$ when the reconstructed topology contains all the false negatives and positives.

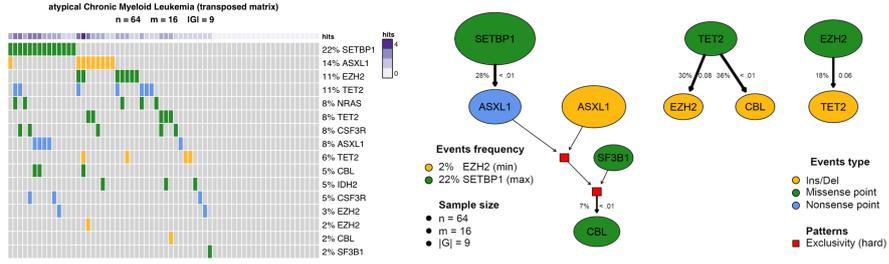

Figure 4.3: **Atypical Chronic Myeloid Leukemia**. *(left)* Mutational profiles of $n = 64$ ACML patients - exome sequencing in [154] - with alterations in $|G| = 9$ genes with either mutation frequency $> 5\%$ or belonging to an hypothesis inputed to CAPRI (§D). Mutation types are classified as *nonsense point*, *missense point* and *insertion/deletions*, yielding $m = 16$ input events. Purple annotations report the frequency of mutations per sample. *(right)* Progression model inferred by CAPRI in supervised mode. Node size is proportional to the marginal probability of each event, edge thickness to the confidence estimated with 1000 non-parametric bootstrap iterations (numbers shown leftmost of every edge). The p-value of the hypergeometric test is displayed too. Hard exclusivity patterns inputed to CAPRI are indicated as red squares. Events without inward/outward edges are not shown.

See §D for a more complete description of the performance evaluation for all the analyzed combinations of parameters. There, we have shown that CAPRI is highly effective when the co-occurrence constraint on confluences is relaxed to *disjunctive* patterns, *even if no input hypotheses are provided*, i.e., $\Phi = \emptyset$. This result hints at CAPRI's robustness to infer patterns with imperfect regularities. Finally, we also show that CAPRI is effective in inferring synthetic lethality relations in this case using the operator $\oplus$ as introduced in §2; when a combination of mutations in two or more genes leads to cell death, while separately, the mutations are viable. In this case, candidate relations are directly input as $\Phi$.

### 4.3.2   Atypical Chronic Myeloid Leukemia (aCML)

As a case study, we applied CAPRI to the mutational profiles of 64 ACML patients described in [154]. Through exome sequencing, the authors identify a recurring *missense point mutation* in the *SET-binding protein 1* (SETBP1) gene as a novel ACML marker.

Among all the genes present in the dataset by Piazza *et al.*, we selected those either (*i*) mutated - considered any mutation type - in at least 5% of the input samples (3 patients), or (*ii*) hypothesised to be part of a functional ACML progression pattern in the literature [9]. The input dataset with selected events is shown in Figure §4.3; notice that somatic mutations are categorised as *indel*, *missense point* and *nonsense point* as in [154]. In Figure §4.3 we show the model reconstructed by CAPRI (supervised mode, execution time ≈ 5 seconds) on this dataset, with confidence assessed via 1000 non-parametric bootstrap iterations. The model highlights several non trivial selectivity relations involving genomic events relevant to ACML development.

First, CAPRI predicts a progression involving mutations in SETBP1, ASXL1 and CBL, consistently with the recent study by [126], in which these genes were shown to be highly correlated and possibly functioning in a synergistic manner for ACML progression. Specifically, CAPRI predicts a selective advantage relation between missense point mutations in SETBP1 and nonsense point mutations in ASXL1. This is in line with recent evidence from [89] suggesting that SETBP1 mutations are enriched among ASXL1-mutated *myelodysplastic syndrome* (MDS) patients, and *in-vivo* experiments point to a driver role of SETBP1 for that leukemic progression. Interestingly, my model seems also to suggest a different role of ASXL1 *missense* and *nonsense* mutation types in the progression, yet more extensive studies (e.g., prospective or systems biology explanation) are needed to corroborate this hypothesis.

Among the hypotheses given as input to CAPRI, the algorithm seems to suggest that the exclusivity pattern among ASXL1 and SF3B1 mutations selects for CBL missense point mutations. The role of the ASXL1/SF3B1 exclusivity pattern is consistent with the study of [115] which shows that, on a cohort of 479 MDS patients, mutations in SF3B1are inversely related to ASXL1 mutations.

Also, in [2] it was recently shown that ASXL1 mutations, in patients with MDS, *myeloproliferative neoplasms* (MPN) and *acute myeloid leukemia*, most commonly occur as nonsense and insertion/deletion in a clustered region adjacent to the highly conserved *PHD* domain (see [61]) and that mutations of any type eventually result in a loss of ASXL1 expression. This observation is consistent with the exclusivity pattern among ASXL1 mutations in the reconstructed model, possibly suggesting alternative trajectories of somatic evolution for ACML (involving either ASXL1 nonsense or indel mutations).

Finally, CAPRI predicts selective advantage relations among TET2 and EZH2 missense point and indel mutations. Even though the limited sample size does not allow to draw definitive conclusions on the ordering of such alterations, we can hypothesize that they may play a synergistic role in ACML progression. Indeed, [133] suggests that the concurrent loss of EZH2 and TET2 might cooperate in the pathogenesis of myelodysplastic disorders, by accelerating the overall tumor development, with respect to both MDSs and *overlap disorders* (MDS/MPN).

---

[9]Two *hard exclusivity* patterns - i.e., mutual exclusivity with "xor" - were tested, involving the mutations of: (*i*) genes ASXL1 and SF3B1 (see [115]), which is present in the inferred progression model in Figure §4.3, and (*ii*) genes TET2 and IDH2 (see [53]). The syntax in which the patterns are expressed is in §D.

# CHAPTER 5

## AN R PACKAGE FOR TRANSLATIONAL ONCOLOGY

TRONCO is an R package for TRanslational ONCOlogy which provides a series of software utilities to support the user in each step of the pipeline described in this thesis (see Chapters §2, §3 and §4), i.e., from data import, through data visualization and, finally, to the inference of cancer progression models. Specifically, in the current version, TRONCO implements CAPRESE and CAPRI algorithms for cancer progression inference, which we extensively described in Chapter §3 and §4. As a reference, see [33, 5].

The core of the two algorithms is a simple quadratic loop[1] that *prunes* arcs from an initially totally connected graph. Each pruning decision is based on the application of Suppes' *probabilistic causation* criteria.

The pseudocode of the two implemented algorithms along with the procedure to evaluate the confidence of the arcs by bootstrap is summarized in Algorithms §3, §4, §5 and §6, which depict the data preparation step, the CAPRESE and CAPRI algorithms and finally the optional *bootstrap* step.

## 5.1    Package implementation

In this Section we will review the structure and implementation of the TRONCO package. For the sake of clarity, we will structure the description through the following functionalities that are implemented in the package.

- **Data import.** Functions for the importation of data both from flat files (e.g., MAF, GISTIC) and from Web querying (e.g., cbioportal).

- **Data exploration and correctness.** Functions for the exploration and visualization of the imported data.

---

[1]For CAPRI the size of the input actually depends on the structural complexity of the input "patterns", i.e., of the boolean formulæ employed in the "lifting operation"

---

**Algorithm 3:** TRONCO Data Import and Preprocessing

**Input**: a data set containing `MAF` or `GISTIC` scores, e.g., as obtained from cBio portal ([24, 15]).

**Result**: a data structure containing boolean flags for "events", relative frequencies and other metadata.

**1** From the dataset (depending on the data format) derive a Boolean matrix $M$, where each entry $\langle i, j \rangle$ is `true` if event $i$ is "present" in sample/patient $j$.

**2 forall the** *events $e$* **do**

**3**  $\quad$ Compute the *frequency* of the event $e$ in the dataset and save it in a map $F$.

**4**  $\quad$ Compute the *joint probability* of co-occurrence of pair of events in the dataset and save it in a map $C$.

**5 end**

**6 return** *A data structure comprising the Boolean matrix $M$, the maps $F$ and $C$ and other metadata.*

---

- **Data editing.** Functions for the preprocessing of the data in order to tidy them.

- **External utilities.** Functions for the interaction with external tools for the analysis of cancer subtypes or groups of mutually exclusive genes.

- **Inference algorithms.** In the current version of TRONCO, the CAPRESE and CAPRI algorithms are provided in a polinomial implementation.

- **Confidence estimation.** Functions for the statistical estimation of the confidence of the reconstructed models.

- **Visualization.** Functions for the visualization of both the input data and the results of the inference and of the confidence estimation.

## Data import

The starting point of TRONCO analysis pipeline, is a dataset of genomics alterations (i.e., somatic mutations and copy number variations) which need to be imported as a TRONCO compliant data structure, i.e., a list R structure containing the required data both for the inference and the visualization. The data import functions take as input such genomic data and from them create a TRONCO compliant data structure.

The core of the data import functionalities from flat files is the function called `import.genotypes(geno, event.type = ``variant'', color = ``Darkgreen'')`.

This function imports a matrix of 0/1 alterations as a TRONCO compliant dataset. The input *"geno"* can be either a dataframe or a file name. In any case the dataframe or the table stored in the file must have a column for each altered gene and a rows for each sample. Colnames will be used to determine gene names, if data are loaded from file the first column will be assigned as rownames.

---

**Algorithm 4:** Pseudocode of the CAPRESE algorithm

**Input**: a dataset of $n$ events, i.e., genomic alterations, and $m$ samples packed in
      a datastructure obtained from Algorithm 3.

**Result**: a *tree model* representing all the relations of selective advantage.

PRUNING BASED ON SUPPES' CRITERIA.

**1** Let $G \leftarrow$ a complete directed graph over the vertices $n$.

**2** **forall the** *arcs $(a, b)$ in $G$* **do**

**3**      Compute a score $S(\cdot)$ for the nodes $a$ and $b$ based on Suppes' criteria.

**4**      **if** *Suppes' criteria are not met* **then**

**5**          Remove the arc $(a, b)$ from $G$

**6**      **else if** *$S(a) > S(b)$ and $S(a) > 0$* **then**

**7**          Keep $(a, b)$ as edge.            I.E., SELECT $a$ AS "CANDIDATE PARENT".

**8**      **else if** *$S(b) > S(a)$ and $S(b) > 0$* **then**

**9**          Keep $(a, b)$ as edge.            I.E., SELECT $b$ AS "CANDIDATE PARENT".

**10** **end**

FIT OF THE *Prima Facie* DIRECTED ACYCLIC GRAPH TO THE BEST TREE MODEL.

**11** Let $\mathcal{T} \leftarrow$ the best tree model obtained by Edmonds' algorithm (see [43]).

**12** **return** *The resulting* tree *model $\mathcal{T}$.*

---

Besides this function, TRONCO implements data import from other file format such as MAF and GISTIC files as wrappers of the function `import.genotypes`. Specifically, the function `import.MAF(file, sep = ''\t'', is.TCGA = TRUE)` imports mutation profiles from a Manual Annotation Format (MAF) file. All mutations are aggregated as a unique event type labeled "Mutation" and are assigned a color accordingly to the default of function `import.genotypes`. If the input is in the TCGA MAF file format, the function also checks for multiple samples per patient and a warning is raised if any are found. Furthermore, the function `import.GISTIC(x)` transforms GISTIC scores for CNAs in a TRONCO compliant object. The input can be a matrix, with columns for each altered gene and rows for each sample; in this case colnames or rownames mut be provided. If the input is a character an attempt to load a table from file is performed. In this case the input table format should be constitent with TCGA data for focal CNA; there should hence be: one column for each sample, one row for each gene, a column Hugo_Symbol with every gene name and a column Entrez_Gene_Id with every genes Entrez ID. A valid GISTIC score should be any value of: "Homozygous Loss" (-2), "Heterozygous Loss" (-1), "Low-level Gain" (+1), "High-level Gain" (+2).

Finally, TRONCO also provides utilities for the query of genomic data from cbioportal. This is implemented in the function `cbio.query(cbio.study = NA, cbio.dataset = NA, cbio.profile = NA, genes)` which is a wrapper for the *CGDS* package. This can work either automatically, if one sets `cbio.study`, `cbio.dataset` and `cbio.profile`, or interactively. A list of genes to query with less than 900 entries should be provided.

---

**Algorithm 5:** Pseudocode of the CAPRI algorithm

**Input**: a dataset of $n$ variables, i.e., genomic alterations or patterns, and $m$ samples.

**Result**: a graphical model representing all the relations of "selective advantage".

PRUNING BASED ON THE SUPPES' CRITERIA

1 Let $G \leftarrow$ a directed graph over the vertices $n$
2 **forall the** *arcs* $(a, b) \in G$ **do**
3     Compute a score $S(\cdot)$ for the nodes $a$ and $b$ in terms of Suppes' criteria.
4     Remove the arc $(a, b)$ if Suppes' criteria are not met.
5 **end**

LIKELIHOOD FIT ON THE PRIMA FACIE DIRECTED ACYCLIC GRAPH

6 Let $\mathcal{M} \leftarrow$ the subset of the remaining arcs $\in G$, that maximize the log-likelihood of the model, computed as: $LL(D \mid \mathcal{M}) - ((\log m)/2)\,\mathsf{dim}(\mathcal{M})$, where $D$ denotes the input data, $m$ denotes the number of samples, and $\mathsf{dim}(\mathcal{M})$ denotes the number of parameters in $\mathcal{M}$ (see [101]).
7 **return** *The resulting* graphical *model* $\mathcal{M}$.

---

This function returns a list with two dataframes: the required genetic profile along with clinical data for the `cbio.study`. The output is also saved to disk as Rdata file. See also the cbioportal page at **http://www.cbioportal.org**.

The function `show(x, view = 10)` prints to console a short report of a dataset "x", which should be a TRONCO compliant dataset.

All the functions described in the following sections will assume as input a TRONCO compliant data structure.

## Data exploration and correctness

TRONCO provides a series of function to explore the imported data and the inferred models. All these functions are named with the "as." prefix.

Concerning the imported data, the function `as.genotypes(x)` returns the 0/1 genotypes matrix. This function can be used in combination with the function `keysToNames(x, matrix)` to translate colnames to event names given the input matrix with colnames or rownames which represent genotypes keys. Also, functions to get the list of genes, events (i.e., each columns in the genotypes matrix, it differs from genes as the same genes of different types are considered different events), alterations (i.e., genes of different types are merged as 1 unique event), samples (i.e., patients or also single cells) and alteration types. See the functions:

```
as.genes(x, types = NA)
as.events(x, genes = NA, types = NA)
as.alterations(x, new.type = "Alteration", new.color = "khaki")
```

---

**Algorithm 6:** Bootstrap Procedure

---

   **Input**: a model $\mathcal{T}$ obtained from CAPRESE or a model $\mathcal{M}$ obtained from
            CAPRI, and the initial dataset.
   **Result**: the *confidence* in the inferred arcs.

**1** Let *counter* $\leftarrow 0$
**2** Let *nboot* $\leftarrow$ the number of bootstrap sampling to be performed.
**3** **while** *counter* $<$ *nboot* **do**
**4**     Create a new dataset for the inference by random sampling of the input data.
**5**     Perform the reconstruction on the sampled dataset and save the results.
**6**     *counter = counter + 1*
**7** **end**
**8** Evaluate the confidence in the reconstruction by counting the number of times
   any arc is inferred in the sampled datasets.
**9** **return** *The inferred model $\mathcal{T}$ or $\mathcal{M}$ augmented with an estimated confidence for
   each arc.*

---

```
as.samples(x)
as.types(x, genes = NA))
```

Functions of this kind are also implemented to explore the results such as the models that have been inferred (see `as.models(x, models = names(x$model))`), the reconstructions (see the functions `as.adj.matrix(x, events = as.events(x), models = names(x$model),type = "fit")`), the considered patters (see `as.patterns(x)`) and the confidence (see `as.confidence(x, conf)`).

Similarly, a set of function to extract the cardinality of the compliant TRONCO data structure are defined (see `nevents(x, genes = NA, types = NA)`, `ngenes(x, types = NA)`, `npatterns(x)`, `nsamples(x)` and `ntypes(x)`).

Furthermore, functions to asses the correctness of the inputes are also provided. With function `is.compliant(x, err.fun = "[ERR]", stage = !(all(is.null(x$stages)) || all(is.na(x$stages))))` it is possible to verify the TRONCO data structure to be compliant with the standards. The function `consolidate.data(x, print = FALSE)` verifies if the input data are consolidated, i.e., if there are events with 0 or 1 probability or indistinguishable in terms of observations. Any indistinguishable event is returned by the function `duplicates(x)`.

Finally, TCGA specific functions are provided. `TCGA.multiple.samples(x)` checks if there are multiple sample in the input, while `TCGA.remove.multiple.samples(x)` remove them accordingly to TCGA barcodes naming.

## Data editing

TRONCO provides a wide range of editing functions. We will describe some of them in the following, for a technical description we refer to the manual.

**Removing and merging.** A set of functions to remove items from the data is provided; such functions are named with the "delete." prefix. Specifically, using these functions it is possible to remove genes (`delete.gene(x, gene)`, events (`delete.event(x, gene, type)`), samples (`delete.samples(x, samples)`), types (`delete.type(x, type)`) as well as patterns (`delete.pattern(x, pattern)`) and inferred progression models (`delete.model(x)`). At the same time, functions to merge events (`merge.events(x, ..., new.event, new.type, event.color)`) and alterations types (`merge.types(x, ..., new.type = "new.type", new.color = "khaki")`) are also provided.

**Binding.** The purpose of the binding functions is to combine different datasets. The function `ebind(...)` combines events from one or more datasets, whose events need be defined over the same set of samples while the function `sbind(...)` combines samples from one or more datasets, whose samples need to be defined over the same set of events. Samples and events of two dataset can also be intersected through the function `intersect.datasets(x, y, intersect.genomes = TRUE)`.

**Changing and renaming.** The two functions `rename.gene(x, old.name, new.name)` and `rename.type(x, old.name, new.name)` can be used respectively to rename genes or alterations types. The function `change.color(x, type, new.color)` can be used to change the color associated to the specified alteration type.

**Selecting and splitting.** Genomics data usually involves a vast number of genes, the most of which is not relevant for cancer development (i.e., such as passenger mutations). For this reason, TRONCO implements the function `events.selection(x, filter.freq = NA, filter.in.names = NA,filter.out.names = NA)` which allows the selection of a set of genes to be analyzed. The selection can be performed by frequency and gene symbols. The 0 probability events can are removed by the function `trim(x)`. Moreover, the functions `samples.selection(x, samples)` and `ssplit(x, clusters, idx = NA)` respectively filters a dataset based on selected samples id and splits the dataset into groups (i.e., groups). This latter function can be used to analyze specific subtypes within a tumor.

## External utilities

TRONCO permits the interaction with external tools to (*i*) reduce inter-tumor heterogeneity by cohort subtyping and (*ii*) detect fitness equivalent exclusive alterations. The first issue can be attacked by adopting clustering techniques to split the dataset in order to analyze each cluster subtype separately. Currently, TRONCO can export and inport data from [82] via the function `export.nbs.input(x, map_hugo_entrez, file = "tronco_to_nbs.mat")` and the previously described splitting functions.

To handle alterations with equivalent fitness, TRONCO implements the interaction with [7] through the functions `export.mutex(x, filename = "tronco_to_mutex", filepath = "./", label.mutation = "SNV", label.amplification = list("High-level Gain"),`

`label.deletion = list("Homozygous Loss"))` and `import.mutex.groups(file, fdr = 0.2, display = TRUE)`. Such exclusivity groups can then be further added as patters (see the next Section).

### Inference algorithms

In the current version of TRONCO are implemented the algorithms CAPRESE [117] and CAPRI [158].

**CAPRESE**   The CAPRESE algorithm [117] can be executed by the function

```
tronco.caprese(data, lambda = 0.5, do.estimation = FALSE, silent = FALSE)
```

with "data" being a compliant TRONCO data structure. The parameter "lambda" can be used to tune the shrinkage-alike estimator adopted by CAPRESE, with the default being 0.5 as suggested in [117].

**CAPRI**   The CAPRI algorithm [158] can be executed by the function

```
tronco.capri(data, command = "hc", regularization = c("bic", "aic"),
                  do.boot = TRUE, nboot = 100, pvalue = 0.05,
                  min.boot = 3, min.stat = TRUE, boot.seed = NULL,
                  do.estimation = FALSE, silent = FALSE)
```

with "data" being a TRONCO compliant data structure. The parameters "command" and "regularization" allows respectively to choose the heuristic search to be performed to fit the network and the regularizer to be used in the likelihood fit (see [158]). CAPRI can be also executed without the bootstrap preprocessing step by the parameter "do.boot"; this is discouraged, but can speed up the execution with big input datasets.

As discussed in [158], CAPRI constrains the seach space using Suppes' prima facie conditions which lead to a subset of possible valid selective advantage relations which are then evaluated by the likelihood fit. Although uncommon, it may so happen (especially when pattern are given as input) that such a resulting prima facie graphical structure may still contain cycles. When this happens, the cycles are removed through the heuristic algorithm implemented in `remove.cycles(adj.matrix, weights.temporal.priority, weights.matrix, not.ordered, hypotheses = NA, silent)`. The function takes as input a set of weights in term of confidence for any selective advantage valid edge, ranks all the valid edges in increasing confidence levels and, starting from the less confident, goes through each edge and remove the ones that can break the cycles.

**Patterns**   CAPRI allows for the input of patterns, i.e., group of events which express possible selective advantage relations. Such patters are given as input with the function `hypothesis.add(data, pattern.label, lifted.pattern, pattern.effect = "*", pattern.cause = "*")`. This function is wrapped by `hypothesis.add.homologous(x,`

`pattern.cause = "*", pattern.effect = "*", genes = as.genes(x), FUN = OR)`
and `hypothesis.add.group(x, FUN, group, pattern.cause = "*", pattern.effect = "*", dim.min = 2, dim.max = length(group), min.prob = 0)` which, respectively, allow addition of analogous patterns (i.e., patterns involving the same gene of different types) and patterns involving a specified group of genes. In the current version of TRONCO, the implemented possible patters are the boolean operators AND, OR and XOR (functions `AND(...)`, `OR(...)` and `XOR(...)`).

## Confidence estimation

To asses the confidence of any selectivity relations TRONCO implements non-parametric and statistical bootstrap. For the non-parametric bootstrap, each event row is uniformly sampled with repetitions from the input genotype and then, on such an input, the inference algorithms are performed. The assesment concludes after $K$ repetitions (e.g., $K = 100$). Similarly, for CAPRI, a statistical bootstrap is provided: in this case the input dataset is kept fixed, but different seeds for the statistical procedures are sampled (see, e.g., [46] for an overview of these methods). The bootstrap is implemented in the function `tronco.bootstrap(reconstruction, type = "non-parametric", nboot = 100, verbose = FALSE)` where "reconstruction" is a compliant object obtained by the inference by one of the implemented algorithms.

## Visualization

During the development of the TRONCO package, a lot of attention was paid to the visualization features which are crucial for the understanding of biological results. Shown below is an overview of the main features; for a detailed description of each function, please refer to the manual.

**OncoPrint.** OncoPrints are compact means of visualizing distinct genomic alterations, including somatic mutations, copy number alterations, and mRNA expression changes across a set of cases. They are extremely useful for visualizing gene set and pathway alterations across a set of cases, and for visually identifying trends, such as trends in mutual exclusivity or co-occurence between gene pairs within a gene set. Individual genes are represented as rows, and individual cases or patients are represented as columns. See http://www.cbioportal.org/. The function `oncoprint` provides such visualizations with a TRONCO compliant data structure as input. The function `oncoprint.cbio` exports the input for the cbioportl visualization, see http://www.cbioportal.org/public-portal/oncoprinter.jsp.

It is also possible to annotate a description (`annotate.description(x, label)`) and tumor stages (`annotate.stages(x, stages, match.TCGA.patients = FALSE)`) to any oncoprint.

**Reconstruction.** The inferred models can be displayed by the function `tronco.plot`. The features included in the plots are multiple, such as the choice of the regularizer(s) to show, editing font of nodes and edges, scaling nodes' size in terms of estimated marginal probabilities, annotating the pathway of each gene and displaying the estimated confidence of each edge. We refer to the manual for a detailed description.

**Reports.** Finally, report utilities are provided. The function `genes.table.report(x, name, dir = getwd(), maxrow = 33, font = 10, height = 11, width = 8.5, fill = "lightblue")` can be used to generate LaTeX code to be used as report, while `genes.table.plot(x, name, dir = getwd())` generates reports histograms.

## 5.2   Use case of TRONCO

In this Section, we will present a case study for the usage of the TRONCO package based on the work presented in [158]. A detailed discussion of the pipeline along with a complete analysis based on it is provided in [33, 5].

### Events selection

We will start by loading the TRONCO package in R along with an example *"dataset"* that comes within the package.

```
> library(TRONCO)
> data(aCML)
> hide.progress.bar <<- TRUE
```

We then use the function `show` to get a short summary of the aCML dataset that has just been loaded.

```
> show(aCML)

Description: CAPRI - Bionformatics aCML data.
Dataset: n=64, m=31, |G|=23.
Events (types): Ins/Del, Missense point, Nonsense Ins/Del, Nonsense point.
Colors (plot): darkgoldenrod1, forestgreen, cornflowerblue, coral.
Events (10 shown):
 gene 4 : Ins/Del TET2
 gene 5 : Ins/Del EZH2
 gene 6 : Ins/Del CBL
 gene 7 : Ins/Del ASXL1
 gene 29 : Missense point SETBP1
 gene 30 : Missense point NRAS
 gene 31 : Missense point KRAS
 gene 32 : Missense point TET2
 gene 33 : Missense point EZH2
 gene 34 : Missense point CBL
```

```
Genotypes (10 shown):
      gene 4 gene 5 gene 6 gene 7 gene 29 gene 30 gene 31 gene 32 gene 33 gene 34
patient 1    0      0      0      0      1       0       0       0       0      0
patient 2    0      0      0      0      1       0       0       0       0      1
patient 3    0      0      0      0      1       1       0       0       0      0
patient 4    0      0      0      0      1       0       0       0       0      1
patient 5    0      0      0      0      1       0       0       0       0      0
patient 6    0      0      0      0      1       0       0       0       0      0
```

Using the function `as.events`, we can have a look at the events in the dataset.

```
> as.events(aCML)

          type              event
gene 4    "Ins/Del"         "TET2"
gene 5    "Ins/Del"         "EZH2"
gene 6    "Ins/Del"         "CBL"
gene 7    "Ins/Del"         "ASXL1"
gene 29   "Missense point"  "SETBP1"
gene 30   "Missense point"  "NRAS"
gene 31   "Missense point"  "KRAS"
gene 32   "Missense point"  "TET2"
gene 33   "Missense point"  "EZH2"
gene 34   "Missense point"  "CBL"
gene 36   "Missense point"  "IDH2"
gene 39   "Missense point"  "SUZ12"
gene 40   "Missense point"  "SF3B1"
gene 44   "Missense point"  "JARID2"
gene 47   "Missense point"  "EED"
gene 48   "Missense point"  "DNMT3A"
gene 49   "Missense point"  "CEBPA"
gene 50   "Missense point"  "EPHB3"
gene 51   "Missense point"  "ETNK1"
gene 52   "Missense point"  "GATA2"
gene 53   "Missense point"  "IRAK4"
gene 54   "Missense point"  "MTA2"
gene 55   "Missense point"  "CSF3R"
gene 56   "Missense point"  "KIT"
gene 66   "Nonsense Ins/Del" "WT1"
gene 69   "Nonsense Ins/Del" "RUNX1"
gene 77   "Nonsense Ins/Del" "CEBPA"
gene 88   "Nonsense point"  "TET2"
gene 89   "Nonsense point"  "EZH2"
gene 91   "Nonsense point"  "ASXL1"
gene 111  "Nonsense point"  "CSF3R"
```

These events account for alterations in the following genes.

```
> as.genes(aCML)
```

```
 [1] "TET2"    "EZH2"    "CBL"     "ASXL1"  "SETBP1" "NRAS"    "KRAS"    "IDH2"
     "SUZ12"   "SF3B1"   "JARID2"  "EED"     "DNMT3A" "CEBPA"  "EPHB3"  "ETNK1"
     "GATA2"   "IRAK4"   "MTA2"    "CSF3R"
[21] "KIT"     "WT1"     "RUNX1"
```

Now we take a look at the alterations of only the gene SETBP1 across the samples.

```
> as.gene(aCML, genes='SETBP1')
```

```
           Missense point SETBP1
patient 1               1
patient 2               1
patient 3               1
patient 4               1
patient 5               1
patient 6               1
patient 7               1
patient 8               1
patient 9               1
patient 10              1
patient 11              1
patient 12              1
patient 13              1
patient 14              1
patient 15              0
patient 16              0
patient 17              0
patient 18              0
patient 19              0
patient 20              0
patient 21              0
patient 22              0
patient 23              0
patient 24              0
patient 25              0
patient 26              0
patient 27              0
patient 28              0
patient 29              0
patient 30              0
```

```
patient 31                    0
patient 32                    0
patient 33                    0
patient 34                    0
patient 35                    0
patient 36                    0
patient 37                    0
patient 38                    0
patient 39                    0
patient 40                    0
patient 41                    0
patient 42                    0
patient 43                    0
patient 44                    0
patient 45                    0
patient 46                    0
patient 47                    0
patient 48                    0
patient 49                    0
patient 50                    0
patient 51                    0
patient 52                    0
patient 53                    0
patient 54                    0
patient 55                    0
patient 56                    0
patient 57                    0
patient 58                    0
patient 59                    0
patient 60                    0
patient 61                    0
patient 62                    0
patient 63                    0
patient 64                    0
```

We consider a subset of all the genes in the dataset to be involved in patters based on the support we found in the literature. See [158] as a reference.

```
> gene.hypotheses = c('KRAS', 'NRAS', 'IDH1', 'IDH2', 'TET2', 'SF3B1', 'ASXL1')
```

Regardless from which types of mutations we include, we select only the genes which appear altered in at least 5% of the patients. Thus, we first transform the dataset into *"Alteration"* (i.e., by collapsing all the event types for the same gene), and then we consider only these events from the original dataset.

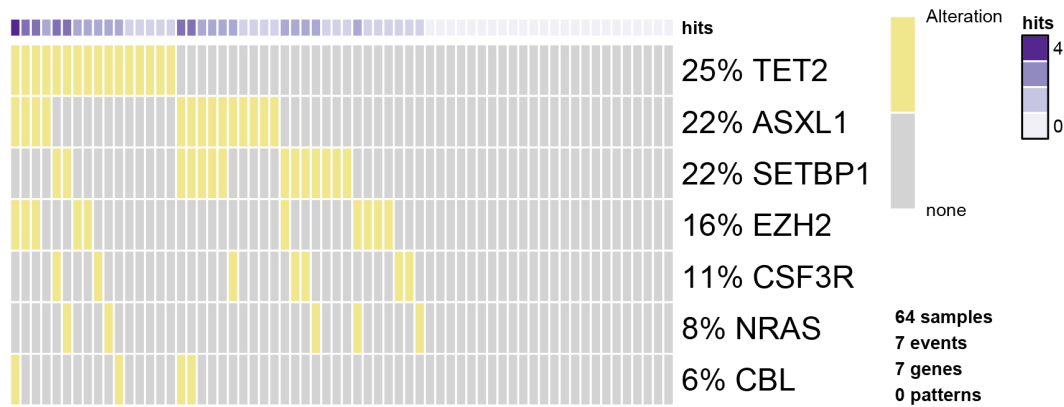

Figure 5.1: **Oncoprint function in** TRONCO. Result of the oncoprint function in TRONCO on the aCML dataset.

```
> alterations = events.selection(as.alterations(aCML), filter.freq = .05)

*** Aggregating events of type(s) Ins/Del, Missense point, Nonsense Ins/Del,
Nonsense point
in a unique event with label "Alteration".
Dropping event types Ins/Del, Missense point, Nonsense Ins/Del, Nonsense point for
23 genes.
*** Binding events for 2 datasets.
*** Events selection: #events=23, #types=1
    Filters freq|in|out = {TRUE, FALSE, FALSE}
Minimum event frequency: 0.05 (3 alterations out of 64 samples).
Selected 7 events.

Selected 7 events, returning.
```

We now show a plot of the selected genes. Note that this plot has no title as by default the function `events.selection` does not add any. The resulting figure is shonw in 5.1.

```
> dummy = oncoprint(alterations,font.row=12,cellheight=20,cellwidth=4)

*** Oncoprint for ""
with attributes: stage=FALSE, hits=TRUE
Sorting samples ordering to enhance exclusivity patterns.
```

## Adding Hypotheses

We now create the `dataset` to be used for the inference of the progression model. We consider the original dataset and from it we select all the genes whose mutations are occurring at least 5% of the times together with any gene involved in any hypothesis. To do so, we use the parameter `filter.in.names` as shown below.

```
> hypo = events.selection(aCML, filter.in.names=c(as.genes(alterations),
        gene.hypotheses))

*** Events selection: #events=31, #types=4 Filters freq|in|out =
{FALSE, TRUE, FALSE}
[filter.in] Genes hold: TET2, EZH2, CBL, ASXL1, SETBP1 ...  [10/14 found].
Selected 17 events, returning.
> hypo = annotate.description(hypo, 'CAPRI - Bioinformatics aCML data
(selected events)')
```

We show a new oncoprint of this latest dataset where we annotate the genes in `gene.hypotheses` in order to identify them 5.2. The sample names are also shown.

```
> dummy = oncoprint(hypo, gene.annot = list(priors= gene.hypotheses),
        sample.id = T, font.row=12, font.column=5, cellheight=20, cellwidth=4)

*** Oncoprint for "CAPRI - Bioinformatics aCML data (selected events)"
with attributes: stage=FALSE, hits=TRUE
Sorting samples ordering to enhance exclusivity patterns.
Annotating genes with RColorBrewer color palette Set1 .
```

We now also add the hypotheses that are described in CAPRI's manuscript. Hypothesis of hard exclusivity (XOR) for NRAS/KRAS events (Mutation). This hypothesis is tested against all the events in the dataset.

```
> hypo = hypothesis.add(hypo, 'NRAS xor KRAS', XOR('NRAS', 'KRAS'))
```

We then try to include also a soft exclusivity (OR) pattern but, since its *"signature"* is the same of the hard one just included, it will not be included. The code below is expected to result in an error.

```
> hypo = hypothesis.add(hypo, 'NRAS or KRAS',  OR('NRAS', 'KRAS'))

Error in hypothesis.add(hypo, "NRAS or KRAS", OR("NRAS", "KRAS")) :
  [ERR] Pattern duplicates Pattern NRAS xor KRAS.
```

To better highlight the perfect (hard) exclusivity among NRAS/KRAS mutations, one can examine further their alterations. See Figure 5.3.

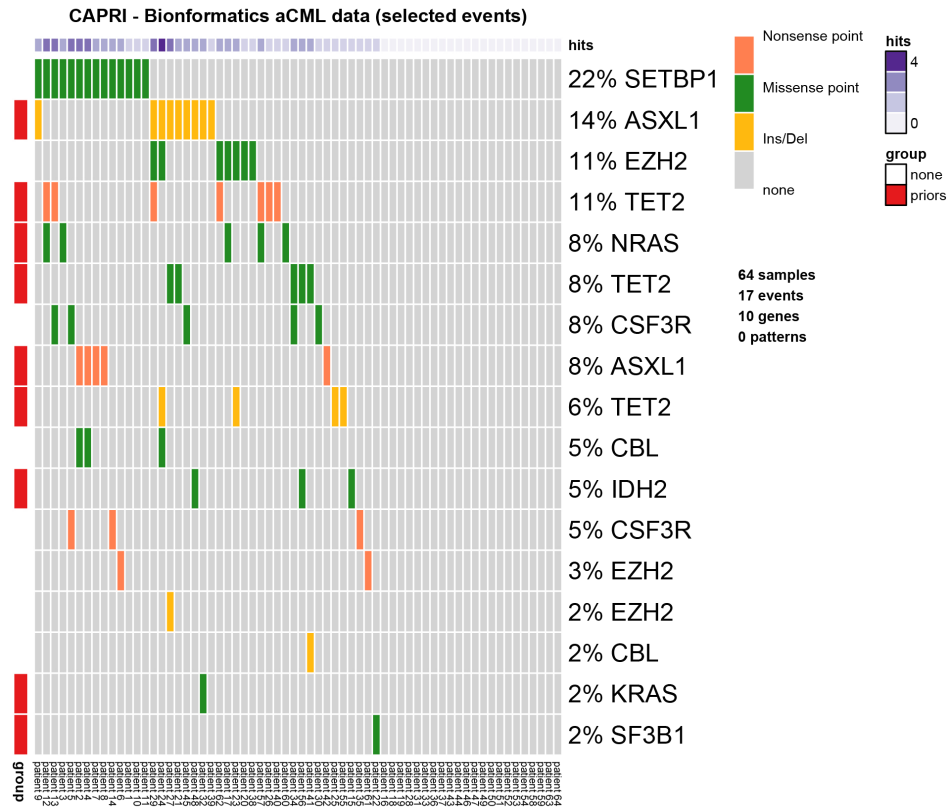

Figure 5.2: **Annotated oncoprint.** Result of the oncoprint function on the selected dataset in TRONCO with annotations.

```
> dummy = oncoprint(events.selection(hypo, filter.in.names = c('KRAS', 'NRAS')),
        font.row=12, cellheight=20, cellwidth=4)

*** Events selection: #events=18, #types=4 Filters freq|in|out =
{FALSE, TRUE, FALSE}
[filter.in] Genes hold: KRAS, NRAS ...  [2/2 found].
Selected 2 events, returning.
*** Oncoprint for ""
with attributes: stage=FALSE, hits=TRUE
Sorting samples ordering to enhance exclusivity patterns.
```

We repeated the same analysis as before for other hypotheses and for the same reasons, we will include only the hard exclusivity pattern.

```
> hypo = hypothesis.add(hypo, 'SF3B1 xor ASXL1', XOR('SF3B1', OR('ASXL1')), '*')
> hypo = hypothesis.add(hypo, 'SF3B1 or ASXL1', OR('SF3B1', OR('ASXL1')), '*')
```

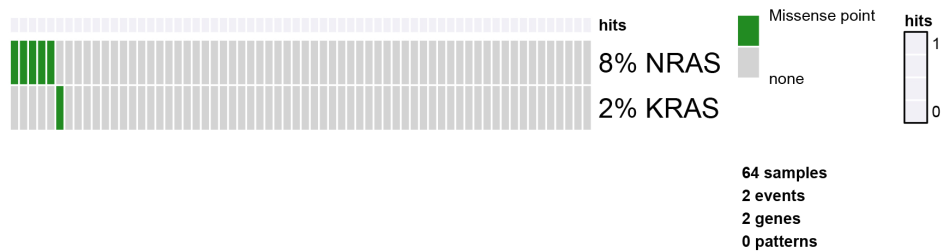

Figure 5.3: **RAS oncoprint.** Result of the oncoprint function in TRONCO for only the RAS genes to better show their hard exclusivity pattern.

<span style="color:red">Error in hypothesis.add(hypo, "SF3B1 or ASXL1", OR("SF3B1", OR("ASXL1")),  :
    [ERR] Pattern duplicates Pattern SF3B1 xor ASXL1.</span>

Finally, we now repeat the same for genes TET2 and IDH2.  In this case 3 events for the gene TET2 are present, that is *"Ins/Del"*, *"Missense point"* and *"Nonsense point"*. For this reason, since we are not specifying any subset of such events to be considered, all TET2 alterations are used.  Since the events present a perfect hard exclusivity, their patters will be included as an *XOR*.  See Figure 5.4.

```
> as.events(hypo, genes = 'TET2')

        type            event
gene 4  "Ins/Del"       "TET2"
gene 32 "Missense point" "TET2"
gene 88 "Nonsense point" "TET2"
> hypo = hypothesis.add(hypo, 'TET2 xor IDH2', XOR('TET2', 'IDH2'), '*')
> hypo = hypothesis.add(hypo,  'TET2 or IDH2', OR('TET2', 'IDH2'), '*')
> dummy = oncoprint(events.selection(hypo, filter.in.names = c('TET2', 'IDH2')),
font.row=12, cellheight=20,cellwidth=4)
*** Events selection: #events=21, #types=4 Filters freq|in|out =
{FALSE, TRUE, FALSE}
[filter.in] Genes hold: TET2, IDH2 ...  [2/2 found].
Selected 4 events, returning.
*** Oncoprint for ""
with attributes: stage=FALSE, hits=TRUE
Sorting samples ordering to enhance exclusivity patterns.
```

We now finally add any possible group of homologous events.  For any gene having more than one event associated we also add a soft exclusivity pattern among them.

```
> hypo = hypothesis.add.homologous(hypo)
```

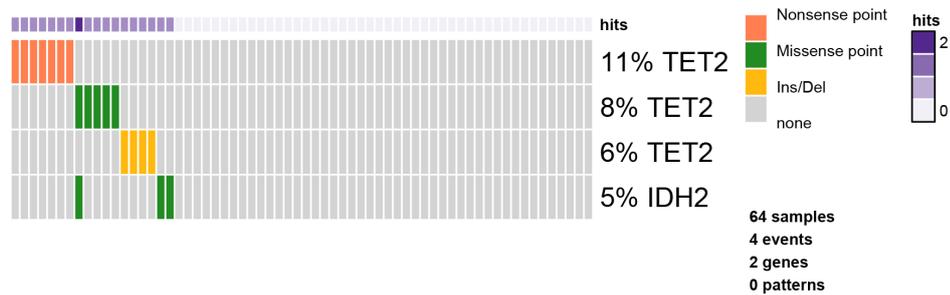

Figure 5.4: **TET/IDH2 oncoprint.** Result of the oncoprint function in TRONCO for only the TET/IDH2 genes.

```
*** Adding hypotheses for Homologous Patterns
 Genes: TET2, EZH2, CBL, ASXL1, CSF3R
 Function: OR
 Cause: *
 Effect: *
Hypothesis created for all possible gene patterns.
```

The final dataset that will be given as input to CAPRI is now finally shown. See Figure 5.5.

```
> dummy = oncoprint(hypo, gene.annot = list(priors= gene.hypotheses),
        sample.id = T, font.row=10, font.column=5, cellheight=15, cellwidth=4)

*** Oncoprint for "CAPRI - Bioinformatics aCML data (selected events)"
with attributes: stage=FALSE, hits=TRUE
Sorting samples ordering to enhance exclusivity patterns.
Annotating genes with RColorBrewer color palette Set1 .
```

## Model reconstruction

We next infer the model by running CAPRI algorithm with its default parameters: we use both AIC and BIC as regularizators, Hill-climbing as heuristic search of the solutions and exhaustive bootstrap (nboot replicates or more for Wilcoxon testing, i.e., more iterations can be performed if samples are rejected), p-value set at 0.05. We set the seed for the sake of reproducibility.

```
> model = tronco.capri(hypo, boot.seed = 12345, nboot=10)

*** Checking input events.
*** Inferring a progression model with the following settings.
```

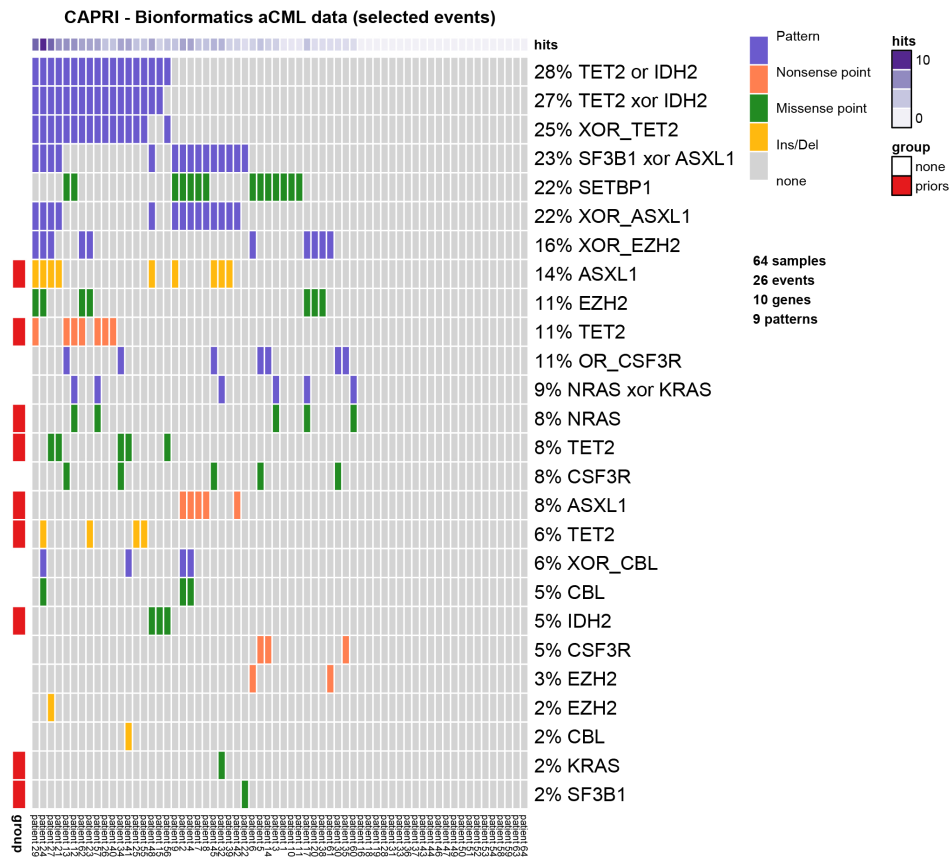

Figure 5.5: **Final dataset for CAPRI.** Result of the oncoprint function in TRONCO on the dataset used in [158].

```
Dataset size: n = 64, m = 26.
Algorithm: CAPRI with "bic, aic" regularization and "hc" likelihood-fit strategy.
Random seed: 12345.
Bootstrap iterations (Wilcoxon): 10.
exhaustive bootstrap: TRUE.
p-value: 0.05.
minimum bootstrapped scores: 3.
*** Bootstraping selective advantage scores (prima facie).
Evaluating "temporal priority" (Wilcoxon, p-value 0.05)
Evaluating "probability raising" (Wilcoxon, p-value 0.05)
*** Loop detection found loops to break.
Removed 26 edges out of 68 (38%)
*** Performing likelihood-fit with regularization bic.
*** Performing likelihood-fit with regularization aic.
```

```
The reconstruction has been successfully completed in 00h:00m:02s
```

We then plot the model inferred by CAPRI with BIC as a regolarizator and we set some parameters to get a good plot; the confidence of each edge is shown both in terms of temporal priority and probability raising (selective advantage scores) and hypergeometric testing (statistical relevance of the dataset of input). See Figure 5.6.

```
> tronco.plot(model, fontsize = 13, scale.nodes = .6, regularization="bic",
        confidence = c('tp', 'pr', 'hg'), height.logic = 0.25, legend.cex = .5,
        pathways = list(priors= gene.hypotheses), label.edge.size=5)

*** Expanding hypotheses syntax as graph nodes:
*** Rendering graphics
Nodes with no incoming/outgoing edges will not be displayed.
Annotating nodes with pathway information.
Annotating pathways with RColorBrewer color palette Set1 .
Adding confidence information: tp, pr, hg
RGraphviz object prepared.
Plotting graph and adding legends.
```

### Bootstrapping data

Finally, we perform non-parametric bootstrap as a further estimation of the confidence in the inferred results. See Figure 5.7.

```
> model.boot = tronco.bootstrap(model, nboot=10)

Executing now the bootstrap procedure, this may take a long time...
Expected completion in approx. 00h:00m:03s
*** Using 7 cores via "parallel"

*** Reducing results

Performed non-parametric bootstrap with 10 resampling and 0.05 as pvalue
for the statistical tests.

> tronco.plot(model.boot, fontsize = 13, scale.nodes = .6, regularization="bic",
        confidence=c('npb'), height.logic = 0.25, legend.cex = .5,
        pathways = list(priors= gene.hypotheses), label.edge.size=10)

*** Expanding hypotheses syntax as graph nodes:
*** Rendering graphics
Nodes with no incoming/outgoing edges will not be displayed.
```

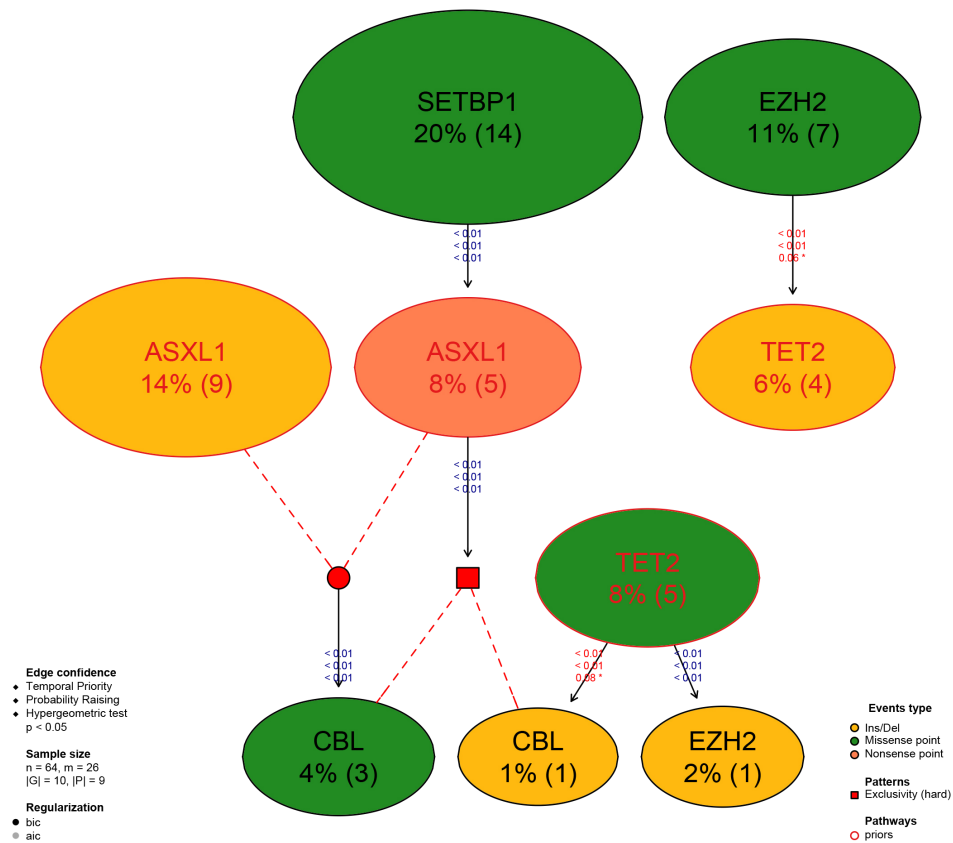

Figure 5.6: **Reconstruction by CAPRI.** Result of the reconstruction by CAPRI on the input dataset.

```
Annotating nodes with pathway information.
Annotating pathways with RColorBrewer color palette Set1 .
Adding confidence information: npb
RGraphviz object prepared.
Plotting graph and adding legends.
```

We now conclude this analysis with an example of inference with the CAPRESE algorithm. As CAPRESE does not consider any patter as input, we use the dataset shown in Figure 5.2. These results are shown in Figure 5.8.

```
> model.boot.caprese = tronco.bootstrap(tronco.caprese(hypo))

*** Checking input events.
```

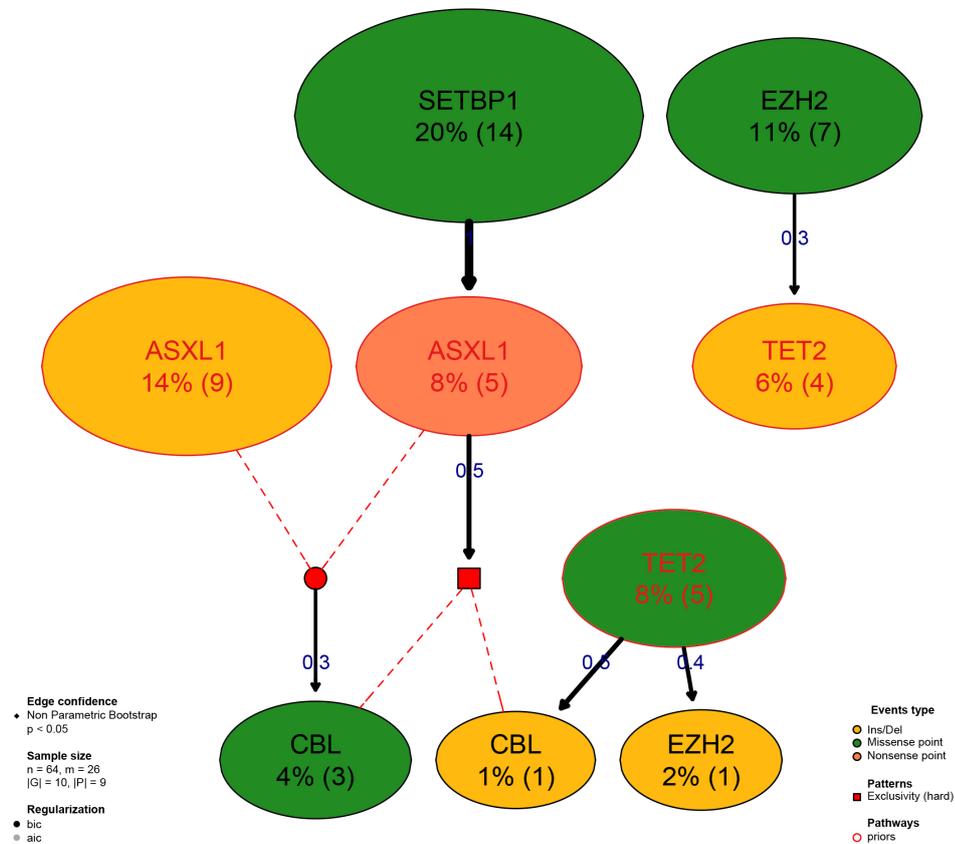

Figure 5.7: **Reconstruction by CAPRI and Bootstrap.** Result of the reconstruction by CAPRI on the input dataset with the assesment by non-parametric bootstrap.

```
*** Inferring a progression model with the following settings.
Dataset size: n = 64, m = 17.
Algorithm: CAPRESE with shrinkage coefficient: 0.5.
The reconstruction has been successfully completed in 00h:00m:00s
Executing now the bootstrap procedure, this may take a long time...
Expected completion in approx. 00h:00m:00s

Performed non-parametric bootstrap with 100 resampling and 0.5
as shrinkage parameter.

> tronco.plot(model.boot.caprese, fontsize = 13, scale.nodes = .6,
        confidence=c('npb'), height.logic = 0.25, legend.cex = .5,
```

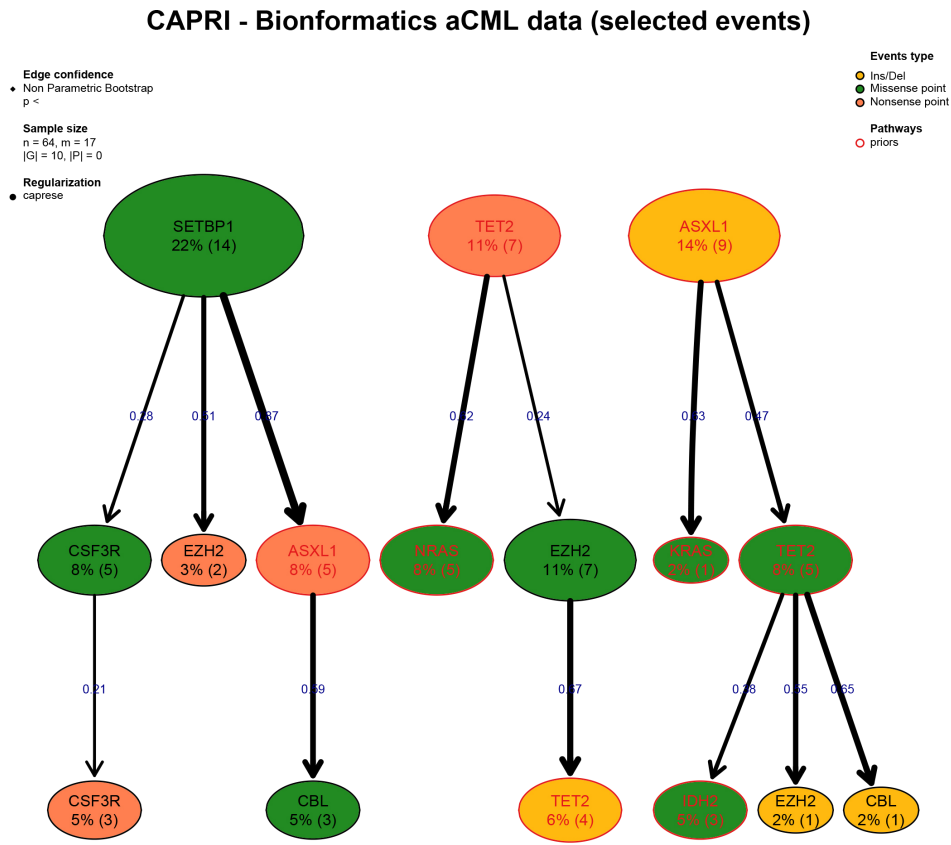

Figure 5.8: **Reconstruction by CAPRESE and Bootstrap.** Result of the reconstruction by CAPRESE on the input dataset with the assessment by non-parametric bootstrap.

```
        pathways = list(priors= gene.hypotheses), label.edge.size=10,
        legend.pos="top")
```

*** Expanding hypotheses syntax as graph nodes:
*** Rendering graphics
Nodes with no incoming/outgoing edges will not be displayed.
Annotating nodes with pathway information.
Annotating pathways with RColorBrewer color palette Set1 .
Adding confidence information: npb
RGraphviz object prepared.
Plotting graph and adding legends.

## 5.3 TCGA MSI/MSS colorectal tumors

We conclude this Chapter with a more elaborate analysis performed with the TRONCO package. As a reference, see Chapter §6 and [21].

We report the acronyms that will be adopted in this Section.

|          |                                           |
|----------|-------------------------------------------|
| MSS      | *Microsatellite Stable*                   |
| MSI      | *Microsatellite Highly-Instable*          |
| CNA      | *Copy Number Alteration*                  |
| TCGA     | *The Cancer Genome Atlas*                 |
| COADREAD | *Human Colon and Rectal Cancer* (TCGA project) |
| MUTEX    | *Mutual Exclusivity* (tool)               |
| CAPRI    | *Cancer Progression Inference* (algorithm)|
| TRONCO   | *Translational Oncology* (tool)           |

### 5.3.1 Summary

The summary and motivation of this study are detailed in the original paper [21]. Here, we just mention that we will (*i*) use samples from the COADREAD project to implement a case/control study, (*ii*) select somatic mutations and focal CNAs in 33 driver genes manually annotated to 5 pathways by the Consortium, (*iii*) scan groups of exclusive alterations with MUTEX ([7]) and, finally, (*iv*) retrieve progression models by running the CAPRI algorithm ([158]) implemented in TRONCO.

**Prerequisites.** Install the TRONCO package either from Bioconductor or its official webpage:

<div align="center">http://bimib.disco.unimib.it</div>

In any case, check that the installed version is updated; this document references TRONCO version 2.0, Mantis Shrimp. This study processes TCGA data originally archived at:

<div align="center">https://tcga-data.nci.nih.gov/docs/publications/coadread_2012/</div>

For conveniency, we have converted such files to formats easily manipulable with R. We host such files at TRONCO's webpage, as of the download performed the 12 March 2015. The files are:

- "crc_clinical_sheet.txt": clinical data file;

- "TCGA_CRC_Suppl_Table2_Mutations_20120719.csv": MAF mutations file, originally Excel file, now converted to csv format with semi-comma separator;

- "crc_gistic.txt": GISTIC file of focal CNAs;

- "TCGA-clusters.csv": clustering results, including MSI/MSS status.

and the code that we provide you with takes care of downloading them autonomously.

For clarity and modularity, this tutorial makes use of different files, reported in sections' titles; the root file is named `main.R`, and starts by loading TRONCO, setting up some variables and folder, then downloading data.

Listing 1: Root file `main.R`

```r
# You might install TRONCO's version from BIMIB Github as
library(devtools)
install_github("BIMIB-DISCo/TRONCO")
library(TRONCO)

# Working directory
workdir = "TCGA-data/"
dir.create(workdir)

# Data files
datafile = 'TCGA-COADREAD-TRONCO.zip'
download.file(
    'https://github.com/BIMIB-DISCo/datasets/raw/master/TCGA-COADREAD-TRONCO.zip',
    destfile=datafile, method='curl')
unzip(datafile, exdir = workdir)

# Name input files
clinical.file = paste0(workdir, "/Clinical/crc_clinical_sheet.txt")
MAF.file = paste0(workdir,
                  "/Mutations/TCGA_CRC_Suppl_Table2_Mutations_20120719.csv")
GISTIC.file = paste0(workdir, "/CNV/crc_gistic.txt")
clusters.file = paste0(workdir, "/Clusters/TCGA-clusters.csv")
MUTEX.msi.file = paste0(workdir, "/MUTEX/msi_results.txt")
MUTEX.mss.file = paste0(workdir, "/MUTEX/mss_results.txt")

# Then this files sources the other scripts
source('scripts/TCGA-import.R', echo = TRUE)
source('scripts/training-samples.R', echo = TRUE)
source('scripts/training-exclusivity.R', echo = TRUE)
source('scripts/training-reconstruction.R', echo = TRUE)
source('scripts/validation-samples.R', echo = TRUE)
source('scripts/validation-pvalues.R', echo = TRUE)
```

## 5.3.2 Import of the TCGA data in TRONCO - `TCGA-import.R`

We begin by loading clinical attributes.

```
clinical.data = TCGA.map.clinical.data(
    file = clinical.file,
    column.samples = 'patient',
    column.map = 'tumor_stage')
head(clinical.data)
```

Now we process separately mutations and CNAs to prepare two `TRONCO` objects and save them as Rdata. One will be the set of all CNAs, and shall be used afterwards to prepare the control samples, the other will be the set of training samples with both CNAs and mutations data.

**Processing mutations.** We can import the MAF file which annotates mutations for the training samples, and augment each patient with its associated stage. We use flag `is.TCGA` to check if any patient has multiple samples associated as we eventually want to select only one sample per patient. These steps generates a `TRONCO` object, on which we use `show` function to get a simple report of the dataset.

```
# Load MAF - use is.TCGA to match samples to patients
MAF = import.MAF(
    file = MAF.file,
    is.TCGA = TRUE,
    sep = ';')

# Add stage annotation - use match.TCGA.patients to match long/short barcodes
MAF = annotate.stages(MAF, clinical.data, match.TCGA.patients = TRUE)
show(MAF)
```

As COADREAD has multiple samples, we use `TRONCO` functions which implement TCGA aliquote disambiguation policies to select one sample per patient

```
# Check for duplicated samples - we find them
TCGA.multiple.samples(MAF)

# Remove duplicated samples according to TCGA criteria,
# shorten barcodes and add stages
MAF = TCGA.remove.multiple.samples(MAF)
MAF = TCGA.shorten.barcodes(MAF)
MAF = annotate.stages(MAF, clinical.data)
```

**Processing Copy Numbers.** As done for mutations, we import CNAs from focal GISTIC profiles. Preprocessing is here minor, as GISTIC profiles are in a matrix which can be imported directly in `TRONCO` after transposition. Editing of the GISTIC information that we do not want to use, is done on the `TRONCO` object as we have functions for its easy manipulation

```r
# Load as a plain table
GISTIC = read.table(
    GISTIC.file,
    check.names = FALSE,
    stringsAsFactors = FALSE,
    header = TRUE)

# Have a look at this table to remove useless information
head(GISTIC[, 1:5])
GISTIC $Entrez_Gene_Id = NULL
rownames(GISTIC) = GISTIC $Hugo_Symbol
GISTIC $Hugo_Symbol = NULL

# Import all GISTIC data
GISTIC = import.GISTIC( t(GISTIC) )
show(GISTIC)

# We want to use only high-confidence scores from the GISTIC,
# renamed as Amplification/Deletion
GISTIC = delete.type(GISTIC, 'Heterozygous Loss') # low-level deletions
GISTIC = delete.type(GISTIC, 'Low-level Gain') # low-level amplifications
GISTIC = rename.type(GISTIC, 'Homozygous Loss', 'Deletion')
GISTIC = rename.type(GISTIC, 'High-level Gain', 'Amplification')
GISTIC = annotate.stages(GISTIC, clinical.data)
show(GISTIC)
```

**Subsetting samples for training.** We now want to select only samples with both somatic mutations and CNAs available - i.e., the intersection of the dataset `GISTIC` and `MAF`. Notice that the GISTIC contains only genes with at least 1 CNV in the dataset, so some well-known CRC genes such as PIK3CA or FAM123B are not annotated in there.

```r
c("PIK3CA", "FAM123B") \%in\% as.genes(GISTIC) # Shall be FALSE
```

Thus, we want to intersect patients as we assume that every patient has at least one CNA, and make the union of all the genes annotated in `MAF`/`GISTIC` datasets. This shall yield a new dataset called `MAF.GISTIC`; this and `GISTIC` will be exported as Rdata.

```r
# We set intersect.genomes to FALSE to take the union of altered genes
MAF.GISTIC = intersect.datasets(GISTIC, MAF, intersect.genomes = FALSE)

# We remove events which have no observations in the dataset,
# and annotate stages
MAF.GISTIC = trim(MAF.GISTIC)
```

```R
MAF.GISTIC = annotate.stages(MAF.GISTIC, clinical.data)
show(MAF.GISTIC)

# Export these datasets as Rdata
save(MAF.GISTIC, file = paste0(workdir, 'MAF.GISTIC.Rdata'))
save(GISTIC, file = paste0(workdir, 'GISTIC.Rdata'))
```

### 5.3.3   Preparing the training datasets - `training-samples.R`

We first create some folders where we will separate `TRONCO` output, then we load the files that we just created, and set some fancy colors.

```R
# Prepare folders
dir.create('./MSS')
dir.create('./MSI')
sub.dir = c('MUTEX', 'Rdata-lifted', 'Rdata-models')
sapply(paste0('./MSS/', sub.dir), dir.create)
sapply(paste0('./MSI/', sub.dir), dir.create)

# Load MAF.GISTIC file, set some fancy colors to get cute visualization
load(paste0(workdir, '/MAF.GISTIC.Rdata'))
MAF.GISTIC = change.color(MAF.GISTIC, 'Mutation', 'darkolivegreen3')
MAF.GISTIC = change.color(MAF.GISTIC, 'Amplification', 'coral')
MAF.GISTIC = change.color(MAF.GISTIC, 'Deletion', 'cornflowerblue')
```

Now we load clustering assignments from TCGA, we define the clustering maps for patients to MSI/MSS status and use that to create the corresponding `TRONCO` objects for each dataset.

```R
# Load table data
file = read.delim(clusters.file, sep = ";")
head(file)

# Select just certain annotations, remove blank lines
tab = file[, c("patient", "MSI_status", "sequenced")]
tab = tab[1:276, ]
rownames(tab) = tab $patient

# Filter out non-sequenced samples, and order them (for console visualization)
tab = tab[tab $sequenced == 1, ]
tab = tab[order(tab $MSI_status), ]
print(tab)

# Define the maps to split samples
map.MSS = tab[tab $MSI_status == "MSS", , drop = FALSE]
```

```
map.MSI.H = tab[tab $MSI_status == "MSI-H", , drop = FALSE]

# These are the samples that we actually use
MSS.samples = rownames(map.MSS)
MSI.H.samples = rownames(map.MSI.H)

# Split is done by using samples.selection with appropriate vectors as input
MSS = trim(samples.selection(MAF.GISTIC, MSS.samples))
MSI.H = trim(samples.selection(MAF.GISTIC, MSI.H.samples))
show(MSS)
show(MSI.H)
```

To produce fancy figures, we shall also define some further groups and colors. Among these, there are the 33 relevant driver genes identified by the Consortium which we include now. Also, as we are not going to use any external clustering tool, we can immediately subset MSI/MSS tumors to include only alterations in those driver genes.

```
# Plotting colors
alteration.color = 'dimgray'
pathways.color = c('firebrick1', 'darkblue', 'darkgreen', 'darkmagenta',
                    'darkorange')

# Driver events - 33 genes mapped to 5 pathways by TCGA
Wnt = c("APC", "CTNNB1", "DKK1", "DKK2", "DKK3", "DKK4", "LRP5", "FZD10",
        "FAM123B", "AXIN2", "TCF7L2", "FBXW7", "ARID1A", "SOX9")
RAS = c("ERBB2", "ERBB3", "NRAS", "KRAS", "BRAF")
PI3K = c("IGF2", "IRS2", "PIK3CA", "PIK3R1", "PTEN")
TGFb = c("TGFBR1", "TGFBR2", "ACVR1B", "ACVR2A", "SMAD2", "SMAD3", "SMAD4")
P53 = c("TP53", "ATM")

# Some variable which will be processed by TRONCO plotting functions
pathway.genes = c(Wnt, RAS, PI3K, TGFb, P53)
pathway.names = c('Wnt', 'RAS', 'PI3K', 'TGFb', 'P53')
pathway.list = list(Wnt = Wnt, RAS = RAS, PI3K = PI3K, TGFb = TGFb, P53 = P53)
```

Finally, we can create the two datasets which we shall use - these contain as driver events only alterations in the 33 TCGA genes. Also, we can use oncoprint function to start making plots out of the data - see Figure 5.9.

```
# MSS tumors
MSS = trim(events.selection(MSS, filter.in.names = pathway.genes))
MSS = annotate.description(MSS, 'MSS subtype')

# MSI-HIGH tumors
MSI.H = trim(events.selection(MSI.H, filter.in.names = pathway.genes))
```

```
MSI.H = annotate.description(MSI.H, 'MSI-HIGH subtype')

# We use TRONCO visualization function - oncoprint - to view these dataset
w = oncoprint(MSS,
    title = 'MSS tumors - with all driver genes',
        legend.cex = .5,    # Legend size for events type
        gene.annot = pathway.list,    # List of mapping to pathways/groups
        gene.annot.color = pathways.color,    # Mapping color
        sample.id = T)    # Sample names

w = oncoprint(MSI.H,
    legend.cex = .5,
        gene.annot = pathway.list,
        gene.annot.color = pathways.color,
        sample.id = T)
```

### 5.3.4  Exclusivity groups in the datasets - `training-exclusivity.R`

`TRONCO` relies on external tools such as `MUTEX` or others to detect exclusivity groups. As the input/output formats of such tools is non-uniform, however, we do not support direct conversion from `TRONCO` objects to every tool input/output.

In this case, we use the `MUTEX` tool to detect exclusivity groups in `MSI/MSS` datasets and, inside `TRONCO`, there are input/output function to use with it, which we update consistently.

```
# Export a file MSS.txt compliant to MUTEX input
export.MUTEX(MSS,
    filename = 'MSS/MUTEX/MSS.txt',
    label.mutation = 'Mutation',
    label.amplification = 'Amplification',
    label.deletion = 'Deletion'
)

export.MUTEX(MSI.H,
    filename = 'MSI/MUTEX/MSI.H.txt',
    label.mutation = 'Mutation',
    label.amplification = 'Amplification',
    label.deletion = 'Deletion'
        )
```

Then we can run the Java `MUTEX` tool by following the routines explained at the following webpage:

https://github.com/BIMIB-DISCo/mutex

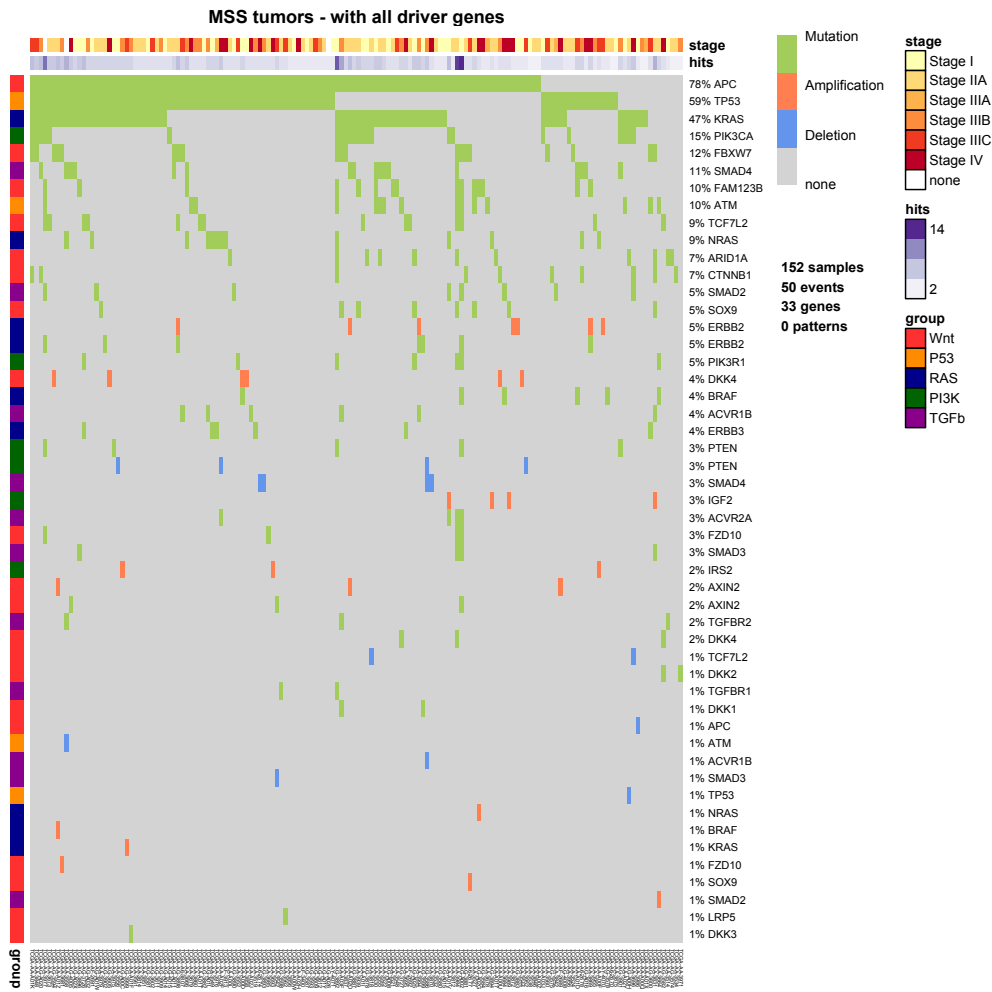

Figure 5.9: MSS training tumors with selected 33 driver genes.

In this case we provide you with the results of running `MUTEX` with its default parameters; we shall then collect results - groups with score below .2 threshold, as usually done - in R's runtime by using function `import.MUTEX.groups`.

```
MSI.H.MUTEX = import.MUTEX.groups(MUTEX.msi.file)
MSS.MUTEX = import.MUTEX.groups(MUTEX.mss.file)
```

The 3 groups detected for MSI tumors can be visualized by arranging GROBS returned by oncoprint and obtain the plot that we show in Figure 5.10.

```r
# Visualize groups via console
print(MSI.H.MUTEX)

# Then combine oncoprint via grid.arrange
grid.arrange(
    oncoprint(
        # Select only events for genes in group 1
        events.selection(MSI.H, filter.in.names = MSI.H.MUTEX[[1]]),
            title = paste("MSI-H - MUTEX group 1"),
                    legend.cex = .3,
                    font.row = 6,
                    ann.hits = FALSE, # Avoid annotating the hits for these groups
                    cellheight = 10,
                    silent = T, # Do not plot the oncoprint
                    gene.annot = pathway.list,
                    gene.annot.color = pathways.color,
        ) $gtable,
        oncoprint(
                events.selection(MSI.H, filter.in.names = MSI.H.MUTEX[[2]]),
                        title = paste("MSI-H - MUTEX group 2"),
                        legend.cex = .3,
                        silent = T,
                        font.row = 6,
                        ann.hits = FALSE,
                        cellheight = 10,
                        gene.annot = pathway.list,
                        gene.annot.color = pathways.color,
        ) $gtable,
        oncoprint(
                events.selection(MSI.H, filter.in.names = MSI.H.MUTEX[[3]]),
                        title = paste("MSI-H - MUTEX group 3"),
                        legend.cex = .3,
                        silent = T,
                        font.row = 6,
                        ann.hits = FALSE,
```

```
                                cellheight = 10,
                                gene.annot = pathway.list,
                                gene.annot.color = pathways.color,
        ) $gtable,
        ncol=1 # Display all plots in a single column
)
```

Other three groups will be used - in each subtype, in principle - according to both literature on CRC, and analysis carried out by the Consortium with the `MEMO` tool (Ciriello *et al*, Curr Protoc Bioinformatics. 2013 Mar;Chapter 8:Unit 8.17). We shall define the following variables.

```
# Apriori CRC knowledge
KNOWLEDGE.PRIOR.WNT = c('APC', 'CTNNB1')
KNOWLEDGE.PRIOR.RAF = c('KRAS', 'NRAS', 'BRAF')

# MEMO group estimated by TCGA for the non-hypermutated tumors:
TCGA.MEMO = c('ERBB2', 'IGF2', 'PIK3CA', 'PTEN')
```

### 5.3.5 Reconstruction of the models - `training-reconstruction.R`

We first implement a selection strategy of driver events. For any dataset we select:

- all genes with an alteration frequency > 5%;

- all genes involved in an exclusivity prior (`MUTEX` / `MEMO` / knowledge-based).

We thus define this function

```
select = function(x, min.freq, forced.genes)
{
        # Collapse multiple events per gene in one unique event
        x.sel = as.alterations(x)

        # Get a list of those with minimum frequency > min.freq
        # but force inclusion of all events for genes in "forced.genes"
        x.sel = events.selection(x.sel, filter.freq = min.freq,
                                        filter.in.names = forced.genes)

        # Subset input - select all events for any gene in "x.sel"
        x = events.selection(x, filter.in.names = as.genes(x.sel))
        return(x)
}

# We set this as a variable
MIN.FREQ = 0.05
```

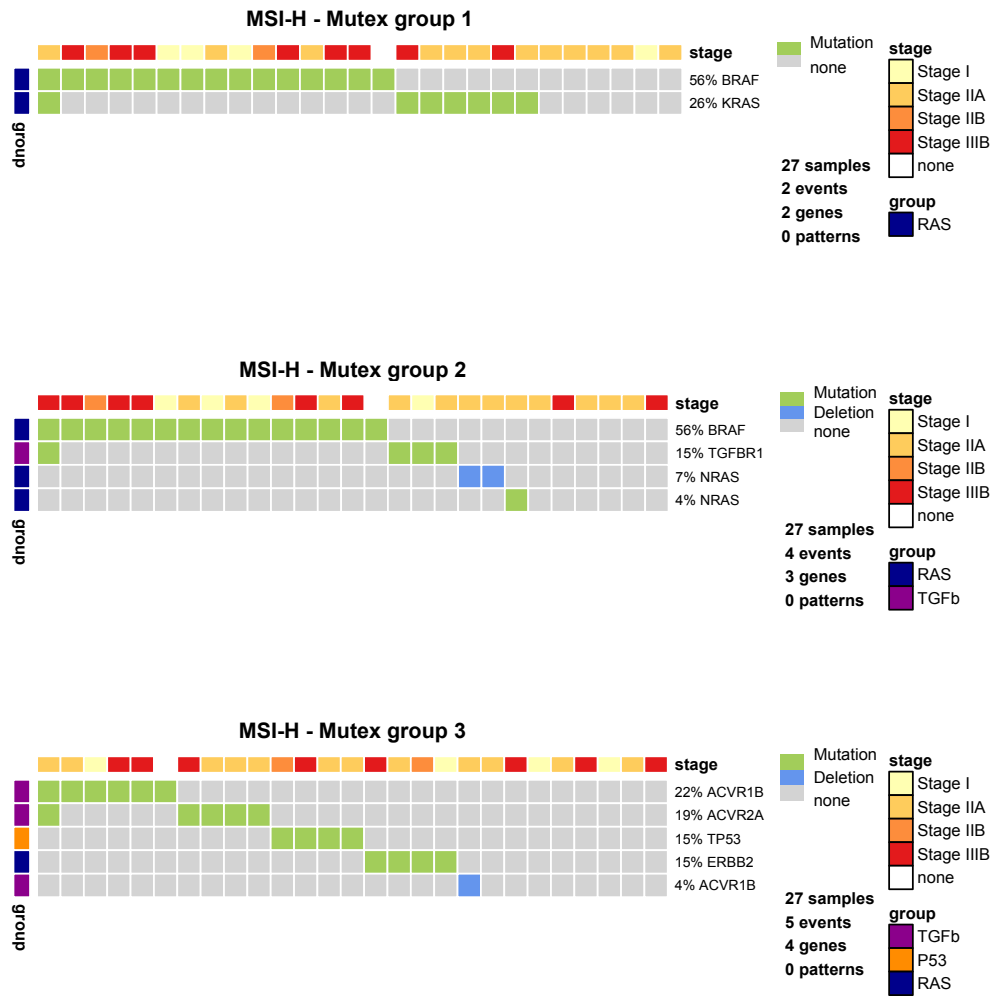

Figure 5.10: MSI exclusivity groups detected with MUTEX.

Which can be used to subset the datasets to define the input objects for CAPRI.

```
# Selected MSS dataset
MSS.select = select(MSS,
        MIN.FREQ,
        unique( # All group genes
                c(TCGA.MEMO,
                KNOWLEDGE.PRIOR.WNT,
                KNOWLEDGE.PRIOR.RAF,
                unlist(MSS.MUTEX))
                )
        )
MSS.select = annotate.description(MSS.select,
                        'TCGA MSS colorectal tumors')

# Selected MSI dataset
MSI.H.select = select(MSI.H,
        MIN.FREQ,
        unique(c(TCGA.MEMO,
                KNOWLEDGE.PRIOR.WNT,
                KNOWLEDGE.PRIOR.RAF,
                unlist(MSI.H.MSS.MUTEX)
                )
                ))
MSI.H.select = annotate.description(MSI.H.select,
                        'TCGA MSI-HIGH colorectal tumors')
```

To visualize the selected data, one can use again `oncoprint` function.

```
oncoprint(MSS.select, gene.annot = pathway.list,
            gene.annot.color = pathways.color)
```

As detailed in CAPRI's definition, we are required to provide no pair of events (rows in the oncoprint) which have the same signature. If that happens, those events are statistically indistinguishable, and we prefer to eliminate one of them - being aware that the inferred model would be equivalent whatever event we decide to delete. Further, we also want that no event appears never (resp. always) the dataset - rows of all 0s or 1s. To detect if `MSS.select` or `MSS.select` has any of these inconsistencies we use a `TRONCO` function which outputs NULL if the dataset has no of these issues.

```
# Selected MSS is ok
cd = consolidate.data( MSS.select, T)

# In MSI we need to manipulate data instead
cd = consolidate.data( MSI.H.select, T)
```

```r
# FBXW7 amplification and PTEN mutations are discarded
MSI.H.select = delete.event(MSI.H.select, gene='FBXW7',
                                        type ='Amplification')
MSI.H.select = delete.event(MSI.H.select, gene='PTEN',
                                        type ='Mutation')
```

Now we really perform inference of selective advantage relations with CAPRI. We first define a function which prepares all formulas that we want to test; these will include exclusivity of genes with multiple alterations, `MUTEX` groups and other priors.

To create formulas from groups we use function `hypothesis.add.group` which creates all possible formulas from a group of genes, or generally alterations identified by their name. The number of alterations determines the combinatorial number of formulas to create, and if a gene in the group has multiple events associated to it, e.g., a mutation and a CNA, a sub-formula is created to combine them. According to the overlap between such events the formula will be written in either hard or soft exclusivity form. Notice also that we use function `hypothesis.add.homologous` which creates, if a gene has multiple events associated, the same sub-formula created by `hypothesis.add.group`, but regardless of the group.

After the creation of formulas we can apply the algorithm CAPRI to the dataset using the function `tronco.capri`. In this case we set the seed to 12345 to get the same results because of bootstrap random shuffling. The obtained progression model is plotted using the function `tronco.plot`.

```r
recon = function(x, folder, MUTEX, ...) {

        lift = x

        # 1. create formulas from every MUTEX group
        if(!is.null(MUTEX))
                # TRONCO function to create formulas from groups.
                # Notice that only 1 formula per group is created
                # by the dim.min constraint
                for (w in MUTEX) {
                        lift = hypothesis.add.group(lift,
                                FUN = OR, # formula of "soft exclusivity" (OR)
                                group = w, # the group
                                dim.min = length(w) # only 1 group is maximal
                                )
                }

        # Now we subset Wnt groups to genes actually present in lift.
        # Notice that formulas for groups with dimension smaller than 2
        # will not be created
```

```
KNOWLEDGE.PRIOR.WNT.subtype = KNOWLEDGE.PRIOR.WNT[
        KNOWLEDGE.PRIOR.WNT %in% as.genes(lift)
        ]

lift = hypothesis.add.group(lift,
        FUN = OR,
        group = KNOWLEDGE.PRIOR.WNT.subtype,
        dim.min = length(KNOWLEDGE.PRIOR.WNT.subtype))

# The same for the RAF group
KNOWLEDGE.PRIOR.RAF.subtype = KNOWLEDGE.PRIOR.RAF[
        KNOWLEDGE.PRIOR.RAF %in% as.genes(lift)
        ]

lift = hypothesis.add.group(lift,
        FUN = OR,
        group = KNOWLEDGE.PRIOR.RAF.subtype,
        dim.min = length(KNOWLEDGE.PRIOR.RAF.subtype))

# And the MEMO group
TCGA.MEMO.subtype = TCGA.MEMO[TCGA.MEMO %in% as.genes(lift)]

lift = hypothesis.add.group(lift,
        FUN = OR,
        group = TCGA.MEMO.subtype,
        dim.min = length(TCGA.MEMO.subtype))

# Homologous in soft exclusivity
lift = hypothesis.add.homologous(lift, FUN = OR)

# Save to file the PDF of the lifted dataset, and its Rdata
lift = annotate.description(lift, as.description(x))
oncoprint(lift, file = paste0(folder, '/Rdata-lifted/lifted.pdf'))
save(lift, file=paste0(folder, '/Rdata-lifted/lifted.Rdata'))

# CAPRI execution with seed set, default parameters
model = tronco.capri(lift, boot.seed = 12345)

# As takes long to bootstrap, we make a simple visualization first
tronco.plot(model,
        pathways = pathway.list,
        confidence = c('tp', 'pr', 'hg'), # Display p-values
        ... )
```

```r
        # Example non-parametric and statistical bootstrap.
        # Note that 5 iterations is purely explanatory,
        # at least 100 should be performed for statistical significance
        model = tronco.bootstrap(model)
        model = tronco.bootstrap(model, type = "statistical", )

        # Save the Rdata
        save(model, file=paste0(folder, '/Rdata-models/model.Rdata'))

        # Plot the TRONCO model
        tronco.plot(model,
                    pathways = pathway.list,
                    edge.cex = 1.5,
                     legend.cex = .5,
                    scale.nodes = .6,
                    confidence = c('tp', 'pr', 'hg'), # Display p-values
                    pathways.color = pathways.color,
                disconnected = F,
                height.logic = .3,
                file = paste0(folder, '/Rdata-models/model.pdf'),
                 ... )

        return(model)
}
```

Now we can apply the function `recon` to `MSS.select` and `MSI.H.select`.

```r
# Work..
MSS.models = recon(x = MSS.select, folder = 'MSS', MUTEX = MSS.MUTEX)
MSI.models = recon(x = MSI.H.select, folder = 'MSI', MUTEX = MSI.H.MUTEX)
```

### 5.3.6   Preparing the models for test - `validation-samples.R`

We create now another folder where we will put `TRONCO`'s output for test dataset, then we load the file which contain the binary matrix with mutation data and we import it into `TRONCO` format using `import.genotypes` function.

```r
# Prepare folders
dir.create('./VALIDATION')
sub.dir = c('MUTEX', 'Rdata-lifted', 'Rdata-models')
sapply(paste0('./VALIDATION/', sub.dir), dir.create)

# Load mutations (binary matrix), process names
```

```
new.data = load(paste0(workdir, "/Mutations/TCGA_fun.Rdata"))
head(tcga.f.ag)
rownames(tcga.f.ag) = tcga.f.ag $gene
tcga.f.ag $gene = NULL

tcga.f.ag = t(tcga.f.ag)
rownames(tcga.f.ag) = gsub('_', '-', rownames(tcga.f.ag))

# Import in TRONCO format
tcga.f.ag = import.genotypes(tcga.f.ag, 'Mutation')
show(tcga.f.ag)
```

Now we separate the list of samples of the dataset just loaded into two sets, one for MSI-HIGH and one for MSS. The discrimination is based on the mutation rate.

```
# Mutation rate to discriminate MSS and MSI-HIGH
plot(sort(rowSums(as.genotypes(tcga.f.ag))), col = 'darkgreen', lty = 3,
        xlab = 'Tumor sample', ylab = 'Mutations (number)')
title('Validation dataset - Mutations per sample')
abline(h = 500, col = 'red', lty = 'dashed')
abline(v = 204, col = 'black', lty = 'dashed')
text(x = 230, y = 10, 'MSI-H (Val.)', cex = .8)
text(x = 170, y = 10, 'MSS (Val.)', cex = .8)
text(x = 30, y = 600, '500 mutations (cutoff)', cex = .5)

# MSS tumors
tcga.f.ag.MSS = rownames(as.genotypes(tcga.f.ag))[
        which(rowSums(as.genotypes(tcga.f.ag)) < 500)
        ]

# MSI-HIGH tumors
tcga.f.ag.MSI = rownames(as.genotypes(tcga.f.ag))[
        which(rowSums(as.genotypes(tcga.f.ag)) > 500)
        ]
```

Then, based on the two list of samples using the function `samples.selection` we produce two dataset: one for MSI-HIGH and one for MSS. After this we trim the new datasets, shorten the barcodes of samples and keep only the genes mapped to selected pathways. At the end of this operation we have two new dataset ready: `tcga.f.ag.MSS` and `tcga.f.ag.MSI`.

```
# TRONCO objects for these tumors
tcga.f.ag.MSS = trim(samples.selection(tcga.f.ag, tcga.f.ag.MSS))
tcga.f.ag.MSI = trim(samples.selection(tcga.f.ag, tcga.f.ag.MSI))
```

```
tcga.f.ag.MSS = TCGA.shorten.barcodes(tcga.f.ag.MSS)
tcga.f.ag.MSI = TCGA.shorten.barcodes(tcga.f.ag.MSI)
show(tcga.f.ag.MSS)
show(tcga.f.ag.MSI)

# Select just genes mapped to pathways
tcga.f.ag.MSS = trim(events.selection(tcga.f.ag.MSS,
                            filter.in.names = pathway.genes))
tcga.f.ag.MSI = trim(events.selection(tcga.f.ag.MSI,
                            filter.in.names = pathway.genes))
tcga.f.ag.MSS = change.color(tcga.f.ag.MSS, 'Mutation', 'darkolivegreen3')
tcga.f.ag.MSI = change.color(tcga.f.ag.MSI, 'Mutation', 'darkolivegreen3')
```

Now we can load the GISTIC of CNA data, change the events color and keep only the events which are present in pathways. The file `GISTIC.RData` contains a dataset already in TRONCO format.

```
# Load CNA data
load(paste0(workdir, 'GISTIC.Rdata'))
show(GISTIC)
GISTIC = change.color(GISTIC, 'Amplification', 'coral')
GISTIC = change.color(GISTIC, 'Deletion', 'cornflowerblue')
GISTIC = events.selection(GISTIC, filter.in.names = pathway.genes)
```

We have now the two dataset with mutations data and the dataset with CNA. The next operation we have to do is to merge rispectively mutations and CNA data of MSS and MSI-HIGH. We can do this using the function **ebind** which merge two datasets with the same samples list and different events. We start with MSS.

```
# Join datasets with both mutations and CNAs: MSS
tcga.f.ag.MSS =
        ebind(
                samples.selection(tcga.f.ag.MSS,
                        intersect(
                                as.samples(tcga.f.ag.MSS),
                                as.samples(GISTIC))
                        ),
                samples.selection(GISTIC,
                        intersect(
                                as.samples(tcga.f.ag.MSS),
                                as.samples(GISTIC))
                        )
                )
tcga.f.ag.MSS = trim(tcga.f.ag.MSS)
```

```
tcga.f.ag.MSS = annotate.stages(tcga.f.ag.MSS, as.stages(GISTIC))
oncoprint(tcga.f.ag.MSS)
```

And we can do the same for MSI-HIGH.

```
# Join datasets with both mutations and CNAs: MSI-HIGH
tcga.f.ag.MSI =
ebind(
        samples.selection(tcga.f.ag.MSI,
                intersect(
                        as.samples(tcga.f.ag.MSI),
                        as.samples(GISTIC))
        ),
        samples.selection(GISTIC,
                intersect(
                        as.samples(tcga.f.ag.MSI),
                        as.samples(GISTIC))
        )
)
tcga.f.ag.MSI = trim(tcga.f.ag.MSI)
tcga.f.ag.MSI = annotate.stages(tcga.f.ag.MSI, as.stages(GISTIC))
oncoprint(tcga.f.ag.MSI)
```

### 5.3.7   Validation of the models - `validation-samples.R`

At this point we can use the previously created `select` function on `tcga.f.ag.MSS` and `tcga.f.ag.MSI` to subset data with the minimum frequency, as we did for training sets.

```
# Selected MSS tumors
tcga.f.ag.MSS.select = select(tcga.f.ag.MSS,
        MIN.FREQ,
        unique(c(TCGA.MEMO,
                KNOWLEDGE.PRIOR.WNT,
                KNOWLEDGE.PRIOR.RAF,
                unlist(MSS.MUTEX)
                )
                ))

# Mnemonic dataset title
tcga.f.ag.MSS.select =
        annotate.description(tcga.f.ag.MSS.select,
                                '(Test) MSS colorectal tumors')

# Selected MSI tumors
```

```
tcga.f.ag.MSI.select = select(tcga.f.ag.MSI,
        MIN.FREQ,
        unique(c(TCGA.MEMO,
                KNOWLEDGE.PRIOR.WNT,
                KNOWLEDGE.PRIOR.RAF,
                unlist(MSI.H.MUTEX)
                )
                ))

# Mnemonic dataset title
tcga.f.ag.MSI.select =
        annotate.description(tcga.f.ag.MSI.select,
                                '(Test) MSI-HIGH colorectal tumors')
```

After this, we can check the consistency of the dataset with the `consolidate.data` function already described before - you should find no output in this case.

Now we can reconstruct the two models with the function **recon** defined before.

```
# Reconstructions
tcga.f.ag.MSS.models = recon(x = tcga.f.ag.MSS.select,
                                folder = 'VALIDATION', MUTEX = MSS.MUTEX)
tcga.f.ag.MSI.models = recon(x = tcga.f.ag.MSI.select,
                                folder = 'VALIDATION', MUTEX = MSS.MUTEX)
```

Finally we list the p-values for MSS training model to be matched to those for test.

```
# P-values for MSS training for
# temporal priority, probability raising and hypergeometric test
as.selective.advantage.relations(MSS.models) # edges used for MLE
as.selective.advantage.relations(MSS.models, type = 'pf') # all edges
as.selective.advantage.relations(tcga.f.ag.MSS.models) # edges used for MLE
as.selective.advantage.relations(tcga.f.ag.MSS.models, type = 'pf') # all edges

# P-values for MSI-HIGH
as.selective.advantage.relations(MSI.models) # edges used for MLE
as.selective.advantage.relations(MSI.models, type = 'pf') # all edges
as.selective.advantage.relations(tcga.f.ag.MSI.models) # edges used for MLE
as.selective.advantage.relations(tcga.f.ag.MSI.models, type = 'pf') # all edges
```

The result of this operation is show for MSI-HIGH in Table §5.1 using BIC as a regularizator and in Table §5.2 using AIC.

| | SELECTS | SELECTED | OBS | OBS | TEMP PRIOR. | PROB RAIS | HYPG |
|---|---|---|---|---|---|---|---|
| 1 | Del NRAS | Del ACVR1B | 2 | 1 | 2.925532e-20 | 1.884409e-40 | 0 |
| 2 | Mut LRP5 | Mut TGFBR2 | 7 | 2 | 8.324422e-74 | 3.418890e-84 | 0 |
| 3 | Mut APC | Del NRAS | 10 | 2 | 1.061600e-80 | 1.178086e-82 | 0 |
| 4 | Mut APC | Mut PIK3CA | 10 | 5 | 2.528354e-65 | 3.810795e-72 | 0.047342 |
| 5 | Mut APC | Mut SMAD3 | 10 | 2 | 9.108542e-81 | 8.615155e-81 | 0 |
| 6 | Mut APC | Pat XOR_NRAS | 10 | 3 | 2.511396e-78 | 5.359681e-89 | 0 |
| 7 | Mut ARID1A | Mut DKK2 | 11 | 5 | 3.623183e-71 | 3.307532e-76 | 0.005722 |
| 8 | Mut AXIN2 | Mut TP53 | 5 | 4 | 1.318165e-07 | 9.078708e-69 | 0.012820 |
| 9 | Mut FAM123B | Mut SMAD4 | 8 | 4 | 4.124423e-51 | 1.977043e-66 | 0.00398 |
| 10 | Mut PIK3CA | Mut DKK4 | 5 | 2 | 4.828289e-53 | 2.649370e-75 | 0.028490 |
| 11 | Mut ERBB2 | Mut NRAS | 4 | 1 | 4.213949e-56 | 1.654684e-58 | 0 |
| 12 | Mut ERBB2 | Mut SMAD2 | 4 | 2 | 3.869181e-28 | 6.690900e-68 | 0 |
| 13 | Mut ERBB3 | Mut KRAS | 7 | 7 | 0.362016 | 4.184342e-77 | 0.004651 |
| 14 | Pat OR_ERBB2... | Mut LRP5 | 8 | 7 | 5.063714e-05 | 2.775605e-79 | 0.011391 |

Table 5.1: *p-values* of **MSI-HIGH** with BIC as regulatizator.

| | SELECT | SELECTED | OBS | OBS | TEMP PRIOR | PROB RAIS | HYPG |
|---|---|---|---|---|---|---|---|
| 1 | Del NRAS | Del ACVR1B | 2 | 1 | 2.925532e-20 | 1.884409e-40 | 0 |
| 2 | Mut ACVR2A | Mut TCF7L2 | 5 | 4 | 2.777518e-08 | 5.573215e-46 | 0.012820 |
| 3 | Mut BRAF | Mut ACVR1B | 15 | 6 | 6.981943e-79 | 6.622440e-67 | 0.138647 |
| 4 | Mut BRAF | Mut DKK1 | 15 | 2 | 2.463469e-81 | 1.180297e-79 | 0 |
| 5 | Mut FBXW7 | Del NRAS | 14 | 2 | 2.226351e-81 | 1.198633e-82 | 0 |
| 6 | Mut LRP5 | Mut TGFBR2 | 7 | 2 | 8.322422e-74 | 3.418890e-84 | 0 |
| 7 | Mut APC | Del NRAS | 10 | 7 | 1.061600e-80 | 1.178086e-82 | 0 |
| 8 | Mut APC | Mut KRAS | 10 | 7 | 8.532849e-34 | 5.804504e-75 | 0.042748 |
| 9 | Mut APC | Mut PIK3CA | 10 | 5 | 2.528354e-65 | 3.810795e-72 | 0.047342 |
| 10 | Mut APC | Mut SMAD3 | 10 | 2 | 9.108542e-81 | 8.615155e-81 | 0 |
| 11 | Mut APC | Pat OR.ERBB2... | 10 | 8 | 3.437078e-18 | 1.925986e-64 | 0.090990 |
| 12 | Pat APC | Pat XOR.NRAS | 10 | 3 | 2.511396e-78 | 5.359681e-89 | 0 |
| 13 | Mut ARID1A | Mut DKK2 | 11 | 5 | 3.623183e-71 | 3.307532e-76 | 0.005722 |
| 14 | Mut ARID1A | Mut ERBB2 | 11 | 4 | 3.154866e-76 | 5.014463e-57 | 0.169230 |
| 15 | Mut ARID1A | Mut TGFBR1 | 11 | 4 | 2.536698e-78 | 3.017138e-68 | 0.169230 |
| 16 | Mut AXIN2 | Mut TP53 | 5 | 4 | 1.318165e-07 | 9.078708e-69 | 0.012820 |
| 17 | Mut KRAS | Mut NRAS | 7 | 1 | 5.145750e-80 | 1.956351e-61 | 0 |
| 18 | Mut FAM123B | Mut SMAD4 | 8 | 4 | 4.124423e-51 | 1.977043e-66 | 0.003988 |
| 19 | Mut FAM123B | Mut SMAD2 | 8 | 2 | 6.316329e-75 | 8.035901e-78 | 0 |
| 20 | Mut PIK3CA | Mut ERBB2 | 5 | 4 | 3.178166e-07 | 1.408422e-40 | 0.012820 |
| 21 | Mut PIK3CA | Mut DKK4 | 5 | 2 | 4.828289e-53 | 2.649370e-75 | 0.028490 |
| 22 | Mut ATM | Del IGF2 | 6 | 1 | 4.687640e-78 | 1.112800e-60 | 0.222222 |
| 23 | Mut ERBB2 | Mut NRAS | 4 | 1 | 4.213949e-56 | 1.654684e-58 | 0 |
| 24 | Mut ERBB2 | Mut SMAD2 | 4 | 2 | 3.869181e-28 | 6.690900e-68 | 0 |
| 25 | Mut ERBB3 | Mut KRAS | 7 | 7 | 0.362016 | 4.184342e-77 | 0.004651 |
| 26 | Pat OR.ERBB2... | Mut LRP5 | 8 | 7 | 5.063771e-05 | 2.775605e-79 | 0.011391 |
| 27 | Pat XOR.ACVR1B | Mut TGFBR2 | 7 | 2 | 9.302768e-73 | 1.92682e-83 | 0 |

Table 5.2: *p-values* of MSI-HIGH with AIC as regulatizator.

# CHAPTER 6

## EVOLUTION OF COLORECTAL CANCER

In this Chapter, we present the design, development and evaluation of an optimal, versatile and modular pipeline which exploits state-of-the-art tools to extract ensemble-level cancer progression models from cross-sectional data. We also show its applications in interpreting colorectal cancer data which, because of its high levels of heterogeneity, may be thought of as one of the most challenging case studies. As a reference, see [21].

## 6.1 A pipeline to infer ensemble-level progression models

We have devised a customizable pipeline to infer ensemble-level cancer progression models from cross-sectional data with CAPRI [158]. To increase the statistical quality of its predictions this pipeline pre-processes data to diminish the confounding effects of inter and intra-tumor heterogeneity. At a high-level, one shall thus identify: (*i*) biologically meaningful subtypes with similar molecular profiles via tumor stratification, (*ii*) the set of driver alterations and (*iii*) the groups of fitness-equivalent (i.e., exclusive) alterations.

Specifically, let us consider $n$ tumors ($n$ patients) and assume the relevant (epi)genetic data to be available. We do not put constraints on data gathering and selection, leaving the user to decide the appropriate "resolution" of the input mutational data. For instance, one might decide whether somatic mutations should be classified by type, or aggregated. Or, one might decide to lift focal CNAs to the wider resolution of cytobands or full arms. These choices depend on data and on the overall understanding of such alterations and their functional effects for the cancer under study, and no single all-encompassing rationale may be provided.

Given these premises, the pipelines consists in the following steps.

**Step 1: Reducing inter-tumor heterogeneity by cohort subtyping.** One might wish to identify cancer subtypes in the *heterogeneous mixture* of input samples. In some cases the classification can benefit from clinical biomarkers, such as evidences of certain cell types [14], but in most cases we will have to rely on multiple *clustering* approaches at once, see, e.g., [137, 139].

Many common approaches cluster expression profiles [119] by relying on non-negative matrix factorization techniques [59] or earlier approaches such as $k$-means, Gaussians mixtures or hierarchical/spectral clustering - see the review in [34]. For glioblastoma and breast cancer, for instance, mRNA expression subtypes provides good correlation with clinical phenotypes [138, 102, 161]. However, this is not always the case as, e.g., in colorectal cancer such clusters mismatch with survival and chemotherapy response [138]. Clustering of *full exome* mutation profiles or smaller panels of genes might be an alternative as it was shown for ovarian, uterine and lung cancers [82, 199].

**Step 2: selection of driver events.** In subtypes detection, with more alterations available it becomes easier to find similarities across $n$ samples, as features selection gains precision. In progression inference, instead, one wishes to focus on $m \ll n$ driver alterations, which ensure also an appropriate statistical ratio between sample size ($n$) and problem dimension ($m$).

Multiple tools filter out driver from passenger mutations. MutSigCV identifies drivers mutated more frequently than background mutation rate, [106]. OncodriveFM, avoids such estimation but looks for functional mutations [66]. OncodriveCLUST scans mutations clustering in small regions of the protein sequence [176]. MuSiC uses multiple types of clinical data to establish correlations among mutation sites, genes and pathways [35]. Some other tools search for driver CNAs that affect protein expression [178]. All these approaches use different statistics to estimate signs of positive selection, and we suggest using them in an orchestrated way, as done in some platforms [70]. Notice that driver genes will likely differ across subtypes, mimicking the different molecular properties of each group of samples.

**Step 3: fitness equivalence of exclusive alterations.** When at the ensemble-level, identification of "groups of equivalent but alternative" mutually exclusivity alterations is crucial, prior to progression inference [158]. A plethora of tools can be used; greedy approaches [194, 129] or their optimizations, such as MEMO, which constrain search-space with network priors [28]. This strategy is further improved in MUTEX, which scans mutations and focal CNAs for genes with a common downstream effect in a curated signalling network, and selects only those genes that significantly contributes to the exclusivity pattern [8]. Other tools, instead, employ advanced statistics or generative approaches without priors [182, 198, 110, 112, 84, 175].

In the fitness equivalent groups, we distinguish between *hard* and *soft* exclusivity, the former assuming strict exclusivity among events, with random errors accounting for possible overlaps, the latter admitting co-occurrences. [8]. CAPRI is the only algorithm where relations among group of genes can be input as *"testable hypotheses"* via *logical Boolean formulas*. In this case, we can use logical connectives such as $\oplus$ (the logical "xor") as a proxy for hard-exclusivity, and $\vee$ (the logical "disjunction") as a proxy for soft-exclusivity[1]. For example, these can be used to test wether colorectal tumors "start" prevalently from $\beta$-catenin deregulation, i.e., APC $\vee$ CTNNB1 , and if they further progress exclusively ($\oplus$) through KRAS or NRAS alterations. In general, as this testing-feature leaves the inference *unbiased* - see [158] - arbitrary hypotheses on significantly mutated subnetworks could be considered as well [111, 181].

**Step 4: progression inference and confidence estimation.** Finally, CAPRI is used to reconstruct cancer progression models of each identified molecular subtype, provided that there exist a reasonable list of driver events and the groups of fitness-equivalent exclusive alterations.

As already discussed in §4 CAPRI's input is a binary $n \times (m + k)$ matrix $\mathbf{M}$ with $n$ samples, $m$ driver alteration events (Bernoulli 0/1 variables) and $k$ testable formulas. CAPRI first *scans pairwise* $\mathbf{M}$ to identify a set of $\mathcal{S}$ plausible selective advantage relations, which then reduces to the most relevant ones, $\mathcal{S}^* \subset \mathcal{S}$.

Construction of $\mathcal{S}$ depends on the number of *non-parametric bootstrap* iterations and confidence p-values for estimating selective advantage among input events $x$ and $y$. CAPRI postulates that "$x$ selects for $y$" if it estimates that "$x$ is earlier than $y$" and that "$x$'s presence increases the probability of observing $y$" [172]. These conditions are implemented with the following inequalities:

$$p(x) > p(y) \qquad\qquad p(y \mid x) > p(y \mid \neg x) \qquad (6.1)$$

for which we get p-values by Mann-Whitney U Testing. Here, $p(\cdot)$ is an empirical marginal probability, $p(\cdot \mid \cdot)$ is a conditional, and $\neg x$ is the negation of $x$.

Optimization of $\mathcal{S}$ is central to the tolerance to *false positives* and *negatives* in $\mathcal{S}^*$. CAPRI's implementation in TRONCO [4] selects from $\mathcal{S}$ a subset of relations by optimizing the *score with regularization*:

$$\mathcal{S}^* = \arg\min_{\hat{\mathcal{S}} \subset \mathcal{S}} \left\{ -2\log[\mathcal{L}(\hat{\mathcal{S}} \mid \mathbf{M})] + \theta|\hat{\mathcal{S}}| \right\} , \qquad (6.2)$$

where $\mathcal{L}(\cdot)$ is the *model likelihood*; the estimated optimal solution is $\mathcal{S}^*$.

Different values of $\theta$ lead to different tolerance to errors in $\mathcal{S}^*$, the *Akaike Information Criterion* (AIC) being for $\theta = 2$, the *Bayesian Information Criterion* (BIC) for $\theta = \log(n)$. Both scores are approximately correct; AIC is more prone to overfitting but likely to provide also good predictions from data and is better when false negatives are more misleading than positive ones. BIC is more prone to underfitting errors, thus is more parsimonious and better in opposite cases. As often done, we suggest to combine both approaches and distinguish which relations are selected by BIC or AIC.

Model confidence can be estimated with *non-parametric*, *parametric* or *statistical bootstrap* [47]. These procedures re-sample datasets to provide a confidence to every selective advantage relation and to the overall model. Bootstrapped datasets are randomly generated by re-shuffling data and seed (non-parametric), just seed (statistical) or by sampling from the model (parametric). CAPRI's other statistics include hypergeometric tests to assess how significant is the overlap between pairs of alterations.

---

[1]Logical disjunction of a set of operands is true if and only if *one or more* of its operands is true. For this reason, if we shall use that as a model of soft-exclusivity, we shall also check that the majority of observations indeed shows an exclusivity trend, meaning that few cases of co-occurent observations happen.

Thus, this pipeline is similar in spirit to those implemented by consortia such as TCGA to analyze huge populations of cancer samples collected [137, 139]. One of the main novelties of this approach, which is only possible by the specific features of hypothesis-testing provided by CAPRI [158], is the exploitation of groups of exclusive alterations as a proxy to detect fitness-equivalent routes of cancer progression. Thus, CAPRI is seen as an ideal tool for the efficient and theoretically-grounded investigations in population-based studies on cancer genomics.

This approach allows one to produce a progression model for virtually every cancer subtype identified in the input cohort, which shall be characteristic of the population trends of cancer initiation and progression. In the following, we empirically characterize the efficacy of the approach in processing colorectal cancer data from TCGA project [137], demonstrating that we were able to re-discover most of the existing body of knowledge about colorectal tumor progression or to propose further experimentally verifiable hypotheses[2].

## 6.2    Clonal evolution of MSI/MSS colorectal tumors

It is common knowledge that *colorectal cancer* (CRC) is a heterogeneous disease comprising different molecular entities. Since similar tumors are most likely to behave in a similar way, grouping tumors with homogeneous characteristics may be useful to define personalized therapies. Indeed, it is currently accepted that colon tumors can be classified according to their global genomic status into two main types: *microsatellite instable tumors* (MSI) and *microsatellite stable* (MSS) tumors (also known as tumors with chromosomal instability). This taxonomy plays a significant role in determining pathologic, clinical and biological characteristics of CRC tumors [146]. Thus, MSS tumors are characterized by changes in chromosomal copy number and show worse prognosis [125, 187]. On the contrary, the less common MSI tumors (about 15% of sporadic CRC) are characterized by the accumulation of a high number of mutations and show predominance in females, proximal colonic localization, poor differentiation, tumor-infiltrating lymphocytes and a better prognosis [183]. In addition, MSS and MSI tumors exhibit different responses to chemotherapeutic agents [99, 189]. Regarding molecular progression, it is also well established that each subtype arises from a distinctive molecular mechanism. While MSS tumors generally follow the classical adenoma-to-carcinoma progression (sequential APC-KRAS-TP53 mutations) described in the seminal work by Vogelstein and Fearon [51], MSI tumors results from the inactivation of DNA mismatch repair genes like MLH-1 [183].

We adopted the discussed pipeline to process MSI and MSS colorectal tumors collected from the The Cancer Genome Atlas project *"Human Colon and Rectal Cancer"*

---

[2]We remark that in-vitro and in-vivo experiments could provide an optimal validation for the newly suggested selective advantage relations and hypotheses, yet this is out of the scope of the current work.

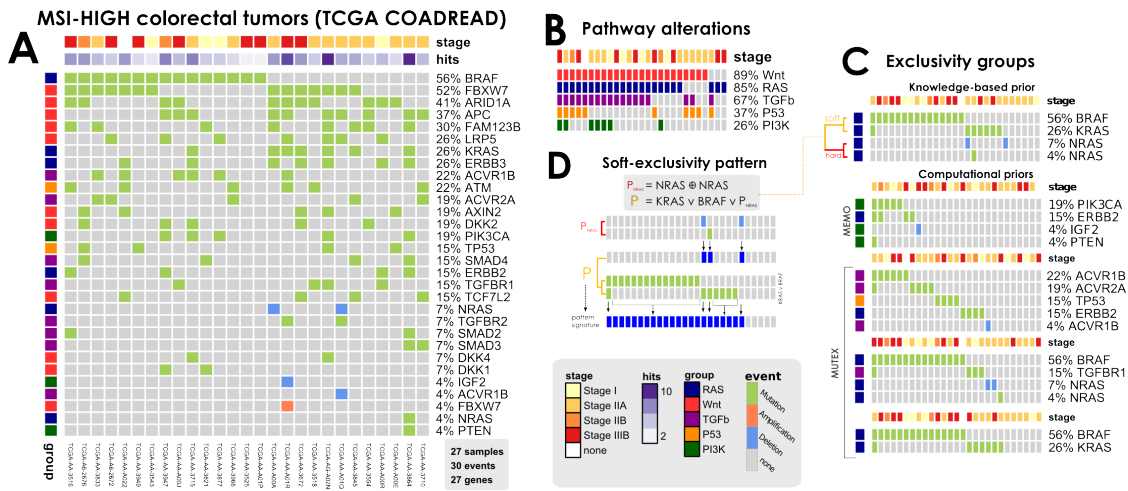

Figure 6.1: .**(A) Selected MSI-HIGH colorectal tumors used for the inference.** Data from the TCGA COADREAD project [137], restricted to 27 samples with both somatic mutations and high-resolution CNA data available and a selection out of 33 driver genes annotated to WNT, RAS, PI3K, TGF-$\beta$ and P53 pathways. This dataset is used to infer the model in Figure 6.3. **(B) Altered pathways.** Mutations and CNAs in these tumors mapped to pathways confirm heterogeneity even at the pathway-level. **(C) Mutually exclusive alterations.** Groups were obtained from [137] - which run the MEMO[28] tool - and by MUTEX[8] tool. Plus, previous knowledge about exclusivity among genes in the RAS pathway was exploited. **(D) Construction of a formula.** A Boolean formula inputed to CAPRI to test the hypothesis that alterations the RAF genes KRAS, NRAS and BRAF confer equivalent selective advantage. The formula accounts for hard exclusivity of alterations in NRAS mutations and deletions, jointly with soft exclusivity with KRAS and NRAS alterations.

(COADREAD, [137]) – also see §E. Details on the implementation are available in §E as well as source code to replicate this study. COADREAD has enough samples to implement a *training/test* statistical validation of these findings - see §E. In brief, we split subtypes by the microsatellite status of each tumor, and select somatic mutations and focal CNAs in 33 driver genes manually annotated to 5 pathways in [137] - WNT, RAS, RAF, TGF-$\beta$, PI3K and P53. Groups of exclusive alterations were scanned by MUTEX [8] (see §E), and fetched by [137] using the MEMO [28] tool; groups were used to create CAPRI's formulas, see §E. The data for MSI-HIGH tumors are shown in Figure 6.1, for MSS tumors see §E. CAPRI was run, on each subtype, by selecting recurrent alterations from the pool of 33 pathway genes and using both AIC/BIC regularizators. The model inferred for MSS tumors is in Figure 6.2, the model for MSI-HIGH ones is in Figure 6.3. Each edge in the graph mirrors selective advantage among the upstream and downstream nodes, as estimated by CAPRI from training datasets (statistics: $p < 0.05$, 100 non-parametric bootstraps); only the minimum amount of edges is selected to maximize the likelihood of data (see Online Methods). In Figure 6.3, for MSI-HIGH tumors, we show how the model predicts alternative routes to somatic evolution for COADREAD samples. As statistical validation of these models, we mark those relations that display significant $p$-values in the test datasets, and rank them if they contribute (or otherwise) to max-likelihood. For some edges it is not possible to provide a validation, as some upstream or downstream event may be missing in the test dataset, while other edges do not show statistical evidence in the test datasets.

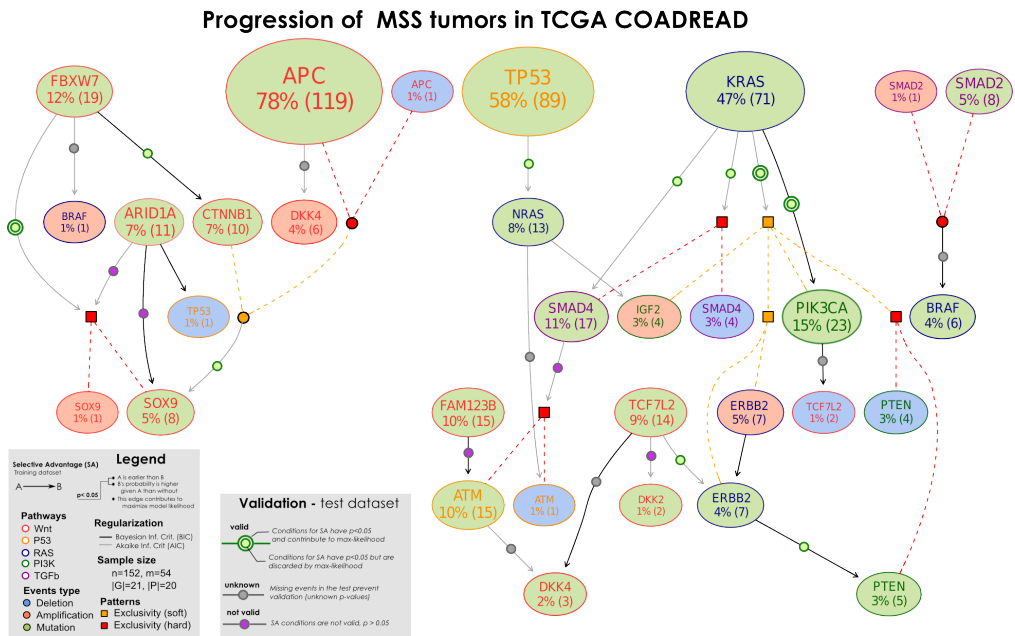

Figure 6.2: **Progression model of MSS colorectal tumors.** Selective advantage relations inferred by CAPRI constitute MSS progression; the input dataset are shown in §E. Formulas written on groups of exclusive alterations are expanded with colors representing the type of exclusivity (red for hard, orange for soft). We mark also those relations that display significant $p$-values in the test dataset, and rank them if they contribute (or otherwise) to max-likelihood. For all MSS tumors in COADREAD, we find at high-confidence selection of SOX9 alterations by FBXW7 mutations (with AIC), as well as selection of alterations in PI3K genes by the KRAS mutations (direct, with BIC, and via the MEMO group, with AIC).

**MSS (Microsatellite Stable).** In agreement with the known literature, the progression model identifies KRAS, TP53 and APC as primary events in the carcinogenesis, as well as NRAS and KRAS determining two independent evolution branches, the former being "selected by" TP53 mutations, i.e. being a downstream event in the model, the latter "selecting for" PIK3CA mutations. The leftmost portion of the model links many WNT genes, in agreement with the observation that multiple concurrent lesions affecting such pathway confer selective advantage. In this respect, the model predicts multiple routes to eventually select alterations in SOX9 gene, a transcription factor known to be active in colon mucosa [1]. Its mutations are indeed selected by APC/CTNNB1 alterations - with APC mutations in 78% of the tumors - or by FBXW7, an early mutated gene that both directly and in a redundant way via CTNNB1 selects for its mutations or amplifications. SOX family of transcription factors have emerged as modulators of canonical WNT/$\beta$-catenin signaling in many disease contexts, with evidences that multiple SOX proteins physically interact with $\beta$-catenin and modulate the transcription of WNT-target genes, as well as with evidences of regulating of SOX's expression by WNT resulting in feedback regulatory loops that fine-tune cellular responses to $\beta$-catenin/TCF activity [103]. Also interestingly, FBXW7 has been previously reported to be involved in the malignant transformation from adenoma to carcinoma [114], and it was recently shown that SCFFbw7, a complex of ubiquitin ligase that contains such gene, targets several oncogenic proteins including SOX9 for degradation [83]; this relation has high-confidence also in the test dataset. The rightmost part of the model involves genes from other pathways, and outlines the relation between KRAS and the PI3K pathway. We indeed find, consistently in the training and test, selection of PIK3CA mutations by KRAS ones, as well as selection of the whole MEMO module, which is responsible for the activation of the PI3K pathway [137]. Further, Mutations in PTEN (in hard exclusivity with its deletions) were found to be a late event acquired by a clone with KRAS mutations directly, or indirectly via ERBB2 mutations. SMAD4 deletions or mutations (in hard exclusivity) are selected from tumors harboring KRAS mutations. In a smaller group of tumors SMAD2 amplification (1%) or mutation (5%) selects for BRAF mutations. To highlight, a sub-group of tumors lacking APC, KRAS or TP53 but exhibiting mutations in FAM123B (10%) or mutations in TCF7L2 (9%) that later converge in DKK2 or DKK4 mutations. Interestingly, these four genes are implicated in the WNT signalling pathway. It is also worth pointing that the model predicts a selection trend among SOX9/ARID1A and ATM/FAM123B; however, for the corresponding events, in the test, we find the opposite model's direction, suggesting a potential confounding effect which can be inputed to these events having very similar frequencies in the training dataset.

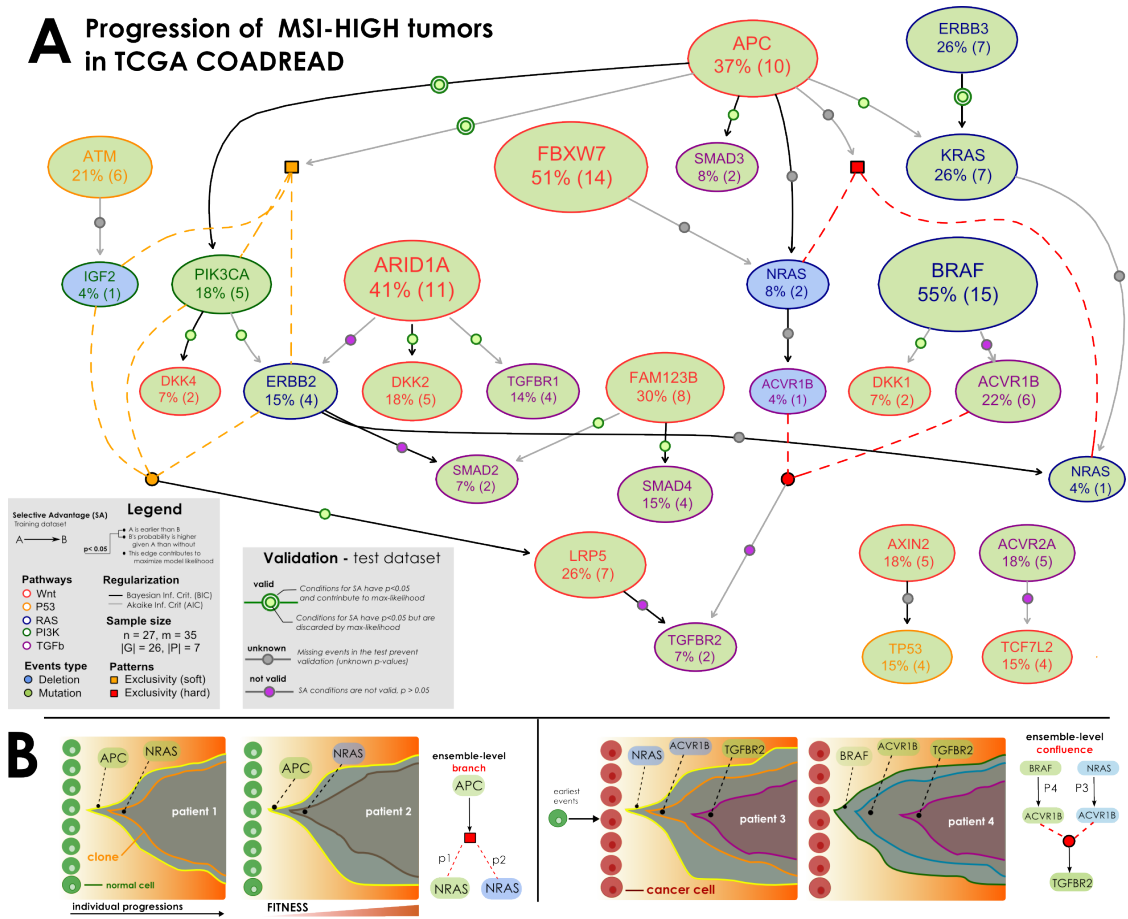

Figure 6.3: **(A) Progression model of MSI-HIGH colorectal tumors.** Selective advantage relations inferred by CAPRI constitute MSI-HIGH progression; input dataset in Figure 6.1. Formulas written on groups of exclusive alterations are expanded as in Figure 6.2. We note the high-confidence in APC mutations selecting for PIK3CA ones, both in training and test via BIC, as well as selection of the MEMO group (ERBB2/PIK3CA mutations or IGF2 deletions) predicted by AIC. Similarly, we find a strong selection trend among mutations in ERBB2 and KRAS. For each relation, we report its status in the test dataset, so to understand whether that is specific of all the COADREAD samples or not. **(B) Predicting clonal expansion from the model.** Evolutionary trajectories from two example selective advantage relations. APC-mutated clones shall enjoy expansion, up to acquisition of further selective advantage via mutations or homozygous deletions in NRAS. These cases should be representative of different individuals in the population, and the ensemble-level interpretation should be that "APC mutations select for NRAS alterations, in hard exclusivity" as no sample harbour both alterations. A similar argument can show that the clones of patients harbouring distinct alterations in ACVR1B - and different upstream events - will enjoy further selective advantage by mutating TGFBR2 gene.

**MSI (Microstaellite Instable).** In agreement with the current literature, BRAF is the most commonly mutated gene in MSI tumors (55%) [95]. CAPRI also predicted convergent evolution of tumors harbouring FBXW7 (51%) or APC (37%) mutations towards deletions of NRAS gene, as well as selection of SMAD2 or SMAD4 mutations by FAM123B mutations, for these tumors. Relevant to all MSI tumors seems again the role of the PI3K pathway. Indeed, a relation among APC and PIK3CA mutations was inferred with a high confidence in both training and test datasets, consistent with recent experimental evidences pointing at a synergistic role of these mutations, which co-occurr in the majority of human colorectal cancers [36]. Similarly, we find consistently a selection trend among APC and the whole MEMO module. Interestingly, both mutations in APC and ERBB3 select for KRAS mutations, which might point to interesting therapeutic implications. On the contrary, mutations in BRAF mostly selects for mutations in ACVR1B, a receptor that once activated phosphorylates SMAD proteins. It forms receptor complex with ACVR2A, a gene mutated in these tumors (18%) that selects for TCF7L2 mutations (also implicated in the progression of MSS tumors). Tumors harbouring TP53 mutations are those selected by exhibit mutations in AXIN2, a gene implicated in WNT signalling pathway, and related to instable gastric cancer development [96].

## 6.3 Inference of patient-specific clonal evolution

We also discovered that the CAPRESE [117] algorithm can be used to successfully reconstruct the clonal architecture in individual patients. This result is indicative of the power of the selective advantage scores à-la-Suppes [172], even outside the scope of cross-sectional data. We performed the analysis on data from Gerlinger *et al.*, which have recently used multi-region targeted exome sequencing ($> 70x$ coverage) to resolve the genetic architecture and evolutionary histories of ten *clear cell renal carcinomas* [62].

Besides quantification of intra-tumor heterogeneity, their work found that loss of the 3p arm and alterations of the Von Hippel-Lindau tumor suppressor gene VHL are the only events ubiquitous among their patients. In Figure 6.4 we show the clonal evolution estimated for one of those patients, RMH004, computed with CAPRESE (shrinkage coefficient $\lambda = 0.5$, time $< 1$ sec) from the Bernoulli 0/1 profiles provided in Supplementary Table 3 and Figure 4 of [62], with non-parametric bootstrap confidence (time $< 6$ sec). This model shall be compared to the one inferred by processing the region-specific VAF with a max-mini optimization of most parsimonious evolutionary trees [156], and performing selection-by-consensus when multiple optimal solutions exist - Supplementary Figure 9 in [62]. CAPRESE requires no arbitrarily defined curation criteria to select the optimal tree, as it constructively searches for a solution which, in this case, is analogous in suggesting *parallel evolution of subclones* via deregulation of the SWI/SNF chromatin-remodeling complex – i.e., as may be noted from multiple clones with distinct PBMR1 mutations. Finally, the approach in [156], estimates also the number of non-synonymous mutations acquired on a certain edge of the tree. While this model is silent about this, it is very likely due to the limitations imposed by the lower-resolution and small sample size of the data – 9 events from 8 regions, and not the VAFs for all alleles.

**Single-cell synthetic data.** We estimate the efficiency of our approach to single-cell sequencing data, as if it was collected from patient RMH004 (synthetic data generated from the clonal phylogeny architecture of Figure 6.4). To mimic a poor reliability of this technology, to each sampled cell a noise model which accounts for false positives and negatives in the calls of their genomic alterations is applied. Performance is measured as the fraction of true-positive and negative ancestry relations inferred among cells (*precision* and *recall*), as a function of the number of sequenced cells and noise level. Results evidence a very good performance even with very small number of cells and reasonable noise levels, hinting at a promising application with this technology. Complete details for synthetic data generation and further performance measures are provided in §E.

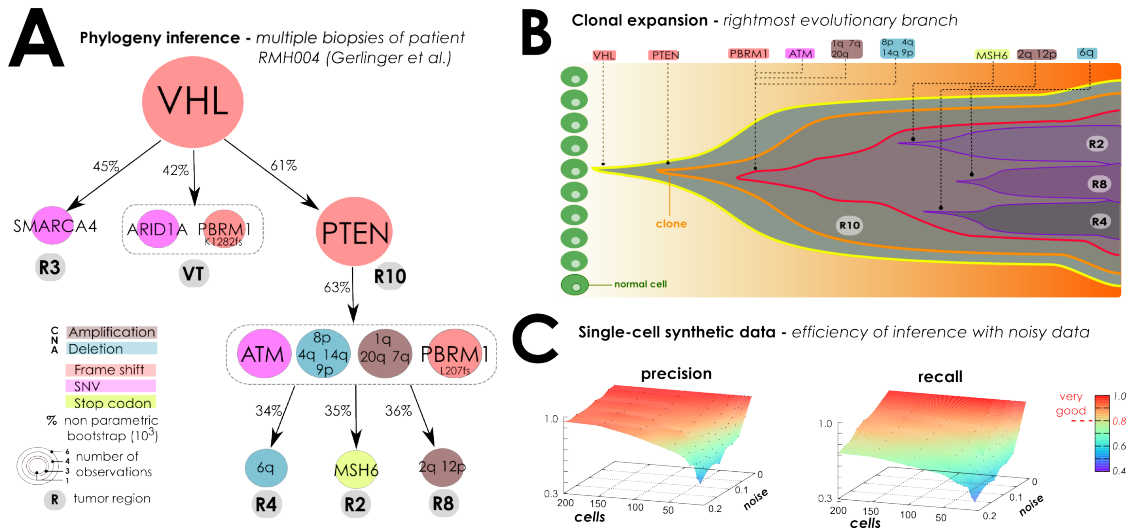

Figure 6.4: **(A) Application of ensemble-level algorithms to individual-patient data.** With data provided by Gerlinger *et al* in [62], we infer a patient-specific clonal evolution from 6 biopsies of a clear cell renal carcinoma (5 primary tumor, 1 from the thrombus in the renal vein, VT). Validated non-synonymous mutations are selected for VHL, SMARCA4, PTEN, PBMR1 (p.Lys1282fs and p.Leu207fs), ARID1A, ATM and MSH6 genes. CNAs are detected on 12 chromosomes. For this patient, both region-specific allele frequencies and Bernoulli profiles are provided. Thus, we can extract a clonal tree, signature and diffusion of each clone, by the unsupervised CAPRESE algorithm [117]. **(B) Clonal expansion in patient RMH004.** The unsupervised model inferred by CAPRESE predicts an analogous clonal expansion observed in [62], and extracted with most parsimonious phylogeny tree reconstruction from allelic frequencies, and hand-curated for selection of the optimal model. For simplicity, we show only expansion of the sub-clones harbouring PTEN's frame shift mutation. **(C) Inference from single-cell data.** We estimate average precision and recall from single-cell sequencing data sampled from the phylogeny history of patient RMH004 (details in §E). Sampled datasets vary for number of sequenced cells, $n \leq 200$, and noise in the data - as a model of potential experimental errors in data collection, manipulation and analysis.

CHAPTER 7 ⎯⎯⎯⎯⎯⎯⎯⎯⎯⎯⎯⎯⎯⎯⎯⎯⎯⎯⎯⎯⎯⎯

CONCLUSIONS

The *reconstruction of cancer progression models* is a pressing problem, as it promises to highlight important clues about the evolutionary dynamics of tumors and to help in better targeting therapy to the disease (see e.g., [118]). In the absence of large longitudinal datasets, progression extraction algorithms rely primarily on *cross-sectional* input data, thus complicating the statistical inference problem.

In this thesis, we explore the nature of somatic evolution in cancer. The proposed model of somatic evolution not only supports the heterogeneity and temporality seen in tumor clonal populations, but it also suggests a selectivity/causality relation that can be used in analyzing (epi)genomic data and exploited in therapy design.

Although strongly supported by empirical studies using synthetic as well as experimental genomic data, the contributions of this thesis are primarily methodological, with the aim of attaching the problem of understanding cancer evolution rigorously, by means of strong foundations built upon the sound theory of probabilistic causality, originally proposed by Patrick Suppes [172] (see §2).

With these premises, we devise efficient approaches, namely *CAPRESE* and *CAPRI* algorithms, which for the first time algorithmize Suppes' formulation to reconstruct respectively tree and directed acyclic graph models of cancer progression (see §3 and §4). These methods, while taming their efficiency satisfactorily, even for many complex situations that are specifically important in cancer studies (e.g., synthetic lethality or oncogene addiction), are kept computational efficient. Furthermore, the aforementioned algorithms along with a set of utilities to support the scientist in his effort of understanding cancer progression are implemented in the *TRONCO R* package (see §5).

Additionally, the framework is shown to be effective in extracting evolutionary trajectories for cancer progression both at the level of populations and individual patients. In the former case a pipeline to minimize the confounding effects imputable to tumor heterogeneity is proposed and applied to a highly-heterogeneous cancer such as colorectal. In the latter we have also shown how the framework can be readily applied to reconstruct clonal phylogeny from multi-sample data, with an application to clear renal cell carcinoma (see §6).

## 7.1 Future works

Despite in the last two decades many specific genes and genetic mechanisms that are involved in different types of cancer have been identified, our understanding of cancer progression is still largely elusive and faces fundamental challenges. Hence, several future research directions are possible.

In particular, the various steps described in the pipeline need further improvements with specific reference to the clustering in subtypes: while a series of state-of-the-art techniques to tackle the problem exist, their results are not always satisfactory and, in fact, in most studies, the identification of subtypes is not clear or it is done manually (see e.g., [137]).

Moreover, with the objective of anticipating the forthcoming technologies, a further development of the proposed framework could involve the introduction of *timed data*, in order to extend the techniques to settings where a temporal information on the samples is available, hence needs not be imputed.

In the current version, a progression model is reconstructed in terms of event (i.e., genetic alterations) at a lower grain than the actual progression in terms of functions (i.e., hallmarks). One further improvement would be to link the inferred models to hallmarks or pathways. This prior knowledge could also be exploited to improve the reconstructions themselves.

Another interesting research direction would be to exploit the reconstructions by *CAPRESE* or *CAPRI* for subsequent quantitative estimations of the progression of the disease such as the waiting time between the occurrence of difference alterations and the survival time of cancer patients.

Finally, as observed multiple times through the thesis, the proposed framework is agnostic of the input data and, hence, it could be adopted also in different contexts. As briefly done in §F with the task of discrimination discovery from data, it would be interesting to explore how the framework proposed in this thesis can be used in different fields of data mining.

FOUNDATIONS OF CAUSATION

In this Chapter, we give an outline of the current state-of-the-art theories of causation, which enjoys a long and colorful history, starting with the work of Aristotle and, more recently, of Avicenna circa 1000 AD. However, we restrict our description only to the main ideas and limitations of these theories, as a more detailed discussion of various topics related to these theories is available elsewhere (see [88] or [80]).

The biological notion of causality proposed in this thesis is firmly grounded on the notions of *Darwinian evolution*: in that, it is about an *ensemble of entities* (e.g., population of cells, organisms, etc.). Within this ensemble, a causal event (say $c$) in a member entity may result in variations (changes in genotypic frequencies); such variations are exhibited in the phenotypic variations within the population, which is subject to Darwinian positive (and subsequently, Malthusian negative) selections, and sets the stage for a new effect event (say $e$) to be selected, should it occur next; we then conclude that "$c \triangleright e$."

While there could be other meaningful extensions of this framework (see [75])[1], we believe that it suffices in describing the causality relations implicit in the somatic evolution responsible for tumor progression. Note further that by its very statistical nature, we capture just those relations that only reflect *"Type-level Causality"*, and relegate *"Token-level Causality"*, – a more nuanced concept – to the future research. Thus, note that, while our framework can estimate for a population of cancer patients of a particular kind (say atypical Chronic Myeloid Leukemia, aCML, patients) whether and with what probability a mutation (such as SETBP1) would cause certain other mutations (such as ASXL1 single nucleotide variants or *in-del*) to occur, it will remain silent as to whether a particular ASXL1 mutation in a particular patient was caused by an earlier SETBP1 mutation.

Based on the afore-mentioned biological framework, we will focus primarily on how to devise efficient and accurate algorithms for extracting causal relations from the patient genomic data; we leave it to the readers to intuit how an inferred causal relation may be verified/refuted by *in vitro* or *in silico* experiments and how it could be used in therapy design that would guide the clocks involved in cancer's natural somatic evolution (more details are forthcoming).

---

[1] Also see the debate between Fisher and Wright in response to Fisher's *fundamental theorem of genetics*.

## A.1   Hume's regularity theory

The modern study of causation begins with the Scottish philosopher David Hume (1711-1776). According to Hume, a theory of causation could be defined axiomatically, using the following ingredients: *temporal priority*, implying that causes are invariably followed by their effects [87], augmented by various constraints, such as contiguity, constant conjunction[2], etc. Theories of this kind, that try to analyze causation in terms of invariable patterns of succession, have been referred to as *regularity theories* of causation.

Nonetheless, the notion of causation has spawned far too many variants and has been a source of acerbic debates. All these theories present well-known limitations and confusion, but have led to a small number of modern versions of commonly accepted (at least among the philosophers) frameworks. See the theories discussed and studied by Suppes et al. §A.2, Lewis et al. §A.3, and Pearl et al. §A.3.1. One of the most prominent among these is Suppes' *probabilistic* causation, whose axioms are expressible in *probabilistic propositional modal logics*, and amenable to algorithmic analysis. It is the framework upon which we build our analyses and algorithms.

We will momentarily discuss the main limitations of regularity theories [80], in order to better prepare the reader for the subsequent discussions of these theories and the algorithms to which they lead. Thus, the next three sections will focus on two issues: (*i*) how the state-of-the-art theories of causation have attempted formulating a *sound and complete* theory of causation, as well as (*ii*) what unsolved problems in this framework still remain open.

**Imperfect regularities.**   In general, one cannot state that causes are *invariably* (i.e., without fail) followed by their effects. For example, while we may state that "smoking is a cause of lung cancer", we do grant that there would be still some smokers who do not develop lung cancer.

Situations such as these are referred to as *imperfect regularities*, and could arise for many different reasons. One of these – which is a very common situation in the context of cancer – involves the heterogeneity of the situations in which a cause resides. For example, some smokers may have a genetic susceptibility to lung cancer, while others do not; moreover, some non-smokers may be exposed to other carcinogens, while others are not. Thus, the fact that not all smokers develop lung cancer can be explained in these terms.

---

[2]Some of these notions have been modernized with the introduction of the machinery from statistical inference, logic and model theory; but they have stayed more or less true to Hume's programme.

**Irrelevance.** An event that is invariably followed by another, can be irrelevant to it. Consider the example in [104]: salt that has been hexed by a sorceror invariably dissolves when placed in water, but hexing does not cause the salt to dissolve. In fact, hexing is irrelevant for this outcome. Probabilistic theories of causation capture exactly this situation by requiring that causes alter the probabilities of their effects, see §A.2.

**Asymmetry.** If we claim that an event $c$ causes another event $e$, then, typically, we would anticipate being able to claim that $e$ does not cause $c$, which would naturally follow from a strict temporal-priority-constraint: *cause precedes effect temporally*. In the context of the preceding example, smoking causes lung cancer, but lung cancer does not cause one to smoke.

**Spurious regularities.** Consider a situation – not very uncommon – where a unique cause is regularly followed by two or more effects. As an example, suppose that one observes the height of the column of mercury in a particular barometer dropping below a certain level. Shortly afterwards, because of the drop in atmospheric pressure (the unobserved cause for falling barometer), a storm occurs. In this settings, a regularity theory could claim that the drop of the mercury column causes the storm when, indeed, it is only correlated to it. Following common terminologies, we will say that such situations are due to *spurious correlations*. There now exists an extensive literature discussing such subtleties that are important in understanding the philosophical foundations of causality theory; see [80].

## A.2 Probabilistic theories of causation

In this Section we will introduce the notion of *probabilistic causation*. The basic idea behind these theories is that "causes alter the probabilities of their effects;" see [80, 98] for more details.

### Suppes' prima facie cause

Patrick Suppes proposed the notion of a *prima facie cause* that represents the core of *probabilistic causation* and also provides the algorithmic foundations of our analysis.

**Definition 8** (Probabilistic causation, [172])**.** *For any two events $c$ and $e$, occurring respectively at times $t_c$ and $t_e$, under the mild assumptions that $0 < \mathcal{P}(c), \mathcal{P}(e) < 1$, the event $c$ is called a* prima facie cause *of $e$ if it occurs* before *and* raises the probability *of $e$, i.e.*

$$t_c < t_e \quad and \quad \mathcal{P}(e \mid c) > \mathcal{P}(e \mid \bar{c}). \tag{A.1}$$

From now on, the first condition will be referred to as *temporal priority*, whereas the second as *probability raising*. This notion of causation has some advantages over the simplest version of a regularity theory of causation, e.g., it deals with various issues usually associated with imperfect regularities (§A.1).

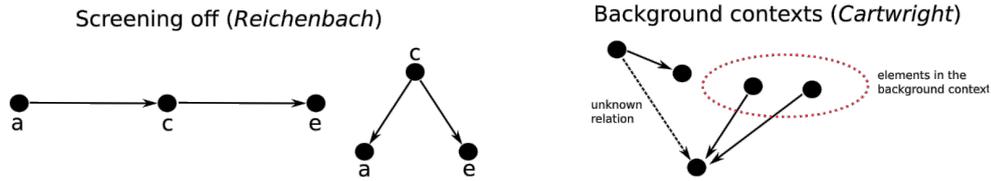

Figure A.1: **Example of screening-off and of background context.** (*left*) Example of Reichenbach's screening-off where $c$ is a genuine cause of $e$ and $a$ is a genuine cause of $c$, and the correlations between $a$ and $e$ are only just manifestations of these known causal connections, and $c$ is a common cause of both $a$ and $e$, that is exactly the situation of spurious correlation described in §A.1. (*right*) Example of Cartwright's background context.

Unfortunately, however, prima facie causality is still *not sufficient* in capturing a causation relationship *in its full generality*. For instance, the problem of spurious regularities still remains, additionally requiring that prima facie causes be refined further into two classes: *genuine* and *spurious*. In the latter case, as discussed, we may observe a prima facie cause to be so labeled only because of spurious correlations. Also, as discussed extensively in the literature (see [80]), one may encounter certain situations, in which Suppes' characterization fails to provide a *necessary* condition. In the next two paragraphs, we will briefly discuss an attempt to make Suppes' conditions sufficient for any causal claims, and another to determine when it is not necessary.

### *Reichenbach's screening-off*

In [160], Reichenbach discussed the notion of *screening-off* to describe a particular type of probabilistic relationship. Consider, e.g., events $a$, $c$ and $e$, and assume to observe $\mathcal{P}(e \mid a \wedge c) = \mathcal{P}(e \mid c)$, then we say that $c$ is *screening $a$ off* from $e$. When $\mathcal{P}(e \wedge c) > 0$, this is equivalent to stating that $\mathcal{P}(a \wedge e \mid c) = \mathcal{P}(a \mid c) \cdot \mathcal{P}(e \mid c)$ – i.e., $a$ and $e$ happen to be probabilistically independent, when conditioned upon $c$. The preceding situation could occur in two cases, see Figure §A.1.

In the first case, $c$ is a genuine cause of $e$ while $a$ is a genuine cause of $c$ as well, and the correlations between $a$ and $e$ are only just manifestations of these known causal connections. For example, unprotected sex ($a$) appears to cause AIDS ($e$) *only* because of sexually transmitted HIV infection ($c$). Then, we would expect that among those who have already been infected with HIV, the probability of contacting AIDS would be the same regardless of whether one is engaged in unprotected sex or not. Here $c$ is a *proximate* cause of $e$ and an *intermediate* cause leading from $a$ to $e$, i.e. an instance of *causal transitivity*. In the second case, $c$ is a common cause of both $a$ and $e$, that is exactly the situation of spurious correlation described in §A.1.

Building upon this idea, Reichenbach formulated the so-called *Common Cause Principle* (CCP) to detect situations leading to "screening-off," and so identify when a spu-rious correlation can be explained in terms of a common cause. Unfortunately, there are situations where such a principle leads to computationally intractable criteria. Since, these issues are not germane to the context of this work, we will not discuss them further, other than pointing the interested readers to appropriate literature [80]. Nevertheless, the idea of screening-off has significantly influenced some of the most widely-used recent theories of causation, and has become central to the topic.

### Simpson's paradox and Cartwright's background context

Up to now, we have discussed the *sufficiency* (or lack of it) of the characterization for causality provided in the Reichenbach-Suppes framework. Conversely, we may also examine those situations where this framework also fails to give all the *necessary* conditions for a causal claim. For example, consider smoking as a cause of lung cancer. But, examine in details a situation where it so happens that smoking is highly correlated with living in the country: those who live in the country are much more likely to smoke than those who do not. Suppose now that city pollution is a second cause of lung cancer, which happens to be a much stronger cause than smoking. Consider now the problem of causal claims on the combination of these two heterogenous populations: including those who live in the country and those who do not. Then, an analysis of those two populations in combination may falsely lead to the conclusion that smokers are, over all, less likely to suffer from lung cancer than non-smokers. This example is an instance of the so-called *Simpson's Paradox*, which has been discussed extensively by various philosophers (see Nancy Cartwright [22] and Brian Skyrms [170]).

Cartwright and Skyrms introduced the concept of *background contexts* to explain and correct this problem. Let us call the set of all the factors that are causes of the event $e$ (a factor can be an *atomic* event but it can also be *the composition of a set* of events), but are not caused by the event $c$, the set of *independent causes* of $e$. A background context for a causal relationship from $c$ to $e$ is the maximal conjunction of factors, each of which is either an independent cause of $e$, or the negation of an independent cause of $e$ (as shown in Figure §A.1). We will denote by variables $b_1, \ldots, b_n$ all the background contexts of a causal relationship. According to Cartwright then, $c$ causes $e$ if and only if $\mathcal{P}(e \mid c \wedge b_i) > \mathcal{P}(e \mid \bar{c} \wedge b_i)$, that is if $c$ raises the probability of $e$ in every background context $b_i \in B$. Skyrms proposed a slightly weaker condition: a cause must raise the probability of its effect in at least one background context, without lowering it in any other.

### Eells' taxonomy

Cartwright defined a cause in terms of raising the probability of its effect. But there are other possible probabilistic relations between $c$ and $e$, as described, for instance, by Eells, who proposes the following taxonomy [44]: (*i*) $c$ is a *cause* of $e$ if and only if it raises its probability in every background context $B$, (*ii*) $c$ is an *inhibition* for $e$ when it lowers such a probability, (*iii*) $c$ is *causally irrelevant* to $e$ when it does not change it and, finally, (*iv*) $c$ is a *mixed cause* of $e$, otherwise.

This thesis will adhere to the basic idea of a cause being a *probability-raiser* of its effect and ignore for the time being all other variants. According to Suppes' probabilistic theories of causation, we can evaluate a causal claim in terms of Definition §8, further augmented by the ideas of screening-off and background contexts; the same algorithmic, inferential and logical tools that we propose here can be used *mutatis mutandis*, should a user wish to explore a variant framework leading to a different axiomatic formulation of causation – provided its expressivity is limited to a probabilistic propositional modal logic – as seems the case to be.

### Issues of probabilistic causation

Next we describe some thorny issues in the theory of probabilistic causation. We also briefly point out some unresolved problems, proposed plans of attack, and ensuing criticisms. For a deeper discussion see [80].

**Pearl's criticism.**    In [150], Pearl argues that the notion that causes "raise the probabilities" of their effects *cannot be expressed in the language of probability theory*. In particular, according to Pearl, the inequality $\mathcal{P}(e \mid c) > \mathcal{P}(e \mid \bar{c})$ fails to capture the intuition behind probability raising, which must be *manipulative* or *counterfactual*. Because of this limit, Pearl argues that it is not possible to rigorously describe the intuitions behind the probability raising theory and, for this reason, the only way to properly assess a causal claim is exclusively by *intervention*. The methods described in this thesis are not negated by these arguments as our model reasons about an ensemble (tumor with heterogeneous cell-types) and type-level causality. Pearl's theory is discussed further in §A.3.1.

**Determining the background context.**    As described, the background contexts of a claim are all the factors causally relevant to the effect, but not to the cause. This assumption appears to prevent Cartwright's theory from being a reductive analysis of causation. In fact, the theory appeals to causal relations to define a set of probabilistic constraints on the possible causal claims compatible with the observations in terms of probabilities. In any case, even if there is no reduction of causation to probability, in practice, it can be difficult (or algorithmically complex) to determine the background contexts without knowing the causal topology in advance. Unfortunately, this argument introduces an unavoidable *circularity*.

## A.3    Counterfactual theories of causation

Here we present a brief discussion of [127] theories of causation where the meaning of causal claims is explained in terms of a *possible-world semantics* and counterfactual conditionals of the form: *had c not occurred, e would not have occurred either*. For detailed discussions see [127].

### Lewis's counterfactuals

The most complete known counterfactual theory of causation is due to David Lewis [113] and exploits a possible world semantics to state truth conditions for counterfactuals in terms of *similarity* among possible worlds: one possible world is closer to actuality than another, if it is more similar to the actual world.

Following this idea, Lewis defined two important constraints on the resulting similarity relation: (*i*) similarity induces an ordering of worlds in terms of closeness to the actual world and (*ii*) the actual world is the closest possible world to actuality. Then, the evaluation of the counterfactual *"if c were the case, e would be the case"* is true just in case it is closer to actuality to make the first term true along with the second – as opposed to making it true without. Therefore, in terms of counterfactuals Lewis defines the following notion of causality: given *c* and *e*, whether *e* occurs or not depends on whether *c* occurs or not, and *e* causally depends on *c* if and only if, if *c* were not to occur *e* would not occur. Thus, the idea of cause is conceptually linked to the idea of *something that makes a difference,* and this concept in turn is naturally described in terms of counterfactuals. Lewis also characterized causation in terms of temporal direction by stating that the direction of causation is the direction of causal dependence and that, typically, events causally depend on earlier events but not on later ones.

**Causal Chains.** In [113], Lewis states that causal dependence between events is *sufficient but not necessary*, i.e., it is possible to have causation without causal dependence. Consider, e.g., when *c* causes *d*, which in turn causes *e*; Lewis argues that *c* must cause *e* as well by means of a transitivity. However, since causal dependence is not transitive as would be the case for causation according to Suppes, the causal relation between *c* and *e* may not be evident. To overcome this problem, Lewis defines a causal chain as the finite sequence of events *c*, *d* and *e* and defines that *c is a cause of e if and only if there exists a causal chain leading from c to e*.

## Issues of counterfactual causation

We briefly describe some issues inherent to these theories; for a deeper discussion, see [127].

**Context-sensitivity.** Lewis's theory assumes that causation is an absolute relation, whose nature does not vary from one context to another. This approach has recently been criticized since it often leads to absurd results [127], as demonstrated by various easy-to-construct counter-examples.

**Transitivity and Preemption.** As discussed above, Lewis incorporates transitivities in his notion of causation by defining them in terms of chains of causal dependence. The transitivity of causation is sound in some contexts, but a number of counter-examples has been shown to cast doubts on this interpretation of causation [127]; the debate surrounding the transitivity of causation is unlikely to be easily settled. Nevertheless, in this work we aim at inferring *minimal models of causation*, in which each cause is sufficient for its child to occur. For this reason, we have opted to remove transitivities.

### A.3.1   Manipulability theories of causation

We now briefly discuss the notion of *intervention* as propounded by Judea Pearl [150]; in general interventionist versions of manipulability theories can be seen as counterfactual theories. For a detailed discussion on this and manipulability theories of causation refer to [192].

   Pearl characterizes his notion of intervention in terms of a primitive notion of causal mechanism. According to him, the world is organized in the form of stable mechanisms (i.e. physical laws) which are autonomous. Therefore, he states that we can change one of them, without changing all the others. Thus an intervention may imply that: *if we manipulate c and nothing happens, then c cannot be cause of e, but if a manipulation of c leads to a change in e, then we know that c is a cause of e, although there might be other causes as well.*

   In other words, when among many events a causal relationship between some $e$ and its parents (i.e. directed causes, say $c_1$, …, $c_n$) is present, the interventions will disrupt completely the relationships between $e$ and $c_1$, …, $c_n$ such that the value of $e$ is determined by the intervention only. Thus, intervention is a surgical operation in the sense that no other causal relationship in the system are changed by it. Hence, Pearl's assumption is that the other variables that change in values under this intervention will do so *only because they are effects of e*. Going back to the barometer example of §A.1: observing the drop of the mercury column increases the probability of a storm coming, but if we manipulate the drop of the mercury column by intervention such that its drop is caused by the intervention only, then we will be able to qualify barometer as a cause of storms instead of the drop itself. Pearl's theory has been very influential among the computational causality theorists, and has generated state-of-the-art algorithms for causal network inference, which we shortly present in §B.

### Issues of interventionist causation

Next, we point the reader to some problems that can arise in practice, when applying intervention in the context of causal inference. For a deeper discussion we refer to [192].

**Circularity.**   An intervention on an event $e$ leaves intact *all* the other *causal* mechanisms besides the ones involving $c$ as a cause. Because of this, Pearl's intervention could lead to circularity problems, i.e., it seems that the causal mechanisms need to be known in advance in order to asses them.

**Possible and impossible interventions.**   Causal claims are described in terms of counterfactuals of what would happen when applying intervention to a given causal relationship. Moreover, the notion of intervention is connected with the possibility of a *human action* to intervene in a system. In some contexts, however, it may be *impossible* to evaluate what would happen by performing a *surgical* intervention. Thus, it should be clear that, regardless of the possible criticisms to Pearl's framework, there are situations where, at least relative to the current human capabilities, it is very complicated, if not impossible, to perform intervention.

## A.4 A simplified framework

In conclusion, each of the existing theories faces various difficulties, which are rooted primarily in the attempt to construct a framework in its full generality: *each theory aims to be both necessary and sufficient for any causal claim, in any context.* In contrast, this thesis simplifies the problem by breaking the task into two: first, define a framework for Suppes' prima facie notion though it admits some spurious causes, but then deal with spuriousness by using a combination of tools, e.g., Bayesian, empirical Bayesian, regularization, which we recall in §B. The framework is based on a set of conditions that are *necessary even though not sufficient* for a causal claim, and is used to refine a prima facie cause to either a genuine or a spurious cause (or even ambiguous ones, to be treated as plausible hypotheses which can be refuted/validated by other means), see §2.

**Statement of assumptions.** Along with the described interpretation of causality, through out this document, we make following simplifying assumptions to guarantee the convergence of the method:

(*i*) *All causes involved in cancer can be expressed by monotonic Boolean formulas*: i.e., all causes are positive and can be expressed in CNF where all literals occur only positively. The size of the formula and each clause therein are bounded by small constants.

(*ii*) *All events are persistent*: i.e., once a mutation has occurred, it cannot disappear. Hence, we do not model situations where $\mathcal{P}(e \mid c) < \mathcal{P}(e \mid \bar{c})$.

(*iii*) *Closed world*: all the events which are causally relevant for the progression are observable and the observation can significantly describe the progressive phenomenon.

(*iv*) *Relevance to the progression*: all the events have probability strictly in the real open interval $(0, 1)$, i.e. it is possible to asses if they are relevant to the progression.

(*v*) *Distinguishability*: no two events appear equivalent, i.e. they are neither both observed nor both missing *simultaneously.*

We conclude by observing and stressing that the assumptions of above are to be intended as theoretical and, given them, we can prove asymptotic convergence of the framework. Nevertheless, as described along the thesis, there can be practical examples were these assumptions may not strictly be true and, yet, the approach still provides valid insights.

LEARNING BAYESIAN NETWORKS

In this Chapter we briefly discuss the notion of *Bayesian Network* (BN) and how to learn both its parameters and structure *ab initio*, with no prior knowledge. For a detailed discussion on the topic, refer to [101, 151]. This Section is intended to be accessible to a non-technical audience, although citations are provided for technical resources on each algorithm discussed.

## B.1 Preliminaries

A BN is a statistical model that succinctly represents a *joint distribution* over $n$ variables and encodes it in a *direct acyclic graph* over $n$ nodes (one per variable)[1]. In BNs, the full joint distribution can be written as a product of conditional distributions on each variable. An edge between two nodes $A$ and $B$ denotes statistical dependence, $\mathcal{P}(A \wedge B) \neq \mathcal{P}(A)\mathcal{P}(B)$, no matter on which other variables we condition on (i.e., for any other set of variables $\mathcal{C}$ it holds $\mathcal{P}(A \wedge B \mid \mathcal{C}) \neq \mathcal{P}(A \mid \mathcal{C})\mathcal{P}(B \mid \mathcal{C})$. In such a graph, the set of variables connected to a node $X$ determines its set of "parent" nodes $\pi(X)$. Note that a node cannot be both ancestor and descendant of another node, as this would cause a directed cycle.

Also, the joint distribution over all the variables can be written as $\prod_X \mathcal{P}(X \mid \pi(X))$. Of course, if a node has no incoming edges (i.e., no parents), we simply use its marginal probability $\mathcal{P}(X)$. Thus, to compute the probability of any combination of values over the variables, we need only parameterize the conditional probabilities of each variable given its parents. If the variables are binary, the number of parameters in each conditional probability table is locally of exponential size: namely, $2^{|\pi(X)|} - 1$. Thus, the total number of parameters needed to compute the full joint distribution is only of size $\sum_X 2^{|\pi(X)|} - 1$, which is considerably less than $2^n - 1$ for sparse networks.

---

[1] In the setting of this thesis, each variable is a modeled event and, for consistency with the BN notation, we will denote these as capital letters in this section.

A useful property of the graph structure is that we can define, for each variable, a set of nodes called the *Markov blanket* so that, conditioned on it, this variable is independent of all other variables in the system. It can be proven that for any BN, the Markov blanket consists of a node's parents, children as well as the parents of the children.

The usage of the symmetrical notion of conditional dependence introduces important limitations of structure learning in BNs. In fact, note that edges $A \rightarrow B$ and $B \rightarrow A$ denote equivalent dependence between $A$ and $B$, thus distinct graphs model the exact same set of independence and conditional independence relations. This yields the notion of *Markov equivalence class* as a *partially directed acyclic graph*, in which the edges that can take either orientation are left undirected. A theorem proves that two BNs are Markov equivalent when they have the same *skeleton* and the same *v-structures*, the former being the set of edges, ignoring their direction (e.g., $A \rightarrow B$ and $B \rightarrow A$ constitute a unique edge in the skeleton) and the latter being all the edge structures in which a variable has at least two parents, but those do not share an edge (e.g., $A \rightarrow B \leftarrow C$)[2] [18].

BNs have an interesting relation to canonical boolean logical operators $\wedge$, $\vee$ and $\oplus$ and formulas over variables. In fact these formulas, which are "deterministic" in principle, in BNs are naturally softened into *probabilistic relations* to allow some degree of uncertainty or noise. This probabilistic approach to modeling logic allows representation of qualitative relationships among variables in a way that is inherently robust to small perturbations by noise. For instance, the phrase *"in order to hear music when listening to an mp3, it is necessary and sufficient that the power is on and the headphones are plugged in"* can be represented by a probabilistic conjunctive formulation that relates power, headphones and music, in which the probability that music is audible depends only on whether power and headphones are present. On the other hand, there is a small probability that the music will still not play (perhaps we forgot to load any songs into the device) even if both power and headphones are on, and there is small probability that we will hear music even without power or headphone (perhaps we are next to a concert and overhear that music).

Note that in this review, we only consider the subset of networks that have discrete random variables that are visible. Networks with latent and continuous variables present their own challenges, although they share most of the mathematical foundations discussed here.

## B.2 Approaches to learn the structure of a BN

In the the literature, there have been two initial families of methods aimed at learning the structure of a BN from data. The methods belonging to the first family seek to explicitly *capture all the conditional independence relations* encoded in the edges, and will be referred to as *constraint based approaches* (§B.2.1). The second family, that of *score based approaches* (§B.2.2), seeks to choose a model that *maximizes the likelihood of the data* given the model. Since both the approaches lead to intractability (NP-hardness) [26, 27], computing and verifying an optimal solution is impractical and, therefore, heuristic algorithms have to be used, which only sometimes guarantee optimality. Recently, a third class of learning algorithms that takes advantage of *specialized logical relations* (mentioned in the previous section) have been introduced (§B.2.3). In the rest of this section we describe in detail some of these approaches.

---

[2]In BN terminology, parent $A$ and $C$ are considered "unwed parents." For this reason, the *v-structure* is often called an immorality or an *unshielded collider*.

## B.2.1   Constraint based approaches

We present an intuitive explanation of several common algorithms used for structure discovery by explicitly considering conditional independence relations between variables. For more detailed explanations and analyses of complexity, correctness and stability, refer to the related references.

The basic idea behind all algorithms is to build a graph structure reflecting the independence relations in the observed data, thus matching as closely as possible the empirical distribution. The difficulty in this approach lies in the number of conditional pairwise independence tests that an algorithm would have to perform to test all possible relations. This is indeed *exponential* requiring to condition on a power set, when testing for the conditional independence between two variables. This inherent intractability requires the introduction of *approximations*.

Here, we focus on two specific constraint based algorithms, the *PC algorithm* [171] and the *Incremental Association Markov Blanket* (IAMB, [179]), because of their proven efficiency and widespread usage. In particular, the PC algorithm solves the aforementioned approximation problem by conditioning on incrementally larger sets of variables, such that most sets of variables will never have to be tested, whereas the IAMB first computes the Markov blanket of all the variables and conditions only on members of the blankets. A few more details about these algorithms follow.

**The PC algorithm.**   The PC algorithm [171] begins with a fully connected graph and, on the basis of pairwise independence tests, iteratively removes all the extraneous edges. It is based on the idea that if a separating set exists that makes two variables independent, we can remove the edge between them. To avoid an exhaustive search of separating sets, these are ordered to find the correct ones early in the search. Once a separating set is found, the search for that pair can end. The PC algorithm orders separating sets of increasing size $l$ starting from 0, the empty set, and incrementing until $l = n - 2$. The algorithm stops when every variable has fewer than $l - 1$ neighbors, since it can be proven that all valid sets must have already been chosen. During the computation, the larger the value of $l$ is, the larger number of separating sets must be considered. However, by the time $l$ gets too large, the number of nodes with degree $l$ or higher must have dwindled considerably. Thus, in practice, we need only consider a small subset of all the possible separating sets.

**Incremental Association Markov Blanket algorithm.** A different type of constraint based learning algorithms uses the Markov blankets to restrict the subset of variables to test for independence. Thus, when this knowledge is available in advance, we do not have to test a conditioning on all possible variables. A widely used and efficient algorithm for Markov blanket discovery is IAMB. In it, for each variable $X$, we keep track of a hypothesis set $\mathcal{H}(X)$. The goal is for $\mathcal{H}(X)$ to equal the Markov blanket of $X$, $\mathcal{B}(X)$, at the end of the algorithm. IAMB consists of a forward and a backward phase. During the forward phase, it adds all possible variables into $\mathcal{H}(X)$ that could be in $\mathcal{B}(X)$. In the backward phase, it eliminates all the false positive variables from the hypotheses set, leaving the true $\mathcal{B}(X)$. The forward phase begins with an empty $\mathcal{H}(X)$ for each $X$. Iteratively, variables with a strong association with $X$ (conditioned on all the variables in $\mathcal{H}(X)$) are added to the hypotheses set. This association can be measured by a variety of non-negative functions, such as *mutual information*. As $\mathcal{H}(X)$ grows large enough to include $\mathcal{B}(X)$, the other variables in the network will have very little association with $X$, conditioned on $\mathcal{H}(X)$. At this point, the forward phase is complete. The backward phase starts with $\mathcal{H}(X)$ that contains $\mathcal{B}(X)$ and false positives, which will have little conditional association, while true positives will associate strongly. Using this test, the backward phase is able to remove the false positives iteratively until all but the true positives are eliminated.

### B.2.2 Score based approaches

This approach to structural learning seeks to maximize the likelihood of a set of observed data. Since we assume that the data are independent and identically distributed, the likelihood of the data $\mathcal{L}(\cdot)$ is simply the product of the probability of each observation. That is,

$$\mathcal{L}(D) = \prod_{d \in D} \mathcal{P}(d)$$

for a set of observations $D$. Since we want to infer a model $\mathcal{G}$ that best explains the observed data, we define the likelihood of observing the data given a specific model $\mathcal{G}$ as

$$\mathcal{LL}(\mathcal{G}, D) = \prod_{d \in D} \mathcal{P}(d \mid \mathcal{G}).$$

The actual likelihood is not used in practice, as this quantity becomes very small and impossible to represent in a computer. Instead, the logarithm of the likelihood is used for three reasons. First, the $log(\cdot)$ function is monotonic. Second, the values that the log-likelihood takes do not cause the same numerical problems that likelihood does. Third, it is easy to compute because the log of a product is simply the sum of the logs (e.g., $\log(xy) = \log x + \log y$), and the likelihood for a Bayesian network is a product of simple terms.

Practically, however, there is a problem in learning the network structure by maximizing log-likelihood alone. Namely, for any arbitrary set of data, the most likely graph is always the fully connected one (i.e. all edges are present), since adding an edge can only increase the likelihood of the data. To correct for this phenomenon, log-likelihood is almost always supplemented with a *regularization term* that penalizes the complexity of the model[3]. There are a plethora of regularization terms, some based on information theory and others on Bayesian statistics (see [23] and references therein), which all serve to promote *sparsity* in the learned graph structure, though different regularization terms are better suited for particular applications.

Also in this case we choose to describe a particularly relevant and known score, the *Bayesian Information Criterion* (BIC, [101]), which will be subsequently compared to the performance of our approach.

**The Bayesian Information Criterion.** BIC uses a score that consists of a log-likelihood term and a regularization term depending on a model $\mathcal{G}$ and data $D$

$$\textsc{bic}(\mathcal{G}, D) = \mathcal{LL}(\mathcal{G}, D) - \frac{\log m}{2} \dim(\mathcal{G}). \tag{B.1}$$

Here, $D$ denotes the data, $m$ denotes the number of samples and $\dim(\mathcal{G})$ denotes the number of parameters in the model. Because, in general, $\dim(\cdot)$ depends on the number of parents each node has, it is a good metric for model complexity. Moreover, each edge added to $\mathcal{G}$ increases model complexity. Thus, the regularization term based on $\dim(\cdot)$ favors graphs with fewer edges and, more specifically, fewer parents for each node. The term $\log m/2$ essentially weighs the regularization term. The effect is that the higher the weight, the more sparsity will be favored over "explaining" the data through maximum likelihood.

Note that the likelihood is implicitly weighted by the number of data points, since each point contributes to the score. As the sample size increases, both the weight of the regularization term and the "weight" of the likelihood increase. However, the weight of the likelihood increases faster than that of the regularization term[4]. Thus, with more data, likelihood will contribute more to the score, and we may trust our observations more and have less need for regularization. Statistically speaking, BIC is a *consistent score* [101]. In terms of structure learning, this observation implies that for sufficiently large sample sizes, the network with the maximum BIC score is *I-equivalent*[5] to the true structure. Consequently, $\mathcal{G}$ contains the same independence relations as those implied by the true structure. As the independence relations are encoded in the edges of the graph, we are guaranteed to learn a Markov-equivalent network, with the same skeleton and the same *v*-structures as the true graph, though not necessarily with the correct orientations for each edge.

---

[3]Note that more edges in a graph require more parameters in the conditional probability distributions, thus increasing model complexity. If it was known that the number of parameters for each node is fixed, then regularization is not necessary.

[4]Specifically, the likelihood weight increases linearly, while the weight of the regularization term grows only logarithmically.

[5]Two networks are I-equivalent if their structures encode the same independence statements.

### B.2.3  Learning logically constrained networks

In §B.1, we noted that an important class of BNs captures common binary logical operators, such as ∧, ∨, and ⊕. Although the learning algorithms mentioned above can be used to infer the structure of such networks, some algorithms employ knowledge of these logical constraints in the learning process.

A widely used approach to learn a monotonic cancer progression network with a directed acyclic graph (DAG) structure and *conjunctive events* are *Conjunctive Bayesian Networks* (see CBNs, [10]). This model is a standard BN over Bernoulli random variables with the constraint that the probability of a node $X$ taking the value 1 is zero if at least one of its parents has value 0. This defines a conjunctive relationship, in that all the parents of $X$ must be 1 for $X$ to possibly be 1. Thus, this model alone cannot represent noise, which is an essential part of any real data. In response to this shortcoming, *hidden CBNs* [65] were developed by augmenting the set of variables: a correspondence to a new variable $Y$ that represents the observed state is assigned to each CBN variable $X$, which captures the "true" state. Thus, each new variable $Y$ takes the value of the corresponding variable $X$ with a high probability, and the opposite value with a low probability. In this model, the variables $X$ are latent, i.e., they are not present in the observed data, and have to be inferred from the observed values for the new variables. Learning is performed using a maximum likelihood approach and is separated into multiple iterations of two steps. First, the parameters for the current hypothesized structure are estimated using the *Expectation-Maximization* algorithm [132] and the likelihood given those parameters is computed. Second, the structure is perturbed to nudge the *hill climbing* scheme used to maximize expectation off local maximum. In their work, the authors used the *Simulated Annealing* algorithm [97] for this step. These two steps are repeated until the score converges. However, the Expectation-Maximization algorithm only guarantees convergence to a likelihood local maximum and, thus, the overall procedure is not guaranteed to converge to the optimal structure.

## B.3   Bayesian interpretation of the proposed framework

The algorithms proposed in this thesis can be placed among the constrained approaches. In particular, the aim is the one of reconstructing Bayesian graphical models whose induced probability distributions are biased in terms of Suppes' criteria as follows. For any pair of nodes for which we have a directed edge from $b$ to $a$,

$$\begin{cases} \mathcal{P}(a \mid b) = \theta \\ \mathcal{P}(a \mid \bar{b}) \leq \epsilon \end{cases}$$

where $\theta, \epsilon \in [0, 1]$ and $\theta \gg \epsilon$. Specifically, $\theta$ represents the conditional probability of any effect to follow its preceding cause and $\epsilon$ models the probability of any noisy observation.

CAPRESE - SUPPLEMENTARY MATERIALS

In this Chapter we will report the supplementary materials related to chapter §3.

## C.1 Proofs

Here the proofs of all the propositions and theorems follow.

**Proof of Proposition §3 (Statistical dependence).**

*Proof.* For $\Rightarrow$ write $\mathcal{P}(\overline{a} \wedge b) = \mathcal{P}(b) - \mathcal{P}(a \wedge b)$, then write the PR as

$$\frac{\mathcal{P}(a \wedge b)}{\mathcal{P}(a)} > \frac{\mathcal{P}(b) - \mathcal{P}(a \wedge b)}{1 - \mathcal{P}(a)}$$

and, since $0 < \mathcal{P}(a) < 1$, the proposition follows by simple algebraic arrangements of $\mathcal{P}(a \wedge b) \cdot [1 - \mathcal{P}(a)] > \mathcal{P}(a)\mathcal{P}(b) - \mathcal{P}(a \wedge b) \cdot \mathcal{P}(a)$. The derivations are analogous but in reverse order for the implication $\Leftarrow$. □

**Proof of Proposition §4 (Mutuality).**

*Proof.* The proof follows by Property §3 and the subsequent implication:

$$\mathcal{P}(b \mid a) > \mathcal{P}(b \mid \overline{a}) \Leftrightarrow \mathcal{P}(a \wedge b) > \mathcal{P}(a)\mathcal{P}(b) \Leftrightarrow \mathcal{P}(a \mid b) > \mathcal{P}(a \mid \overline{b}).$$

□

**Proof of Proposition §5 (Natural ordering).**

*Proof.* We first prove the forward direction $\Rightarrow$. Let $x = \mathcal{P}(\overline{a} \wedge b)$, $y = \mathcal{P}(a \wedge b)$ and $z = \mathcal{P}(a \wedge \overline{b})$. We have two assumptions we will use later on:

1. $\mathcal{P}(a) > \mathcal{P}(b)$ which implies $\mathcal{P}(\overline{a} \wedge b) < \mathcal{P}(a \wedge \overline{b})$, i.e., $x < z$.

2. $\mathcal{P}(a \mid b) > \mathcal{P}(a \mid \overline{b})$ which, when $0 < x + y < 1$, implies by simple algebraic rearrangements the inequality

$$y[1 - x - y - z] > xz \,. \tag{C.1}$$

We proceed by rewriting $\mathcal{P}(b \mid a)/\mathcal{P}(b \mid \overline{a}) > \mathcal{P}(a \mid b)/\mathcal{P}(a \mid \overline{b})$ as

$$\frac{\mathcal{P}(a \wedge b)\mathcal{P}(\overline{a})}{\mathcal{P}(\overline{a} \wedge b)\mathcal{P}(a)} > \frac{\mathcal{P}(a \wedge b)\mathcal{P}(\overline{b})}{\mathcal{P}(a \wedge \overline{b})\mathcal{P}(b)}$$

which means that

$$\frac{\mathcal{P}(b \mid a)}{\mathcal{P}(b \mid \overline{a})} > \frac{\mathcal{P}(a \mid b)}{\mathcal{P}(a \mid \overline{b})} \iff \frac{\mathcal{P}(\overline{a})}{\mathcal{P}(\overline{a} \wedge b)\mathcal{P}(a)} > \frac{\mathcal{P}(\overline{b})}{\mathcal{P}(a \wedge \overline{b})\mathcal{P}(b)} \tag{C.2}$$

We can rewrite the right-hand side of (§C.2) by using $x$, $y$, $z$ where $\mathcal{P}(a) = \mathcal{P}(a, b) + \mathcal{P}(a, \overline{b}) = y + z$ and $\mathcal{P}(b) = \mathcal{P}(a, b) + \mathcal{P}(\overline{a}, b) = x + y$, and then do suitable algebraic manipulations. We have

$$\frac{1 - y - z}{x(y + z)} > \frac{1 - x - y}{z(x + y)} \iff yz - y^2z - xz^2 - yz^2 > xy - x^2y - x^2z - xy^2 \tag{C.3}$$

when $x(y + z) \neq 0$ and $z(x + y) \neq 0$. To check that the right side of (§C.3) holds we show that

$$(xy - x^2y - x^2z - xy^2) - (yz - y^2z - xz^2 - yz^2) < 0 \,.$$

First, we rearrange it as $(x - z)[y - y^2 - xz - y(x + z)] < 0$ to show that

$$(x - z)[y(1 - y - x - z) - zx] < 0 \tag{C.4}$$

is always negative. By observing that, by assumption 1 we have $z > x$ and thus $(x - z) < 0$, and, by equation (§C.1) we have $y(1 - y - x - z) - zx > 0$, we derive

$$\frac{\mathcal{P}(b \mid a)}{\mathcal{P}(b \mid \overline{a})} > \frac{\mathcal{P}(a \mid b)}{\mathcal{P}(a \mid \overline{b})}$$

which concludes the $\Rightarrow$ direction.

The other direction $\Leftarrow$ follows immediately by contraposition: assume that $\mathcal{P}(a \mid b) > \mathcal{P}(a \mid \overline{b})$, $\mathcal{P}(b \mid a)/\mathcal{P}(b \mid \overline{a}) > \mathcal{P}(a \mid b)/\mathcal{P}(a \mid \overline{b})$ and $\mathcal{P}(b) \leq \mathcal{P}(a)$. We distinguish two cases:

1. $\mathcal{P}(b) = \mathcal{P}(a)$, then $\mathcal{P}(b \mid a)/\mathcal{P}(b \mid \overline{a}) = \mathcal{P}(a \mid b)/\mathcal{P}(a \mid \overline{b})$.

2. $\mathcal{P}(b) < \mathcal{P}(a)$, then by symmetry $\mathcal{P}(b \mid a) > \mathcal{P}(b \mid \overline{a})$, and by the $\Rightarrow$ direction of the proposition it follows that $\mathcal{P}(b \mid a)/\mathcal{P}(b \mid \overline{a}) < \mathcal{P}(a \mid b)/\mathcal{P}(a \mid \overline{b})$.

In both cases we have a contradiction. This completes the proof. $\qquad\square$

**Proof of Proposition §6 (Monotonic normalization).**

*Proof.* We prove the forward direction $\Rightarrow$, the converse follows by a similar argument. Let us assume

$$\frac{\mathcal{P}(b \mid a)}{\mathcal{P}(b \mid \bar{a})} > \frac{\mathcal{P}(a \mid b)}{\mathcal{P}(a \mid \bar{b})} \tag{C.5}$$

then $\mathcal{P}(b \mid a)\mathcal{P}(a \mid \bar{b}) > \mathcal{P}(a \mid b)\mathcal{P}(b \mid \bar{a})$. Now, to show the righthand side of the implication, we will show that

$$\Big[\mathcal{P}(b \mid a) - \mathcal{P}(b \mid \bar{a})\Big]\Big[\mathcal{P}(a \mid b) + \mathcal{P}(a \mid \bar{b})\Big] > \Big[\mathcal{P}(b \mid a) + \mathcal{P}(b \mid \bar{a})\Big]\Big[\mathcal{P}(a \mid b) - \mathcal{P}(a \mid \bar{b})\Big]$$

which reduces to show

$$\mathcal{P}(b \mid a)\mathcal{P}(a \mid \bar{b}) - \mathcal{P}(b \mid \bar{a})\mathcal{P}(a \mid b) > \mathcal{P}(b \mid \bar{a})\mathcal{P}(a \mid b) - \mathcal{P}(b \mid a)\mathcal{P}(a \mid \bar{b}).$$

By (§C.5), two equivalent inequalities hold

$$\mathcal{P}(b \mid a)\mathcal{P}(a \mid \bar{b}) - \mathcal{P}(b \mid \bar{a})\mathcal{P}(a \mid b) > 0$$
$$\mathcal{P}(b \mid \bar{a})\mathcal{P}(a \mid b) - \mathcal{P}(b \mid a)\mathcal{P}(a \mid \bar{b}) < 0$$

and hence the implication holds. $\qquad\square$

**Proof of Proposition §7 (Coherence in dependency and temporal priority).**

*Proof.* We make two assumptions:

1. $\mathcal{P}(b \mid a) > \mathcal{P}(b \mid \bar{a})$ which implies $\alpha_{a \to b} > 0$.

2. $\mathcal{P}(a, b) > \mathcal{P}(a)\mathcal{P}(b)$ which implies $\beta_{a \to b} > 0$.

The proof for dependency follows by Proposition §3 and its implication:

$$\mathcal{P}(b \mid a) > \mathcal{P}(b \mid \bar{a}) \Leftrightarrow \mathcal{P}(a \wedge b) > \mathcal{P}(a)\mathcal{P}(b) \Leftrightarrow \alpha_{a \to b} > 0 \Leftrightarrow \beta_{a \to b} > 0.$$

Moreover, being $\beta$ symmetric by definition, the proof for temporal priority follows directly by Proposition §4 $\qquad\square$

**Proof of Theorem §1 (Independent progressions).**

*Proof.* For any $a_i \in G^*$ it holds that

$$\mathcal{P}(a_i) > \mathcal{P}(b) \qquad m_{a_i \to b} > 0$$

being prima facie, also $a^* \to b$ is the edge selected by CAPRESE being the $\max\{\cdot\}$ over $G^*$. Thus, CAPRESE selects $\diamond \to b$ instead of $a^* \to b$ if, for any $a_i$, it holds

$$\frac{1}{1 + \mathcal{P}(b)} > \frac{\mathcal{P}(a_i)}{\mathcal{P}(a_i) + \mathcal{P}(b)} \frac{\mathcal{P}(a_i \wedge b)}{\mathcal{P}(a_i)\mathcal{P}(b)}.$$

With some algebraic manipulations we rewrite this as

$$\mathcal{P}(a_i)\mathcal{P}(b) + \mathcal{P}(b)^2 > \mathcal{P}(a_i \wedge b)(1 + \mathcal{P}(b)),$$

which gives the inequality in the theorem statement. $\qquad\square$

**Proof of Theorem §2 (Algorithm correctness).**

*Proof.* It is clear that CAPRESE does not create disconnected components since, to each node in $G$, a unique parent is attached (either from $G$ or $\diamond$). For the same reason, no transitive connections can appear.

The absence of cycles results from Properties §5, §6 and §7. Indeed, suppose for contradiction that there is a cycle $(a_1, a_2), (a_2, a_3), \ldots, (a_n, a_1)$ in $E$, then by the three propositions we have

$$\mathcal{P}(a_1) > \mathcal{P}(a_2) > \ldots > \mathcal{P}(a_n) > \mathcal{P}(a_1)$$

which is a contradiction. □

**Proof of Theorem §3 (Asymptotic convergence).**

*Proof.* For any $u \in G$, when $s \to \infty$ the *observed probability* $\mathcal{P}(u)$ (evaluated from $D$) is equivalent to the product of the probabilities (in $T$) obtained by traversing the forest from the root $\diamond$ to $u$ (Definition §3). Thus, $\mathcal{P}(u) \in (0, 1)$ since the traversal probabilities are in $(0, 1)$ too, hence all events are distinguishable and Algorithm 1 reconstructs a tree with the same events set $G$ of $T$.

We now observe that the distribution induced by $T$ (Definition §3) respects a *single-cause prima facie topology* where to each event is assigned *at most* a single cause. In other words, Definition §8 holds for any edge $(u, v) \in E$:

- by the event-persistence property usually assumed in cancer (fixating mutations are present in the progeny of a clone) the occurring times satisfy $t_u < t_v$ which, in a frequentist sense, implies $\mathcal{P}(u) > \mathcal{P}(v)$;

- it holds by construction (Definition §3) that $\mathcal{P}(v \wedge u) = \mathcal{P}(v)$, thus $\mathcal{P}(v \mid u) = \mathcal{P}(v)/\mathcal{P}(u)$ which is strictly positive since $\mathcal{P}(v)$ and $\mathcal{P}(u)$ are, and that $\mathcal{P}(v \wedge \overline{u}) = 0$, thus $\mathcal{P}(v \mid \overline{u}) = 0$.

To correctly reconstruct $T$ we rely on the fact that our score $m_{u \to v}$ is consistent with the prima facie probabilistic causation because of:

- Proposition §5, which states that PR (embedded as $\alpha_{u \to v}$ in $m$) subsumes a good temporal priority model of occurring times, as stated above;

- Proposition §6 and §7 which ensure the monotonicity and sign coherency among $\alpha_{u \to v}$ and $\beta_{u \to v}$ in $m$.

Thus, $m$ is consistent with a single-cause prima facie topology. We now show that Algorithm 1 reconstructs correctly a generic edge in $E$, and hence also $T$.

Consider an event $v \in G$ and edge $(u, v) \in E$. The set of its "candidate" parent events is $G \setminus \{v\}$, we partition it in three disjoint sets $\mathcal{G}$, $\mathcal{S}$ and $\mathcal{N}$:

- $\mathcal{G}$, *genuine:* all the backward-reachable events, in $G \setminus \{v\}$, from $v$;

- $\mathcal{S}$, *spurious (or ambiguous):* all the events (but $v$ ) in the sub-forest which includes the path from $\diamond$ to $v$, which are not in $\mathcal{G}$;

- $\mathcal{N}$, *non prima facie:* all other events, i.e., $G \setminus (\{v, \} \cup \mathcal{S} \cup \mathcal{G})$;

Notice that $G = \{v\} \cup \mathcal{G} \cup \mathcal{S} \cup \mathcal{N}$, that $u \in \mathcal{G}$ and that all the effects of $v$ are non prima facie to $v$ because of the temporal priority, as shown below.

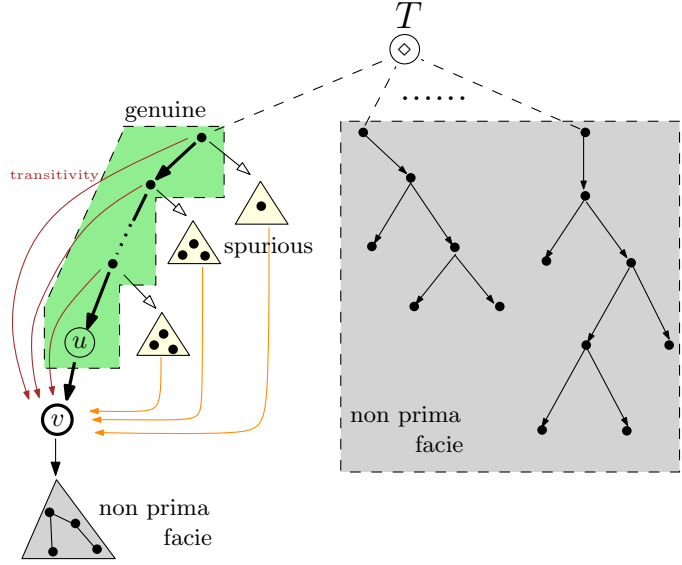

This way of partitioning events according to the structure of $T$ subsumes a equivalent partitioning based on the score $\alpha \in [0, 1]$, which we use to prove correctness of our algorithm: for any $x \in G$ it holds that $\alpha_{x \to v} = 1$ if $x \in \mathcal{G}$, $0 < \alpha_{x \to v} < 1$ if $x \in \mathcal{S}$ and $\alpha_{x \to v} < 0$ if $x \in \mathcal{N}$.

We now show that CAPRESE correctly selects $u \in \mathcal{G}$:

- a non prima facie event $x \in \mathcal{N}$ either satisfies (Proposition §3)

$$\mathcal{P}(x \wedge v) \leq \mathcal{P}(x)\mathcal{P}(v)$$

  which means that $\alpha_{x \to v} < 0$, $\beta_{x \to v} < 0$ (Proposition §7) and thus $m_{x \to v} < 0$, or it is a descendant of $v$, which means that $\mathcal{P}(x) < \mathcal{P}(v)$. By construction, CAPRESE considers as candidate parents of $v$ only not descendant events with positive score (see step 3);

- a spurious event $x \in \mathcal{S}$ is prima facie to $v$ but $\alpha_{x \to v} < 1$ since:

  - $\mathcal{P}(x)\mathcal{P}(v) < \mathcal{P}(v \wedge x) < \mathcal{P}(v)$, otherwise $x$ would be backward reachable from $x$ and thus in $\mathcal{G}$;

  - $0 < \mathcal{P}(v \wedge \overline{x}) = \mathcal{P}(v) - \mathcal{P}(v \wedge x)$ which means that $\mathcal{P}(v \wedge \overline{x}) < \mathcal{P}(\overline{x})\mathcal{P}(v)$;

  - by all of the above $\mathcal{P}(v \mid \overline{x}) > 0$ which implies that $\alpha_{x \to v} < 1$.

Recall now that $\lambda \to 0$, which means that $m_{x \to v} \approx \alpha_{x \to v} < 1$. CAPRESE will thus not select any of these events as cause of $v$ if there exist an event with $m_{x \to v} = 1$, which is actually the case with genuine causes;

- genuine causes are the real cause of $v$, $u$, plus *all* the transitive backward-reachable events. Any $x$ of these has maximum score $\alpha_{x \to v} = 1$ since:
  - $\mathcal{P}(x)\mathcal{P}(v) < \mathcal{P}(v \wedge x) = \mathcal{P}(v)$ and $0 = \mathcal{P}(v \wedge \overline{x})$;
  - by the above $\mathcal{P}(v \mid \overline{x}) = 0$ which implies $\alpha_{x \to v} = 1$.

Thus, CAPRESE will pick an event from $\mathcal{G}$, and not from $\mathcal{S}$. We need to show that $u$ is the event with maximum score.

Enumerate the events in $\mathcal{G}$ as $g_1$ (which is $u$), ..., $g_k$ in a way that

$$\mathcal{P}(g_1) < \ldots < \mathcal{P}(g_k)$$

and recall that this is a *total ordering* induced by the temporal priority, and that this is consistent with coefficient $\beta$, which means that

$$\beta_{g_1 \to v} > \ldots > \beta_{g_k \to v}.$$

Thus, in the limit $\lambda \to 0$

$$\max\{m_{g_i \to v} \mid g_i \in \mathcal{G}\} \overset{\lambda \to 0}{\approx} 1 + \max\{\beta_{g_i \to v} \mid g_i \in \mathcal{G}\} \approx 1 + \beta_{g_1 \to v} \approx m_{u \to v}$$

is the event closer in time to $v$, with respect to $\beta$. This event, namely $u$, is chosen by the algorithm as the real cause of $v$.

Finally, we show that the last step of the algorithm (the independent progression filter, step 4), does not invalidate the edge $(u, v)$. In fact, the algorithm would replace such an edge with $(\diamond, v)$ if, for all nodes $x$ backward-reachable from $v$ (i.e., those in $\mathcal{G} \cup \mathcal{S}$) it was

$$\frac{1}{1 + \mathcal{P}(v)} > \frac{\mathcal{P}(x)}{\mathcal{P}(x) + \mathcal{P}(v)} \frac{\mathcal{P}(x \wedge v)}{\mathcal{P}(x)\mathcal{P}(v)}.$$

It suffices thus to show that the above inequality is violated just by one of the backward-reachable nodes. We pick just $u \in \mathcal{G}$ and note that

$$\frac{\mathcal{P}(u)}{\mathcal{P}(u) + \mathcal{P}(v)} \frac{\mathcal{P}(u \wedge v)}{\mathcal{P}(u)\mathcal{P}(v)} = \frac{\mathcal{P}(u \mid v)}{\mathcal{P}(u) + \mathcal{P}(v)}.$$

Also, we have that $\mathcal{P}(u) < 1$, $\mathcal{P}(v) < 1$ and, by construction, $\mathcal{P}(u \mid v) = 1$ because all the instances of $v$ are co-occurring with those of $u$ (but not the converse). Thus, inequality

$$\frac{1}{1 + \mathcal{P}(v)} < \frac{1}{\mathcal{P}(u) + \mathcal{P}(v)},$$

is always true and ensures that edge $(u, v)$ is maintained, which concludes the proof. $\quad \square$

**Proof of Corollary §1 (Uniform noise).**

*Proof.* As shown in [173], the uniform rates $\epsilon_+$ and $\epsilon_-$ affect the observed probabilities as follows

$$\mathcal{P}(i)^* = \mathcal{P}(i)(1 - \epsilon_-) + (1 - \mathcal{P}(i))\epsilon_+ \tag{C.6}$$

$$\mathcal{P}(i \wedge j)^* = \mathcal{P}(i \wedge j)(1 - \epsilon_-)^2 + [\pi_{ij} - \mathcal{P}(i \wedge j)](1 - \epsilon_-)\epsilon_+ + (1 - \pi_{ij})\epsilon_+^2, \tag{C.7}$$

where $\pi_{ij} = \mathcal{P}(i) + \mathcal{P}(j) - \mathcal{P}(i \wedge j)$. It is important to note (Lemma 1, [173]) that

$$\mathcal{P}(i) > \mathcal{P}(j) \implies \mathcal{P}(i)^* > \mathcal{P}(j)^*,$$

namely uniform noise is still implying temporal priority. Because of this, and since the raw estimate $\alpha$ is monotonic relative to temporal priority, all the derivations for Theorem §3 are still valid in this context, and the algorithm selects the correct genuine cause for each effect.

To guarantee that no valid connection is broken by the independent progressions filter, we again rely on Szabo's result (Reconstruction Theorem 1, [173]). In particular, for any correctly selected edge $(u, v)$ in our algorithm, since we implement Desper's filter (or, analogously, Szabo's) for independent progressions we do not mistake by deleting $(u, v)$ unless also their algorithms do. Since this is not the case when $\epsilon_+ < \sqrt{p_{\min}}(1 - \epsilon_+ - \epsilon_-)$ the proof is concluded. □

## C.2    Synthetic data generation

A set of random trees is generated to prepare synthetic tests. Let $n$ be the number of considered events and let $p_{\min} = 0.05 = 1 - p_{\max}$, a *single tree* with maximum depth $\log(n)$ is generated as follows:

1: pick an event $r \in G$ as the tree root;
2: assign to each event but $r$ an integer value in $[2, \log(n)]$ representing its depth in the tree, ensure that for each level there is at least one event (0 is reserved for $\diamond$, 1 for $r$);
3: **for all** events $e \neq r$ **do**
4:     let $l$ be the level assigned to $e$;
5:     assign a father to $e$ selecting an event among those at which level $l-1$ was assigned;

6:     add the selected pair to the set of edges $E$;
7: **end for**
8: **for all** edges $(i, j) \in E$ **do**
9:     assign $\alpha((i, j))$ a random value in $[p_{\min}, p_{\max}]$;
10: **end for**
11: **return**  the generated tree;

When a forest is to be generated, we repeat the above algorithm to create its constituent trees. These trees (or forests), in turn, are used to sample the input matrix for the reconstruction algorithms, with the parameters described in the main text.

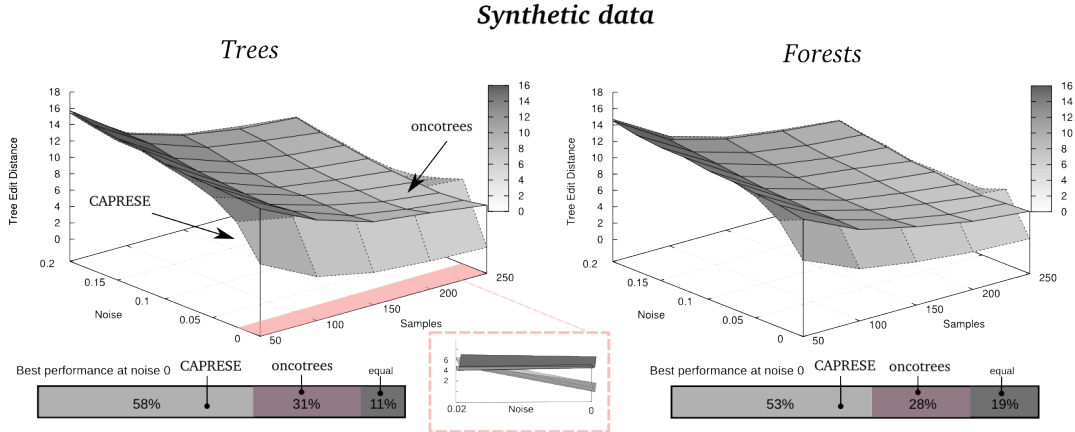

Figure C.1: **Reconstruction with noisy synthetic data and $\lambda \to 0$.** The settings of the experiments are the same as those used in Figure §3.6, but in this case the estimator is shrank by $\lambda \to 0$, i.e., $\lambda = 0.01$. In the magnified image one can sees that the performance of CAPRESE converges to Desper's one already for $\nu \approx 0.01$, hence largely faster than in the case of $\lambda \approx 1/2$ (Fig. §3.6).

## C.3 Further results

We show here the results of the experiments discussed but not presented in the main text.

**Reconstruction of noisy synthetic data with $\lambda \to 0$.** Although we know that $\lambda \to 0$ is not the optimal value of the shrinkage-like coefficient for noisy data, we show in Figure §C.1 the analogue of Figure §3.6 when the estimator is shrank by $\lambda \to 0$, i.e., $\lambda = 0.01$. When compared to Figure §3.6 it is clear that a best performance of CAPRESE is obtained with $\lambda \approx 1/2$, as suggested by Figure §3.2.

**Comparison with hidden Conjunctive Bayesian Networks, h-CBNs.** We here compare the performance of CAPRESE to hidden Conjunctive Bayesian Networks (h-CBN) [65], as well as to oncotrees. The settings of the experiment are slightly different from those of the previous analyses: we used 100 distinct random trees of 10 events each. We ranged the number of samples available for reconstruction from 50 to 200, with a step size of 50. The settings used for running h-CBNs are relatively standard settings: we allowed for 50 annealing iterations with initial temperature equal to 1. Since h-CBNs reconstruct DAGs, it is not possible to quantify its performance using Tree Edit Distance, as we did in the comparison with oncotrees. Instead, we here adopt Hamming Distance (computed on the connection adjacency matrix), as a closely related and computationally feasible alternative for measuring performance [76].

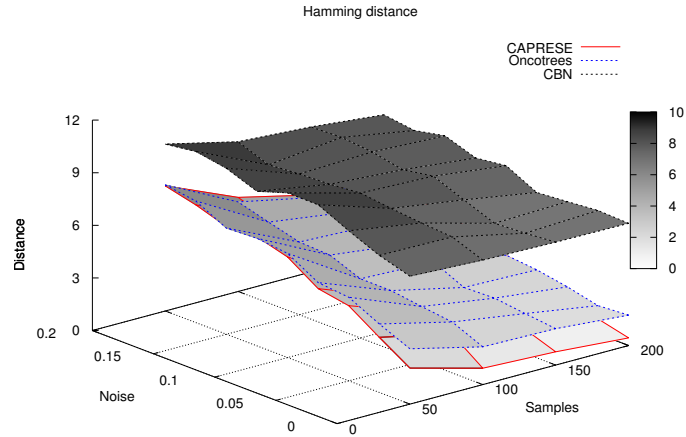

Figure C.2: **Reconstruction with CAPRESE compared to oncotrees and h-CBN with noisy synthetic data.** Performance of CAPRESE compared to oncotrees and h-CBNs as a function of the number of samples and noise $\nu$. The $\lambda$ parameter used for CAPRESE is $1/2$, and the reconstructed topologies contain 10 nodes each.

The results of the experiment can be found in Figure §C.2, and show that CAPRESE clearly outperforms h-CBNs. In particular, it is possible to notice that, for all the analyzed values of noise and sample sizes, both CAPRESE and oncotrees display a (average) Hamming Distance between the reconstructed model and the original tree topology that is significantly lower than h-CBNs, with the largest differences observed in the noise-free case. This result would point at a much faster convergence of CAPRESE with respect to the number of samples, also in presence of moderate levels of noise.

A few remarks are warranted about this experiment. First, in contrast to the comparison with oncotrees, we ran each experiment exactly once rather than averaging the results over 10 repetitions, and on relatively smaller trees. These limitations are a consequence of the extremely high time complexity of the simulated annealing step of h-CBNs. However, the comparison between CAPRESE and h-CBNs shows a so large difference in the performance that we do not expect this to be have significant impact. Second, the results obtained by h-CBNs are perhaps worse than expected based on results in the absence of noise presented in [72], which were however based on a unique tree topology. Yet, this outcome may have been potentially influenced by either the estimation procedure of the noise parameter in h-CBN, the adopted annealing procedure or by the used number of iterations. In future work we plan to extend our algorithm to extract more general topologies and to compare both methods in a greater detail.

**Inference of models with multiple conjunctive parents.** CAPRESE is specifically tailored to reconstruct models with independent progressions and a unique cause for each event (i.e., trees or forests), while other approaches such as CBNs can recon-struct models where multiple conjunctive parents co-occur to cause an effect (i.e., $a \wedge b$ cause $c$). It is thus reasonable to use such conjunctive approaches to infer more complex model, in spite of CAPRESE.

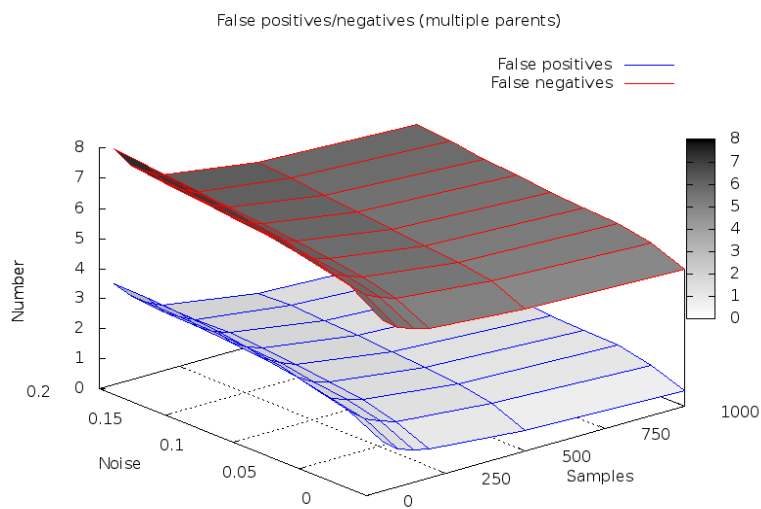

Figure C.3: **Performance of CAPRESE to reconstruct models with conjunctive parents and noisy data.** Performance of CAPRESE measured in terms of the number of *false positives/negatives* in the reconstructed model, when data are generated from directed acyclic graphs with 10 nodes and where each event is caused by at most 3 *conjunctive events* (randomly assigned). The λ parameter is set to 1/2.

However, it is interesting to asses CAPRESE's performance when (synthetic) data are sampled from a model with multiple parents and noise. By sampling input data from random *directed acyclic graphs* with 10 nodes and where each event is caused by at most 3 conjunctive events (randomly assigned), we assess the number of *false positives* and *false negatives* retrieved in the model reconstructed with CAPRESE. We show the results in Figure §C.3. Our results indicate that for increasing sample size, the number of false positives approaches 0. Thus, for sufficiently large number of samples, all the causal claims returned by CAPRESE are true. In addition, the number of false negatives is always higher and proportional to the connectivity of the target model. This is to be expected since CAPRESE assigns at most one parent (the cause) to every node.

# APPENDIX D

## CAPRI - SUPPLEMENTARY MATERIALS

In this Chapter we will report the supplementary materials related to chapter §4.

## D.1 Theorems

The statements and proofs of the theorems mentioned in the main text follow.

### D.1.1 Complexity

Let $\mathcal{U}$ denote the *universe* of all possible patterns over a set $G$ of $n$ events, as before. Since $|\mathcal{U}|$ is exponential in $|G|$, then the following theorem holds.

**Theorem 4** (Asymptotic complexity). *Let $|G| = n$ and $D \in \{0,1\}^{m \times n}$ where $m \gg n$, and let $N$ be the nodes in the DAG returned by CAPRI, the* worst case *time and space complexity (ignoring the cost of bootstrap) of building a selectivity topology is:*

- *$\Theta(mn)$ time and $\Theta(n^2)$ space, if $\Phi = \emptyset$;*

- *$\Theta(|\Phi|mn)$ time and $\Theta(|\Phi|m)$ space, if $\Phi \subset \mathcal{U}$ and $|N| \ll m$ (i.e., there are sufficiently many samples to characterize the input hypotheses);*

- *$\mathcal{O}(2^{2^n})$ time and space, if $\Phi = \mathcal{U}$.*

*Thus, the overall complexity of CAPRI is one of the above, as suitable in each case, plus the complexity of likelihood fit with regularization.*

*Proof.* Recall that $k = |\Phi|$, $n = |G|$ and $D \in \{0,1\}^{m \times n}$, thus $D(\Phi)$ has $K = (n+k)m$ entries. We now analyze the complexity of CAPRI step-by-step.

- The cost of lifting depends on the input set $\Phi$, if $\Phi = \emptyset$ it is $\mathcal{O}(1)$ both in time and space since $D(\emptyset) = D$. For non-empty sets, it is necessary to evaluate $k \cdot m$ entries, after each hypothesis $\varphi \triangleright e$ is evaluated. Given that every $\varphi$ has at worst $n$ events included, its evaluation cost is at most $\mathcal{O}(n)$, even if lazy evaluation is performed. Thus, the cost of lifting is $\Theta(k \cdot m \cdot n)$, for a single bootstrap, which amplifies the bootstrap cost, as discussed in the previous section, and does so in a multiplicative fashion. In terms of space, if $\Phi \neq \emptyset$ the overhead is $\Theta(K)$ if one copies $D$ in $D(\Phi)$, $\Theta(km)$ otherwise.

- The cost of computing the parent function for the DAG requires a pair-wise calculation of the probabilistic scores, plus the cost of testing the $\sqsubseteq$ relation[1]. Let $w = |N|$, where $N$ is the set of nodes in the DAG returned by CAPRI. The score matrices for temporal priority and probability raising are $n \times w$, i.e., have columns for both atomic events and the disjunctive patterns in the formulas of $\Phi$, since CAPRI disregards patterns of the form $\varphi_i \triangleright \varphi_j$ and $a \triangleright \varphi$ (differently, it would have been $w \times w$). With the simplest membership test algorithm, checking whether an atomic event is present in a patterns is logarithmic in the size of the pattern, if we lexicographically order its atomic events, thus bounded from above by $\log n$. Thus if we perform lazy evaluation for $\sqsubseteq$ the total number of comparison to select the parent function is at most

$$n[(n-1) + (w-n)\log n],$$

yielding a $\Theta(n^2)$ cost in time and space, if $w-n$ is small (it is 0 if $\Phi = \emptyset$), $\mathcal{O}(n(w-n)\log n)$ otherwise. In terms of space, the complexity is $\Theta(n[(n-1) + (w-n)])$, for a general $\Phi$.

- As explained in CAPRI's definition, sometimes, albeit extremely rarely, a few extra operations might have to be performed when degenerate scores and loops are present. The procedure we suggested in CAPRI's definition requires sorting plus scan, thus its worst-case time complexity is $\mathcal{O}(n \log n)$. Clearly, as this term is omitted in the worst-case complexity analysis of the steps discussed above, this unlikely scenario does not alter the complexity of the algorithm.

- Note that the cost of this analysis does not include the cost of BIC/likelihood - or any regularization strategy one might adopt, as spelled out in the theorem statement[2].

The overall complexity follows, since:

- $\Phi = \emptyset$ then the major cost is that of evaluating $\mathcal{P}(\cdot)$ since usually $m \gg n$, thus $mn > n^2$. With regard to space, the only cost is that of book-keeping the scores.

---

[1] Relation $\sqsubseteq$ represents the usual syntactical ordering relation among atomic events, e.g., $a$, $b$, and formulas, e.g., $a \sqsubseteq (a \vee b) \vee c \vee d$.

[2] Since in the current version of CAPRI, the likelihood fit is computed by a *hill climbing* heuristic algorithm, the overall cost of CAPRI is still polynomial.

- Let $m \gg n$ and $w - n > k$, in this case since $km \gg n$ and, under the mild assumption that $m > w$ and that $k$ and $\log n$ are not relevant (in size) for $m$ and $w$, then $km \gg (w - n)\log n$ which is the cost of lifting; thus is $\Theta(kmn)$ in time. Similarly, it follows that $mk \gg n[(n - 1) + (w - n)]$.

- By computations similar to those carried out, it is indeed possible to see that $\mathcal{U}$, which is clearly finite since $G$ is, grows *double-exponentially* in size with $|G|$ (i.e. the number of $n$-ary boolean functions, defined over the atomic events in any pattern, possibly with negated literals). Thus the bound follows.

$\square$

### D.1.2   Correctness and expressivity

Let $\mathcal{W} \subseteq \mathcal{U}$ be the set of true patterns, which we seek to infer. Here, we investigate the relation between $\mathcal{W}$ and the patterns retrieved by CAPRI, as a function of sample size $m$ and error present as false positives/negatives, which are assumed to occur at rates $\epsilon_+$ and $\epsilon_-$.

Hereafter, $\Sigma$ denotes the set of patterns, implicit in the DAG returned by our algorithm for an input set $\Phi$ and a matrix $D$; we write this fact as $D(\Phi) \Vdash \Sigma$. We prove the following theorems[3].

**Theorem 5** (Soundness and completeness). *Let the sample size $m \to \infty$ and the data be uniformly randomly corrupted by false positives and negatives rates $\epsilon_- = \epsilon_+ \in [0, 1)$. If the given input is a superset of the true patterns, then CAPRI reconstructs exactly the true patterns in $\mathcal{W}$, that is, $\mathcal{W} \subset \Phi \Rightarrow D(\Phi) \Vdash \mathcal{W} \cap \Phi$.*

*Proof.* We first prove the case with $\epsilon_+ = \epsilon_- = 0$, that is, the case where data have no noise. Some notations, used below: *(i)* we denote with $\varphi \rhd e$ true patterns (i.e. in $\mathcal{W}$), and *(ii)* with $\varphi^* \rhd e$ false ones. We divide the proof into several steps:

- First, we show that a selectivity DAG contains all the true patterns, which is

$$\forall_{\varphi \rhd e \in \mathcal{W}} \ \pi(e) = \{\varphi\}.$$

   By the event-persistence property usually valid for cancer genomes (fixating mutations are present in the progeny of a clone) the occurring times satisfy $t_\varphi < t_e$ which, in a frequentist sense, implies $\mathcal{P}(\varphi) > \mathcal{P}(e)$. In addition, it holds by construction that $\mathcal{P}(\varphi \wedge e) = \mathcal{P}(e)$ when $\epsilon_+ = \epsilon_- = 0$, thus $\mathcal{P}(e \mid \varphi) = \mathcal{P}(e)/\mathcal{P}(\varphi)$, which is strictly positive since $\mathcal{P}(\varphi)$ and $\mathcal{P}(e)$ are, and that $\mathcal{P}(\overline{\varphi} \wedge e) = 0$ , thus $\mathcal{P}(e \mid \overline{\varphi}) = 0$. Notice that $e \not\sqsubseteq \varphi$ by hypothesis.

- Now, we show that it might contain also spurious patterns, which is

$$\exists_{\varphi^* \rhd e \notin \mathcal{W}} \ \pi(e) \subseteq \text{clauses}\,(\varphi^*) \cup \{\varphi^*\}.$$

---

[3]These results assume a BIC regularisation but hold for any convergent regularization score.

These $\varphi^* \rhd e$ are of two types: sub-formulas spurious or topologically spurious (which include transitivities, as we may recall). For the former case note that

$$\forall_{\varphi \rhd e \in \mathcal{W}} \ \forall_{\hat{\varphi}^* \in \text{clauses}(\varphi)} \ \hat{\varphi}^* \rhd e \notin \mathcal{W},$$

but satisfies both temporal priority and probability raising. Also, consider any other $\hat{\varphi}^*_\star \sqsubseteq \hat{\varphi}^*$ and note that even this might satisfy both temporal priority and probability raising. For the latter case, it might be that there exists some other $\varphi^*$ such that, it is positively statistically correlated to a real pattern, and that might satisfy Suppe's conditions as well.

Thus, for any $e \in G$ such that $\varphi \rhd e \in \mathcal{W}$

$$\pi(e) = \{\varphi\} \cup \mathcal{S},$$

where $\mathcal{S}$ is a set of spurious patterns. We now examine the relation holding between the selectivity DAG and its modification performed via BIC. The derivations shown in the following hold regardless of the type of regularization which enjoys convergency.

We denote these DAGs as $\mathcal{D}_{\text{pf}}$ and $\mathcal{D}_{\text{BIC}}$.

(i) First, we show that all true patterns in $\mathcal{D}_{\text{pf}}$ are in $\mathcal{D}_{\text{BIC}}$, i.e.

$$\forall_{\varphi \rhd e \in \mathcal{W}} \ \pi_{\text{BIC}}(e) = \{\varphi\} .$$

Note that, although in general $\mathcal{P}(a \wedge b) \leq \min\{\mathcal{P}(a), \mathcal{P}(b)\}$, for the true patterns the following holds: $\mathcal{P}(\varphi \wedge e) = \mathcal{P}(e)$, when $\epsilon_+ = \epsilon_- = 0$; it is the maximum value for this joint probability, thus ensuring the maximum-likelihood fit. Thus the pattern is maintained in $\mathcal{D}_{\text{BIC}}$.

(ii) Second, we need to show that if $\forall \varphi^* \rhd e \notin \mathcal{W}$ but present in $\mathcal{D}_{\text{pf}}$, there exists a pattern $\varphi \rhd e \in \mathcal{W}$, which is present in $\mathcal{D}_{\text{pf}}$ and in $\mathcal{D}_{\text{BIC}}$ and any $\varphi^* \rhd e$ is not in $\mathcal{D}_{\text{BIC}}$.

Note that $\mathcal{P}(\varphi \wedge e) = \mathcal{P}(e)$, as above. Instead, $\mathcal{P}(\varphi^* \wedge e) < \mathcal{P}(e)$ since it is spurious, hence $\mathcal{P}(\varphi \wedge \varphi^* \wedge e) < \mathcal{P}(\varphi \wedge e)$, thus the likelihood fit of $\varphi \rhd e$ is maximal with respect to any of the patterns $\varphi^* \rhd e$.

To extend the proof to $\epsilon_+ = \epsilon_- \in [0, 1)$ with uniform noise, it suffices to note that the marginal and joint probabilities change monotonically as a consequence of the assumption that the noise is uniform. Thus, all inequalities used in the preceding proof still hold, which concludes the proof. $\qquad \square$

Notice that if it could be assumed that $\Phi$ characterizes $\mathcal{W}$ well, then all true patterns would be in $\Phi$, and the corollaries below follows immediately.

**Corollary 2** (Exhaustivity). *Assuming the same hypothesis as for the theorem above, $D(\mathcal{U}) \Vdash \mathcal{W}$.*

□

**Corollary 3** (Least Fixed Point). $\mathcal{W}$ *is the* lfp *of the monotonic transformation*

$$\bigsqcup_{\Phi} D(\Phi) \equiv D\Big(\bigsqcup_{\Phi}\Phi\Big) \Vdash \mathcal{W}. \qquad \Box.$$

Since a direct application of this theorem incurs a prohibitive computational cost, it only serves to idealize the ultimate power of the framework we have proposed. That is, the theorem only states that CAPRI is able to select only the true patterns asymptotically (in the sample size), regardless of how the putative hypotheses size $\mathcal{U}$ grows, e.g., in the worst-case exponentially. It also clarifies that the algorithm is able to "filter out" all the spurious patterns (true negatives), and produces the true positives more and more reliably as a function of the computational and data resources.

Now we restrict our attention to co-occurrence types of patterns so as to enable a fair comparison with [10]. We denote with $\mathcal{C} \subset \mathcal{U}$ the set of all possible such patterns, and we prove the following.

**Theorem 6** (Inference of co-occurrence patterns). *Suppose $\Phi = \emptyset$; as before, let the sample size $m \to \infty$ and let the data be uniformly corrupted by false positives and negatives rates $\epsilon_- = \epsilon_+ \in [0, 1)$. Then only co-occurrence patterns on atomic events are inferred, which are either true or spurious for general CNF formulas. That is: if $D(\emptyset) \Vdash \Sigma$ then $\Sigma \subseteq \mathcal{C}$. Furthermore,*

1. *$\Sigma \cap \mathcal{W}$ are true patterns and*

2. *For any other pattern $\alpha \triangleright e \in (\Sigma \setminus \Sigma \cap \mathcal{W})$ there exist $\beta \triangleright e \in \mathcal{W} \setminus \mathcal{C}$ such that $\beta$ screens off $\alpha$ from $e$.*

*Proof.* Consider the proof of the previous theorem. In this case, we am dealing with formulas such that clauses $(\varphi) \subseteq G$, i.e., formulas do not have any disjunctive component. All the derivations for Theorem 2 can be carried out in this context, notice that: formulas considered in step $(i)$ of such a proof are those which are purely conjunctive and correctly inferred. Similarly, formulas in $(ii)$ are those that screen off the false patterns, but are incorrectly present in $\mathcal{D}_{\text{BIC}}$. □

This theorem states that, even if one is neither willing to pay the cost of augmenting CAPRI's input with patterns nor able to find any suitable one, the algorithm is still capable of inferring singleton and conjunctive instances of $\triangleright$ relation, whose members are either true or part of a more complex types of patterns that fall outside CAPRI's scope. An immediate corollary of these two theorems is that CAPRI works as specified, when it is fed with all possible co-occurrence patterns.

**Corollary 4.** *Under the hypothesis of the above theorems, $D(\emptyset) \Vdash \Sigma \iff D(\mathcal{C}) \Vdash \Sigma$.*
□

In practice, this algorithm, though still exponential, is certainly less computationally intensive. For instance, when using $\mathcal{C}$ than with $\mathcal{U}$, it can trade off computational complexity against expressivity of the inferred patterns.

## D.2   Results: synthetic data

**Setting for comparison.**   The performance of all the algorithms were evaluated empirically with four different types of topologies: (*i*) *trees*, (*ii*) *forests*, (*iii*) *DAGs without disconnected components* and (*iv*) *DAGs with disconnected components*. Irrespective of the topology considered, we exclusively used atomic events, which implies that either singleton or co-occurrence patterns were used in the experiments. Based on Corollary 3, it sufficed to run CAPRI with $\Phi = \emptyset$. This strategy is consistent with the fact that our algorithm can infer more general formulas if an input "set of putative causes, $\Phi \neq \emptyset$" is given in addition – a fact which, without the care taken, could have unfairly and favorably biased our analysis in the more general situation. For the sake of completeness, however, we also tested specific CNF formulas, as shown in the next sections.

Type (*i* − *ii*) topologies are DAGs constrained to have nodes with a unique parent; condition (*i*) further restricts such DAGs to have no disconnected components, meaning that all nodes are reachable from a starting root $r$. Practically, condition (*i*) satisfies $|\pi(j)| = 1$ for $j \neq r$, and $\pi(r) = \emptyset$, while in (*ii*) we allow more roots to be present. This kind of topologies can be reconstructed with either ad-hoc algorithms [117, 38, 173] or general DAG-inference techniques [171, 179, 10, 167, 79]. Type (*iii* − *iv*) topologies are DAGs which have either a unique starting node $r$, or a set of independent sub-DAGs. Similarly, condition (*iii*) satisfies $|\pi(j)| \geq 1$ for $j \neq r$, and $\pi(r) = \emptyset$, while in (*iv*) we allow more roots to be present, as it was in (*ii*). This kind of topologies are not reconstructible with tree-specific algorithms, and thus only algorithms in [171, 179, 10, 167, 79] could be used for comparison. The algorithm for the synthetic data generation is described in the following paragraph.

**Generating synthetic data.**   Let $n$ be the number of events we want to include in a DAG and let $p_{\min} = 0.05$, $p_{\max} = 0.95$, $p_{\min} = 1 - p_{\max}$. A *DAG without disconnected components* (i.e. an instance of type (*iv*) topology) with maximum depth $\log n$ and where each node has at most $w^*$ parents (i.e. $|\pi(j)| \leq w^*$, for $j \neq r$) is generated as follows:

1: pick an event $r \in G$ as the root of the DAG;
2: assign to each $j \neq r$ an integer in the interval $[2, \lceil \log n \rceil]$ representing its depth in the DAG (1 is reserved for $r$), ensure that each level has at least one event;
3: **for all** events $j \neq r$ **do**
4:     let $l$ be the level assigned to $e$;
5:     pick $|\pi(j)|$ uniformly over $(0, w^*]$, and accordingly define $\pi(j)$ with events selected among those at which level $l - 1$ was assigned;
6: **end for**
7: assign $\alpha(r)$ a random value in the interval $[p_{\min}, p_{\max}]$;
8: **for all** events $j \neq r$ **do**
9:     let $y$ be a random value in the interval $[p_{\min}, p_{\max}]$, assign

$$\alpha(j) = y \prod_{x \in \pi(j)} \alpha(x) \, ;$$

10: **end for**
11: **return** the generated DAG;

When an instance of type $(iv)$ topology is to be generated, we repeat the above algorithm to create its constituent DAGs. In this case, if multiple DAGs are generated, each one with randomly sampled $n_i$ events we require that $|G| = \sum n_i = n$. When instances of type $(i)$ topology are required $w^* = 1$, and by iterating multiple independent sampling instances of type $(ii)$ topology are generated. When required DAGs were sampled, these are used to generate an instance of the input matrix $D$ for the reconstruction algorithms.

### D.2.1   Performance with different topologies and small datasets

Here we estimate the performance of CAPRI for datasets with sizes that are likely to be found in currently available cancer databases, such as The Cancer Genome Atlas, TCGA [136], i.e. $m \approx 250$ samples, and 15 events. The results are shown in Figure §D.1, for topologies $(i)$ and $(ii)$, and Figure §D.2, for topologies $(iii)$ and $(iv)$. There, we show all the results obtained by running the algorithm with bootstrap resampling, although results (data not shown) without this pre-processing leave the conclusions unaffected.

Results suggest a trend, as to be expected: namely, performance degrades as noise increases and sample size diminishes. However, it is particularly interesting to notice that, in various settings, CAPRI almost converges to a perfect score even with these small datasets. This happens for instance with type $(i - ii)$ topologies, where the Hamming distance almost drops to 0 for $m \geq 150$. In general, it is also clear that reconstructing forests is easier than trees, when the same number of events $n$ is considered. This is a consequence of the fact that, once $n$ is fixed, forests are likely to have less branches since every tree in the forest has less nodes. When reconstructing type $(iii - iv)$ topologies, instead, the convergence-speed of CAPRI to lower Hamming distance is slower, as one might reasonably expect. In fact, in those settings the distance never drops below 3, and more samples would be required to get a perfect score. We consider this to be a remarkable result, when compared to the worst-case Hamming distance value of $15 \cdot 14 = 210$. Panels of Figure §D.2 also suggest that disconnected DAGs are easier to reconstruct than connected ones, when a fixed number of events is considered. Similarly to the above, this could be credited to the fact that the size of the conjunctive claims is generally smaller, for fixed $n$. With respect to the precision and recall scores, one may note that CAPRI seems to be quite robust to noise, since the loss in the score-values appear nearly unaffected by any increase in the noise parameter.

### D.2.2 Comparison with other reconstruction techniques

We compare now with state-of-the-art approaches mentioned in the main text[4], which we divide into three categories: *structural* - Incremental Association Markov Blanket (IAMB) and PC algorithm -, *likelihood* - Bayesian Information Criterion (BIC) and Bayesian Dirichlet (BDE) and *hybrid* - Conjunctive Bayesian Networks (CBN) and Cancer Progression Inference with Single Edges (CAPRESE). For all the algorithms we used their standard R implementations: for IAMB, BDE and BIC we used package `scutari2009learning` [168], for the PC algorithm we used package `pcalg`, for CAPRESE we used `TRONCO` [4] (first release) and for CBN we used `h-cbn` [11].

Clearly, other algorithms exist in the literature, but we selected those which satisfied at least one of the following criteria: earlier, they have proven to be more effective in inferring "causal" claims, i.e., they are considered the best algorithms to infer "causal networks" (i.e., IAMB and PC); they regularize the Bayesian over-fit (i.e., BDE and BIC); they assume a prior (i.e. BDE) or they were developed specifically for cancer progression inference (i.e., CBN and CAPRESE). Prominent among the ones absent in this study are the following: *Grow and Shrink* [124], which preliminary analysis have shown to be very similar to IAMB, and the *DiProg algorithm* [50], which unrealistically requires advanced knowledge of input error rate to reconstruct a model; note that this kind of information is not generally available *a priori*.

Notice that we selected all the algorithms capable of inferring generic DAGs but CAPRESE [117], which can only be applied to infer trees or forests (i.e., type $(i - ii)$ topologies). In the literature there exist other approaches specifically tailored for such topologies, e.g., [38, 173]; however, since in [117] it is shown that CAPRESE performs better than other approaches, we assume no loss of information in restricting our study. We place CAPRI in the *Hybrid* category, though we clearly compare its performance with all the other approaches in order to quantify its suitability for reconstruction of all classes of topologies, as defined earlier.

The general trend is summarized in Figure §D.3, where we rank all of these algorithms according to their median performance, estimated as a function of noise and sample size, and provide the parameters used for comparison. In Figure §D.4, we compare CAPRI with the structural approaches (IAMB and PC). In Figure §D.5, we compare it with the likelihood approaches (BIC and BDE) and, finally, in Figure §D.6, we compare it with the hybrid algorithms. We remark that, because of the high computational cost of running CBNs, which relies on a nested Expectation-Maximization algorithm with Simulated Annealing, the number of ensembles performed is limited to 100 for CBNs, while it is 1000 for all other algorithms. Though this strategy provides less robust statistics for CBNs (i.e., less "smooth" performance surfaces), it is still sufficiently accurate to indicate the general comparative trends and relative performance efficiency.

---

[4] Classic versions of the IAMB and PC algorithm were further subjected to log-likelihood optimization to assign a direction to all of the computed non-oriented edges. This additional feature is necessary to permit a fair comparison against various structural approaches, which, otherwise, would be penalized with a worse Hamming distance, since these algorithms, in principle, can return non-oriented edges. Note that progression models, by their very nature, consist only of oriented structures.

### D.2.3   Reconstruction without hypotheses: disjunctive patterns

Recall that our algorithm expects as input all the hypothesized patterns to infer more expressive logical formulas, i.e., hypotheses with pure CNF formulas or even disjunctive patterns over atomic events. Nonetheless, it is instructive to investigate its performance under two specific conditions, especially to clarify the robustness with respect to imperfect regularities (the, e.g, "noisy and"): namely, (*i*) without hypotheses ($\Phi = \emptyset$) and (*ii*) for datasets sampled from topologies with *disjunctive* patterns.

To generate the input dataset, we have to modify the generative procedure used for the other tests, thus reflecting the switch from co-occurrence to disjunctive patterns. This task is actually rather simple, since we just change the labeling function $\alpha$ to account for the probability of picking any subset of the clauses in the disjunctive pattern, while omitting the others. We use DAGs with 10 events and disjunctive patterns with at most 3 atomic events involved, which is a reasonable size, given the events considered. Clearly, this setting is generally harder than the one shown in Figures §D.4– §D.6, thus we expect performance to be somewhat inferior.

Here we compare CAPRI with all the algorithms used so far, and we show the result of this comparison in Figure §D.7, where $\Phi = \emptyset$, as noted earlier. The plot clearly confirms the trends suggested by previous analyses: namely, CAPRI infers the correct patterns more often than the others. Note also that the performance is measured on the reconstructed topology only, since, without input hypotheses, the algorithm evaluates only co-occurrence types of patterns, and does not allow different types of relations (e.g. disjunctions) to be inferred automatically. However, as anticipated, observed performance improvement is now much lower, and the Hamming distance fails to rise above 4. Furthermore, convergence to optimal performance was not observed for $m \leq 1000$, and it appears not to be reachable even for $m \gg 1000$ (at least so, when no hypotheses are used). It is also possible that, as $n$ and the number of maximum disjunctive patterns increase, the result could be an even less satisfactory speed of convergence.

### D.2.4   Reconstruction with hypotheses: synthetic lethality

We wondered whether CAPRI would be able to infer synthetic lethality relations, when these are directly hypothesized in the input set $\Phi$. We started with a test of the simplest form: e.g., $[a \oplus b \; \triangleright \; c]$, for a set of events $G = \{a, b, c\}$, where we force progression from $a$ to $c$ to be preferential, i.e. it appears with 0.7 probability, whereas $b$ to $c$ does so with only 0.3 probability, thus implying that samples involving $(a \wedge \bar{b})$ will be more abundant than those involving $(\bar{a} \wedge b)$. Despite this being the smallest possible synthetically lethal pattern, the goal was to estimate the *probability* of such a pattern being robustly inferable, when $\Phi = \{a \oplus b \triangleright c\}$, and its dependence on the sample size and noise. We measured the performance of all the algorithms, with an input lifted according to the pattern so that all algorithms start with the same initial pieces of information. The performance metric estimates how likely an edge from $a \oplus b$ to $c$ could be found in the reconstructed structures.

We show the results of this comparison in Figure §D.8. We note that CAPRI suc-ceeds in inferring the synthetic lethality relation more frequently than 93% of the times, irrespective of the noise and sample size used. More precisely, with $m \geq 60$ the algorithm infers the correct pattern under any execution, thus suggesting that CAPRI, with the correct input hypotheses, is able to infer complicated structures, many of which could have high biological significance. Naturally, it would be reasonably expected that the performance of any of these algorithms would drop, were the target relations part of a bigger model.

| **Setting A** ($\nu = 0$) | mean | median | standard deviation |
|---|---|---|---|
| CAPRESE | 0.006 | 0.005 | 0.005 |
| BIC | 0.023 | 0.022 | 0.011 |
| IAMB | 0.028 | 0.027 | 0.005 |
| CAPRI without bootstrap | 0.029 | 0.029 | 0.003 |
| BDE | 0.041 | 0.032 | 0.063 |
| PC | 0.144 | 0.112 | 0.154 |
| CAPRI with bootstrap | 1.143 | 1.056 | 0.360 |

| **Setting B** ($\nu = .10$) | mean | median | standard deviation |
|---|---|---|---|
| CAPRESE | 0.005 | 0.005 | 0.001 |
| BIC | 0.022 | 0.022 | 0.003 |
| IAMB | 0.029 | 0.028 | 0.004 |
| CAPRI without bootstrap | 0.030 | 0.029 | 0.004 |
| BDE | 0.030 | 0.028 | 0.010 |
| PC | 0.103 | 0.094 | 0.034 |
| CAPRI with bootstrap | 0.719 | 0.689 | 0.138 |

Table D.1: **Comparison in the execution time.**

### D.2.5 Execution time

We report in table §D.1 an evaluation of the execution time for all the algorithms we tested, but CBN - which computation time is more than one order of magnitude higher than the competing techniques. Two distinct settings of experiments were used: Setting ($A$): $n = 10$ events, $m = 100$ samples, $\nu = 0$ noise; Setting ($B$): $n = 10$, $m = 100$, $\nu = .10$. Results account for the average time of execution as of 100 randomly generated topologies (one dataset sampled per topology). Time unit is *second* and the test was performed on a MacBook with 2.3 GHz Intel i7 processor, 16 Gb of RAM and Yosemite 10.9 OS.

To allow a fair comparison of CAPRI against the other algorithms we both executed the algorithm with and without bootstrap preprocessing, in order to asses the prima facie condition (Mann-Withney U test being performed in the former case). Execution timings are sorted according to mean time.

## D.3    Biological examples

In this Section two examples of analysis of genomic alterations will be presented.

### D.3.1    Atypical Chronic Myeloid Leukemia

#### Input hypotheses for CAPRI (supervised mode)

By fetching the literature we selected the following patterns to input as CAPRI's hypotheses:

(1) "exclusivity among ASXL1 and SF3B1 mutations" [115]:

$$(\text{ASXL1 } \textit{Nonsense point} \oplus \text{ASXL1 } \textit{Ins/del}) \oplus \text{SF3B1 } \textit{Missense point}$$

(1) "exclusivity among TET2 and IDH2 mutations" [53]:

$$(\text{TET2 } \textit{Nonsense point} \oplus \text{TET2 } \textit{Missense point} \oplus \text{TET2 } \textit{Ins/del}) \oplus \text{IDH2 } \textit{Missense point}$$

These patterns were used to build CAPRI's hypotheses which were tested against all events which do not appear in the above pattern itself, e.g., pattern (1) was tested against all input events but those involving ASXL1 and SF3B1 genes.

As shown in the main text, among all, the following hypothesis gets selected by CAPRI

$$(\text{ASXL1 } \textit{Nonsense point} \oplus \text{ASXL1 } \textit{Ins/del}) \oplus \text{SF3B1 } \textit{Missense point} \triangleright \text{CBL } \textit{Missense point}$$

#### aCML progression model with different techniques

In Figure §D.9 one can find the progression models reconstructed on the the ACML dataset [154], with 3 different algorithms: (*i*) CAPRESE, (*ii*) BIC and (*iii*) IAMB. These three techniques were chosen for this comparative study because of the overall better performance on synthetic tests (see Section 3.2-3.4 of the SI). The reconstruction obtained with CAPRI can be found in Figure 5 in the main text. For a biological interpretation of the results please refer to Section 4.2 in the main text.

Note that all the progression model share some specific selective advantage relations, yet being substantially different. Relations involving SETBP1 and ASXL1 and those involving TET2 and EZH2 are, in fact, inferred by all the four algorithms, yet with different confidences and, sometimes, edge direction. In addition, IAMB does not include CBL in the path involving SETBP1 and ASXL1, and none of the algorithms but CAPRI can infer the complex pattern involving ASXL1 mutations of both types and SF3B1 (Figure 5 in the main text). Finally, note that IAMB and BIC are often not able to disambiguate the edge direction and this represent a major limit of these techniques with respect to CAPRI and CAPRESE.

### D.3.2 Ovarian cancer

**Ovarian cancer progression model with different techniques**

We analyzed an ovarian cancer dataset reporting chromosome-level amplifications and deletions detected via Comparative Genome Hybridization in [100]. Similar to the case of aCML, we used 4 different techniques to infer a progression models for events included in the dataset: CAPRI (unsupervised), CAPRESE, BIC and IAMB. Models and input dataset are shown in Figure §D.10. Like with aCML extraction, the progression models share only some of the inferred relations. Among the most relevant differences is the conjunctive pattern inferred by CAPRI between the loss on chromosome *5q* (5q−) and the gain on chromosome *8q* (8q+) which is predicted to select for a loss on 8p; note also the aforementioned limitation of BIC and IAMB in disambiguating the direction of some of the inferred relations. Note that CAPRI infers a co-occurrence pattern of selective advantage which is not input a priori as hypothesis - unsupervised execution. In summary, CAPRI displays a better overall confidence on the reconstructed model.

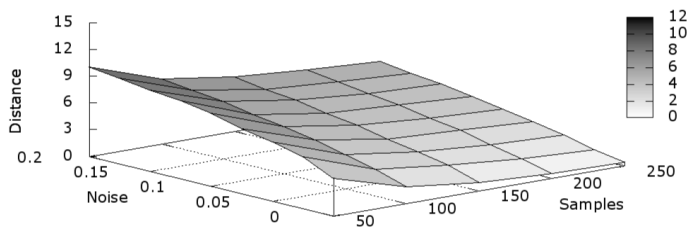

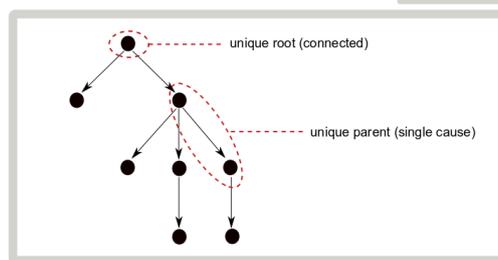

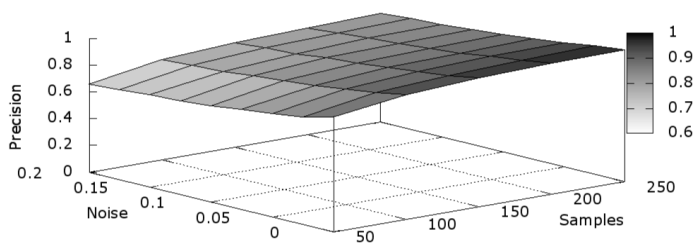

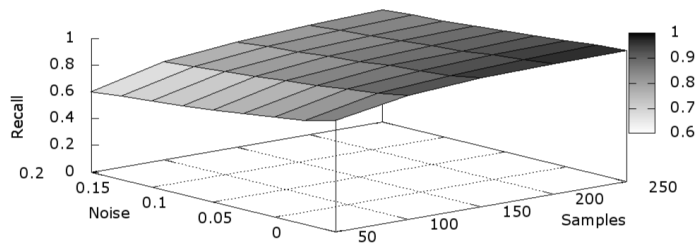

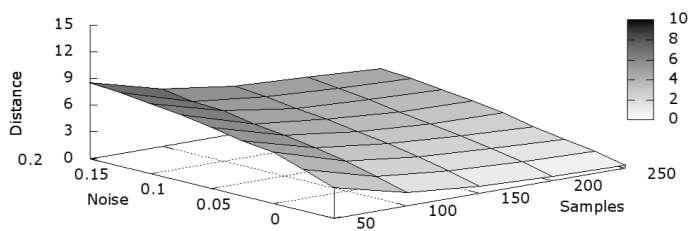

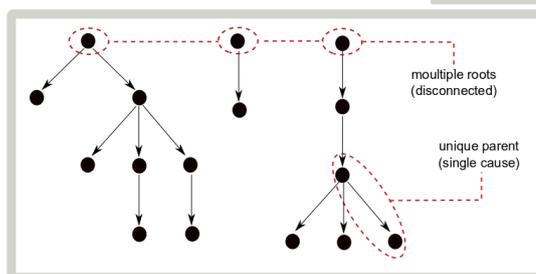

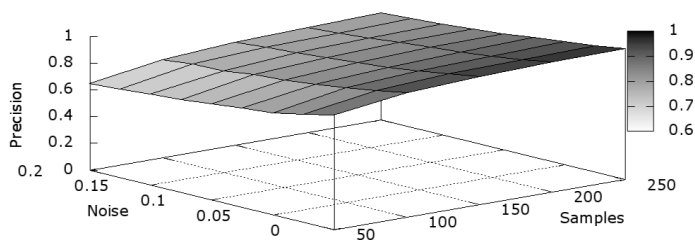

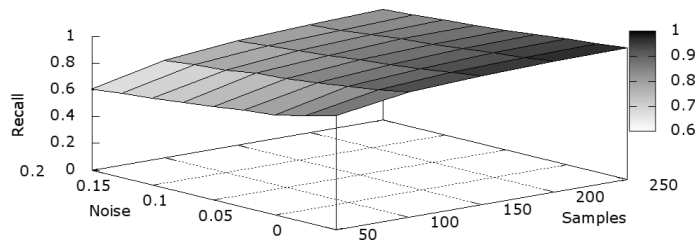

Figure D.1: **Reconstruction of trees and forests with small datasets.** Hamming distance, precision and recall of CAPRI for synthetic data generated by trees (i.e., models with a singleton pattern per event and a unique progression), in top panels, and by forests (i.e., models with a singleton pattern per event but multiple independent progressions), in bottom panels. In both cases $n = 15$ events are considered, $m$ ranges from 50 to 250 and the noise rate ranges from 0% to 20%. To have a reliable statistics, for each type of topology, we generate 100 distinct progression models and, for each value of sample size and noise rate, we sample 10 datasets from each topology. Thus, every performance entry is the average of 1000 reconstruction results. Notice that Hamming distance almost drops to 0 for $m \geq 150$ and that precision and recall decrease very little as noise increases.

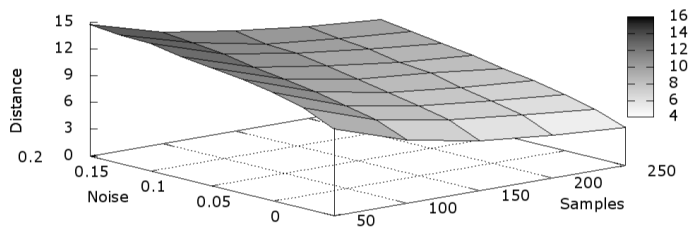

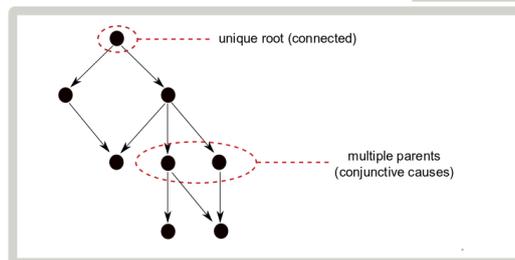

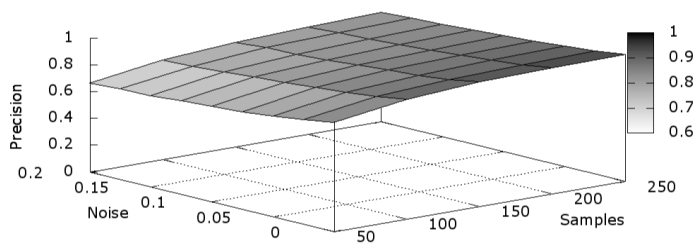

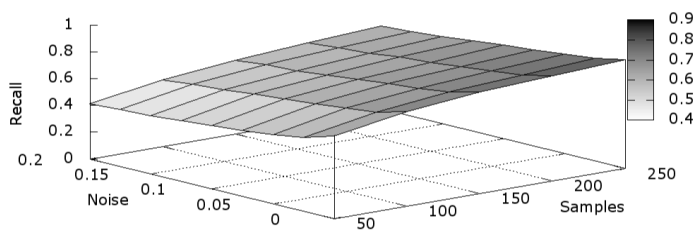

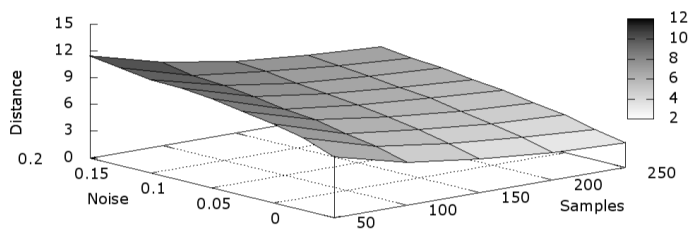

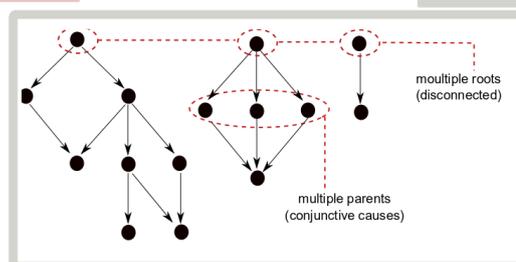

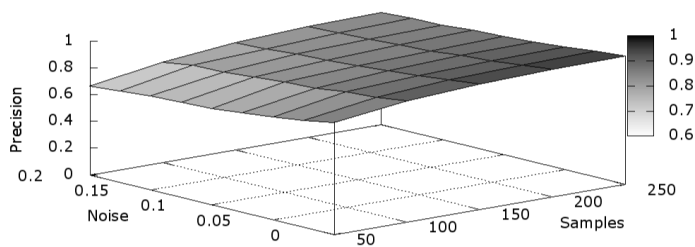

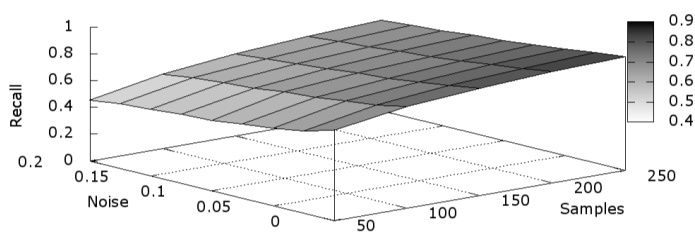

Figure D.2: **Reconstruction of DAGs with small datasets.** Hamming distance, precision and recall of CAPRI for synthetic data generated by connected DAGs (i.e., models with either a singleton or co-occurrence pattern per event and a unique progression), in top panels, and by disconnected DAGs (i.e., models with either a singleton or co-occurrence pattern per event and multiple progressions), in bottom panels. In both cases the same parameters as in Figure §D.1 are used ($n = 15$, $50 \leq m \leq 250$, $0\% \leq \nu \leq 20\%$ and every performance entry is the average of 1000 reconstructions). In this setting, which is harder than the one shown in Figure §D.1, Hamming distance does not reach values below 3 – a reasonably small number for our purposes – while precision and recall still suffer very little as noise increases.

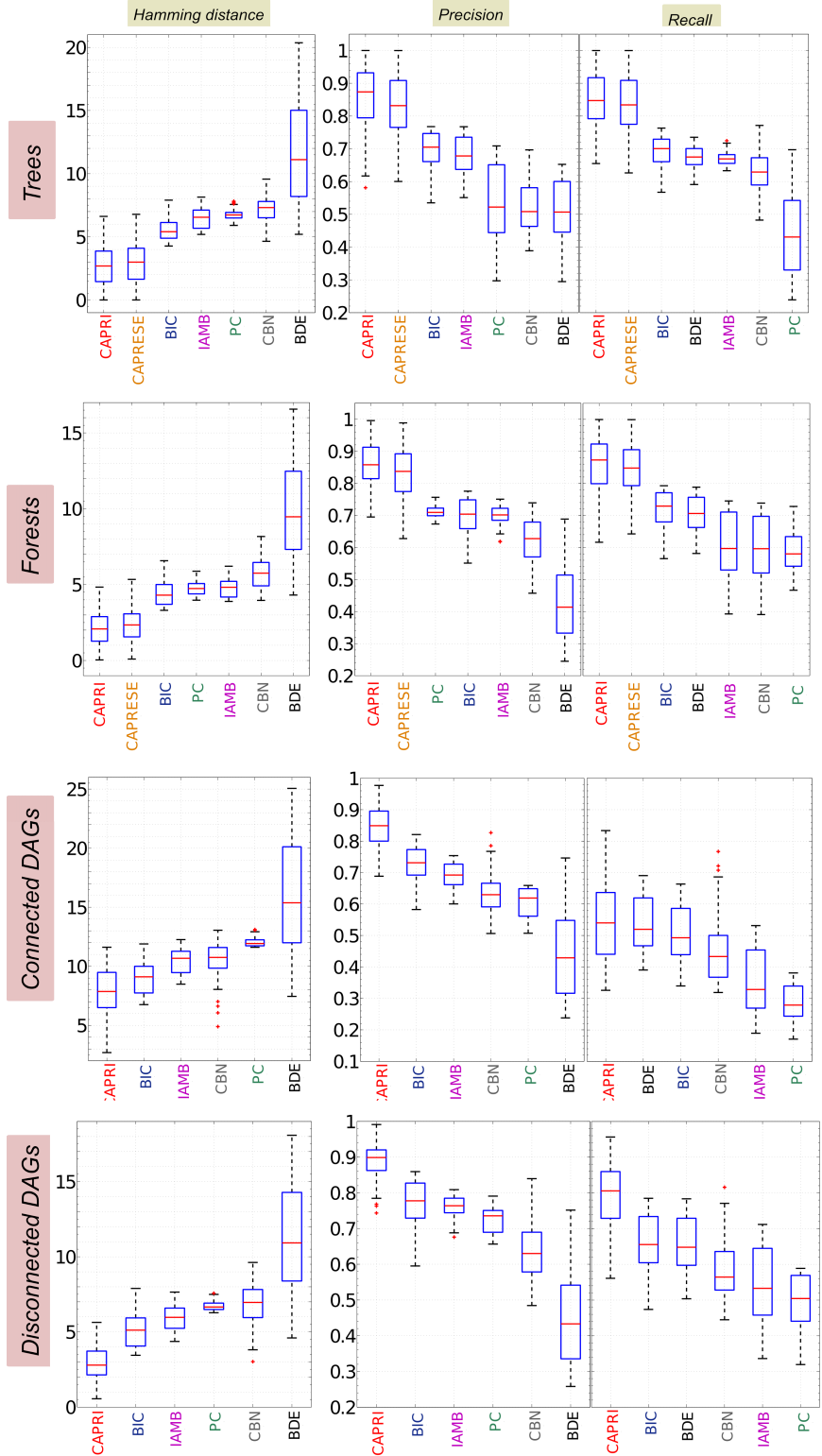

**Parameter values**

| $n$ | number of events | 10 |
|---|---|---|
| $m$ | number of samples | $[50, 1000]$ |
| $\nu$ | rate of false positives $\epsilon_+$ and negatives $\epsilon_-$ | $[0, 0.2]$ (0%-20% noise rate) |
| $-$ | ensemble size | 1000 (100 for CBN) |

Figure D.3: **Co-occurrence patterns: performance ranking.** We rank the algorithms we compared in Figure §D.4, §D.5 and §D.6 according to their performance for the parameters in the table. Rankings are divided according to the topology type and sorted according to the median performance.

# Structural algorithms

Hamming distance    Precision    Recall

**Trees**

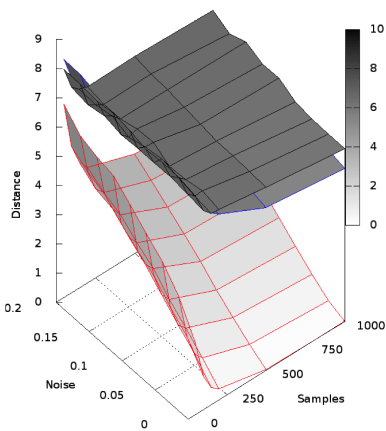 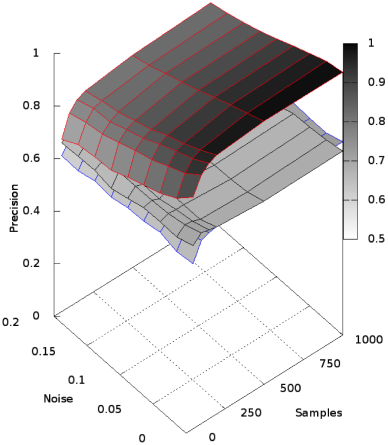 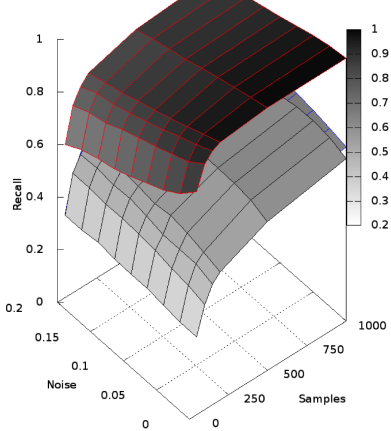

**Forests**

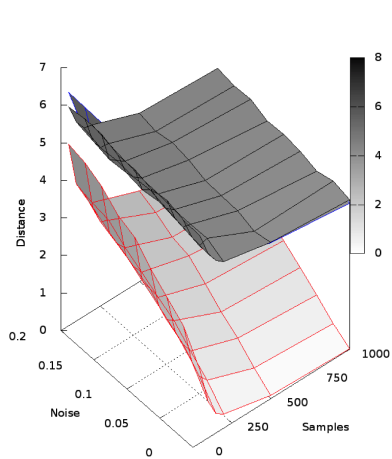 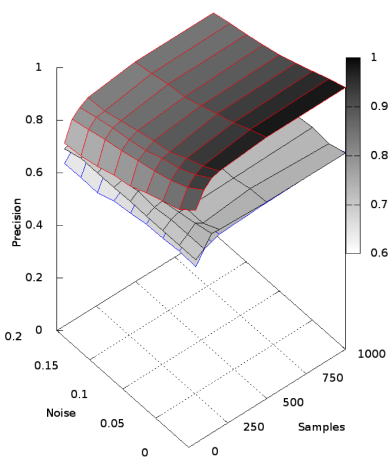 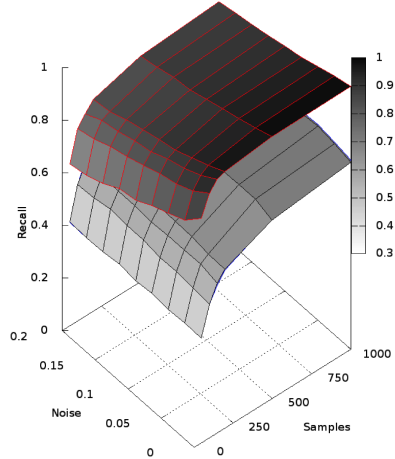

**Connected DAGs**

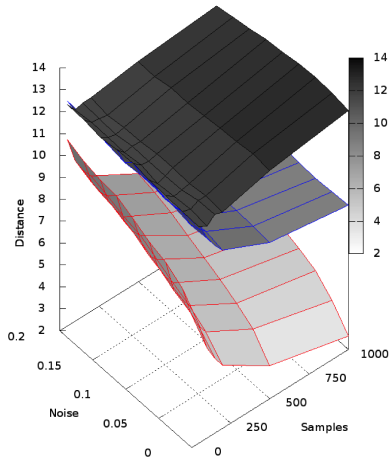 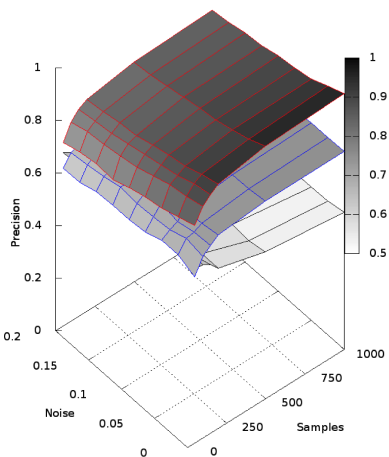 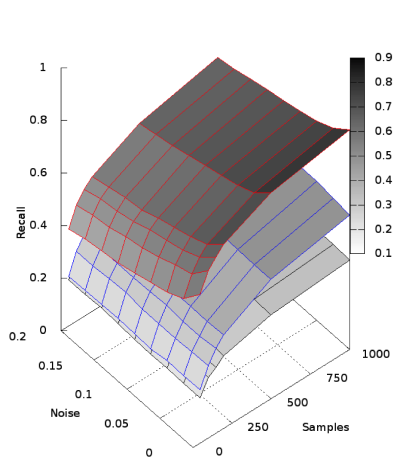

**Disconnected DAGs**

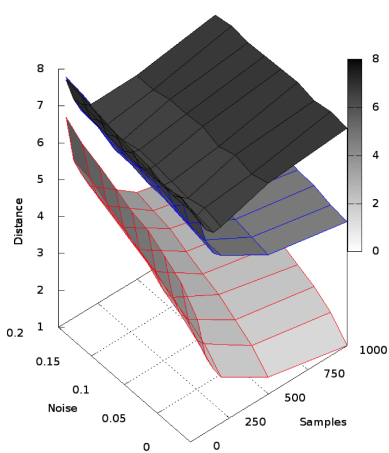 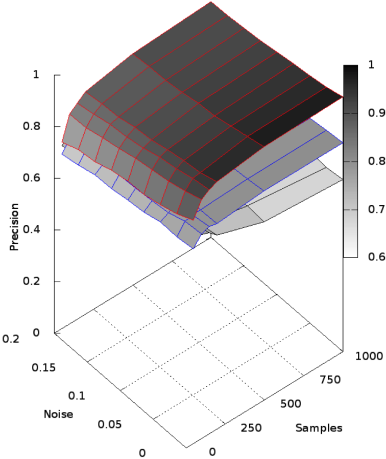 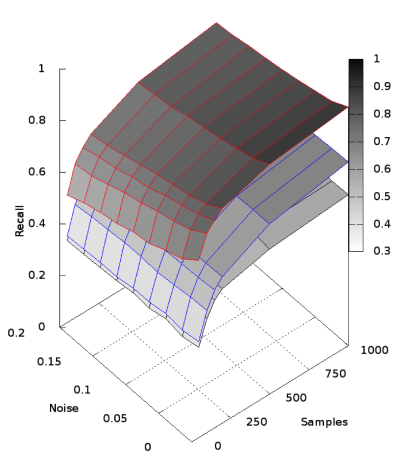

Figure D.4: **Comparison with related works: structural algorithms.** We compare CAPRI, IAMB and the PC algorithm to infer *trees*, *forests*, *connected DAGs* and *disconnected DAGs* with the parameters described in Table §D.3. Average Hamming distance, precision and recall are shown.

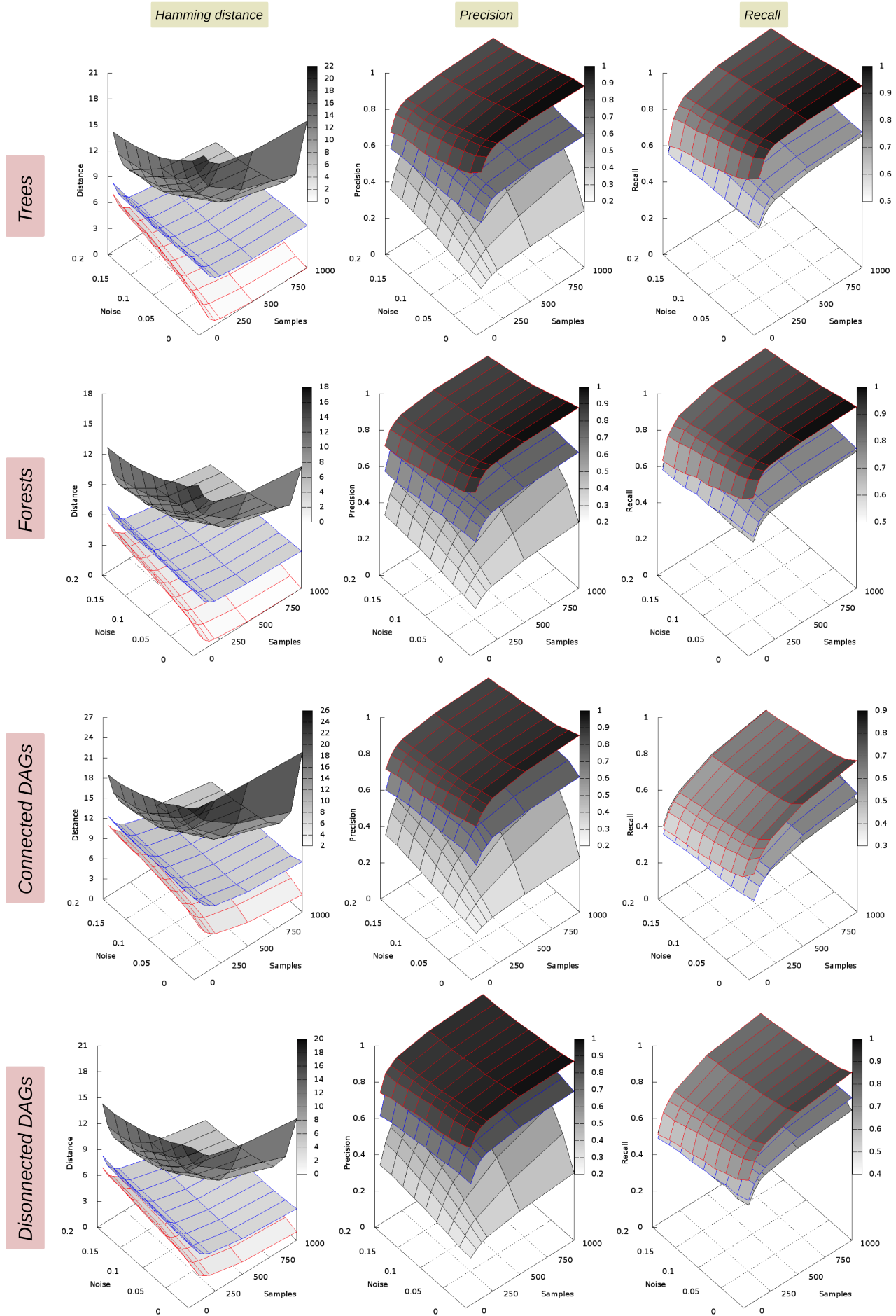

Figure D.5: **Comparison with related works: likelihood-based algorithms.** We compare CAPRI against likelihood-based methods optimizing BIC and BDE scores to infer *trees*, *forests*, *connected DAGs* and *disconnected DAGs* with the parameters described in Table §D.3. Average Hamming distance, precision and recall are shown.

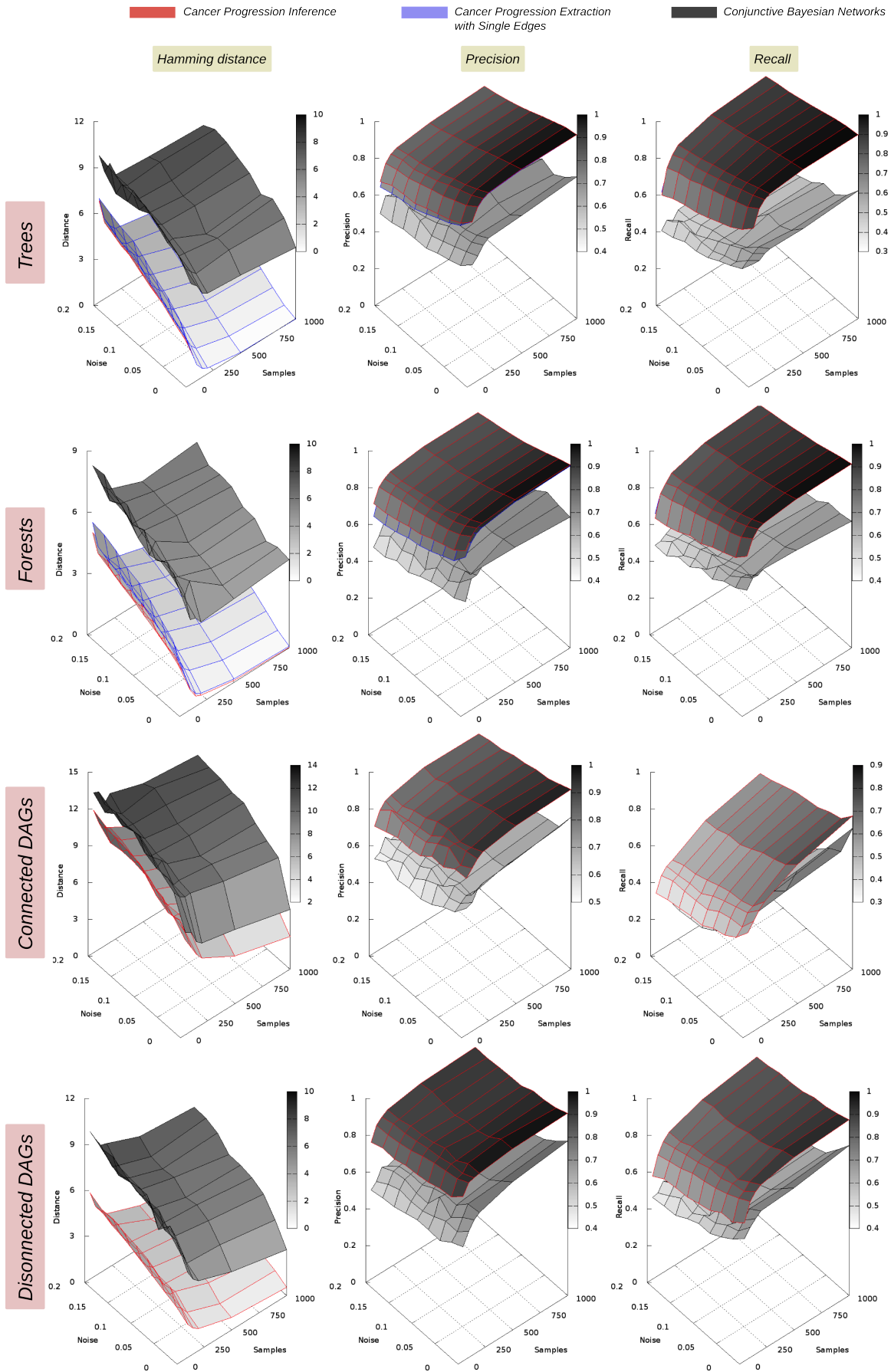

Figure D.6: **Comparison with related works: hybrid algorithms.** We compare CAPRI, CBNs and CAP-RESE to infer *trees*, *forests*, *connected* and *disconnected DAGs* with the parameters of Table §D.3 but, because of the computational cost of running CBNs with 100 annealing steps, we reduced the number of ensembles performed as: 100 for CBNs, 1000 for CAPRESE and, for CAPRI, 100 for DAGs and 1000 otherwise. Average Hamming distance, precision and recall are shown.

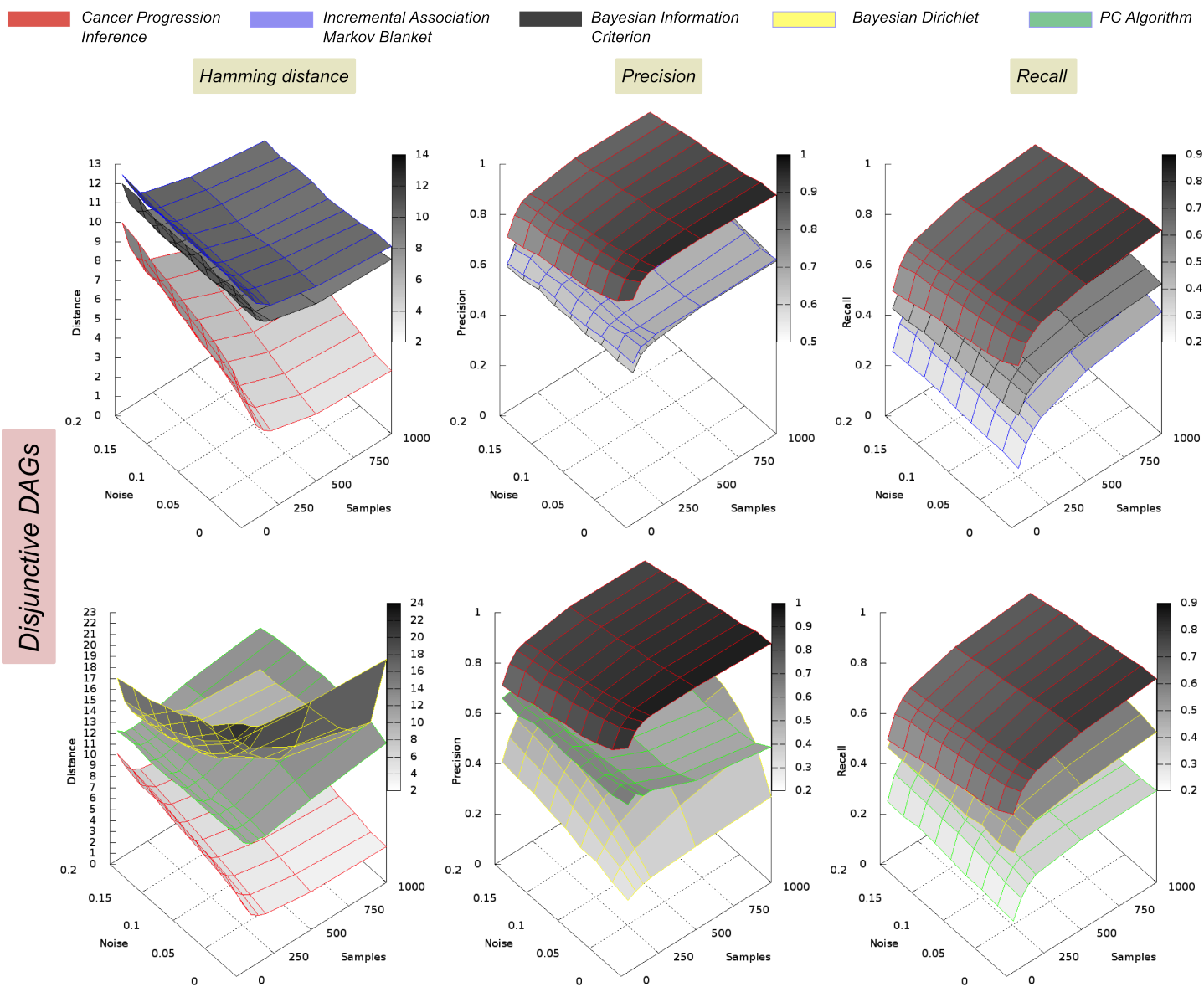

Figure D.7: **Reconstruction of disjunctive patterns with no hypotheses.** We compare CAPRI against all the algorithms to infer progressions with disjunctive patterns. In top panel we show IAMB as the best structural algorithm, and the BIC score as the best among likelihood-based methods, according to Table §D.3. In bottom panel we compare the other algorithms. No hypotheses ($\Phi = \emptyset$) are given as input to CAPRI. Input data is generated by DAGs with 10 atomic events and disjunctive patterns with at most 3 atomic events involved. Sample size ranges from 50 to 1000, noise rate from 0% to 20% and 1000 ensembles are generated for each configuration of noise and sample size. This setting is generally harder than the one shown in Figures §D.4– §D.6. Hamming distance, precision and recall are shown and confirm that this type or pattern is harder than the co-occurrent one to be inferred, hinting at the difficulty of modeling unbalanced confluent progressions.

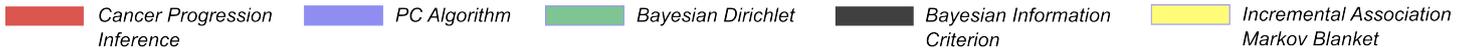

**Inference of a synthetic lethality relation**

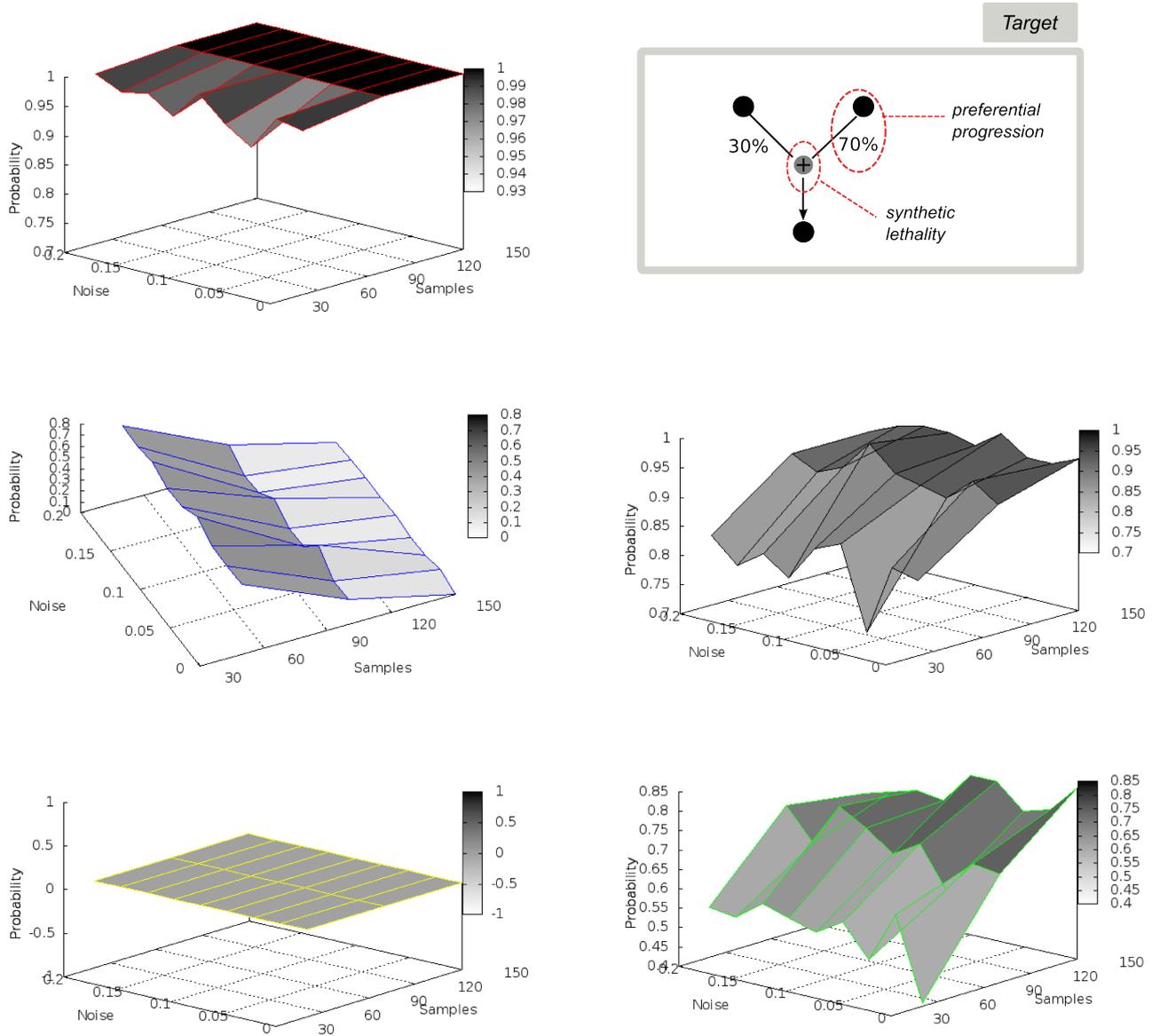

Figure D.8: **Reconstruction with hypotheses: synthetic lethality.** We show the *average probability* of inferring a claim $a \oplus b \rhd c$ (*synthetic lethality*), when this is provided in the input set $\Phi$. We show such a probability for CAPRI, the likelihood-based algorithms with BIC and BDE scores, and the structural IAMB and PC Algorithm. Data is generated from the model in the upper left panel (unbalanced "exclusive or" with a preferential progression), samples size ranges from 30 to 120, noise rate from 0% to 20% and 1000 ensembles are generated for each configuration of noise and sample size. Results suggest that a threshold level on the number of samples exists such that CAPRI infers the correct claim when $\Phi = \{a \oplus b \rhd c\}$. We executed all the algorithms with an input matrix lifted to contain the target claim.

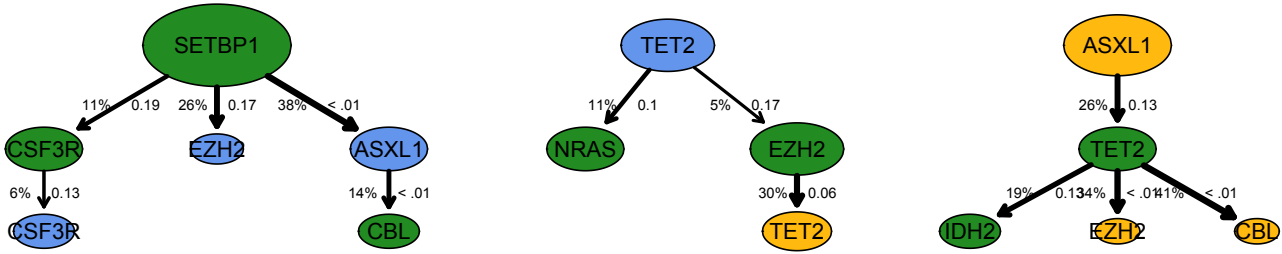

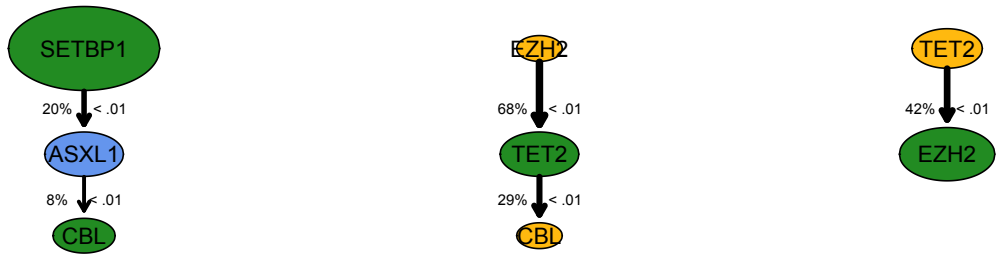

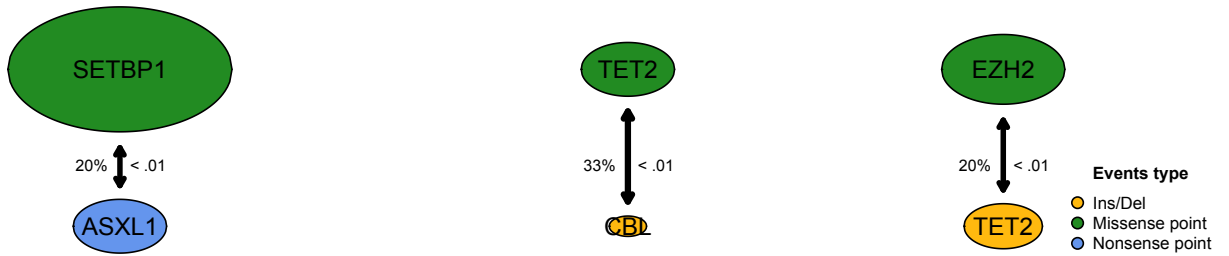

Figure D.9: **CAPRESE, IAMB and BIC progression models of aCML.** Progression models reconstructed from the ACML dataset described in the main text - taken from [154] - obtained with the following algorithms: CAPRESE, IAMB and BIC. The model inferred by CAPRI is shown in the Main Text. Confidence shown is assessed as for the CAPRI algorithm. Nodes are scaled differently to better layout the graphs reconstructed by every algorithm.

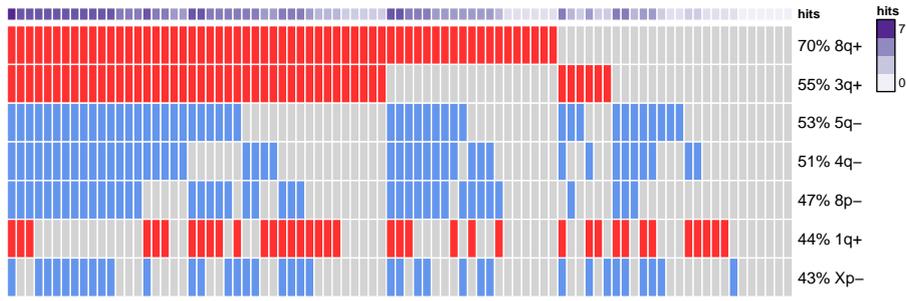

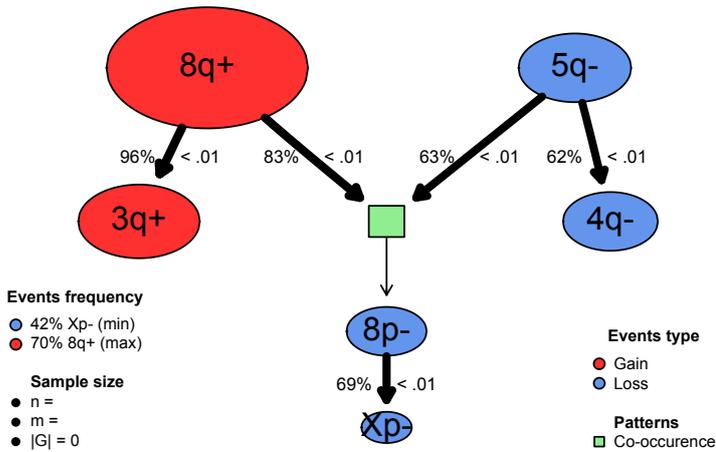

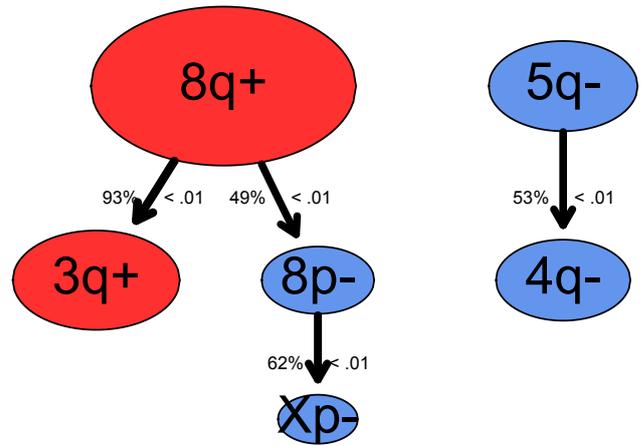

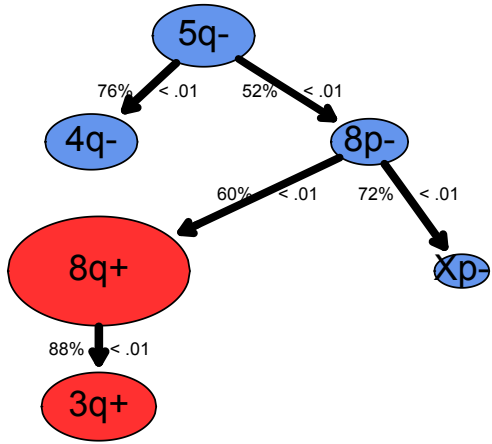

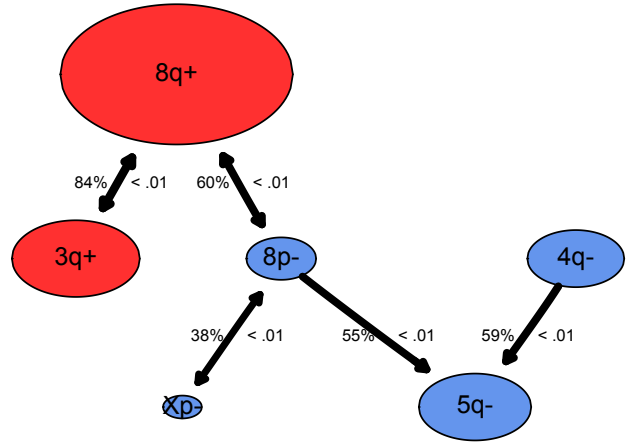

Figure D.10: **CAPRI, CAPRESE, IAMB and BIC inferred progression models of ovarian cancer.** Progression models reconstructed from the ovarian cancer Comparative Genome Hybridization dataset shown in top [100]. Algorithms used to infer the models are CAPRI, CAPRESE, IAMB and BIC. Confidence is shown as non-parametric bootstrap and hypergeometric test (p-values). Nodes are scaled differently to better layout the graphs reconstructed by every algorithm.

# CRC STUDY - SUPPLEMENTARY MATERIALS

Here we detail all the steps implemented to perform the pipeline widely discussed in Chapter §6. The source code to replicate this study is available for download along with the documentation detailing all the implementation at:

<div align="center">

http://bimib.disco.unimib.it/index.php/Tronco

</div>

## E.1 TCGA COADREAD project data

COADREAD provides genome-scale analysis of samples with exome sequence, DNA copy number, promoter methylation, messenger RNA and microRNA expression data which we used to define "training" and "control" datasets. In both validation and control datasets, only samples with both mutations and CNAs profiles were used. Table §E.1 details each of the four datasets.

**Training dataset.** Samples published in [137] were used as training; for these samples, TCGA provides somatic mutation profiles and high-resolution focal CNAs via GISTIC. These are obtained from TCGA data freeze as of 2 February 2012, downloaded on 12 March 2015, from the repository:

<div align="center">

https://tcga-data.nci.nih.gov/docs/publications/coadread_2012/

</div>

The following files were processed to produce the training data:

- `TCGA_CRC_Suppl_Table2_Mutations_20120719.xlsx`. *Somatic mutations* profiles obtained via whole-exome sequencing of 224 colorectal tumors[1]. All annotated mutations were considered for analysis;

---

[1]15995 mutations in 228 samples are annotated in a *Manual Annotation Format* (MAF). Samples were selected to univocally match the 224 patients as of the TCGA guidelines for aliquote disambiguation, see `https://wiki.nci.nih.gov/display/TCGA/TCGA+barcode`.

- `crc_gistic.txt.zip`. Focal *Copy Number Alterations* (CNAs) for 564 patients derived from whole-genome sequencing using the Illumina HiSeq platform. *High-level gains* and *homozygous deletions* were considered for analysis by selecting entries with GISTIC scores $\pm 2$;

- `crc_clinical_sheet.txt`. *Clinical data* summary with patient stage and *Micro Satellite Stabe/Instable* (MSS/MSI) status being any of: MSS, MSI-high and MSI-low.

The list of patients used was first reduced to those having *both* CNAs and somatic mutation data, and then was split in two groups: MSI-HIGH and MSS. The training cohort has 152 MSS and 27 MSI-HIGH samples; samples flagged as low MSI were excluded from the study as they have not been shown to differ in their clinicopathologic features or in most molecular features from MSS tumors [149].

**Validation dataset.** For samples collected afterwards the consortium provides raw sequencing data, and CNAs (also in the file `crc_clinical_sheet.txt`). Thus, reads were processed to produce mutation profiles.

Bowtie 2.0 software was used to align sequences over the human reference genome HG19 [105]. To refine the data, reads unmapped, reads with unmapped mate, not primary alignments, and reads that were PCR or optical duplicates were discarded from the study (http://picard.sourceforge.net/). We also executed a local realignment around indels defined in SnpDB [169] and 1000G [29]. Variant calling was executed with GATK software and low quality variants (mapping quality below 30 or read depth below 10), were discarded [37]. Germline variants were also removed, i.e., variants that were present in non-tumoral samples, and variants reported in the 1000G project. Finally, variants were annotated using the SeattleSeq Variant Annotation web tool [140].

The input cohort of patients was partitioned according to the number of mutations per sample $r$, see Figure §E.2; a curated cutoff of 500 mutations per sample was used to determine 203 samples with MSS status ($r < 500$) and 36 MSI-HIGH samples ($r \geq 500$). The former group had mutations in 14580 genes, and the latter in 15071.

### E.1.1 Driver events selection

We have selected 33 genes annotated to 5 pathways as drivers of colorectal tumorigenesis [137]. These are well-known cancer genes, frequently reported as relevant to colorectal progression and to the major pathways involved in CRC. Driver events are alterations in:

- WNT genes (14): APC, DKK-4, TCF7L2, CTNNB1, LRP5, FBXW7, DKK-1, FZD10, ARID1A, DKK-2, FAM123B, SOX9, DKK-3 and AXIN2;

- RTK/RAS genes (5): ERBB2, ERBB3, NRAS, KRAS and BRAF;

- TGF-$\beta$ genes (5): TGFBR1, SMAD3, TGFBR2, SMAD4, ACVR1B, ACVR2A and SMAD2;

- IGF2/PI3K genes (5): IGF2, IRS2, PIK3CA, PIK3R1 and PTEN;

- P53 genes (2): TP53 and ATM.

In Chapter §6,RTK/RAS and IGF2/PI3K pathways are shortly denoted as RAS and PI3K.

## E.1.2 Mutual exclusivity groups of alterations

Groups of alterations showing a trend of mutual exclusivity were scanned with MUTEX and mutations and CNA hitting any of the 33 selected genes as input. MUTEX was run independently on MSS and MSI-HIGH groups (Supplementary Table §E.2, running times: approximately 6 and 3.5 hours, respectively, on a standard Desktop machine).

We selected only groups with score $< 0.2$, where the *score is derived from p-values corrected for false discovery rate.* 3 groups are found for MSI-HIGH tumors and 6 for MSS. For MSI-HIGH tumors, the three predicted groups consists of genes ACVR1B, ACVR2A, TP53 and ERBB2, of genes BRAF, NRAS and TGFBR2, and of genes KRAS and BRAF.

Further groups of exclusive alterations were considered consistent with results reported in [137]. These include groups derived by consolidated knowledge of colorectal progression: the well-known Wnt alterations in APC/CTNNB1 [64], as well as RAS alterations in KRAS, NRAS and BRAF genes [194]. Similarly, we used also a group collected by scanning non-hypermutated tumors with the MEMO tool in [137] - this group includes PIK3CA, PTEN, ERBB2 and IGF2 genes. These groups were restricted to account only for genes actually altered in a certain subtype, e.g., MSI-HIGH tumors lack CTNNB1 mutation, making the Wnt group irrelevant. Groups for MSS tumors are shown as Supplementary Figure §E.3.

## E.1.3 CAPRI's execution

CAPRI was run, on each group of tumors, by selecting alterations from the pool of 33 pathway genes; every alteration on a gene $x$ is included if *any* of these apply:

- the alteration frequency of $x$ - sum of mutation and CNA frequency - is greater than 5%;

- $x$ it is part of an exclusivity group.

The set of selected events for MSI-HIGH training tumors is shown in Chapter §6, the analogous for MSS tumors is shown in Figure §E.4.

CAPRI was executed in its supervised mode by writing formula over groups and genes with multiple alterations associated, as explained in Chapter §6. For instance, for MSI-HIGH tumors with alterations in RAS pathway we grouped hard exclusivity of NRAS mutations and deletions, with soft exclusivity of KRAS and BRAF mutations. Our aim was to account for a small subset of samples with concurrent KRAS and NRAS alterations (see Chapter §6). The list of all Boolean formulas written over groups is in Tables §E.3 and §E.4; this approach was adopted also when a gene harbors multiple alterations in a subtype, e.g., ERBB2 in MSS training samples which shows a trend of soft exclusivity between mutations and amplifications. We used both AIC and BIC scores to regularize inference after 100 non-parametric bootstrap iterations for estimation of the preliminary selective advantage relations. Statistical significance was determined in terms of p-values using the Mann-Whitney U test for CAPRI's inequalities. In most cases these are orders of magnitude below significance threshold - exact values reported as Additional File `CRC-pvalues.xslx`. CAPRI's models with such p-values and non-parametric bootstrap confidence are shown in Figures §E.5 and §E.6.

## E.2   Single-cell synthetic data

We sampled single-cell data from the clonal phylogeny tree (see Chapter §6). Such a tree reports the presence of the following clones in patient $RMH004$:

| clone | signature |
|:-----:|:---------:|
| $\mathbf{c}_1$ | VHL frame-shift |
| $\mathbf{c}_2$ | VHL frame-shift, SMARCA4 SNV |
| $\mathbf{c}_3$ | VHL frame-shift, ARID1A SNV |
| $\mathbf{c}_4$ | VHL, PTEN frame-shift |
| $\mathbf{c}_5$ | VHL, PTEN frame-shift, ATM SNV |
| $\mathbf{c}_6$ | VHL, PTEN frame-shift, ATM SNV, 6Q deletion |
| $\mathbf{c}_7$ | VHL, PTEN frame-shift, ATM SNV, MSH6 stop codon |
| $\mathbf{c}_8$ | VHL, PTEN frame-shift, ATM SNV, 2Q amplification |

where when there are multiple undistinguishable alterations we report the first appearing in the plot, from left to right. Single-cell sampling consists in sampling any of these clones, where a clone-sampling probability is a function of the probability to sample each of its constituting alterations. The probabilities of sampling any of the considered alterations is given by the shown in Chapter §6, where marginal and conditional probabilities are estimated by the observed frequencies in the input data.

So, for instance, the probability of sampling clone $\mathbf{c}_4$ is

$$
\begin{aligned}
p(\mathbf{c}_4) = & \, p(\text{VHL:fs}) & & \textit{sample the trunk event} \\
& \times p(\text{PTEN:fs}) & & \textit{and its downstream event,} \\
& \times \Big(1 - p(\text{ARID1A:SNV})\Big) & & \textit{do not sample private events of } \mathbf{c}_2, \\
& \times \Big(1 - p(\text{SMARCA4:SNV})\Big) & & \textit{do not sample private events of } \mathbf{c}_3, \\
& \times \Big(1 - p(\text{ATM:SNV})\Big) & & \textit{do not sample private events of } \mathbf{c}_5. \quad \text{(E.1)}
\end{aligned}
$$

where 'fs' is a short-hand for frame-shift. In this way, we can easily transform a clone sample in a binary signature which can be inputed to CAPRESE's reconstruction, being a 0/1 Bernoulli model. So, for instance a sample of clone $\mathbf{c}_4$ will be the binary vector:

| VHL | SMARCA4 | ARID1A | PTEN | ATM | 6Q | MSH6 | 2Q |
|-----|---------|--------|------|-----|----|------|----|
| 1   | 0       | 0      | 1    | 0   | 0  | 0    | 0  |

To make the problem more realistic, we included a noise parameter which is the probability to assign a random value to any entry of such vector, regardless its actual value. This aims at mimicking the problem of missing coverage for variant calling or other artefacts such as error in measurements. In Chapter §6, we sampled datasets with number of cells $n$ spanning from 5 to 200 (with discretization at different densities), with noise ranging from 0 to 20% (i.e., probability .2, discretization .05). For each of these settings we generate 100 independent datasets, and average the performance. In Figure §E.7, we show further results with $n \leq 1000$ - to report method's convergence for increased sample size. The performance of CAPRESE was measure as precision and recall, computed as standard, and Hamming distance, which provides a standard approach to measure similarity between the phylogeny tree reconstructed by running the algorithm on a single dataset, and the tree shown in Chapter §6.

## E.3 Supplementary tables and figures

**Training dataset**

| cancer[†] | statistics | | | alteration type | | |
|-----------|-----|-------|-------|-----------|----------------|-----------|
|           | $n$ | $m$   | $|G|$ | *mutations* | *amplifications* | *deletions* |
| MSI-HIGH  | 27  | 16100 | 13798 | 11556     | 2888           | 1656      |
| MSS       | 152 | 21317 | 16371 | 12417     | 6925           | 1975      |

[†] Samples were classified as MSI-HIGH/LOW and MSS by TCGA; see flag `MSI_status` in clinical data available for the COADREAD project.

**Validation dataset**

| cancer[†] | statistics | | | alteration type | | |
|-----------|-----|-------|-------|-----------|----------------|-----------|
|           | $n$ | $m$   | $|G|$ | *mutations* | *amplifications* | *deletions* |
| MSI-HIGH  | 36  | 16891 | 15779 | 15071     | 199            | 1621      |
| MSS       | 202 | 24957 | 18158 | 14567     | 5846           | 4544      |

[†] Samples were classified as MSI-HIGH if they had more than 500 mutations, as MSS otherwise; see Figure §E.2.

Table E.1: **COADREAD Datasets.** Data used in this study, derived from the TCGA COADREAD project [137].

**MUTEX parameters**

| Parameter | Value | Description |
|---|---|---|
| `signalling-network` | - | *MUTEX network*[†] |
| `max-group-size` | 5 | *maximum size of a result group* |
| `first-level-random-iteration` | 10000 | *number of randomisation to estimate null distribution of member p-values in groups* |
| `second-level-random-iteration` | 100 | *number of runs to estimate the null distribution of final scores* |
| `fdr-cutoff` | - | *false-discovery-rate cutoff maximising the expected value of true positives - false positives is estimated from data* |
| `search-on-signaling-network` | TRUE | *reduce the search space using the signalling network* |

[†] Manually curated from Pathway Commons, SPIKE and SignaLink databases. Provided with the tool; available for download at `https://code.google.com/p/mutex/`.

**MUTEX groups with score $< .2$**

| | **MSI-HIGH Groups** | *score* | *q-value* |
|---|---|---|---|
| **1** | KRAS, BRAF, | 0.095 | 0.48 |
| **2** | NRAS, BRAF, TGFBR1 | 0.1677 | 0.45 |
| **3** | ERBB2, TP53, ACVR1B, ACVR2A | 0.1703 | 0.355 |
| | | | |
| | **MSS Groups** | *score* | *q-value* |
| **1** | TP53, ATM, | 0.051 | 0.34 |
| **2** | ARID1A, TP53 | 0.075 | 0.193 |
| **3** | KRAS, NRAS, BRAF, | 0.0864 | 0.1975 |
| **4** | CTNNB1, APC, DKK2, | 0.098 | 0.144 |
| **5** | DKK1, TP53, ATM, DKK2 | 0.1387 | 0.176 |
| **6** | PIK3CA, TP53, ATM | 0.164 | 0.207 |

Table E.2: **MUTEX: parameters and results.** Top: Parameters used to run MUTEX on the training MSS/MSI-HIGH datasets with input CNA and somatic mutations in the pathway genes described in Text. Bottom: MUTEX identified 3 and 6 groups of alterations showing a trend of mutual exclusivity in these groups with score below the suggested cutoff of 0.2.

## CAPRI formulas inputed for testing [†]

| | MSI-HIGH tumors | description |
|---|---|---|
| 1 | `(NRAS:m ⊕ NRAS:d) ∨ KRAS:m ∨ BRAF:m` | `RAF` exclusivity |
| 2 | `PIK3CA:m ∨ ERBB2:m ∨ PTEN:m ∨ IGF2:d` | MEMO group |
| 3 | `(ACVR1B:m ⊕ ACVR1B:d) ∨ ACVR2A:m ∨ TP53:m ∨ ERBB2:m` | MUTEX group |
| 4 | `(NRAS:m ⊕ NRAS:d) ∨ TGFBR1:m ∨ BRAF:m` | MUTEX group |
| 5 | `KRAS:m ∨ BRAF:m` | MUTEX group |
| 6 | `ACVR1B:m ⊕ ACVR1B:a` | multiple alterations |
| 7 | `NRAS:m ⊕ NRAS:a` | multiple alterations |
| 8 | `FBXW7:m ∨ FBXW7:a` | multiple alterations [‡] |

[†] Events type: mutation (m), deletion (d), amplification (a). Hard (⊕) and soft (∨) exclusivity.

[‡] Formula not included as it creates a duplicated signature in the dataset.

Table E.3: **CAPRI formulas for MSI-HIGH.** Formulas created for the groups, and inputed to CAPRI for testing. These are either derived from exclusivity groups or from genes involved in different types of alterations.

**CAPRI formulas inputed for testing** †

| | MSS tumors | description |
|---|---|---|
| 1 | (APC:m ⊕ APC:d) ∨ CTNNB1:m | WNT exclusivity |
| 2 | (KRAS:m ∨ KRAS:a) ∨ (NRAS:m ⊕ NRAS:a) ∨ (BRAF:m ⊕ BRAF:a) | RAF excl and MEMO group |
| 3 | PIK3CA:m ∨ (ERBB2:m ∨ ERBB2:a) ∨ (PTEN:m ⊕ PTEN:d) ∨ IGF2:a | MEMO group |
| 4 | (TP53:m ⊕ TP53:d) ∨ (ATM:m ⊕ ATM:d) | MUTEX group |
| 5 | (TP53:m ⊕ TP53:d) ∨ ARID1A:m | MUTEX group |
| 6 | (TP53:m ⊕ TP53:d) ∨ ARID1A:m | MUTEX group |
| 7 | (APC:m ⊕ APC:d) ∨ CTNNB1:m ∨ DKK2:m | MUTEX group |
| 8 | (TP53:m ⊕ TP53:d) ∨ (ATM:m ⊕ ATM:d) ∨ DKK2:m ∨ DKK1:m | MUTEX group |
| 9 | (TP53:m ⊕ TP53:d) ∨ (ATM:m ⊕ ATM:d) ∨ PIK3CA:m | MUTEX group |
| 10 | (APC:m ⊕ APC:d) | multiple alterations |
| 11 | (TP53:m ⊕ TP53:d) | multiple alterations |
| 12 | (SMAD4:m ⊕ SMAD4:d) | multiple alterations |
| 13 | (TCF7L2:m ⊕ TCF7L2:d) | multiple alterations |
| 14 | (ATM:m ⊕ ATM:d) | multiple alterations |
| 15 | (NRAS:m ⊕ NRAS:d) | multiple alterations |
| 16 | (ERBB2:m ∨ ERBB2:a) | multiple alterations |
| 17 | (PTEN:m ⊕ PTEN:d) | multiple alterations |
| 18 | (SMAD2:m ⊕ SMAD2:a) | multiple alterations |
| 19 | (DKK4:m ⊕ DKK4:a) | multiple alterations |
| 20 | (SOX9:m ⊕ SOX9:d) | multiple alterations |
| 21 | (BRAF:m ⊕ BRAF:a) | multiple alterations |

† Events type: mutation (m), deletion (d), amplification (a). Hard (⊕) and soft (∨) exclusivity.

‡ Formula not included as it creates a duplicated signature in the dataset.

Table E.4: **CAPRI formulas for MSS.** Formulas created for the groups, and inputed to CAPRI for testing. These are either derived from exclusivity groups or from genes involved in different types of alterations.

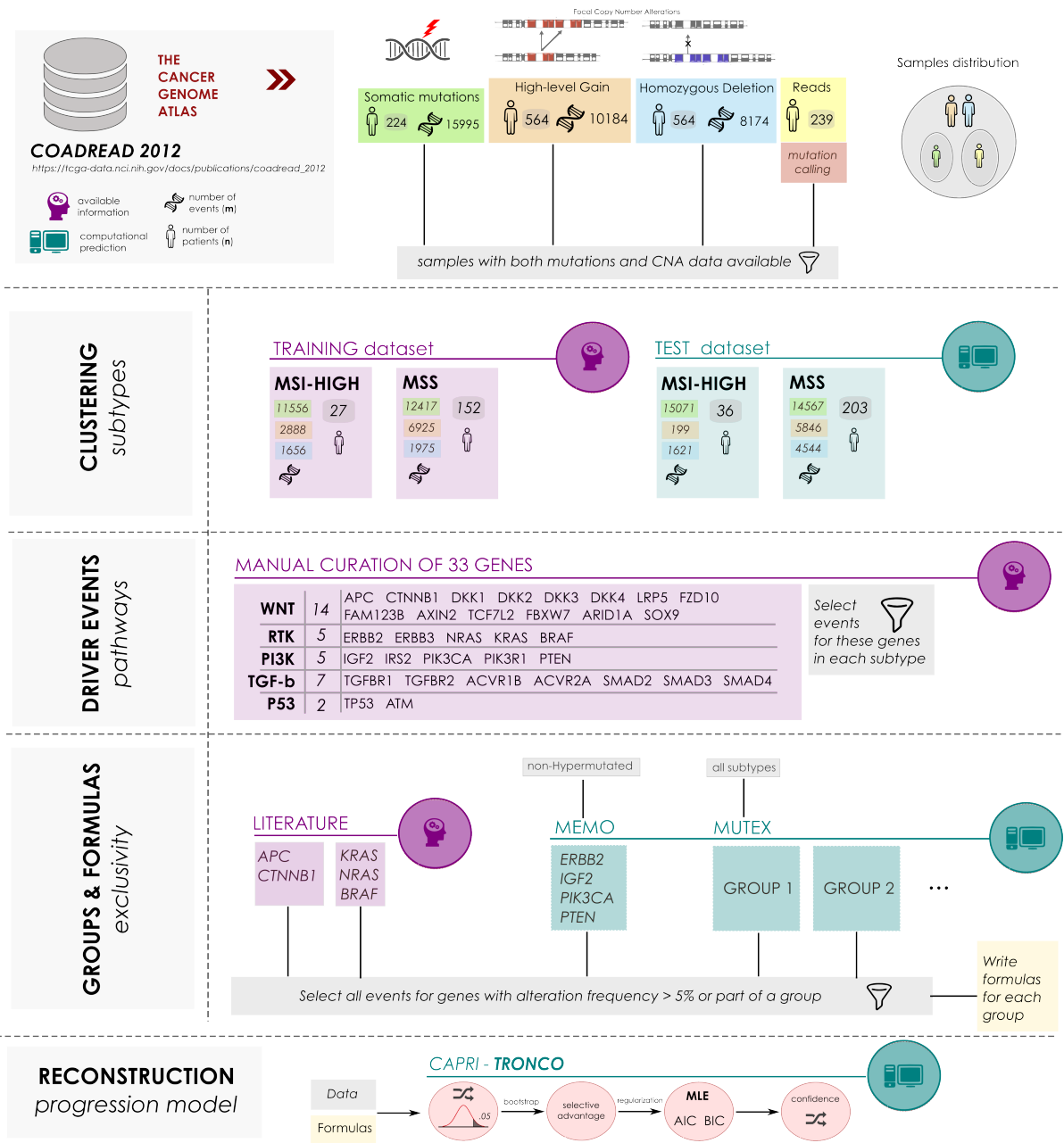

Figure E.1: **CRC pipeline processing MSI/MSS tumors.** we process Microsatellite Stable and highly Instable tumors collected from the The Cancer Genome Atlas project "Human Colon and Rectal Cancer". We implement a test/training study on selected somatic mutations and focal CNAs in 33 driver genes manually annotated to 5 pathways in the project. We scan groups of exclusive alterations with computational tools and from the project results, and we select which alterations we input to CAPRI. Then, inference is performed with various settings of regularization and confidence.

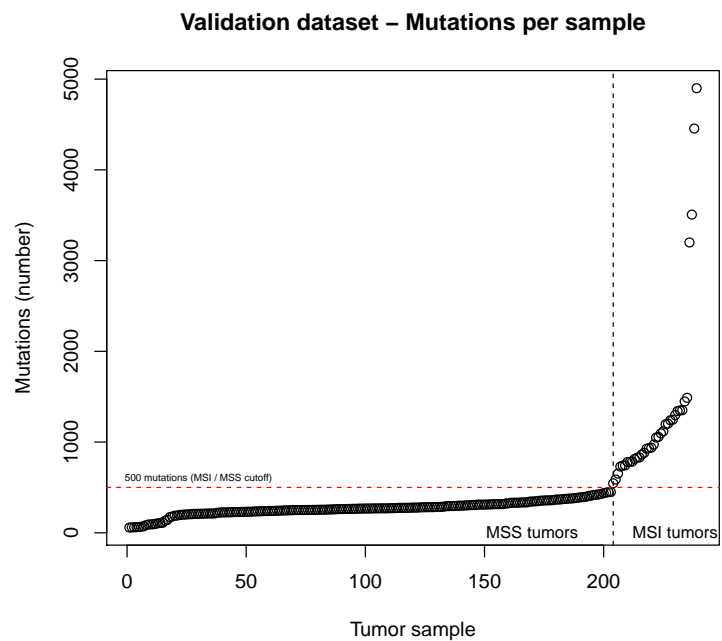

Figure E.2: **MSI/MSS samples in the test dataset.** Mutation rate per sample, $r$, in data collected for testing (which do not have clinical annotation from TCGA COAD-READ project). To distinguish between tumors with MSS/MSI-HIGH status we set an empirical cutoff of 500 mutations to define two groups of 203 ($r \leq 500$) and 36 ($r > 500$) samples respectively.

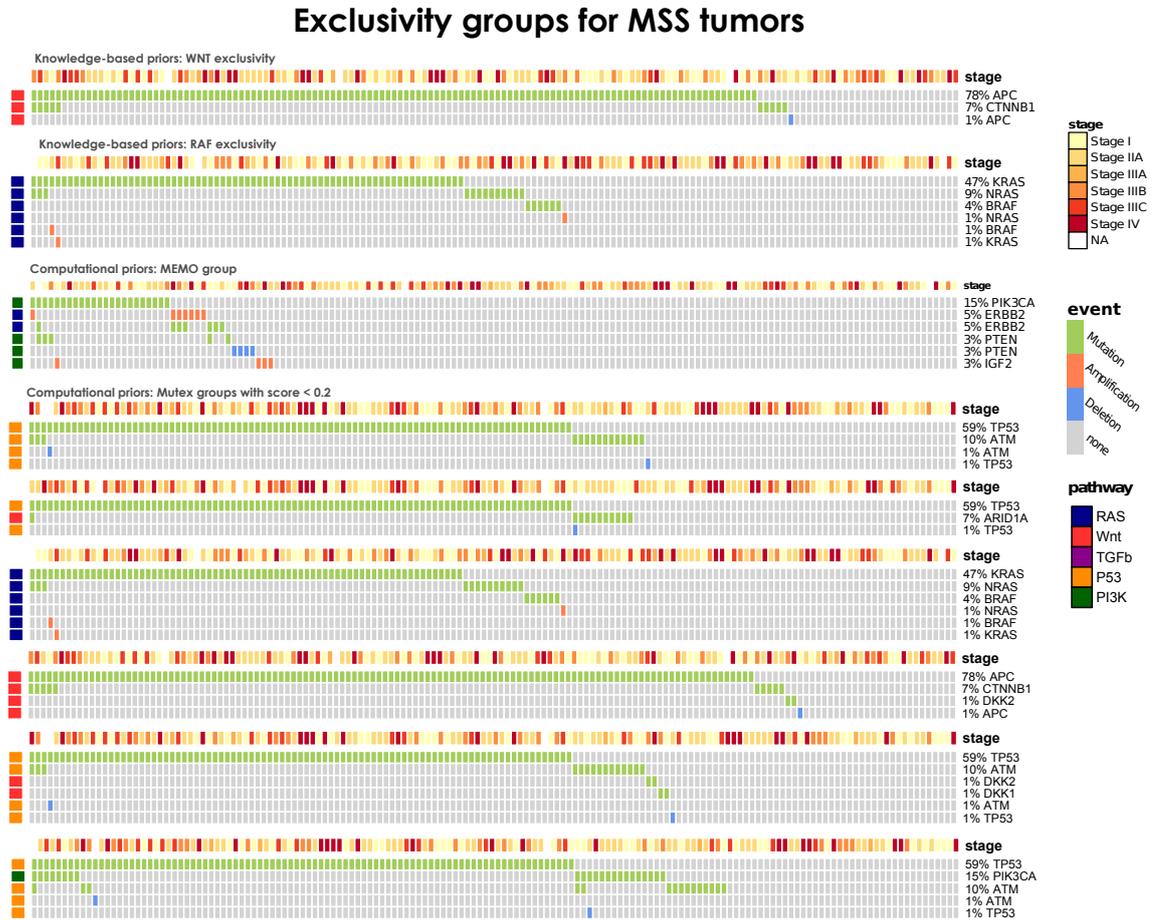

Figure E.3: **Groups of exclusive alterations for MSS tumors.** Knowledge-based groups of exclusive alterations consist of: KRAS, NRAS and BRAF genes (RAF pathway) and APC and CTNNB1 genes (WNT pathway). The MEMO[28] group identified in [137] in this cohort consists of genes PIK3CA, ERBB2, IGF2 and PTEN. Finally, 6 groups are predicted by MUTEX [8] with score below .2, one of these is equivalent to the known exclusive alterations in RAF pathway.

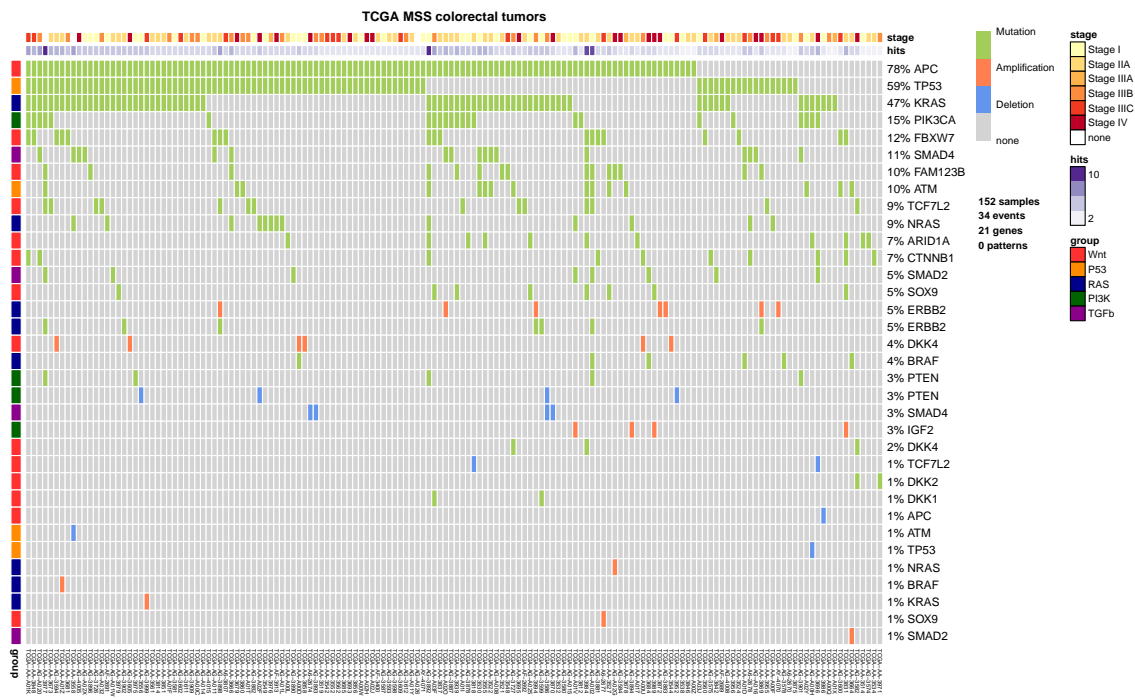

Figure E.4: **Selected data for MSS tumors.** Colorectal tumors with Microsatellite Stable clinical status in the TCGA COADREAD project, restricted to 152 samples with both somatic mutations and CNA data available. 33 driver genes annotated to 5 pathways are selected as of the list published in [137] to automatically detect groups of mutually exclusive alterations. Events selected for reconstruction are those involving genes altered in at least 5% of the cases, or part of group of alterations showing an exclusivity trend (see Figure §E.3). This dataset is used to infer the set of selective advantage relations which constitute the MSS progression model presented in Chapter §6.

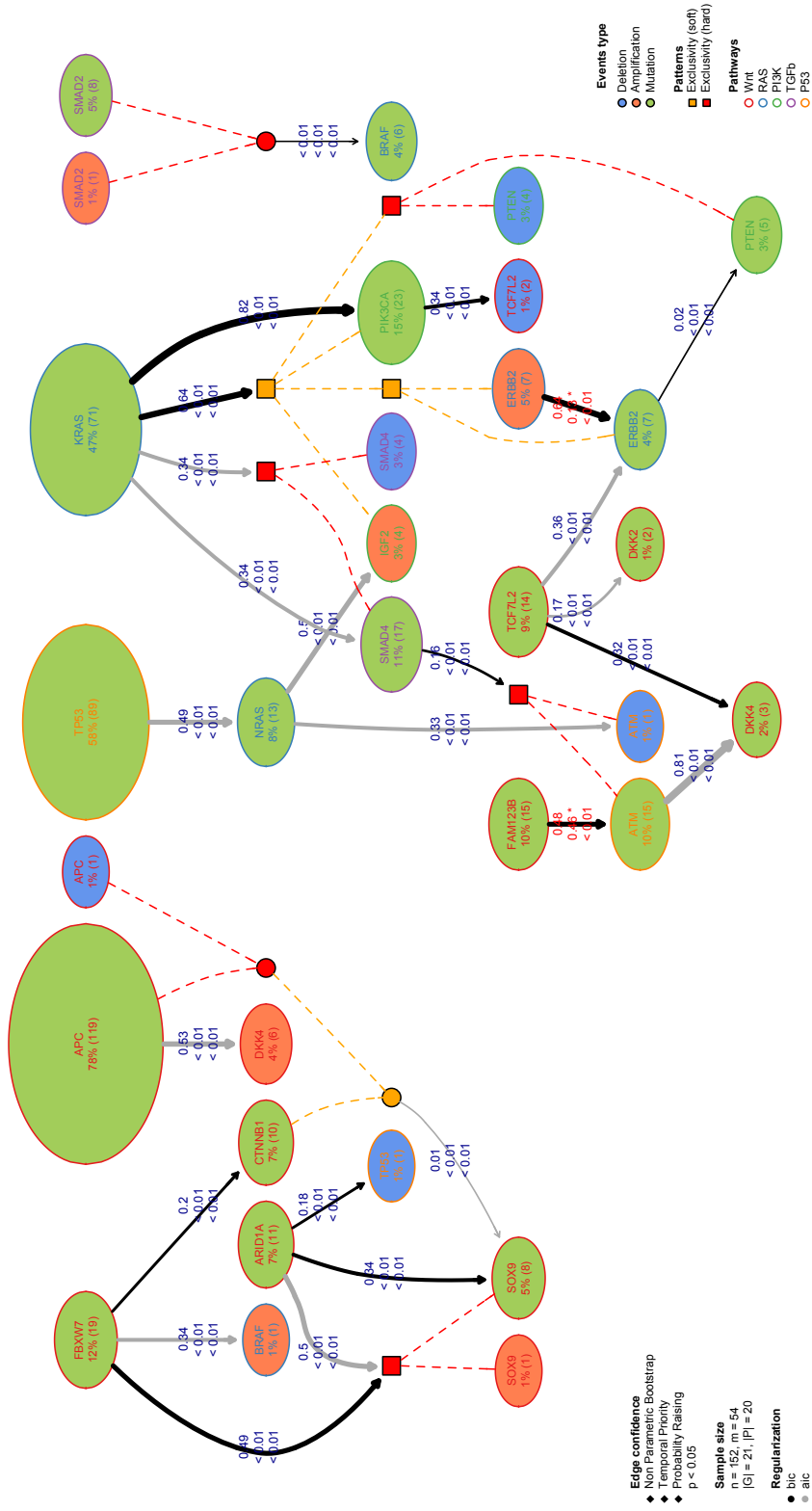

Figure E.5: **MSS training tumors: bootstrap confidence.** Progression model for MSS tumors with confidence shown as edge labels. The first label represents the relation confidence estimated with 100 non-parametric bootstrap iterations, the second and third are p-values for temporal priority and probability raising. Red p-values are above the minimum significance threshold of .005.

Figure E.6: **MSI training tumors: bootstrap confidence.** Progression model for MSS tumors with confidence shown as edge labels. The first label represents the relation confidence estimated with 100 non-parametric bootstrap iterations, the second and third are p-values for temporal priority and probability raising. Red p-values are above the minimum significance threshold of .005.

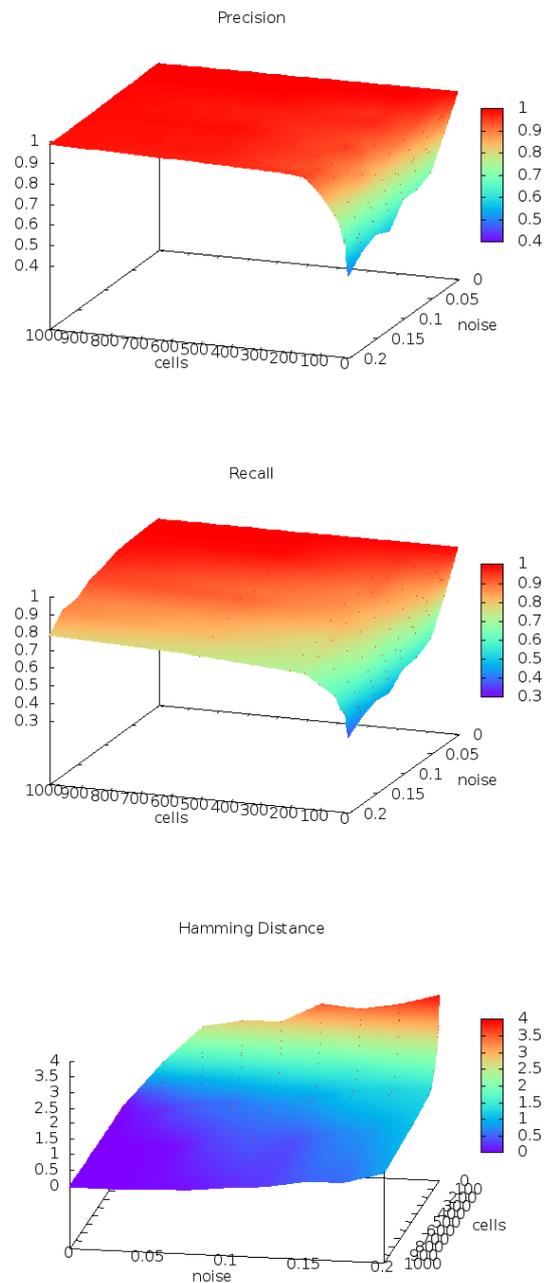

Figure E.7: **Single-cell synthetic data: performance.** Reconstruction performance with CAPRESE and $n \leq 1000$ as sample size. Precision and recall are reported as well as Hamming Distance between the phylogeny tree which generated the data (see Chapter §6) and the inferred by the algorithm.

# THE CAUSAL STRUCTURE OF DISCRIMINATION

*Discrimination discovery* from data is an important task aiming at identifying patterns of illegal and unethical discriminatory activities against protected-by-law groups, e.g., ethnic minorities. While any legally-valid proof of discrimination requires evidence of causality, the state-of-the-art methods are essentially correlation-based, albeit, as it is well known, correlation does not imply causation.

In this Chapter we present how the framework proposed in this thesis can be also adopted to tackle the data mining problem of discrimination detection in databases. Following Suppes' *probabilistic causation theory*, we define a method to extract, from a dataset of historical decision records, the causal structures existing among the attributes in the data. The result is a type of constrained Bayesian network, which we dub *Suppes-Bayes Causal Network* (SBCN). Next, we develop a toolkit of methods based on random walks on top of the SBCN, addressing different anti-discrimination legal concepts, such as direct and indirect discrimination, group and individual discrimination, genuine requirement, and favoritism. Finally experiments on real-world datasets confirm the inferential power of this approach in all these different tasks. As a reference for this work, see [17].

## F.1   Introduction

**The importance of discrimination discovery.** At the beginning of 2014, as an answer to the growing concerns about the role played by data mining algorithms in decision-making, USA President Obama called for a 90-day review of big data collecting and analysing practices. The resulting report[1] concluded that *"big data technologies can cause societal harms beyond damages to privacy"*. In particular, it expressed concerns about the possibility that decisions informed by big data could have discriminatory effects, even in the absence of discriminatory intent, further imposing less favorable treatment to already disadvantaged groups.

---

[1] http://www.whitehouse.gov/sites/default/files/docs/big_data_privacy_report_may_1_2014.pdf

Discrimination refers to an unjustified distinction of individuals based on their membership, or perceived membership, in a certain group or category. Human rights laws prohibit discrimination on several grounds, such as gender, age, marital status, sexual orientation, race, religion or belief, membership in a national minority, disability or illness. Anti-discrimination authorities (such as equality enforcement bodies, regulation boards, consumer advisory councils) monitor, provide advice, and report on discrimination compliances based on investigations and inquiries. A fundamental role in this context is played by *discrimination discovery in databases*, i.e., the data mining problem of unveiling discriminatory practices by analyzing a dataset of historical decision records.

**Discrimination is causal.** According to current legislation, discrimination occurs when a group is treated "less favorably" [107] than others, or when "a higher proportion of people not in the group is able to comply" with a qualifying criterion [109]. Although these definitions do not directly imply causation, as stated in [56] all discrimination claims require plaintiffs to demonstrate a causal connection between the challenged outcome and a protected status characteristic. In other words, in order to prove discrimination, authorities must answer the counterfactual question: what would have happened to a member of a specific group (e.g., nonwhite), if he or she had been part of another group (e.g., white)?

"The Sneetches", the popular satiric tale[2] against discrimination published in 1961 by Dr. Seuss, describes a society of yellow creatures divided in two races: the ones with a green star on their bellies, and the ones without. The Star-Belly Sneetches have some privileges that are instead denied to Plain-Belly Sneetches. There are, however, Star-On and Star-Off machines that can make a Plain-Belly into a Star-Belly, and viceversa. Thanks to these machines, the causal relationship between race and privileges can be clearly measured, because stars can be placed on or removed from any belly, and multiple outcomes can be observed for an individual. Therefore, one could readily answer the counterfactual question, saying with certainty what would have happened to a Plain-Belly Sneetch had he or she been a Star-Belly Sneetch.

In the real world however, proving discrimination episodes is much harder, as one cannot manipulate race, gender, or sexual orientation of an individual. This highlights the need to assess discrimination as a causal inference problem [32] from a database of past decisions, where causality can be inferred probabilistically. Unfortunately, *the state of the art of data mining methods for discrimination discovery in databases does not properly address the causal question*, as it is mainly based on correlation-based methods (surveyed in Section §F.2).

**Correlation is not causation.** It is well known that correlation between two variables does not necessarily imply that one causes the other. Consider a unique cause $X$ of two effects, $Y$ and $Z$: if we do not take in account $X$, we might derive wrong conclusions because of the observable correlation between $Y$ and $Z$. In this situation, $X$ is said to act as a *confounding factor* for the relationship between $Y$ and $Z$.

---

[2] http://en.wikipedia.org/wiki/The_Sneetches_and_Other_Stories

Variants of the complex relation just discussed can arise even if, in the example, $X$ is not the actual cause of either $Y$ or $Z$, but it is only correlated to them, for instance, because of how the data were collected. Consider for instance a credit dataset where there exists high correlation between a variable representing low income and an other variable representing loan denial and let us assume that this is due to an actual legitimate causal relationship in the sense that, legitimately, a loan is denied if the applicant has low income. Let us now assume that high correlation between low income and being female is also observed, which, for instance, can be due to the fact that the women represented in the specific dataset in analysis, tend to be underpaid. Given these settings, in the data we would also observe high correlation between the variable gender being female and the variable representing loan denial, due to the fact that we do not account for the presence of the variable low income. Following common terminologies, we will say that such situations are due to *spurious correlations*.

However, the picture is even more complicated: it could be the case, in fact, that being female is the actual cause of the low income and, hence, be the *indirect cause* of loan denial *through* low income. This would represent a causal relationship between the gender and the loan denial, that we would like to detect as discrimination.

Disentangling these two different cases, i.e., female is only correlated to low income in a spurious way, or being female is the actual cause of low income, is at the same time important and challenging. This highlights the need for a principled causal approach to discrimination detection.

Another typical pitfall of correlation-based reasoning is expressed by what is known as Simpson's paradox[3] according to which, correlations observed in different groups might disappear when these heterogeneous groups are aggregated, leading to *false positives* cases of discrimination discovery. One of the most famous false-positive examples due to Simpson's paradox occurred when in 1973 the University of California, Berkeley was sued for discrimination against women who had applied for admission to graduate schools. In fact, by looking at the admissions of 1973, it first appeared that men applying were significantly more likely to be admitted than women. But later, by examining the individual departments carefully, it was discovered that none of them was significantly discriminating against women. On the contrary, most departments had exercised a small bias in favor of women. The apparent discrimination was due to the fact that women tended to apply to departments with lower rates of admission, while men tended to apply to departments with higher rates [16]. Later in Section §F.5.5 we will use the dataset from this episode to highlight the differences between correlation-based and causation-based methods.

Spurious correlations can also lead to *false negatives* (i.e., discrimination existing but not being detected) as is commonly seen in *"reverse-discrimination"*. The typical case is when authorities take affirmative actions, e.g., with compensatory quota systems, in order to protect a minority group from a potential discrimination. Such actions, while trying to erase the supposed discrimination (i.e., the spurious correlation), fail to address the real underlying causes for discrimination, potentially ending up denying individual members of a privileged group from access to their rightful shares of social goods. In the early 70's, a case involving the University of California at Davis Medical School highlighted one such incident as the school's admissions program reserved 16 of the 100 places in its entering class for "disadvantaged" applicants, thus unintentionally reducing the chances of admission for a qualified applicant.[4]

---

[3]http://en.wikipedia.org/wiki/Simpson's_paradox

These are just few typical examples of the pitfalls of correlation-based reasoning in the discovery of discrimination. Later in Section §F.5.5 we show concrete examples from real-world datasets where correlation-based methods to discrimination discovery are not satisfactory.

**The proposal and contributions.** In this Chapter we take a principled causal approach to the data mining problem of discrimination detection in databases. Following Suppes' *probabilistic causation theory* [80, 172] we define a method to extract, from a dataset of historical decision records, the causal structures existing among the attributes in the data.

In particular, we define the *Suppes-Bayes Causal Network* (SBCN), i.e., a directed acyclic graph (DAG) where we have a node representing a Bernulli variable of the type $\langle attribute = value \rangle$ for each pair attribute-value present in the database. In this DAG an arc $(A, B)$ represents the existence of a causal relation between $A$ and $B$ (i.e., $A$ causes $B$). Moreover, each arc is labeled with a score, representing the strength of the causal relation.

The SBCN is a constrained Bayesian network reconstructed by means of maximum likelihood estimation (MLE) from the given database, where we force the conditional probability distributions induced by the reconstructed graph to obey Suppes' constraints: i.e., *temporal priority* and *probability rising*. Imposing Suppes' temporal priority and probability raising we obtain what we call the *prima facie causes* graph [172], which might still contain spurious causes (false positives). In order to remove these spurious case we add a bias term to the likelihood score, favoring sparser causal networks: in practice we sparsify the *prima facie causes* graph by extracting a minimal set of edges which best explain the data. This regularization is done by means of the Bayesian Information Criterion (BIC) [167].

The obtained SBCN provides a clear summary, amenable to visualization, of the probabilistic causal structures found in the data. Such structures can be used to reason about different types of discrimination. In particular, we show how using several random-walk-based methods, where the next step in the walk is chosen proportionally to the edge weights, we can address different anti-discrimination legal concepts. The experiments show that the measures of discrimination produced by the methods are very strong, almost binary, signals: the measures are very clearly separating the discrimination and the non-discrimination cases.

*To the best of our knowledge this is the first proposal of discrimination detection in databases grounded in probabilistic causal theory.*

---

[4] http://en.wikipedia.org/wiki/Regents_of_the_University_of_California_v._Bakke

**Roadmap.** The rest of the Chapter is organized as follows. Section §F.2 discusses the state of the art in discrimination detection in databases. In Section §F.3 we formally introduce the SBCN and we present the method for extracting such causal network from the input dataset. Once extracted the SBCN, in Section §F.4 we show how to exploit it for different concepts of discrimination detection, by means of random-walk methods. Finally Section §F.5 presents the experimental assessment and comparison with correlation-based methods on two real-world datasets.

## F.2 Related work

Discrimination analysis is a multi-disciplinary problem, involving sociological causes, legal reasoning, economic models, statistical techniques [31, 162]. Some authors [73, 94] study how to prevent data mining from becoming itself a source of discrimination. In this Chapter instead we focus on the data mining problem of detecting discrimination in a dataset of historical decision records, and in the rest of this section we present the most related literature.

Pedreschi et al. [153, 180, 165] propose a technique based on extracting classification rules (inductive part) and ranking the rules according to some legally grounded measures of discrimination (deductive part). The result is a (possibly large) set of classification rules, providing local and overlapping niches of possible discrimination. This model only deals with group discrimination.

Luong et al. [120] exploit the idea of *situation-testing* [163] to detect individual discrimination. For each member of the protected group with a negative decision outcome, testers with similar characteristics ($k$-nearest neighbors) are considered. If there are significantly different decision outcomes between the testers of the protected group and the testers of the unprotected group, the negative decision can be ascribed to discrimination.

Zliobaite et al. [200] focus on the concept of *genuine requirement* to detect that part of discrimination which may be explained by other, legally grounded, attributes. In [41] Dwork et al. address the problem of fair classification that achieves both group fairness, i.e., the proportion of members in a protected group receiving positive classification is identical to the proportion in the population as a whole, and individual fairness, i.e., similar individuals should be treated similarly.

The above approaches assume that the dataset under analysis contains attributes that denote protected groups (i.e., direct discrimination). This may not be the case when such attributes are not available, or not even collectable at a micro-data level as in the case of the loan applicant's race. In these cases we talk about indirect discrimination discovery. Ruggieri et al. [152] adopt a form of rule inference to cope with the indirect discovery of discrimination. The correlation information is called background knowledge, and is itself coded as an association rule.

Mancuhan and Clifton [123] propose Bayesian networks as a tool for discrimination discovery. Bayesian networks consider the dependence between all the attributes and use these dependencies in estimating the joint probability distribution without any strong assumption, since a Bayesian network graphically represents a factorization of the joint distribution in terms of conditional probabilities encoded in the edges. Although Bayesian networks are often used to represent causal relationships, this needs not be the case, in fact a directed edge from two nodes of the network does not imply any causal relation between them. As an example, let us observe that the two graphs $A \rightarrow B \rightarrow C$ and $C \rightarrow B \rightarrow A$ impose exactly the same conditional independence requirements and, hence, any Bayesian network would not be able to disentangle the direction of any causal relationship among these events.

The work departs from this literature as ($i$) it is grounded in probabilistic causal theory instead of being based on correlation, ($ii$) it proposes a holistic approach able to deal with different types of discrimination in a single unifying framework, while the methods in the state of the art usually deal with one and only one specific type of discrimination. This is also the first work to adopt graph theory and social network analysis concepts, such as random-walk-based centrality measures and community detection, for discrimination detection. The proposed methods also have low computational cost compared to the existing approaches in the literature.

## F.3   Suppes-Bayes Causal Network

In order to study discrimination as a causal inference problem, we exploit the criteria defined in the theories of *probabilistic causation* [80]. In particular, we follow [172], where Suppes proposed the notion of *prima facie causation* that is at the core of probabilistic causation. Suppes' definition is based on two pillars: ($i$) any cause must happen before its effect (*temporal priority*) and ($ii$) it must raise the probability of observing the effect (*probability raising*).

In the rest of this Section we introduce the method to construct, from a given relational table $\mathcal{D}$, a type of causal Bayesian network constrained to satisfy the conditions dictated by Suppes' theory, which we dub *Suppes-Bayes Causal Network* (SBCN).

In the literature many algorithms exist to carry out structural learning of general Bayesian networks and they usually fall into two families [101]. The first family, *constraint based learning*, explicitly tests for pairwise independence of variables conditioned on the power set of the rest of the variables in the network. These algorithms exploit structural conditions defined in various approaches to causality [80, 127, 192]. The second family, *score based learning*, constructs a network which maximizes the likelihood of the observed data with some regularization constraints to avoid overfitting. Several hybrid approaches have also been recently proposed [19].

The framework can be considered a hybrid approach exploiting *constrained maximum likelihood estimation* (MLE) as follows: ($i$) we first define all the possible causal relationship among the variables in $\mathcal{D}$ by considering only the oriented edges between events that are consistent with Suppes' notion of probabilistic causation and, subsequently, ($ii$) we perform the reconstruction of the SBCN by a score-based approach (using BIC), which considers only the valid edges.

The idea of adopting Suppes' theory to reconstruct the causal structure subsumed by a progression model is the main contribution of this thesis and, as already deeply discussed, it was introduced for the first time in [117, 158], albeit in a completely different context, i.e., modelling somatic evolution in cancer.

We next present in details the whole learning process.

### F.3.1 Suppes' constraints

We start with an input relational table $\mathcal{D}$ defined over a set $A$ of $h$ categorical attributes and $s$ samples. In case continuous numerical attributes exists in $\mathcal{D}$, we assume they have been discretized to become categorical. From $\mathcal{D}$, we derive $\mathcal{D}'$, an $m \times s$ binary matrix representing $m$ Bernoulli variables of the type $\langle attribute = value \rangle$, where an entry is 1 if we have an observation for the specific variable and 0 otherwise.

**Temporal priority.** The first constraint, temporal priority, cannot be simply checked in the data as we have no timing information for the events. In particular, in this context the events for which we want to reason about temporal priority are the Bernoulli variables $\langle attribute = value \rangle$.

The idea here is that, e.g., $income = low$ cannot be a cause of $gender = female$, because the time when the gender of an individual is determined is antecedent to that of when the income is determined. This intuition is implemented by simply letting the data analyst provide as input to the framework a partial temporal order $r : A \to \mathbb{N}$ for the $h$ attributes, which is then inherited from the $m$ Bernoulli variables. Note that the learning technique requires the input order $r$ to be correct and complete in order to guarantee its convergence. Nevertheless, if this is not the case, it is still capable of providing valuable insights about the underlying causal model, although with the possibility of false positive or false negative causal claims.

Based on the input dataset $\mathcal{D}$ and the partial order $r$ we produce the first graph $G = (V, E)$ where we have a node for each of the Bernoulli variables, so $|V| = m$, and we have an arc $(u, v) \in E$ whenever $r(u) \leq r(v)$. This way we will immediately rule out causal relations that do not satisfy the temporal priority constraint.

**Probability raising.** Given the graph $G = (V, E)$ built as described above the next step requires to prune the arcs which do not satisfy the second constraint, probability raising, thus building $G' = (V, E')$, where $E' \subseteq E$. In particular we remove from $E$ each arc $(u, v)$ such that $\mathcal{P}(v \mid u) \leq \mathcal{P}(v \mid \neg u)$. The graph $G'$ so obtained is called *prima facie* graph.

As previously proved in [117], we recall that the probability raising condition is equivalent to constraining for positive statistical dependence: in the prima facie graph we model *all and only* the positive correlated relations among the nodes already partially ordered by temporal priority, consistently with Suppes' characterization of causality in terms of relevance.

### F.3.2 Network simplification

As proved in [158], Suppes' conditions are necessary but not sufficient to evaluate causation: especially when the sample size is small, the model may have false positives (spurious causes), even after constraining for Suppes' temporal priority and probability raising criteria (which aim at removing false negatives). Consequently, although we expect all the statistically relevant causal relations to be modelled in $G'$, we also expect some spurious ones in it.

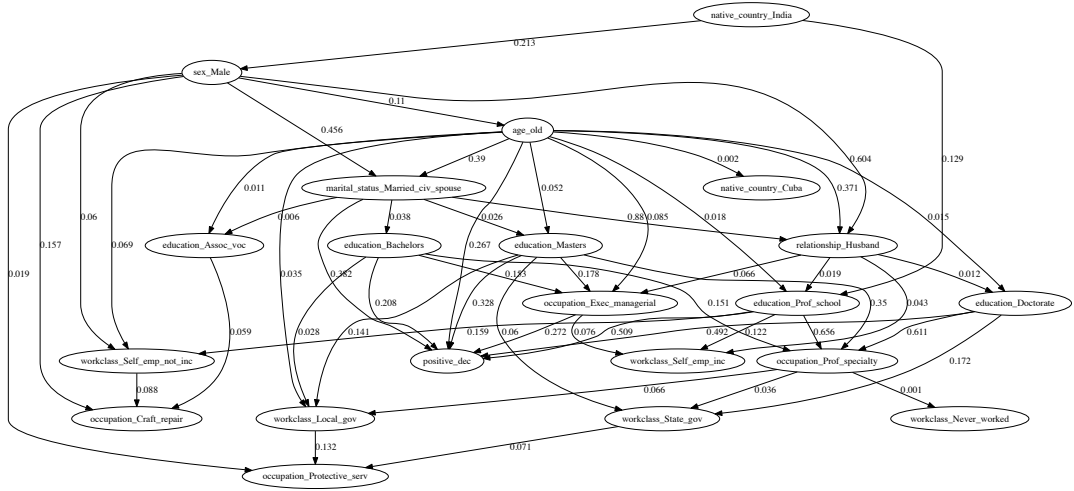

Figure F.1: One portion of the SBCN extracted from the Adult dataset. This subgraph corresponds to the $C_2$ community reported later in Table §F.3 (Section §F.5) extracted by a community detection algorithm.

In this proposal, in place of other structural conditions used in various approaches to causality, (see e.g., [80, 127, 192]), we perform a network simplification (i.e., we sparsify the network by removing arcs) with a score based approach, specifically by relying on the Bayesian Information Criterion (BIC) as the regularized likelihood score [167].

We consider as inputs for this score the graph $G'$ and the dataset $\mathcal{D}'$. Given these, we select the set of arcs $E^* \subseteq E'$ that maximizes the score:

$$score_{\mathrm{BIC}}(\mathcal{D}', G') = LL(\mathcal{D}'|G') - \frac{\log s}{2}\dim(G').$$

In the equation, $G'$ denotes the graph, $\mathcal{D}'$ denotes the data, $s$ denotes the number of samples, and $\dim(G')$ denotes the number of parameters in $G'$. Thus, the regularization term $-\dim(G')$ favors graphs with fewer arcs. The coefficient $\log s/2$ weighs the regularization term, such that the higher the weight, the more sparsity will be favored over "explaining" the data through maximum likelihood. Note that the likelihood is implicitly weighted by the number of data points, since each point contributes to the score.

Assume that there is one *true* (but unknown) probability distribution that generates the observed data, which is, eventually, uniformly randomly corrupted by false positives and negatives rates (in $[0, 1)$). Let us call *correct model*, the statistical model which best approximate this distribution. The use of BIC on $G'$ results in removing the false positives and, asymptotically (as the sample size increases), converges to the correct model. In particular, BIC is attempting to select the candidate model corresponding to the highest Bayesian Posterior probability, which can be proved to be equivalent to the presented score and its $log(s)$ penalization factor.

We denote with $G^* = (V, E^*)$ the graph that we obtain after this step. We note that, as for general Bayesian network, $G^*$ is a DAG by construction.

### F.3.3 Confidence score

Using the reconstructed SBCN, we can represent the probabilistic relationships between any set of events (nodes). As an example, suppose to consider the nodes representing respectively $income = low$ and $gender = female$ being the only two direct causes (i.e., with arcs toward) of $loan = denial$. Given SBCN, we can estimate the conditional probabilities for each node in the graph, i.e., probability of $loan = denial$ given $income = low \ AND \ gender = female$ in the example, by computing the conditional probability of only the pair of nodes directly connected by an arc. For an overview of state-of-the-art methods for doing this, see [101]. However, we expect to be mostly dealing with full data, i.e., for every directly connected node in the SBCN, we expect to have several observations of any possible combination $attribute = value$. For this reason, we can simply estimate the node probabilities by counting the observations in the data. Moreover, we will exploit such conditional probabilities to define the confidence score of each arc in terms of their causal relationship.

In particular, for each arc $(v, u) \in E^*$ involving the causal relationship between two nodes $u, v \in V$, we define a confidence score $W(v, u) = \mathcal{P}(u \mid v) - \mathcal{P}(u \mid \neg v)$, which, intuitively, aims at estimating the observations where the cause $v$ is followed by its effect $u$, that is $\mathcal{P}(u \mid v)$, and the ones where this is not observed, i.e., $\mathcal{P}(u \mid \neg v)$, because of imperfect causal regularities. We also note that, by the constraints discussed above, we require $\mathcal{P}(u \mid v) \gg \mathcal{P}(u \mid \neg v)$ and, for this reason, each weight is positive and no larger than 1, i.e., $W : E^* \rightarrow (0, 1]$.

Combining all of the concepts discussed above, we conclude with the following definition.

**Definition 9** (Suppes-Bayes Causal Network). Given an input dataset $\mathcal{D}'$ of $m$ Bernoulli variables and $s$ samples, and given a partial order $r$ of the variables, the Suppes-Bayes Causal Network $SBCN = (V, E^*, W)$ subsumed by $\mathcal{D}'$ is a weighted DAG such that the following requirements hold:

- **[Suppes' constraints]** for each arc $(v, u) \in E^*$ involving the causal relationship between nodes $u, v \in V$, under the mild assumptions that $0 < \mathcal{P}(u), \mathcal{P}(v) < 1$:

$$r(v) \leq r(u) \quad and \quad \mathcal{P}(u \mid v) > \mathcal{P}(u \mid \neg v).$$

- **[Simplification]** let $E'$ be the set of arcs satisfying the Suppes' constraints as before; among all the subsets of $E'$, the set of arcs $E^*$ is the one whose corresponding graph maximizes BIC:

$$E^* = \underset{E \subseteq E', G=(V,E)}{\arg\max} \left( LL(\mathcal{D}', |G) - \frac{\log s}{2} dim(G) \right).$$

- **[Score]** $W(v, u) = \mathcal{P}(u \mid v) - \mathcal{P}(u \mid \neg v),\ \forall(v, u) \in E^*$

An example of a portion of a SBCN extracted from a real-world dataset is reported in Figure §F.1. Algorithm §7 summarized the learning approach adopted for the inference of the SBCN .

---

**Algorithm 7:** Learning the Suppes-Bayes Causal Network

1: Inputs: $\mathcal{D}'$ an input dataset of $m$ Bernoulli variables and $s$ samples,
    and $r$ a partial order of the variables
2: Output: $SBCN(V, E^*, W)$ as in Definition 2
3: **[Suppes' constraints]**
4: **for all** the arcs $(v, u)$ between each pair of the $m$ Bernoulli variables **do**
5:    **if** $r(v) \leq r(u)$ **and** $\mathcal{P}(u \mid v) > \mathcal{P}(u \mid \neg v)$ **then**
6:       Set to the arc $(v, u)$ its weight, being $W(v, u) = \mathcal{P}(u \mid v) - \mathcal{P}(u \mid \neg v)$.
7:       Add the arc $(v, u)$ to $SBCN$.
8:    **end if**
9: **end for**
10: **[Simplification]**
11: Consider $G(V, E^*, W)_{fit} = \emptyset$.
12: **while** $!StoppingCriterion()$ **do**
13:    Let $G(V, E^*, W)_{neighbors}$ be the neighbor solutions of $G(V, E^*, W)_{fit}$.
14:    Remove from $G(V, E^*, W)_{neighbors}$ any solution whose arcs are
       not included in $SBCN$.
15:    Consider a random solution $G_{current}$ in $G(V, E^*, W)_{neighbors}$.
16:    **if** $score_{BIC}(\mathcal{D}', G_{current}) > score_{BIC}(\mathcal{D}', G_{fit})$ **then**
17:       $G_{fit} = G_{current}$.
18:       Assign to the arcs of $G_{fit}$ the related weights of $SBCN$.
19:    **end if**
20: **end while**
21: $SBCN = G_{fit}$.
22: **return** $SBCN$.

---

Given $\mathcal{D}'$ an input dataset over $m$ Bernoulli variables and $s$ samples, and $r$ a partial order of the variables, Suppes' constraints are verified (Lines 4-9) to construct a DAG as described in Section §F.3.1. The likelihood fit is performed by hill climbing (Lines 12-21), an iterative optimization technique that starts with an arbitrary solution to a problem (in this case an empty graph) and then attempts to find a better solution by incrementally visiting the neighbourhood of the current one. If the new candidate solution is better than the previous one it is considered in place of it. The procedure is repeated until the stopping criterion is matched. The $!StoppingCriterion$ occurs (Line 12) in two situations: ($i$) the procedure stops when we have performed a large enough number of iterations or, ($ii$) it stops when none of the solutions in $G_{neighbors}$ is better than the current $G_{fit}$. Note that $G_{neighbors}$ denotes all the solutions that are derivable from $G_{fit}$ by removing or adding at most one edge.

**Time and space complexity.** The computation of the valid DAG according to Suppes' constraints (Lines 4-10) requires a pairwise calculation of the probabilistic scores leading to a polynomial cost. After that, the likelihood fit by hill climbing (Lines 11-21) is performed[5]. Hence, let $m$ denotes the number of the Bernoulli variables and $s$ the number of records in $\mathcal{D}'$, and $l$ the maximum number of iterations required for the hill climbing, the total computational complexity of Algorithm §7 is $O(s \cdot m)$ in time and $m^2$ in space.

### F.3.4 Expressivity of a SBCN

We conclude this Section with a discussion on the causal relations that we model by a $SBCN$.

Let us assume that there is one true (but unknown) probability distribution that generates the observed data whose structure can be modelled by a DAG. Furthermore, let us consider the causal structure of such a DAG and let us also assume each node with more then one cause to have conjunctive parents: any observation of the child node is preceded by the occurrence of all its parents. As before we call correct model, the statistical model which best approximate the distribution. On these settings, we can prove the following theorem.

**Theorem 1.** Let the sample size $s \to \infty$, the provided partial temporal order $r$ be correct and complete and the data be uniformly randomly corrupted by false positives and negatives rates (in $[0, 1)$), then the $SBCN$ inferred from the data is the correct model.

*Proof. [Sketch]* Let us first consider the case where the observed data have no noise. On such an input, we observe that the prima facie graph has no false negatives: in fact $\forall [c \to e]$ modelling a genuine causal relation, $\mathcal{P}(e \wedge c) = \mathcal{P}(e)$, thus the probability raising constraint is satisfied, so it is the temporal priority given that we assumed $r$ to be correct and complete.

Furthermore, it is know that the likelihood fit performed by $BIC$ converges to a class of structures equivalent in terms of likelihood among which there is the correct model: all these topologies are the same unless the directionality of some edges. But, being the prima facie graph already ordered by temporal priority, we can conclude that in this case the $SBCN$ coincides with the correct model.

To extend the proof to the case of data uniformly randomly corrupted by false positives and negatives rates (in $[0, 1)$), we note that the marginal and joint probabilities change monotonically as a consequence of the assumption that the noise is uniform. Thus, all inequalities used in the preceding proof still hold, which concludes the proof. $\qquad \square$

---

[5]Note that being an heuristic, the computational cost of hill climbing depends on the sopping criterion. However, constraining by Suppes' criteria tends to regularize the problem leading on average to a quick convergence to a good solution.

In the more general case of causal topologies where any cause of a common effect is independent from any other cause (i.e., we relax the assumption of conjunctive parents), the $SBCN$ is not guaranteed to converge to the correct model but it coincides with a subset of it modeling all the edges representing statistically relevant causal relations (i.e., where the probability raising condition is verified).

## F.4   Discrimination discovery by random walks

In this Section we propose several random-walk-based methods over the reconstructed SBCN, to deal with different discrimination-detection tasks.

### F.4.1   Group discrimination and favoritism

The basic problem in the analysis of direct discrimination is precisely to quantify the degree of discrimination suffered by a given protected group (e.g., an ethnic group) with respect to a decision (e.g., loan denial). In contrast to discrimination, favoritism refers to the case of an individual treated better than others for reasons not related to individual merit or business necessity: for instance, favoritism in the workplace might result in a person being promoted faster than others unfairly. In the following we denote favoritism as positive discrimination in contrast with negative discrimination.

Given an SBCN we define a measure of group discrimination (either negative or positive) for each node $v \in V$. Recall that each node represents a pair $\langle attribute = value \rangle$, so it is essentially what we refer to as a group, e.g., $\langle gender = female \rangle$. The task is to assign a score of discrimination $ds^- : V \to [0, 1]$ to each node, so that the closer $ds^-(v)$ is to 1 the more discriminated is the group represented by $v$.

We compute this score by means of a number $n$ of random walks that start from $v$ and reaches either the node representing the positive decision or the one representing the negative decision. In these random walks the next step is chosen proportionally to the weights of the out-going arcs. Suppose a random walk has reached a node $u$, and let $deg_{out}(u)$ denote the set of outgoing arcs from $u$. Then the arc $(u, z)$ is chosen with probability

$$p(u, z) = \frac{W(u, z)}{\sum_{e \in deg_{out}(u)} W(e)}.$$

When a random walk ends in a node with no outgoing arc before reaching either the negative or the positive decision, it is restarted from the source node $v$.

**Definition 10** (Group discrimination score). Given a $SBCN = (V, E^*, W)$, let $\delta^- \in V$ and $\delta^+ \in V$ denote the nodes indicating the negative and positive decision, respectively. Given a node $v \in V$, and a number $n \in \mathbb{N}$ of random walks to be performed, we denote as $rw_{v \to \delta^-}$ the number of random walks started at node $v$ that reach $\delta^-$ earlier than $\delta^+$. The discrimination score for the group corresponding to node $v$ is then defined as

$$ds^-(v) = \frac{rw_{v \to \delta^-}}{n}.$$

This implicitly also defines a score of positive discrimination (or favoritism): $ds^+(v) = 1 - ds^-(v)$.

Taking advantage of the SBCN we also propose two additional measures capturing how far a node representing a group is from the positive and negative decision respectively. This is done by computing the average number of steps that the random walks take to reach the two decisions: we denote these scores as $as^-(v)$ and $as^+(v)$.

### F.4.2 Indirect discrimination

The European Union Legislation [109] provides a broad definition of indirect discrimination as occurring "where an apparently neutral provision, criterion or practice would put persons of a racial or ethnic origin at a particular disadvantage compared with other persons". In other words, the actual result of the apparently neutral provision is the same as an explicitly discriminatory one. A typical legal case study of indirect discrimination is concerned with *redlining*: e.g., denying a loan because of ZIP code, which in some areas is an attribute highly correlated to race. Therefore, even if the attribute race cannot be required at loan-application time (thus would not be present in the data), still race discrimination is perpetrated. Indirect discrimination discovery refers to the data mining task of discovering the attributes values that can act as a proxy to the protected groups and lead to discriminatory decisions indirectly [153, 180, 73].

In the considered setting, indirect discrimination can be detected by applying the same method described in Section §F.4.1.

### F.4.3 Genuine requirement

The legal concept of genuine requirement refers to detecting that part of the discrimination which may be explained by other, legally-grounded, attributes; e.g., denying credit to women may be explainable by the fact that most of them have low salary or delay in returning previous credits. A typical example in the literature is the one of the "genuine occupational requirement", also called "business necessity" in [108, 49]. In the state of the art of data mining methods for discrimination discovery, it is also known as *explainable discrimination* [74] and *conditional discrimination* [200].

The task here is to evaluate to which extent the discrimination apparent for a group is "explainable" on a legal ground. Let $v \in V$ be the node representing the group which is suspected of being discriminated, and $u_l \in V$ be a node whose causal relation with a negative or positive decision is legally grounded. As before, $\delta^-$ and $\delta^+$ denote the negative and positive decision, respectively. Following the same random-walk process described in Section §F.4.1, we define the *fraction of explainable discrimination* for the group $v$:

$$fed^-(v) = \frac{rw_{v \to u_l \to \delta^-}}{rw_{v \to \delta^-}},$$

i.e., the fraction of random walks passing trough $u_l$ among the ones started in $v$ and reaching $\delta^-$ earlier than $\delta^+$. Similarly we define $fed^+(v)$, i.e., the fraction of explainable positive discrimination.

### F.4.4   Individual and subgroup discrimination

Individual discrimination requires to measure the amount of discrimination for a specific individual, i.e., an entire record in the database. Similarly, subgroup discrimination refers to discrimination against a subgroup described by a combination of multiple protected and non-protected attributes: personal data, demographics, social, economic and cultural indicators, etc. For example, consider the case of gender discrimination in credit approval: although an analyst may observe that no discrimination occurs in general, it may turn out that older women obtain car loans only rarely.

Both problems can be handled by generalizing the technique introduced in Section §F.4.1 to deal with a set of starting nodes, instead of only one. Given an $SBCN = (V, E^*, W)$ let $v_1, \ldots, v_n$ be the nodes of interest. In order to define a discrimination score for $v_1, \ldots, v_n$, we perform a *personalized PageRank*[90] computation with respect to $v_1, \ldots, v_n$. In personalized PageRank, the probability of jumping to a node when abandoning the random walk is not uniform, but it is given by a vector of probabilities for each node. In this case the vector will have the value $\frac{1}{n}$ for each of the nodes $v_1, \ldots, v_n \in V$ and zero for all the others. The output of personalized PageRank is a score $ppr(u|v_1, \ldots, v_n)$ of proximity/relevance to $\{v_1, \ldots, v_n\}$ for each other node $u$ in the network. In particular, we are interested in the score of the nodes representing the negative and positive decision: i.e., $ppr(\delta^-|v_1, \ldots, v_n)$ and $ppr(\delta^+|v_1, \ldots, v_n)$ respectively.

**Definition 11** (Generalized discrimination score). Given an $SBCN = (V, E^*, W)$, let $\delta^- \in V$ and $\delta^+ \in V$ denote the nodes indicating the negative and positive decision, respectively. Given a set of nodes $v_1, \ldots, v_n \in V$, we define the generalized (negative) discrimination score for the subgroup or the individual represented by $\{v_1, \ldots, v_n\}$ as

$$gds^-(v_1, \ldots, v_n) = \frac{ppr(\delta^-|v_1, \ldots, v_n)}{ppr(\delta^-|v_1, \ldots, v_n) + ppr(\delta^+|v_1, \ldots, v_n)}.$$

This implicitly also defines a generalized score of positive discrimination: $gds^+(v_1, \ldots, v_n) = 1 - gds^-(v_1, \ldots, v_n)$.

## F.5   Experimental Evaluation

This section reports the experimental evaluation of this approach on four datasets, *Adult*, *German credit* and *census-income* from the UCI Repository of Machine Learning Databases[6], and *Berkeley Admissions Data* from [57]. These are well-known real-life datasets typically used in discrimination-detection literature.

**Adult**: consists of 48,842 tuples and 10 attributes, where each tuple correspond to an individual and it is described by personal attributes such as age, race, sex, relationship, education, employment, etc. Following the literature, in order to define the decision attribute we use the income levels, ≤50K (negative decision) or >50K (positive decision). We use four levels in the partial order for temporal priority: age, race, sex, and native country are defined in the first level; education, marital status, and relationship are defined in the second level; occupation and work class are defined in the third class, and the decision attribute (derived from income) is the last level.

---

[6] http://archive.ics.uci.edu/ml

**German credit**: consists of 1000 tuples with 21 attributes on bank account holders applying for credit. The decision attribute is based on repayment history, i.e., whether the customer is labeled with good or bad credit risk. Also for this dataset the partial order for temporal priority has four orders. Personal attributes such as gender, age, foreign worker are defined in the first level. Personal attributes such as employment status and job status are defined in the second level. Personal properties such as savings status and credit history are defined in the third level, and finally the decision attribute is the last level.

**Census-income**: consists of 299,285 tuples and 40 attributes, where each tuple correspond to an individual and it is described by demographic and employment attributes such as age, sex, relationship, education, employment, ext. Similar to **Adult** dataset, the decision attribute is the income levels and we define four levels in the partial order for temporal priority.

The main characteristics of the extracted **SBCN** are reported in Table §F.1, while the distribution of the edges scores $W(e)$ is plotted in Figure §F.2.

| Dataset | $|V|$ | $|A|$ | avgDeg | maxInDeg | maxOutDeg |
|---|---|---|---|---|---|
| Adult | 92 | 230 | 2.5 | 7 | 19 |
| German credit | 73 | 102 | 1.39 | 3 | 7 |
| Census-income | 386 | 1426 | 3.69 | 8 | 54 |

Table F.1: **SBCN** main characteristics.

As discussed in the Introduction we also use the dataset from the famous 1973 episode at University of California at Berkeley, in order to highlight the differences between correlation-based and causation-based methods.

**Berkeley Admissions Data**: consists of 4,486 tuples and three attributes, where each tuple correspond to an individual and it is described by the gender of applicants and the department that they apply for it. For this dataset the partial order for temporal priority has three orders. Gender is defined in the first level, department in the second level, and finally the decision attribute is the last level. Table §F.2 is a three-way table that presents admissions data at the University of California, Berkeley in 1973 according to the variables department (A, B, C, D, E), gender (male, female), and outcome (admitted, denied). The table is adapted from data in the text by Freedman, et al. [57].

### F.5.1 Community detection on the **SBCN**

Given that the **SBCN** is a directed graph with edge weight, as a first characterization we try to partition it using a random-walks-based community detection algorithm, called

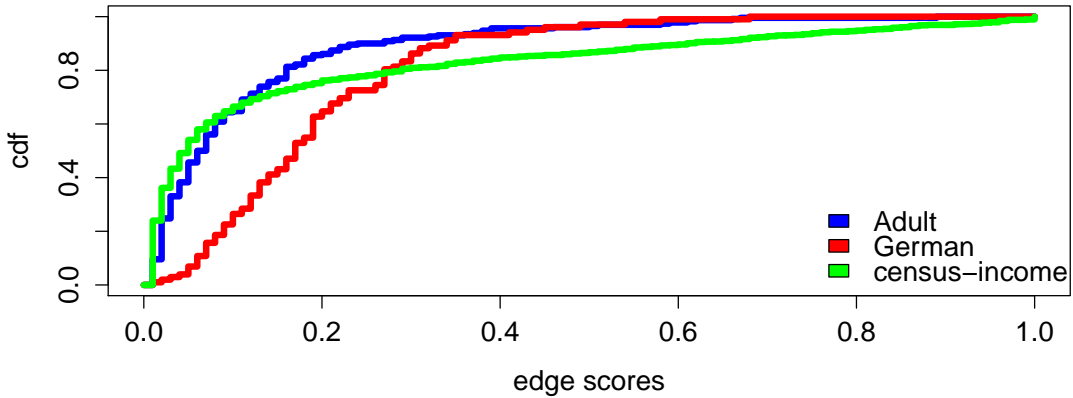

Figure F.2: Distribution of the edge scores.

|        | Male   |          | Female   |        |            |
|--------|--------|----------|----------|--------|------------|
| Admitted | Denied | Admitted | Denied | Department |
| 512    | 313    | 89       | 19     | A          |
| 313    | 207    | 17       | 8      | B          |
| 120    | 205    | 202      | 391    | C          |
| 138    | 279    | 131      | 244    | D          |
| 53     | 138    | 94       | 299    | E          |
| 22     | 351    | 24       | 317    | F          |

Table F.2: Berkeley Admission Data

*Walktrap* and proposed in [155], whose unique parameter is the maximum number of steps in a random walk (we set it to 8), and which automatically identifies the right number of communities. The idea is that short random walks tend to stay in the same community (densely connected area of the graph). Using this algorithm over the reconstructed SBCN from Adult dataset, we obtain 5 communities: two larger ones and three smaller ones (reported in Table §F.3). Interestingly, the two larger communities seem built around the negative ($C_1$) and the positive ($C_2$) decisions.

Figure §F.1 in Section §F.3 shows the subgraph of the SBCN corresponding to $C_2$ (that we can call, the favoritism cluster): we note that such cluster also contains nodes such as sex_Male, age_old, relationship_Husband. The other large community $C_1$, can be considered the discrimination cluster: beside the negative decision it contains other nodes representing disadvantaged groups such as sex_Female, age_young, race_Black, marital_status_Never_married. This good separability of the SBCN in the two main clusters of discrimination and favoritism, highlights the goodness of the causal structure captured

by the SBCN.

| $C_1$ |
|---|
| **negative_dec**, wc:Private, ed:Some_college, ed:Assoc_acdm, **ms:Never_married**, ms:Divorced, **ms:Widowed**, ms:Married_AF_spouse, oc:Sales, oc:Other_service, oc:Priv_house_serv, re:Own_child, re:Not_in_family, re:Wife, **re:Unmarried**, re:Other_relative, **ra:Black**, oc:Armed_Forces, oc:Handlers_cleaners, oc:Tech_support, oc:Transport_moving, ed:7th_8th, ed:10th, ed:12th, ms:Separated, ed:HS_grad,ed:11th, nc:Outlying_US_Guam_USVI_etc, nc:Haiti, **ag:young**, **sx:Female**, ra:Amer_Indian_Eskimo, nc:Trinadad_Tobago, nc:Jamaica, oc:Machine_op_inspct, ms:Married_spouse_absent, oc:Adm_clerical, |

| $C_2$ |
|---|
| **positive_dec**, oc:Prof_specialty, wc:Self_emp_not_inc, ms:Married_civ_spouse, oc:Craft_repair,oc:Protective_serv, **re:Husband**, ed:Prof_school, wc:Self_emp_inc, **ag:old** , wc:Local_gov, **oc:Exec_managerial**, ed:Bachelors, ed:Assoc_voc, ed:Masters, wc:Never_worked, wc:State_gov, ed:Doctorate, **sx:Male**, nc:India, nc:Cuba |

| $C_3$ |
|---|
| oc:Farming_fishing, wc:Without_pay, nc:Mexico, nc:Canada, nc:Italy, nc:Guatemala, nc:El_Salvador, ra:White, nc:Poland, ed:1st_4th, ed:9th,ed:Preschool, ed:5th_6th |

| $C_4$ |
|---|
| nc:Iran, nc:Puerto_Rico, nc:Dominican_Republic, nc:Columbia, nc:Peru, nc:Nicaragua, ra:Other |

| $C_5$ |
|---|
| nc:Philippines, nc:Cambodia, nc:China, nc:South, nc:Japan, nc:Taiwan, nc:Hong, nc:Laos, nc:Thailand, nc:Vietnam, ra:Asian_Pac_Islander |

Table F.3: Communities found in the SBCN extracted from the Adult dataset by *Walktrap*[155]. In the table the attributes are shortened as in parenthesis: age (ag), education (ed), marital_status (ms), native_country (nc), occupation (oc), race(ra), relationship (re), sex (sx), workclass (wc).

## F.5.2 Group discrimination and favoritism

We next focus on assessing the discrimination score $ds^-$ we defined in Section §F.4.1, as well as the average number of steps that the random walks take to reach the negative and positive decisions, denote $as^-(v)$ and $as^+(v)$ respectively.

Tables §F.4, §F.5 and §F.6 report the top-5 and bottom-5 nodes w.r.t. the discrimination score $ds^-$, for datasets Adult, German and Census-income, respectively. The first and most important observation is that this discrimination score provides a very clear signal, with some disadvantaged groups having very high discrimination score (equal to 1 or very close), and similarly clear signals of favoritism, with groups having $ds^-(v) = 0$, or equivalently $ds^+(v) = 1$. This is more clear in the Adult dataset, where the positive and negative decisions are artificially derived from the income attribute. In the German credit dataset, which is more realistic as the decision attribute is truly about credit, both discrimination and favoritism are less palpable. This is also due to the fact that German credit contains less proper causal relations, as reflected in the higher sparsity of the SBCN. A consequence of this sparsity is also that the random walks generally need more steps to reach one of the two decisions. In Census-income dataset, we observe favoritism with respect to married and asian_pacific individuals.

### F.5.3 Genuine requirement

We next focus on genuine requirement (or explainable discrimination). Table §F.7 reports some examples of fraction of explainable discrimination (both positive and negative) on the Adult dataset. We can see how some fractions of discrimination against protected groups, can be "explained" by intermediate nodes such as having a low education profile, or a simple job. In the case these intermediate nodes are considered legally grounded, then one cannot easily file a discrimination claim.

Similarly, one can observe that the favoritism towards groups such as married man, is not just simple favoritism but it is explainable, to a large extent, by higher education and good working position, such as managerial or executive roles.

### F.5.4 Subgroup and Individual Discrimination

We next turn the attention to subgroup and individual discrimination discovery. Here the problem is to assign a score of discrimination not to a single node (a group), but to multiple nodes (representing the attributes of an individual or a subgroup of citizens). In Section §F.4.4 we have introduced based on the PageRank of the positive and negative decision, $ppr(\delta^+)$ and $ppr(\delta^-)$ respectively, personalized on the nodes of interest. Figure §F.3 presents a scatter plot of $ppr(\delta^+)$ versus $ppr(\delta^-)$ for each individual in the German credit dataset. One can observe the perfect separation between individuals corresponding to a high personalized PageRank with respect to the positive decision, and those associated with a high personalized PageRank relative to the negative decision.

Such good separation is also reflected in the *generalized discrimination score* (Definition 4) that we obtain by combining $ppr(\delta^+)$ versus $ppr(\delta^-)$.

In Figure §F.4 we report the distribution of the generalized discrimination score $gds^-$ for the population of the German credit dataset: one can make a note of the clear separation between the two subgroups of the population.

In the Adult dataset (Figure §F.5) we do not observe the same neat separation in two subgroups as in the German credit dataset, also due to the much larger number of points. Nevertheless, as expected, $ppr(\delta^+)$ and $ppr(\delta^-)$ still exhibit anticorrelation. In Figure §F.5 we also use colors to show two different groups: red dots are for age_Young and blue dots are for age_Old individuals. As expected we can see that the red dots are distributed more in the area of higher $ppr(\delta^-)$.

The plots in Figure §F.6 have a threshold $t \in [0, 1]$ on the X-axis, and the fraction of tuples having $gds^-() \geq t$ on the Y-axis, and they show this for different subgroups. The first plot, from the Adult dataset, shows the group female, young, and young female. As we can see the individuals that are both young and female have a higher generalized discrimination score. Similarly, the second plot shows the groups old, single male, and old single male from the German credit dataset. Here one can observe much lower rates of discrimination with only 1/5 of the corresponding populations having $gds^-() \geq 0.5$, while in the previous plot it was more than 85%.

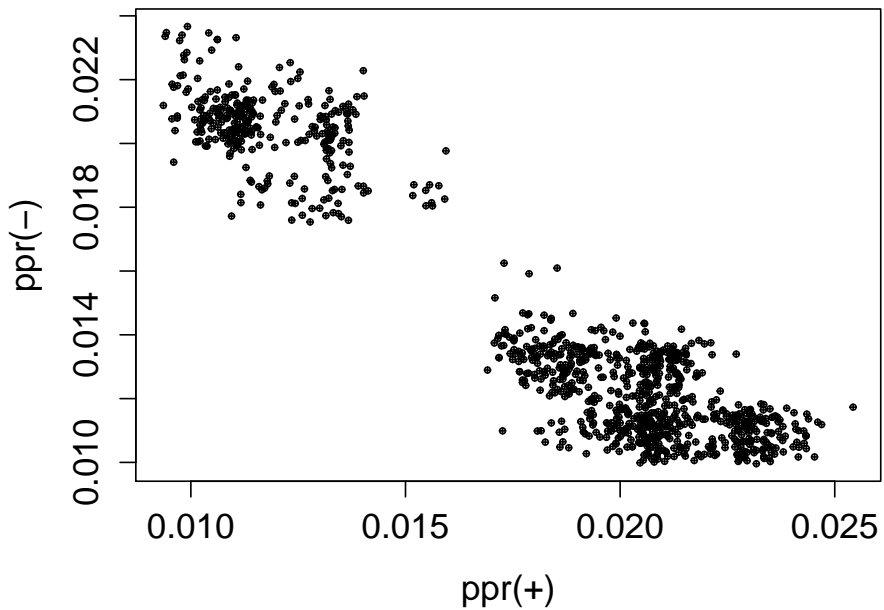

Figure F.3: Scatter plot of $ppr(\delta^+)$ versus $ppr(\delta^-)$ for each individual in the German credit dataset.

### F.5.5 Comparison with prior art

We next discuss examples in which the causation-based method draws different conclusions from the correlation-based methods presented in [153, 180, 165] using the same datasets and the same protected groups.[7]

The first example involves the foreign_worker group from German Credit dataset, whose contingency table is reported in Figure §F.8. Following the approaches of [153, 180, 165] the foreign_worker group results strongly discriminated. In fact Figure §F.8 shows an $RD$ value (*risk difference*) of 0.244 which is considered a strong signal: in fact $RD > 0$ is already considered discrimination [165].

However, we can observe that the foreign_worker group is per se not very significant, as it contains 963 tuples out of 1000 total. In fact the causal approach does not detect any discrimination with respect to foreign_worker which appears as a disconnected node in the SBCN.

The second example is in the opposite direction. Consider the race_black group from Adult dataset whose contingency table is shown in Figure §F.9. The causality-based approach detects a very strong signal of discrimination ($ds^-() = 0.994$), while the approachs of [153, 180, 165] fail to discover discrimination against black minority when the value of minimum support threshold used for extracting classification rules is more than 10%. On the other hand, when such minimum support threshold is kept lower, the number of extracted rules might be overwhelming. Moreover, the value of $RD$ is not very strong, while in this method the discrimination reported is strong, regardless of the small size of the black population contained in the dataset.

---

[7]We could not compare with [123] due to repeatability issues: we contacted the authors asking their help, but we didn't get a reply.

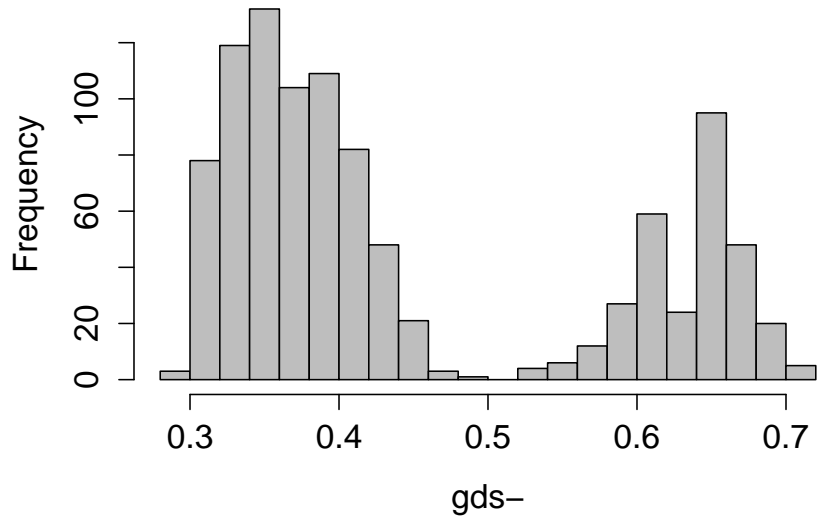

Figure F.4: Individual discrimination: histogram representing the distribution of the values of the generalized discrimination score $gds^-$ for the population of the German credit dataset.

Figure §F.7 presents the SBCN extracted by this approach from Berkeley Admission Data. Interestingly, we observe that there is no direct edge between node sex_Female and Admission_No. And sex_Female is connected to node Admission_No through nodes of Dep_C, Dep_D, Dep_E, and Dep_F, which are exactly the departments that have lower admission rate. By running the random walk-based methods over SBCN we obtain the value of 1 for the score of explainable discrimination confirming that apparent discrimination in this dataset is due the fact that women tended to apply to departments with lower rates of admission.

Similarly, we observe that is no direct edge between node sex_Male and Admission_Yes. And sex_Male is connected to node Admission_Yes through nodes of Dep_A, and Dep_B, which are exactly the departments that have higher admission rate. By running the random walk-based methods over SBCN we obtain the value of 1 for the score of explainable discrimination confirming that apparent favoritism towards men is due to the fact that men tended to apply to departments with higher rates of admission.

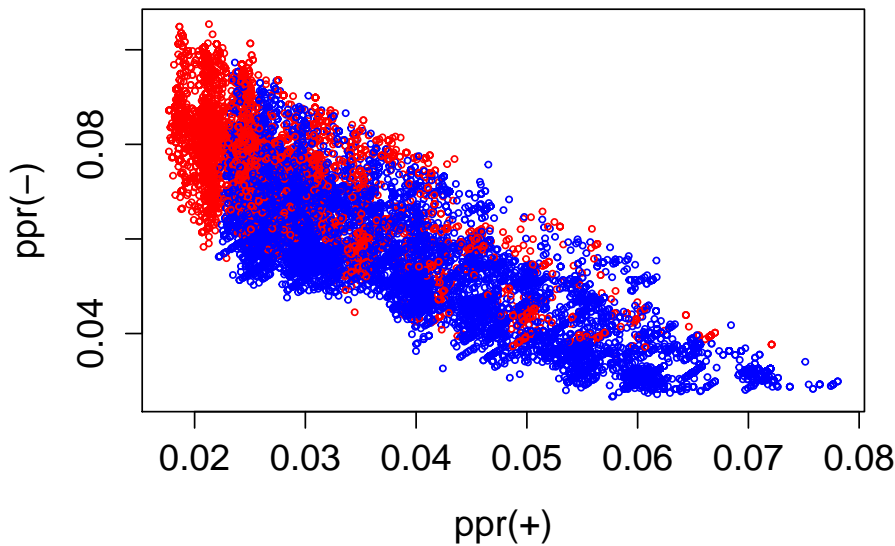

Figure F.5: Individual discrimination: scatter plot of $ppr(\delta^+)$ versus $ppr(\delta^-)$ for each individual in the Adult dataset. Red dots are for age_Young and blue dots are for age_Old.

However, following the approaches of [153, 180, 165], the contingency table shown in Figure §F.10 can be extracted from Berkeley Admission Data. As shown in Figure §F.10, the value of $RD$ suggests a very strong signal of discrimination versus women. This highlights once more the pitfalls of correlation-based approaches to discrimination detection and the need for a principled causal approach.

## F.6 Conclusions

Discrimination discovery from databases is a fundamental task in understanding past and current trends of discrimination, in judicial dispute resolution in legal trials, in the validation of micro-data before they are publicly released. While discrimination is a causal phenomenon, and any discrimination claim requires to prove a causal relationship, the bulk of the literature on data mining methods for discrimination detection is based on correlation reasoning.

In this Chapter we propose the first discrimination detection method grounded in probabilistic causal theory. We first define a method to extract a graph representing the causal structures found in the database, and then we propose several random-walk-based methods over the causal structures, addressing a range of different discrimination problems.

The experimental assessment confirmed the great flexibility of the proposal in tackling different aspects of the discrimination detection task, and doing so with very clean signals, clearly separating discrimination cases.

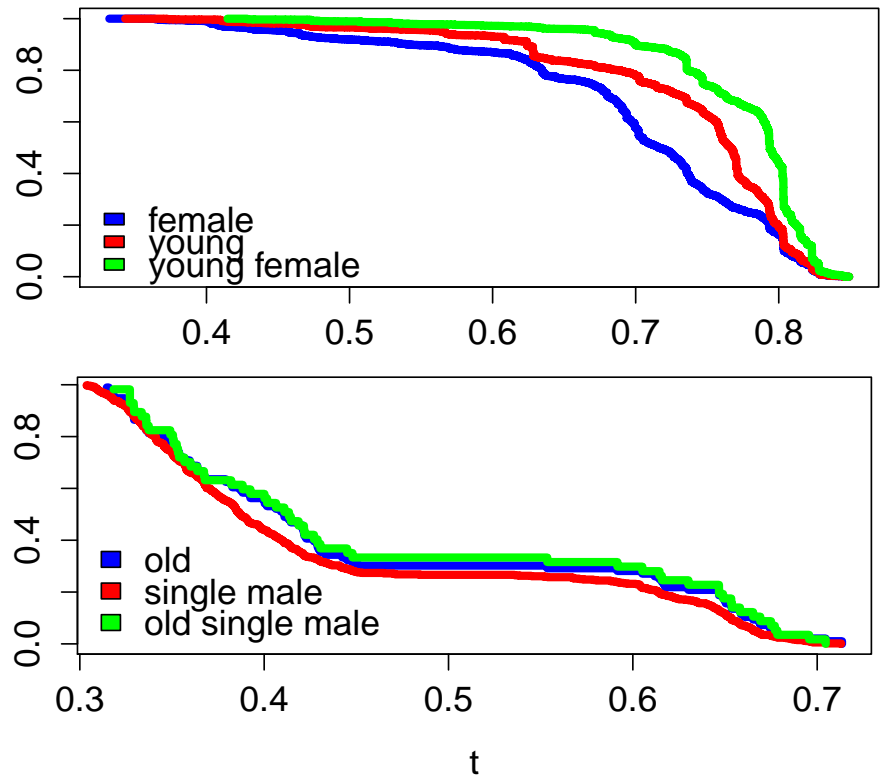

Figure F.6: Subgroup discrimination: plots reporting a threshold $t \in [0, 1]$ on the X-axis, and the fraction of tuples having $gds^-() \geq t$ on the Y-axis. The top plot is from Adult, while the bottom is from German credit.

|  | decision | | |
|---|---|---|---|
|  | - | + |  |
| foreign_worker=yes | 298 | 667 | 968 |
| foreign_worker=no | 2 | 30 | 32 |
|  | 300 | 700 | 1000 |

$p_1 = 298/968 = 0.307$
$p_2 = 2/32 = 0.0625$
$RD = p_1 - p_2 = 0.244$

Table F.8: Contingency table for foreign_worker in the German credit dataset.

|  | decision | | |
|---|---|---|---|
|  | - | + |  |
| race=black | 4119 | 566 | 4685 |
| race≠black | 33036 | 11121 | 44157 |
|  | 37155 | 11687 | 48842 |

$p_1 = 4119/4685 = 0.879$
$p_2 = 33036/44157 = 0.748$
$RD = p_1 - p_2 = 0.13$

Table F.9: Contingency table for race_black in the Adult dataset.

|  | $ds^-(v)$ | $as^-(v)$ | $as^+(v)$ |
|---|---|---|---|
| relationship_Unmarried | 1 | 1.164 | - |
| marital_status_Never_married | 0.996 | 1.21 | 2.14 |
| age_Young | 0.995 | 2.407 | 3.857 |
| race_Black | 0.994 | 2.46 | 4.4 |
| sex_Female | 0.98 | 2.60 | 3.76 |

|  | $ds^-(v)$ | $as^-(v)$ | $as^+(v)$ |
|---|---|---|---|
| relationship_Husband | 0 | - | 2 |
| marital_status_Married_civ_spouse | 0 | - | 2.06 |
| sex_Male | 0 | - | 3.002 |
| native_country_India | 0.002 | 4.0 | 3.25 |
| age_Old | 0.018 | 2.062 | 2.14 |

Table F.4: Top-5 and bottom-5 groups by discrimination score $ds^-(v)$ in **Adult** dataset.

|  | $ds^-(v)$ | $as^-(v)$ | $as^+(v)$ |
|---|---|---|---|
| residence_since_le_1d6 | 1 | 6.0 | - |
| residence_since_gt_2d8 | 1 | 2.23 | - |
| residence_since_from_1d6_le_2d2 | 1 | 6.0 | - |
| age_gt_52d6 | 0.86 | 3.68 | 4.0 |
| personal_status_male_single | 0.791 | 5.15 | 5.0 |

|  | $ds^-(v)$ | $as^-(v)$ | $as^+(v)$ |
|---|---|---|---|
| job_unskilled_resident | 0 | - | 2.39 |
| personal_status_male_mar_or_wid | 0.12 | 8.0 | 4.4 |
| age_le_30d2 | 0.186 | 7.0 | 3.34 |
| personal_status_female_ | 0.294 | 6.48 | 4.4 |
| div_or_sep_or_mar | | | |

Table F.5: Top-5 and bottom-4 groups by discrimination score $ds^-(v)$ in **German credit**. We report only the bottom-4, because there are only 4 nodes in which $ds^+(v) > ds^-(v)$.

|  | $ds^-(v)$ | $as^-(v)$ | $as^+(v)$ |
|---|---|---|---|
| MIGSAME_Not_in_universe_under_1_year_old | 0.71 | 4.09 | 8.82 |
| WKSWORK_94_5_inf | 0.625 | 3.0 | 6.76 |
| AWKSTAT_Not_in_labor_force | 0.59 | 2.0 | 6.16 |
| VETYN_0_5_20_5 | 0.58 | 1.01 | 5.17 |
| MARSUPWT_3188_455_4277_98 | 0.55 | 5.0 | 9.25 |

|  | $ds^-(v)$ | $as^-(v)$ | $as^+(v)$ |
|---|---|---|---|
| AHGA_Doctorate_degreePhD_EdD | 0 | - | 3.07 |
| AMARITL_Married_A_F_spouse_present | 0 | - | 4.49 |
| AMJOCC_Sales | 0 | - | 2.0 |
| ARACE_Asian_or_Pacific_Islander | 0 | - | 6.47 |
| VETYN_20_5_32_5 | 0 | - | 5.89 |

Table F.6: Top-5 and bottom-5 groups by discrimination score $ds^-(v)$ in **Census-income** dataset.

| Source node | Intermediate | $fed^-(v)$ |
|---|---|---|
| race_Amer_Indian_Eskimo | education_HS_grad | 0.481 |
| sex_Female | occupation_Other_service | 0.310 |
| age_Young | occupation_Other_service | 0.193 |
| relationship_Unmarried | education_HS_grad | 0.107 |
| race_Black | education_11th | 0.083 |

| Source node | Intermediate | $fed^+(v)$ |
|---|---|---|
| relationship_Husband | occupation_Exec_managerial | 0.806 |
| sex_Male | occupation_Exec_managerial | 0.587 |
| native_country_Iran | education_Bachelors | 0.480 |
| native_country_India | education_Prof_school | 0.415 |
| age_Old | occupation_Exec_managerial | 0.39 |

Table F.7: Fraction of explainable discrimination for some exemplar pair of nodes in the Adult dataset.

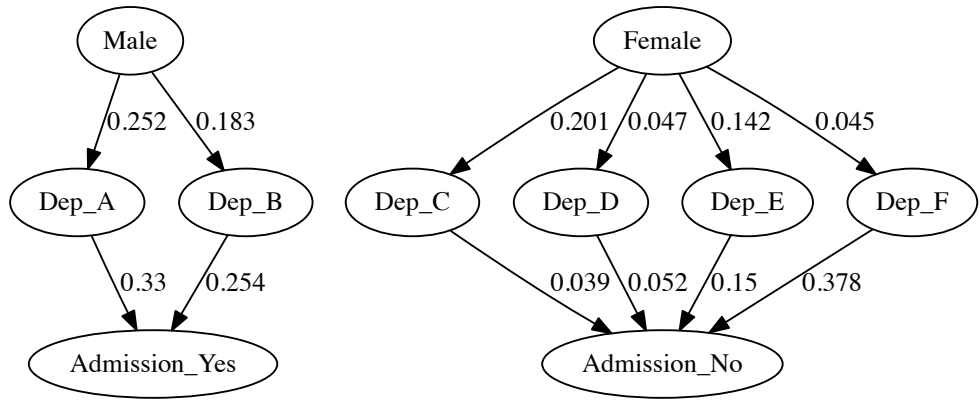

Figure F.7: The SBCN constructed from Berkeley Admission Data dataset.

| | decision | | |
|---|---|---|---|
| | − | + | |
| gender=female | 1278 | 557 | 1835 |
| gender=male | 1493 | 1158 | 2651 |
| | 2771 | 1715 | 4486 |

$p_1 = 1278/1835 = 0.696$
$p_2 = 1493/2651 = 0.563$
$RD = p_1 - p_2 = 0.133$

Table F.10: Contingency table for female in the Berkeley Admission Data dataset.